\documentclass{article}

\usepackage{microtype}
\usepackage{graphicx}
\usepackage{subcaption}
\usepackage{booktabs} 
\usepackage{comment}
\excludecomment{comment}

\usepackage{enumitem}
\usepackage{multirow}

\usepackage[preprint]{icml2026}

\usepackage{amsmath}
\usepackage{amssymb}
\usepackage{mathtools}
\usepackage{amsthm}

\usepackage{algorithm}
\usepackage{algorithmic}

\theoremstyle{plain}

\theoremstyle{definition}

\theoremstyle{remark}

\usepackage[textsize=tiny]{todonotes}

\usepackage[utf8]{inputenc} 
\usepackage[T1]{fontenc}    
\usepackage{url}            
\usepackage{booktabs}       
\usepackage{amsfonts}       
\usepackage{nicefrac}       
\usepackage{microtype}      
\usepackage[dvipsnames,svgnames,x11names,table]{xcolor}
\definecolor{CalGoldHex}{HTML}{d38102}
\definecolor{UWPurple}{HTML}{4b2e83}
\definecolor{IronGrey}{HTML}{6D6E71}

\usepackage{graphicx}
\usepackage{amsmath}
\usepackage{multicol}
\usepackage{multirow}
\usepackage{enumitem}
\usepackage{caption}
\usepackage{subcaption}
\usepackage{tcolorbox}
\usepackage{fancyvrb}
\RequirePackage[colorlinks=true, allcolors=RoyalBlue4]{hyperref}
\usepackage[capitalize,noabbrev]{cleveref}
\usepackage{xcolor} 
\tcbuselibrary{skins,breakable} 
\usepackage{subcaption}
\usepackage{wrapfig}
\usepackage{minitoc}
\usepackage{amssymb}
\usepackage{listings}
\usepackage{titletoc}
\usepackage{makecell}
\usepackage{lipsum}

\hypersetup{
    colorlinks = true,
    citecolor = {UWPurple},
}

\newcommand{\qwenmath}[0]{Qwen2.5-Math-7B\xspace}
\newcommand{\qwenmathfamily}[0]{Qwen2.5-Math\xspace}
\newcommand{\qwenbasefamily}[0]{Qwen2.5\xspace}

\newcommand{\qwenmathsmall}[0]{Qwen2.5-Math-1.5B\xspace}
\newcommand{\qwen}[0]{Qwen2.5-7B\xspace}
\newcommand{\qwensmall}[0]{Qwen2.5-1.5B\xspace}
\newcommand{\llama}[0]{Llama3.1-8B-Instruct\xspace}
\newcommand{\llamasmall}[0]{Llama3.2-3B-Instruct\xspace}
\newcommand{\llamabase}[0]{Llama3.1-8B\xspace}
\newcommand{\llamabasesmall}[0]{Llama3.2-3B\xspace}
\newcommand{\olmo}[0]{OLMo2-7B\xspace}
\newcommand{\olmosft}[0]{OLMo2-7B-SFT\xspace}

\definecolor{forestgreen}{rgb}{0.13, 0.55, 0.13}

\newcommand{\cblock}[3]{
  \mbox{
  \protect\hspace{-1.5mm}
  \protect\begin{tikzpicture}
    \protect\node[draw, minimum size=2.5mm, thick, line width=0.5pt, 
          fill={rgb,255:red,#1;green,#2;blue,#3}] () {};
  \protect\end{tikzpicture}%
  }
}
  
\usepackage{xspace}

\begin{document}

\twocolumn[
  \icmltitle{Spurious Rewards: Rethinking Training Signals in RLVR}

  \icmlsetsymbol{equal}{*}

  \begin{icmlauthorlist}
    \icmlauthor{Rulin Shao}{equal,uw}
    \icmlauthor{Shuyue Stella Li}{equal,uw}
    \icmlauthor{Rui Xin}{equal,uw}
    \icmlauthor{Scott Geng}{equal,uw}
    \icmlauthor{Yiping Wang}{uw}
    \icmlauthor{Sewoong Oh}{uw} 
    \icmlauthor{Simon Shaolei Du}{uw}
    \icmlauthor{Nathan Lambert}{ai2}
    \icmlauthor{Sewon Min}{ucb}
    \icmlauthor{Ranjay Krishna}{uw,ai2}
    \icmlauthor{Yulia Tsvetkov}{uw}
    \icmlauthor{Hannaneh Hajishirzi}{uw,ai2}\\
    \icmlauthor{Pang Wei Koh}{uw,ai2}
    \icmlauthor{Luke Zettlemoyer}{uw}
  \end{icmlauthorlist}

  \begin{center}
    \vspace{-0.3\baselineskip}
    \href{https://github.com/ruixin31/Spurious_Rewards}{%
      \raisebox{-0.15em}{\includegraphics[height=1em]{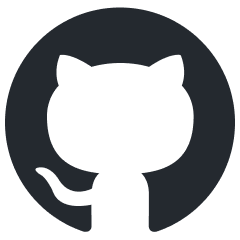}}%
      \hspace{0.45em}\texttt{Spurious\_Rewards}%
    }
  \end{center}

  \icmlaffiliation{uw}{University of Washington, Seattle, WA, USA}
  \icmlaffiliation{ai2}{Allen Institute for Artificial Intelligence, Seattle, WA, USA}
  \icmlaffiliation{ucb}{University of California, Berkeley, Berkeley, CA, USA}

  \icmlcorrespondingauthor{}{\{rulins,stelli,rx31,sgeng\}@cs.washington.edu}

  \vskip 0.3in
]

\printAffiliationsAndNotice{\icmlEqualContribution}

\begin{abstract}
We show that reinforcement learning with verifiable rewards (RLVR) can elicit strong mathematical reasoning in certain language models even with \emph{spurious rewards} that have little, no, or outright negative correlation with the correct answer.
For example, RLVR training with GRPO improves MATH-500
performance for \qwenmath in absolute points by 21.4\% using randomly assigned rewards, nearly matching the 29.1\% gained with ground truth rewards. 
To explain this counterintuitive observation, we show
that GRPO exhibits a \emph{clipping bias} arising from the clip term, which can amplify high-prior behaviors learned during pre-training even without informative rewards.
As a case study, we identify one such high-prior behavior for \qwenmathfamily models, which we term \emph{code reasoning}---reasoning in code without actual code execution;
code reasoning frequency increases from 65\% to over 90\% with spurious rewards.
However, the presence of such amplifiable behaviors is highly model-dependent. 
In practice, spurious rewards that are effective for Qwen models often fail to produce gains for other model families, such as Llama3 or OLMo2.
Our results highlight the importance of validating RL methods across diverse models rather than relying on a single de facto choice: large performance gains can arise on Qwen models even from random rewards that do not reflect genuine capability improvements.
\end{abstract}

\section{Introduction}

Reinforcement learning with verifiable rewards (RLVR) is highly effective in improving language model reasoning~\citep{lambert2024tulu3, deepseekteam2024deepseek, zeng2025simplerl, deepscaler2025}, but the mechanisms underlying these gains remain poorly understood.
In this work, we show counterintuitively that RLVR can yield substantial performance improvements on math reasoning even under a family of \emph{spurious rewards} that carry little to no task-relevant signal and that this effect depends critically on the base model.
For example, on the \qwenmathfamily models~\citep{yang2024qwen2,Yang2024Qwen25TR}, which are widely used in the RLVR literature, randomly assigned rewards yield a 21.4\% absolute accuracy gain on MATH-500, compared to a 29.1\% gain from ground-truth rewards (\S\ref{sec:qwen_works}). We observe similar trends on more challenging math benchmarks, including AMC and AIME.
In contrast, for other model families, including \qwenbasefamily~\citep{Yang2024Qwen25TR}, OLMo2~\citep{OLMo20242O2}, and Llama3~\citep{Dubey2024TheL3}, training with spurious rewards yields minimal improvement or even performance degradation (\S\ref{sec:others_dont}).
This divergence in outcomes suggests that pretraining priors, different in each model, strongly shape RL training dynamics.

\begin{figure*}[t!]
    \centering
    \includegraphics[width=0.9\linewidth]{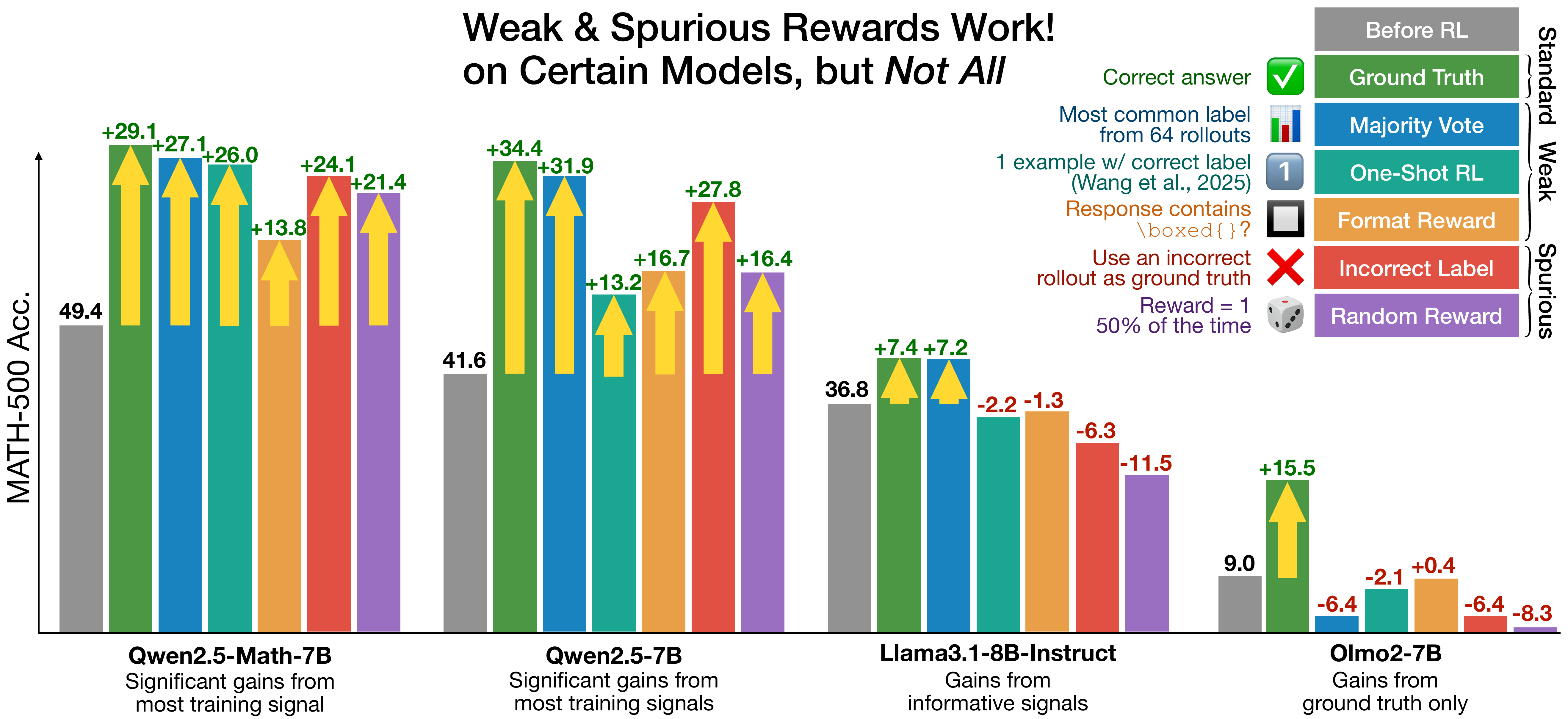}\vspace{-1mm}
    \caption{MATH-500 accuracy after 300 steps of RLVR on various training signals.
    We show that even ``spurious rewards'' (e.g., rewarding \textit{incorrect} labels or with completely random rewards) can yield strong MATH-500 gains on Qwen models. Notably, these reward signals do not work for other models like \llama and \olmo, which have different reasoning priors. 
    }\vspace{-2mm}
    \label{fig:overview}
\end{figure*}

To further investigate the underlying mechanism contributing to such spurious improvements, we unpack the GRPO training objective \citep{shao2024deepseekmath} and find that the clipping function systematically amplify tokens with high prior probabilities in the base model (\S\ref{sec:random_reward}). 
To help explain this discrepancy, we also analyze what reasoning patterns that RLVR is learning to favor in these cases (\S\ref{sec:analysis}). 
In particular, we find a majority of \qwenmath answers on MATH-500 contain reasoning chains expressed in Python---a behavior we call {\em code reasoning}---despite having no access to code execution. Code reasoning is highly predictive of overall performance; answers with it have an accuracy of 60.9\%,  much higher than without (28.0\% accuracy). 
Code reasoning also correlates with MATH-500 accuracy over the course of RLVR training. Both metrics increase consistently during training with any spurious reward, leading to roughly $90\%$ or higher code frequency after training. 
Besides code reasoning, we find that lexical repetitions show similar but less substantial correlation with MATH performance in \qwenmath models.
Based on this observation, we further hypothesize that intervening with other methods that increase any beneficial behavior should similarly increase test performance. Our experiments validate this hypothesis: we design prompt-based and RL-based elicitation methods to increase code reasoning and lexical repetition; all such methods significantly increase \qwenmath's performance.
We show that these results are robust to prompt variations, despite the initial model performance being sensitive to prompts in sometimes unexpected ways (Appendix~\ref{app:prompt_effects}).

Our findings not only open new questions but also have practical implications. 
We should generally be more aware that reasoning patterns instilled during pretraining heavily impact the behavior of downstream RLVR training.
Qwen models, with open weights and high performance on reasoning tasks, have become the de facto choice for RLVR research in the open-source community---a range of recent research on RLVR drew conclusions on \qwenmath-centric experiments~\citep{zuo2025ttrl, zhao2025learning, wang2024openr, xie2025logic, hu2025open, zhang2025right,shafayat2025largereasoningmodelsselftrain,prabhudesai2025rent,gao2025oneshotentropyminimization,wang20258020rulehighentropyminority}.
However, we show that it is easy to get significant performance improvements on Qwen models even with completely spurious reward signals. Thus, we suggest that future RLVR research should be confirmed on other models and using spurious rewards as dummy baselines. 
We emphasize that spurious rewards are proposed purely for analytical purposes and should not be interpreted as a recommended approach for developing true model capabilities.

\begin{figure*}[t!]
    \centering
    \cblock{94}{149}{78} Ground Truth
    \cblock{70}{129}{188} Majority Vote
    \cblock{221}{162}{88} Format
    \cblock{211}{98}{83} Incorrect
    \cblock{147}{112}{188} Random
    \begin{subfigure}[t]{0.49\textwidth}
        \centering
        \includegraphics[width=0.49\linewidth]{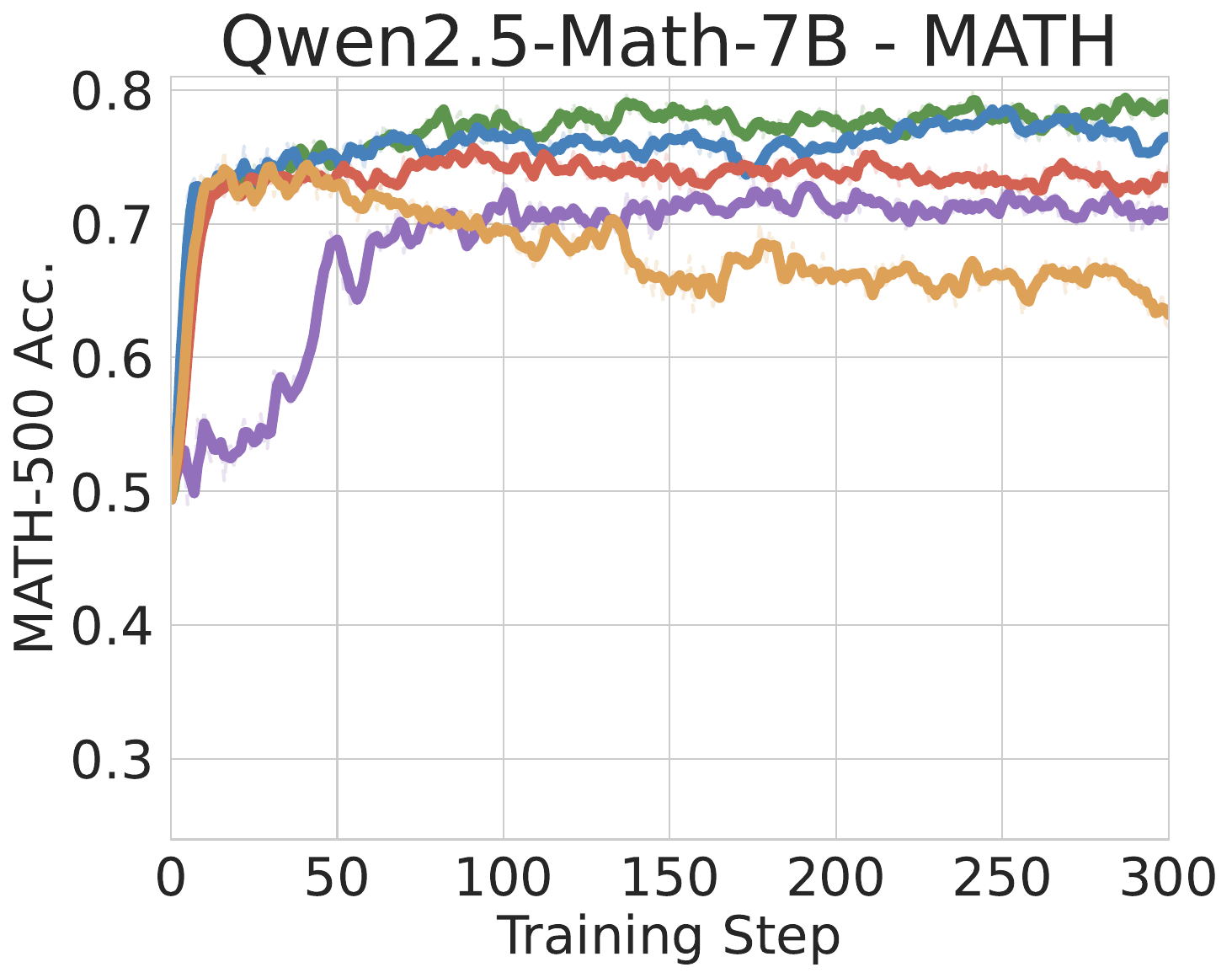}
        \includegraphics[width=0.49\linewidth]{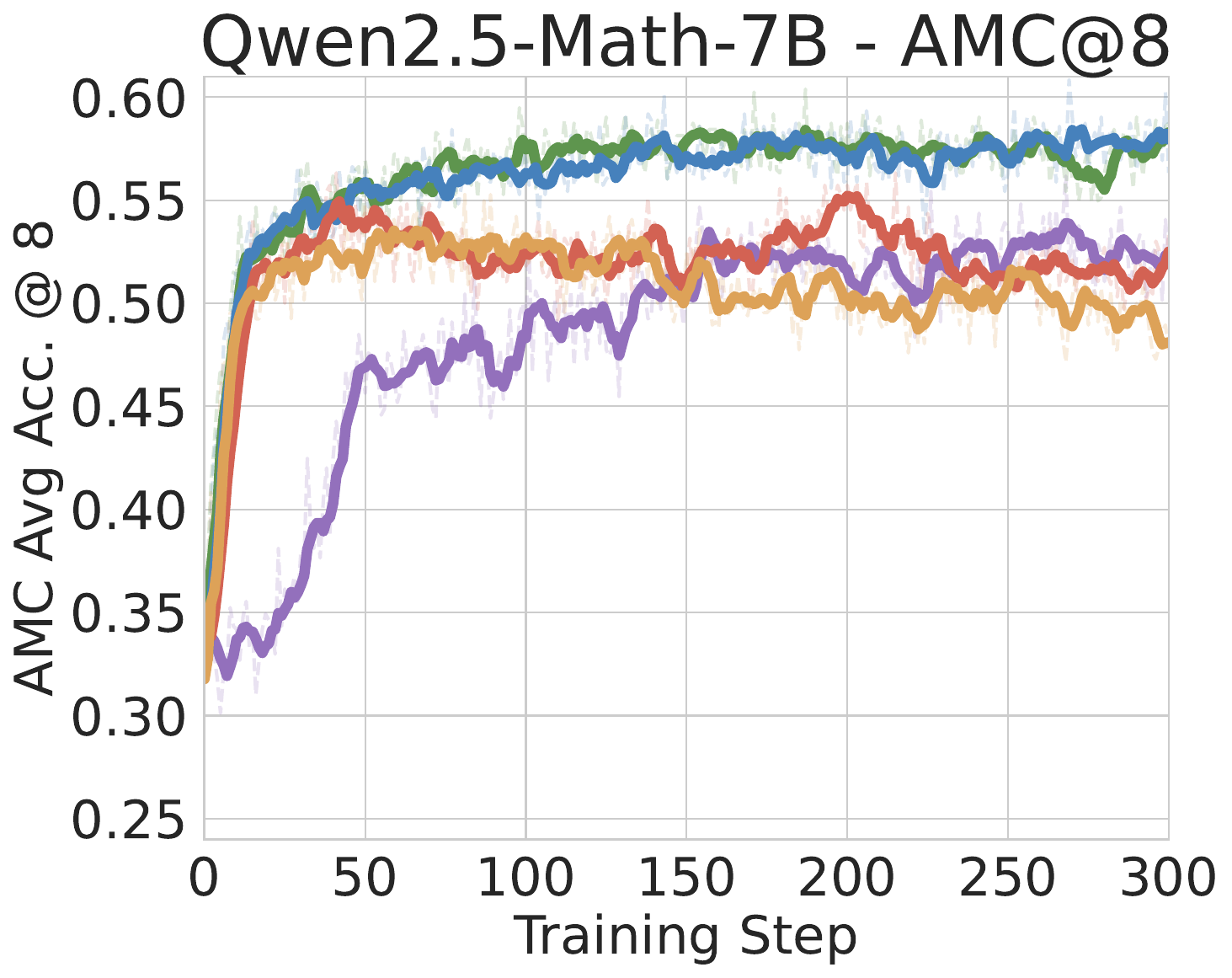}
        \caption{\qwenmath}
        \label{fig:qwen_results}
    \end{subfigure}%
    ~
    \begin{subfigure}[t]{0.49\textwidth}
        \centering
        \includegraphics[width=0.49\linewidth]{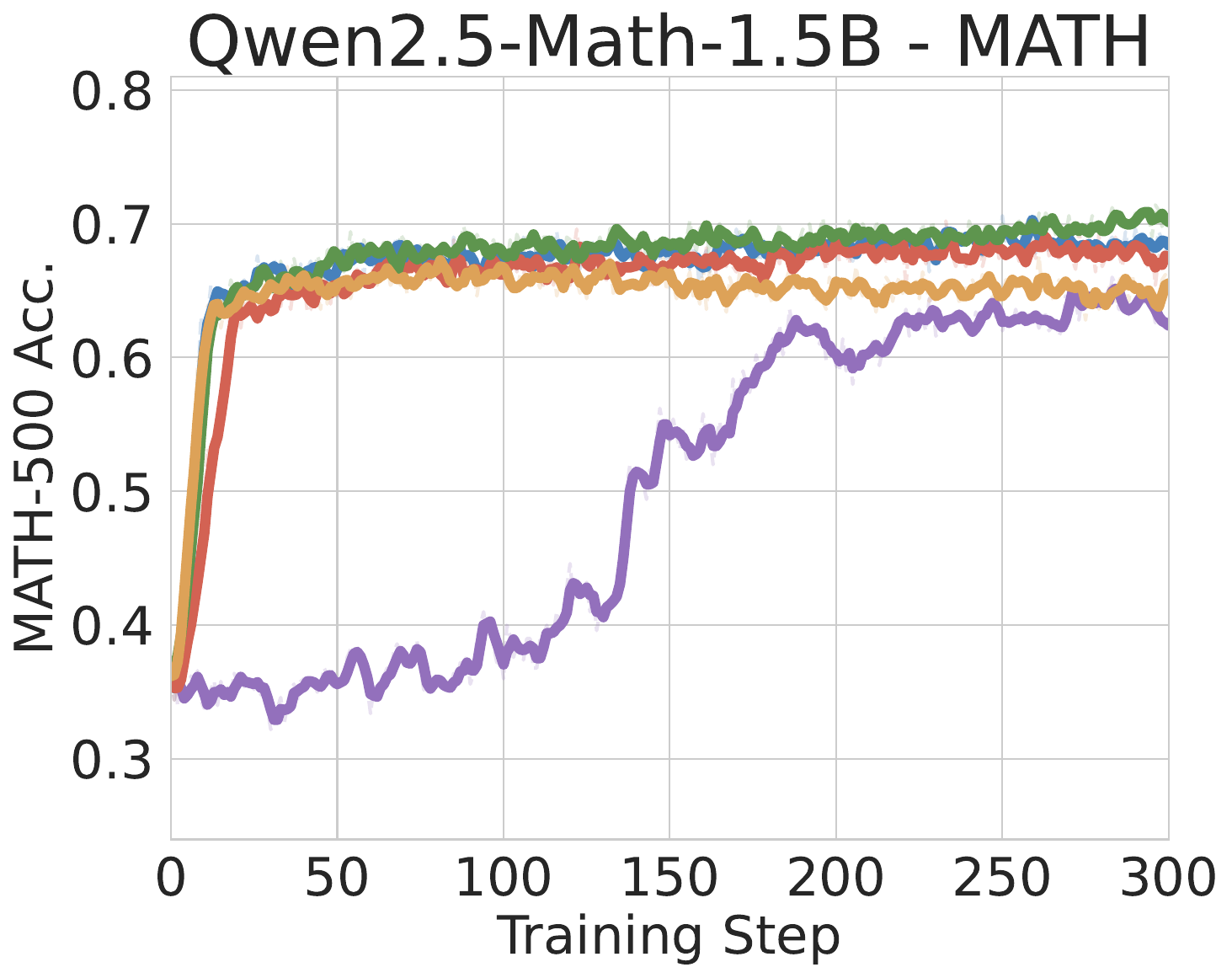}
        \includegraphics[width=0.49\linewidth]{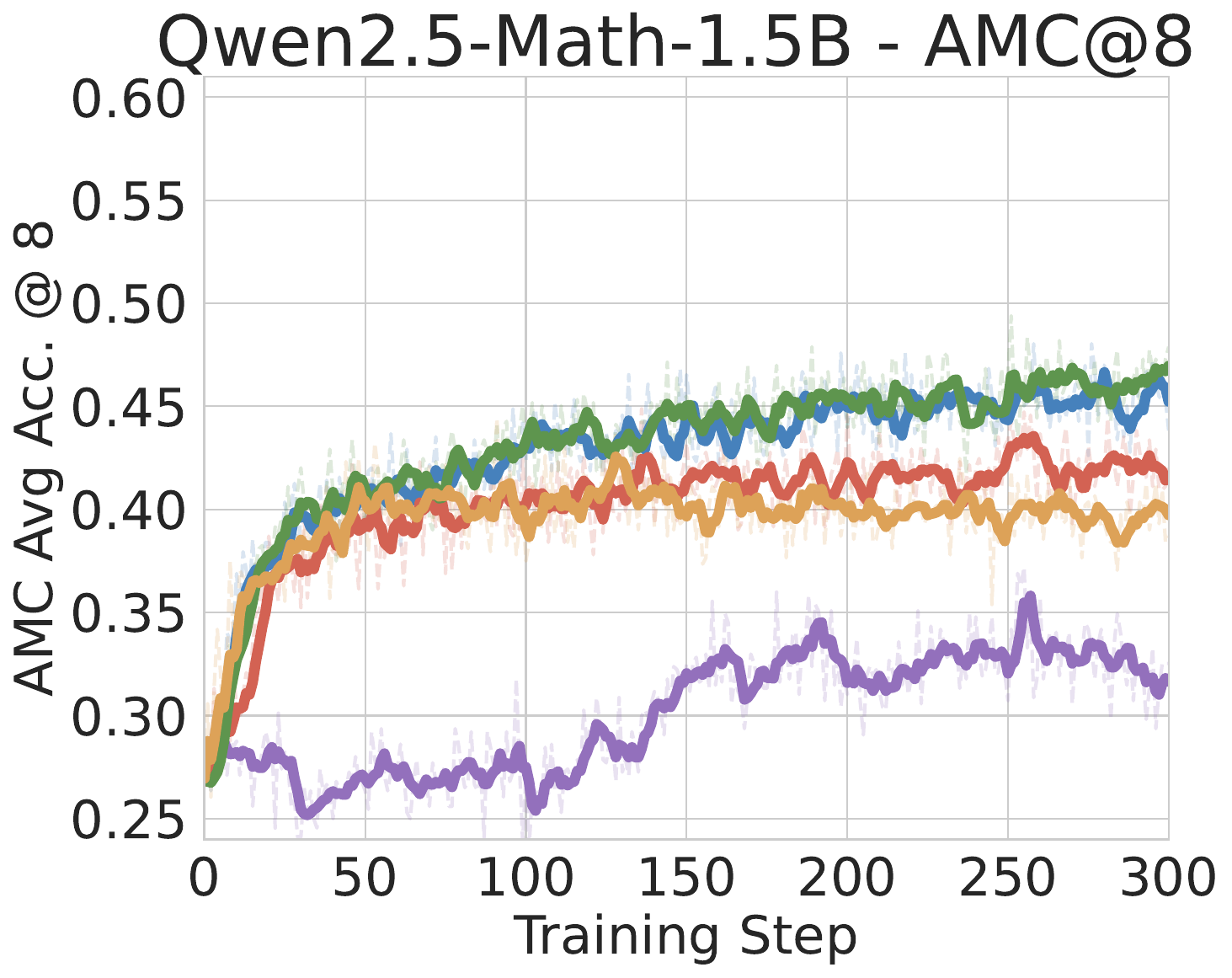}
        \caption{\qwenmathsmall}
        \label{fig:qwen_1b_results}
    \end{subfigure}
    \caption{
    Model performance on MATH and AMC with varied training rewards smoothed over window size of 10 (dotted lines are unsmoothed values). We report pass@1 for MATH and average@8 for AMC. Both \qwenmath and \qwenmathsmall significantly improve after RLVR on a range of reward signals from meaningful to spurious. 
    We note that the random reward converges slower than the other spurious rewards, but the fact that it leads to significant gains at all is surprising.
    }
    \label{fig:spurious_rewards_qwen}
\end{figure*}

\section{Spurious Rewards Yield Significant RLVR Gains}\label{sec:qwen_works}

We design a progression of reward functions to replace the standard ground-truth reward: \textit{weak rewards} (majority vote reward and format reward) and \textit{spurious rewards} (random reward and incorrect reward).
Remarkably, we find that all weak and spurious rewards suffice for RLVR to significantly improve the math performance of Qwen2.5-Math, a popular starting point for RLVR training.

\subsection{Experimental Setup}\label{sec: exp setup}
Following recent RLVR work~\citep{wang2025reinforcement, zuo2025ttrl, zeng2025simplerl}, we use GRPO~\citep{deepseekteam2024deepseek} to finetune \qwenmathfamily models~\citep{yang2024qwen2}.
The standard RLVR approach uses a dataset of questions paired with ground truth labels. During training, model rollouts are given a binary (0-1) reward based on whether the generated answer is verifiably correct. 
We replace this ground truth-based reward with a variety of increasingly spurious binary 0-1 reward functions that do not require access to ground truth labels. 
We design these alternative rewards to investigate the limits of how little supervision is needed for effective RLVR training. We train on DeepScaleR data~\citep{deepscaler2025} with our rewards; all other experimental details are kept constant.

In the main paper, we evaluate performance as pass@1 and average@8 accuracy on two standard math reasoning benchmarks: MATH-500~\citep{hendrycks2021measuring} and AMC~\citep{li2024numinamath}, respectively. See Appendix~\ref{app:aime_results} for additional results on AIME 2024 and 2025. 
Following the default evaluation setup in the popular RL framework OpenRLHF~\citep{hu2024openrlhf}, we use the default chat template for \qwenmathfamily and all instruct models and chat template provided by Olmo~\citep{OLMo20242O2} for other base models in our main experiments.
Additional analysis on the effect of different prompts can be found in Appendix~\ref{app:prompt_effects}, where we show \qwenmath is very sensitive to prompts---even a task-irrelevant prompt (which we name \textit{spurious prompt}) can sometimes result in high initial performance.
See Appendix~\ref{app:setup} for full details of our training and evaluation setup.

\subsection{Standard to Weak to Spurious Rewards}

We consider the following rewards: 

\begin{enumerate}[itemsep=0pt,topsep=0pt,leftmargin=12pt]
\item \textbf{Ground Truth Rewards:} To establish a baseline, we consider the standard RLVR approach~\citep{lambert2024tulu3} of using ground truth labels to reward responses with verifiably correct answers. This setting serves as an upper bound for reward supervision quality.

\item \textbf{Majority Vote Rewards:}
Instead of using ground truth labels for computing rewards, we use the model prior to RLVR training to pseudo-label the training set by selecting the majority answer from 64 sampled responses per prompt. These (potentially wrong) labels are then used to reward responses during standard online RLVR training.

\item \textbf{Format Rewards:} We further weaken the reward signal to disregard the responses' mathematical correctness altogether. We instead heuristically reward all responses containing at least one non-empty \verb|\boxed{}| expression, regardless of the correctness of the enclosed answer. Including \verb|\boxed{}| is specified in \qwenmathfamily's system prompt; this reward incentivizes some degree of prompt following.

\item \textbf{Random Rewards:} 
We study whether providing \textit{no} guidance in the rewarding process is sufficient to provide meaningful math performance gains. 
To do so, we assign rewards randomly. Given a fixed probability hyperparameter $\gamma$, all responses receive a reward of $1$ with chance indicated by the parameter, and receive $0$ otherwise. In our main experiments, we present $\gamma = 0.5$; in Appendix~\ref{app:random_reward}, we show that using $\gamma \in \{0.001, 0.3, 0.7\}$ obtains similar improvements with varying convergence speed, and verify that $\gamma = 0$ results in no change as expected analytically (with $\gamma = 0$, loss is constant, and all gradients are zero).

\item \textbf{(Majority-voted) Incorrect Rewards:}
We furthermore deliberately provide incorrect supervision and reward only incorrect answers. 
We first label all training data using majority voting and select the subset with incorrect labels for training,
obtaining incorrect labels that are still probable outputs of the models. 
During training, we reward responses whose answers verifiably match these incorrect labels.
This design disentangles whether majority vote rewards work because labels are more likely correct or because they represent high-probability model outputs. 
We reward specific incorrect answers rather than any incorrect answer to maintain comparable reward sparsity with our other conditions.
\end{enumerate}

\textbf{Note:} We emphasize that spurious rewards---particularly random and incorrect rewards---are proposed purely for analytical purposes and should not be interpreted as a recommended approach for developing true model capabilities.

Figure \ref{fig:spurious_rewards_qwen} presents the performance of Qwen2.5-Math models after RLVR training with each reward function. Overall, all reward functions, even pathologically designed ones, lead to significant improvements in math performance within the first 50 steps across all benchmarks compared to the untuned baseline. 
One exception is that \qwenmathsmall sees gains with random rewards much slower (after 100 steps) and less so on AMC (only 4.9\%).
Remarkably, performance gains from spurious rewards are often within a few points of the gain from RLVR with ground truth labels. 
For example, training with incorrect label reward yields 24.1\% gains over \qwenmath on MATH-500, compared to a 29.1\% gain from RLVR with ground truth answers. Even random rewards---which by design provide pure noise in the rewarding process
---still produce a 21.4\% performance boost. We observe similar trends in AMC, where training on format, incorrect, or random rewards yields a gain of $13.8\%$, $24.1\%$, or $21.4\%$, respectively, approaching the $\sim27\%
$--$29\%$ improvement gained from training on majority voted and ground truth labels.

In Appendix~\ref{app:aime_results}, we supplement results on AIME24 and AIME25.
On AIME2024, format reward (+10.3\%) approaches ground truth rewards (+15.3\%), and spurious rewards (incorrect \& random) still lead to high performance gains of 10.2\% and 10.2\% respectively on \qwenmath.
Ground truth labels show a clear advantage compared to other rewards on AIME2025, which contains questions written after the knowledge cutoff of all models we consider. Nonetheless, other rewards still lead to a -0.4\% to 4.5\% gain in performance.

Our findings with these simple rewards provide additional evidence for a nascent hypothesis in the literature: that RLVR, at least at the compute scales of open-source post-training pipelines~\citep{lambert2024tulu3}, does not teach models new reasoning capabilities, but instead triggers latent ones already present in the base model~\citep{wang2025reinforcement, liu2025understanding, gandhi2025cognitive, yue2025does, shah2025rethinking,choshen2020weaknessesreinforcementlearningneural}. Whereas prior work has hinted at this effect using limited-quantity ground-truth labels~\citep{wang2024openr} or noisy labels~\citep{zuo2025ttrl}, our results push this idea to its limit: we show that even outright incorrect rewards or information-free rewards (i.e., random) can elicit performance gains in \qwenmathfamily models. In the remainder of our paper, we show that this elicitation effect is model-dependent (\S\ref{sec:others_dont}), and trace the specific properties of \qwenmathfamily models that could enable spurious rewards to induce this elicitation (\S\ref{sec:analysis}).

\begin{figure*}[t!]
    \centering
    \cblock{94}{149}{78} Ground Truth
    \cblock{70}{129}{188} Majority Vote
    \cblock{221}{162}{88} Format
    \cblock{211}{98}{83} Incorrect
    \cblock{147}{112}{188} Random
    \begin{subfigure}[t]{0.245\textwidth}
        \centering
        \includegraphics[width=\linewidth]{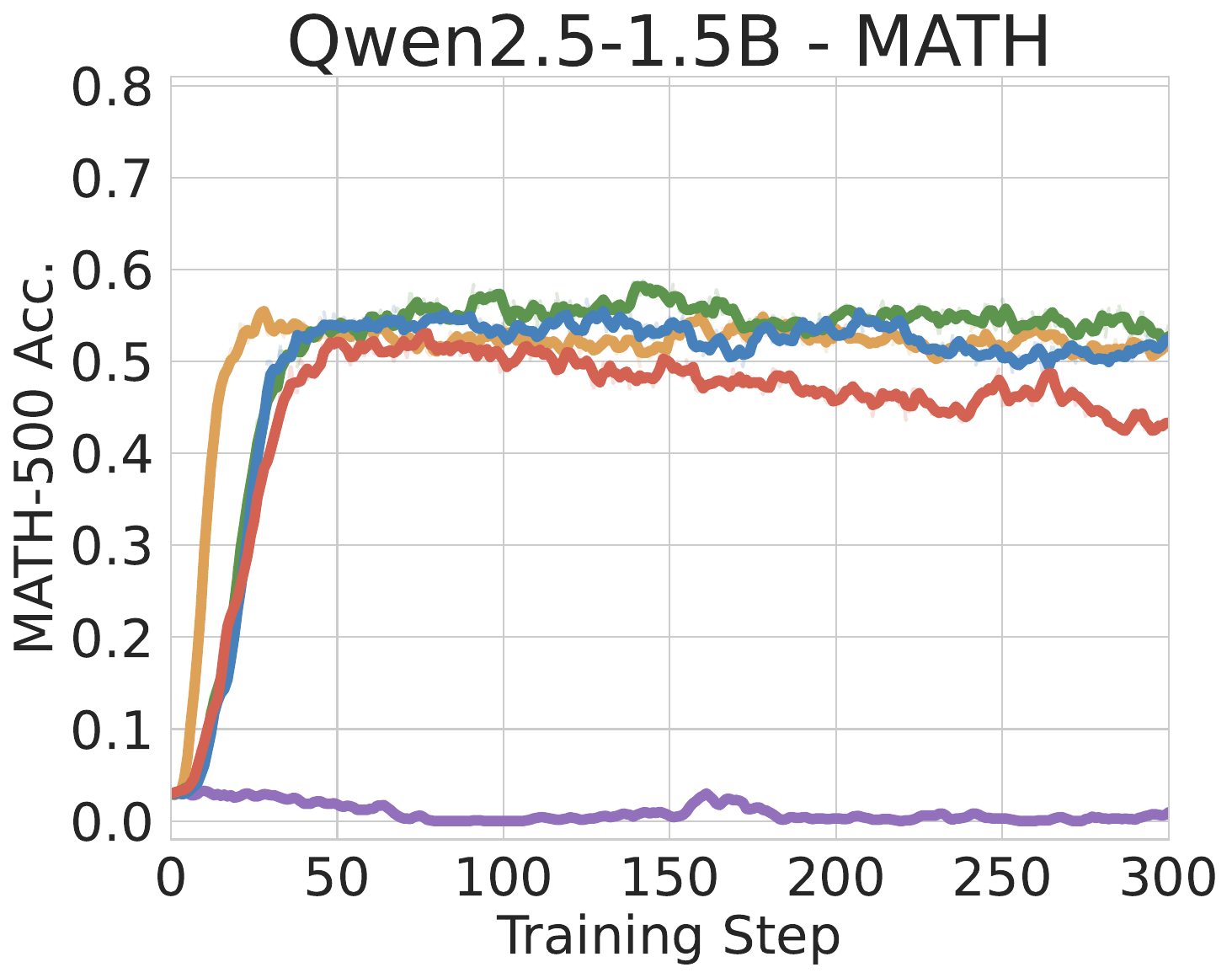}\\
        \caption{Qwen2.5-1.5B}
        \label{fig:qwen1.5b_results}
    \end{subfigure}%
    ~
    \begin{subfigure}[t]{0.245\textwidth}
        \centering
        \includegraphics[width=\linewidth]{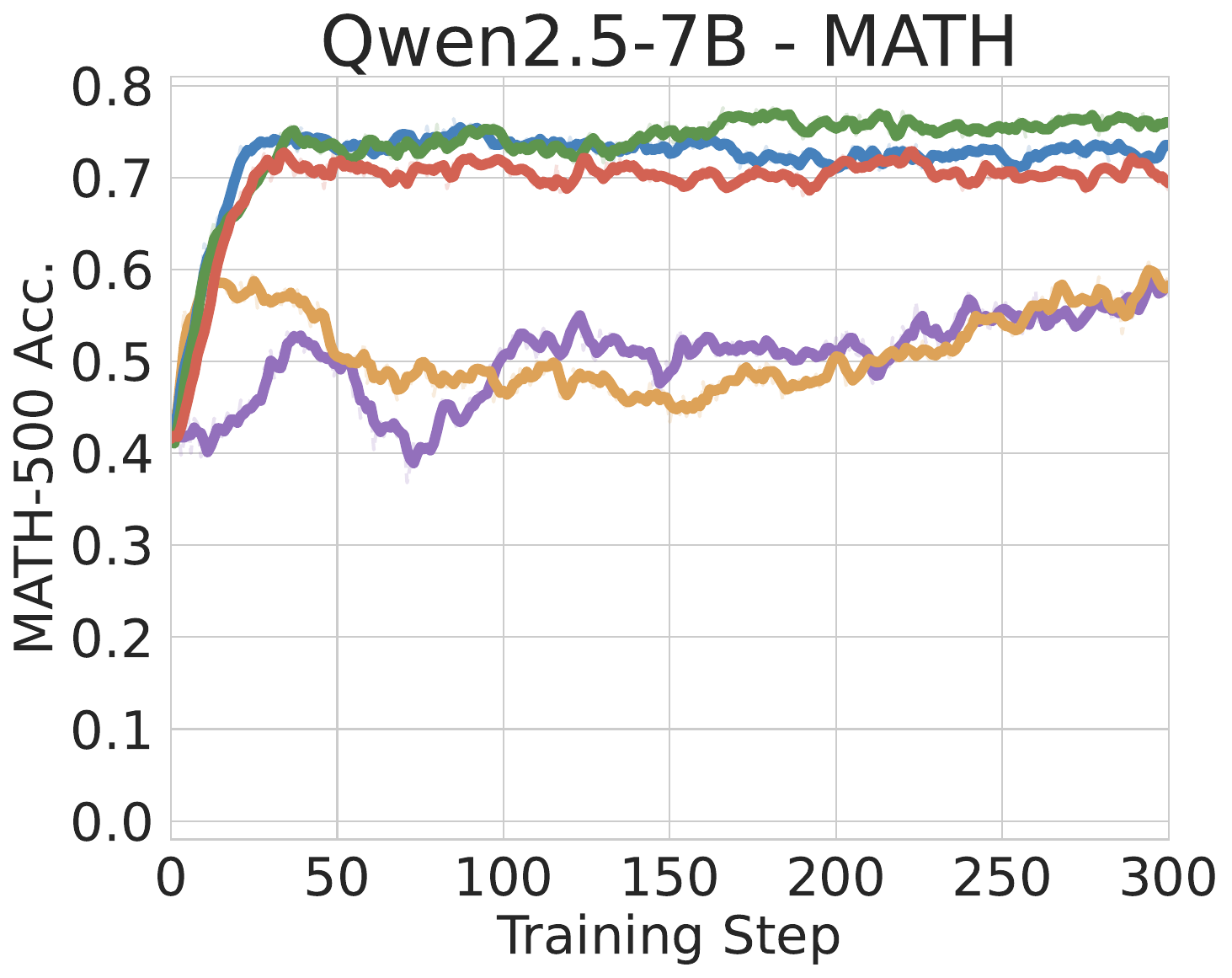}
        \caption{Qwen2.5-7B}
        \label{fig:qwen_results}
    \end{subfigure}%
    ~
    \begin{subfigure}[t]{0.245\textwidth}
        \centering
        \includegraphics[width=\linewidth]{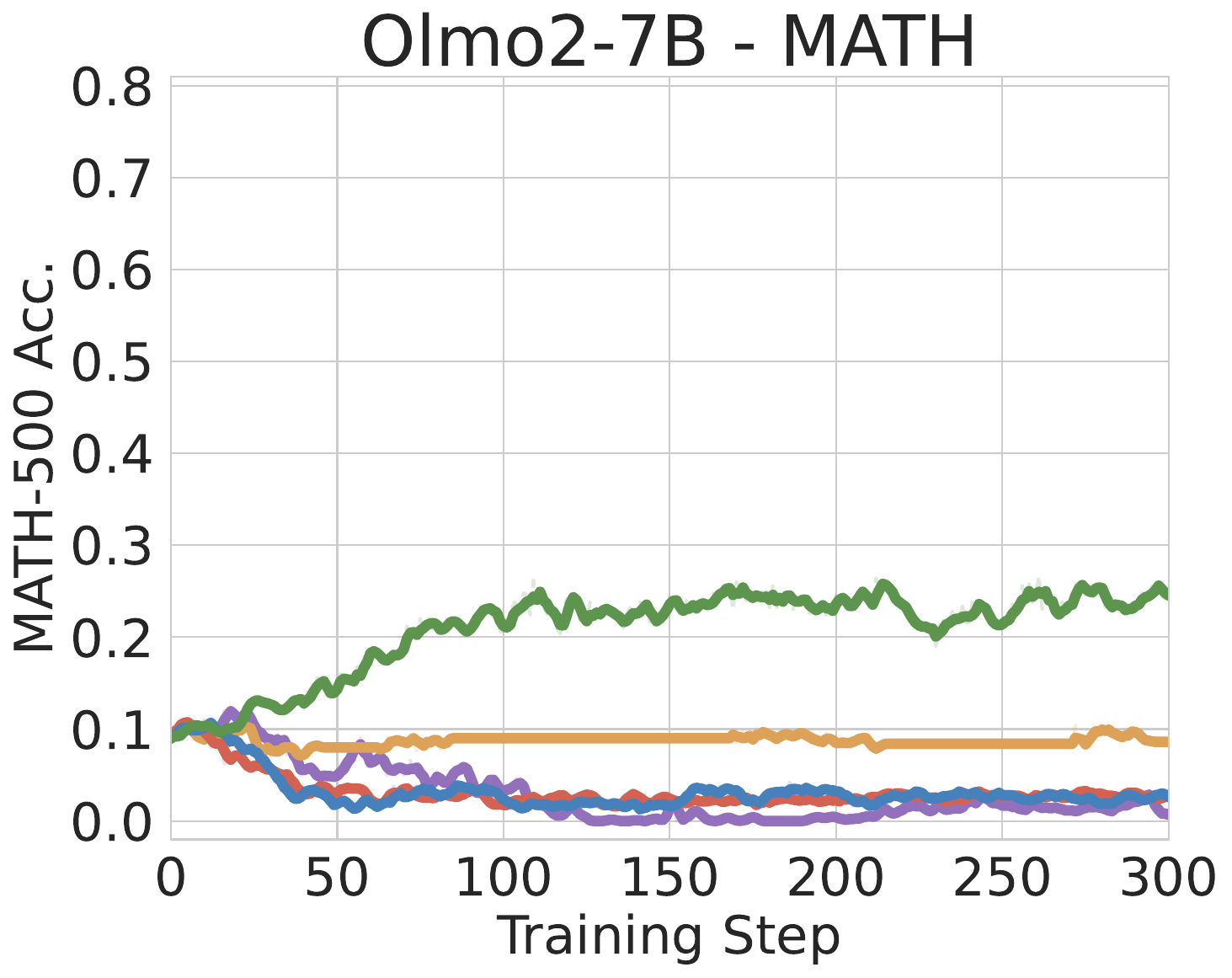}
    \caption{Olmo-2-1124-7B}
    \label{fig:olmo_results}
    \end{subfigure}%
    ~ 
    \begin{subfigure}[t]{0.245\textwidth}
        \centering
        \includegraphics[width=\linewidth]{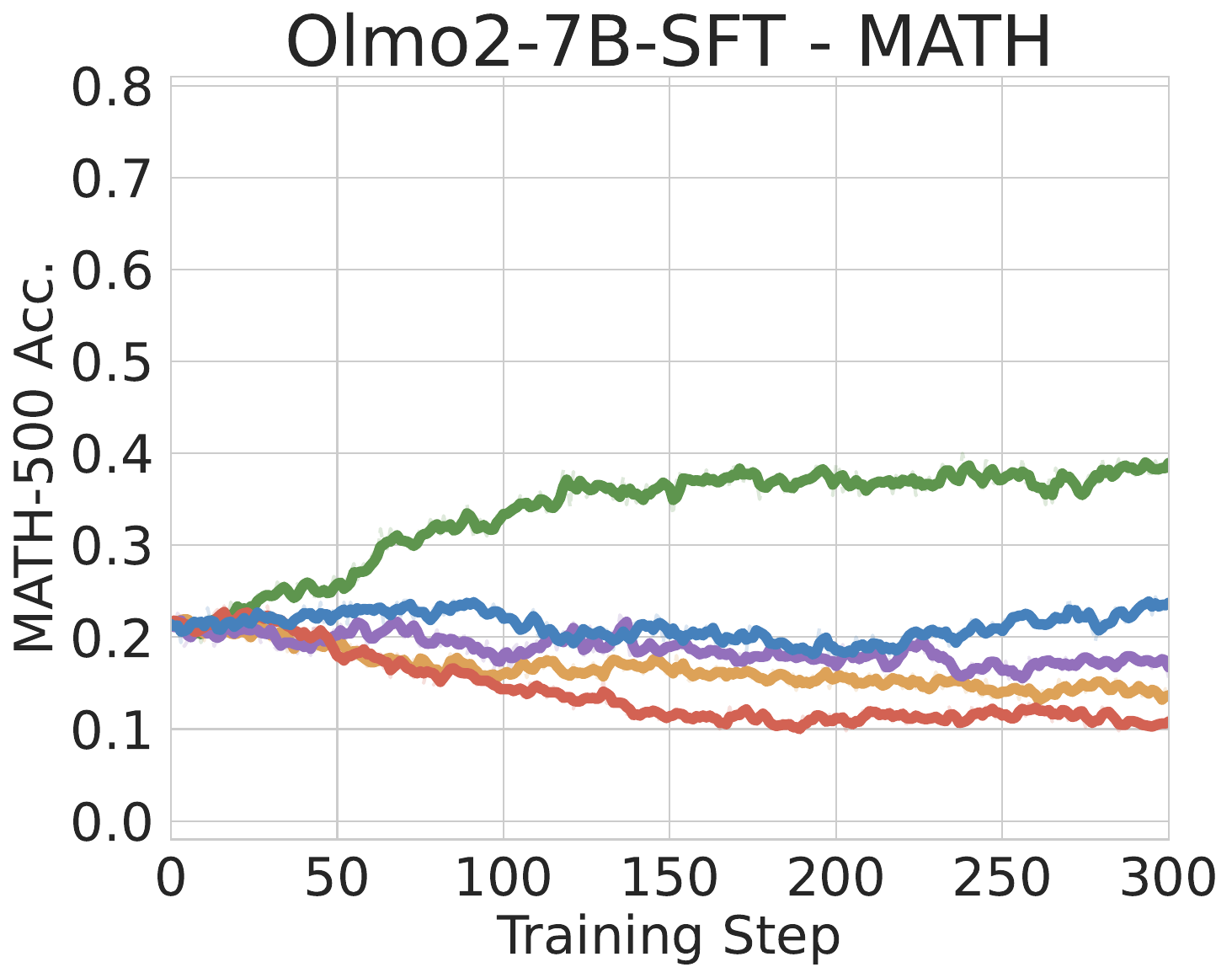}
    \caption{Olmo-2-1124-7B-SFT}
    \label{fig:olmo_sft_results}
    \end{subfigure}

    \begin{subfigure}[t]{0.245\textwidth}
        \centering
        \includegraphics[width=\linewidth]{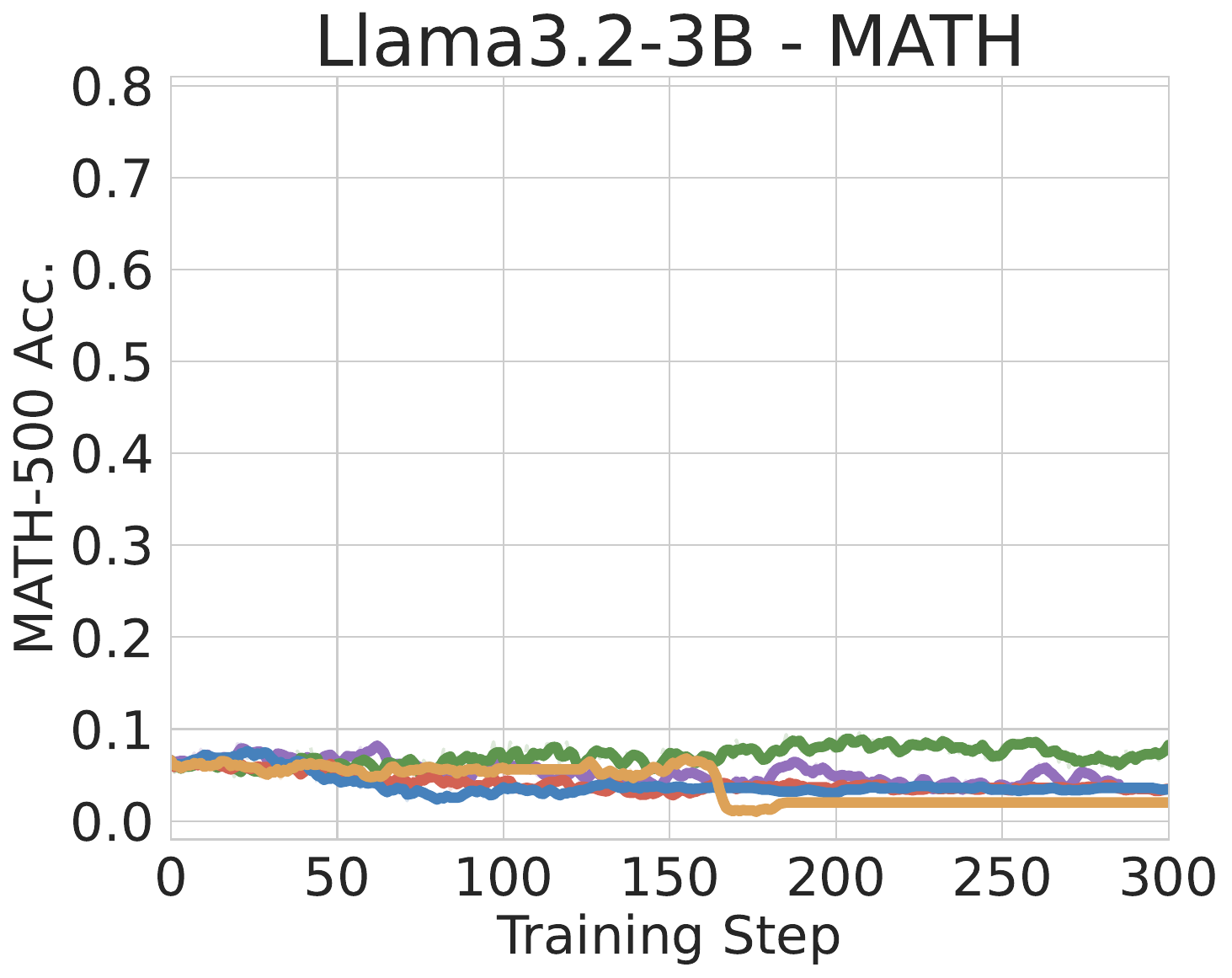}
    \caption{Llama3.2-3B}
    \label{fig:llama3_base_results}
    \end{subfigure}%
    ~
    \begin{subfigure}[t]{0.245\textwidth}
        \centering
        \includegraphics[width=\linewidth]{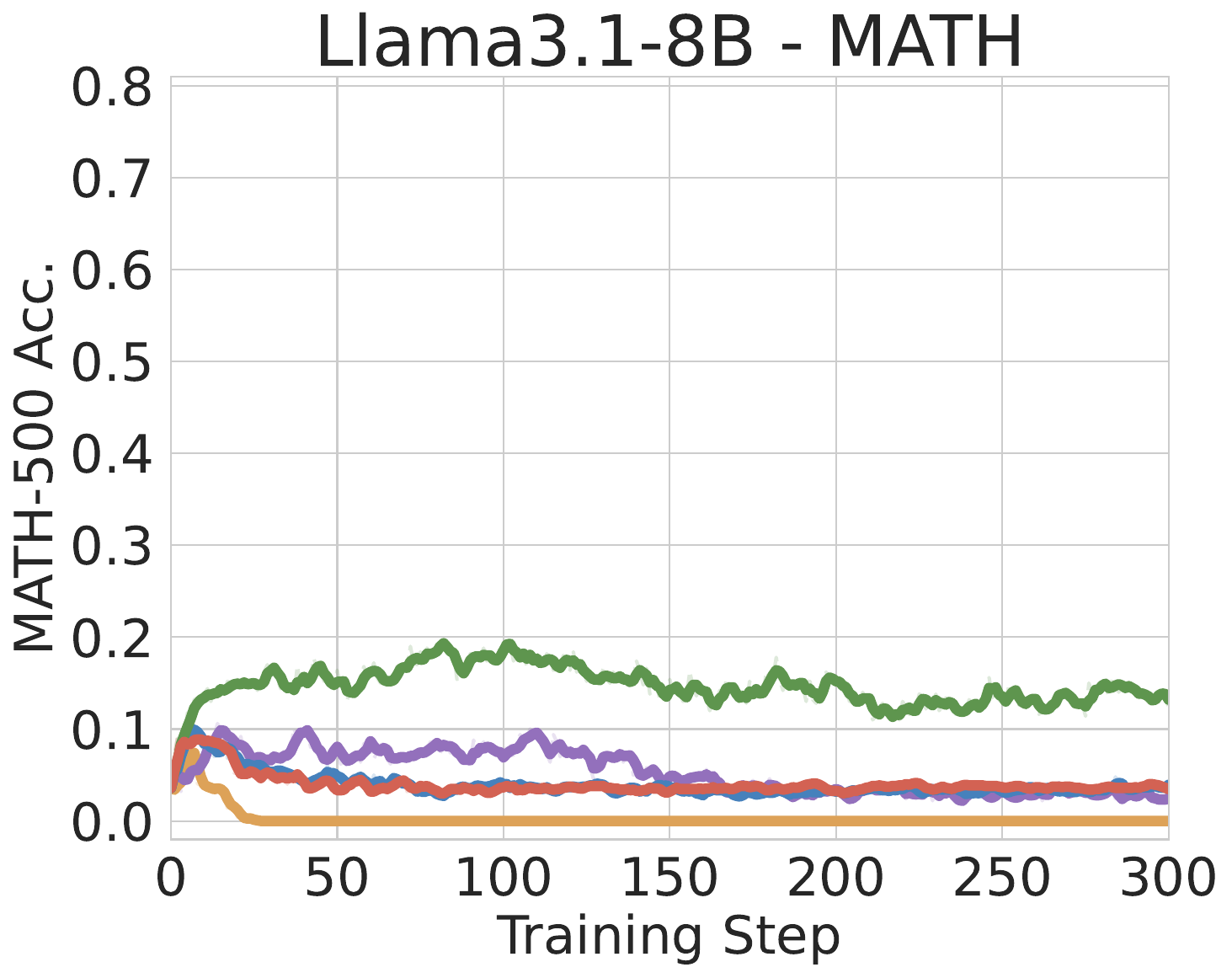}
    \caption{Llama3.1-8B}
    \label{fig:llama3_base_results}
    \end{subfigure}%
    ~
    \begin{subfigure}[t]{0.245\textwidth}
        \centering
        \includegraphics[width=\linewidth]{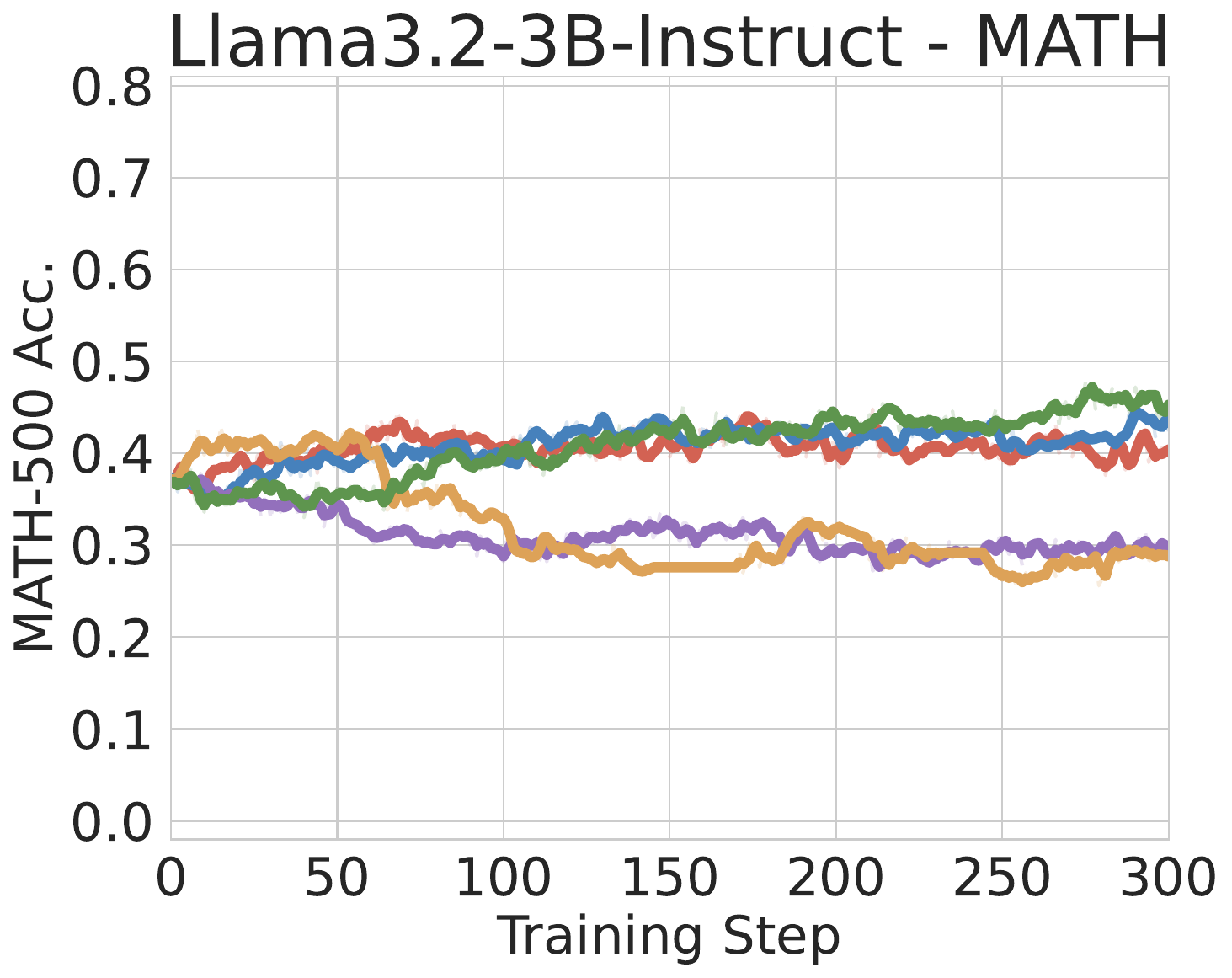}
    \caption{Llama3.2-3B-Instruct}
    \label{fig:llama3_results}
    \end{subfigure}%
    ~
    \begin{subfigure}[t]{0.245\textwidth}
        \centering
        \includegraphics[width=\linewidth]{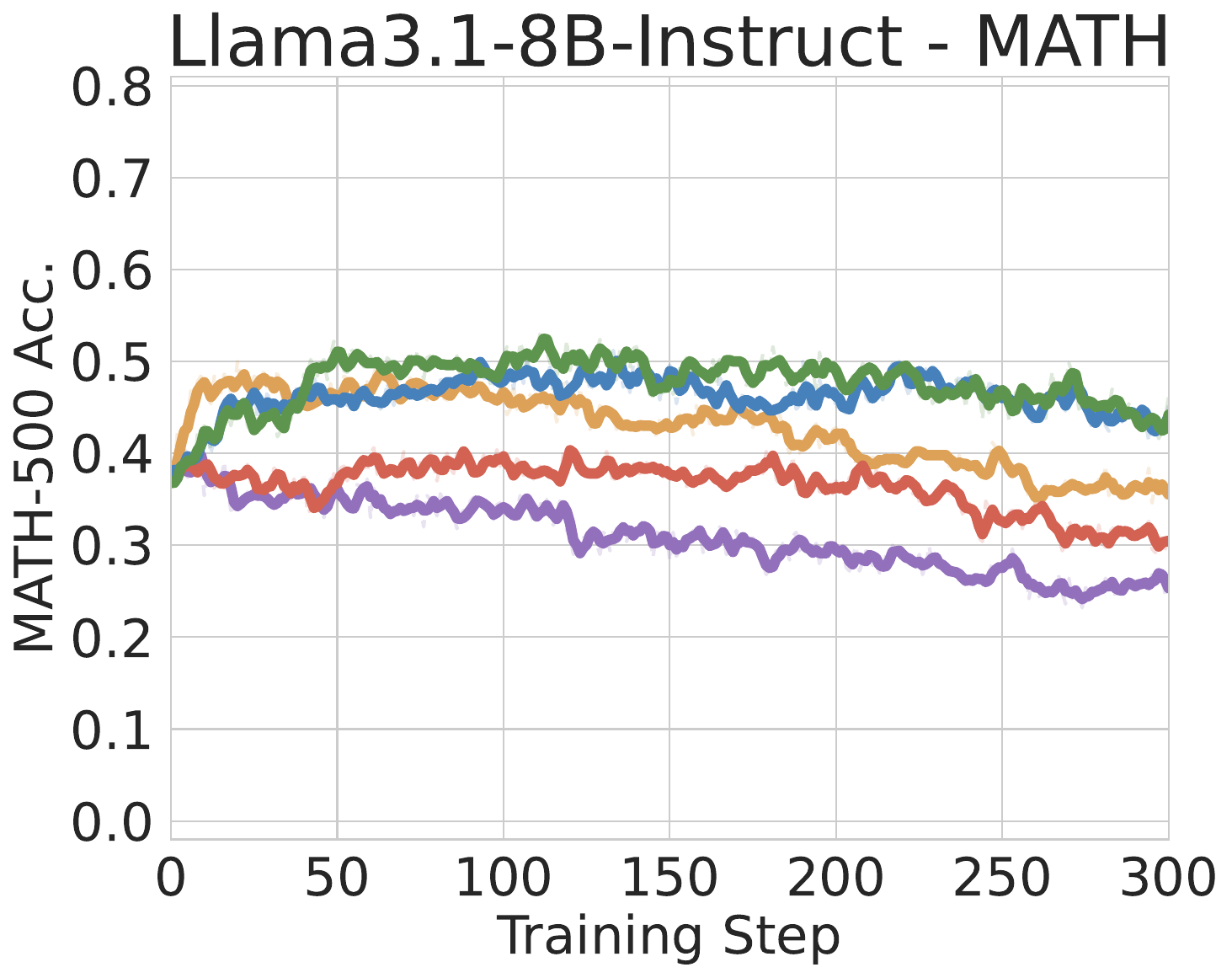}
    \caption{Llama3.1-8B-Instruct}
    \label{fig:llama8_results}
    \end{subfigure}
    
    \caption{Varying rewards across additional model classes on MATH-500. Spurious rewards remain effective on general-purpose \qwenbasefamily models but generally fail to yield any gains on other model families. The performance improvements on non-\qwenbasefamily models are substantially smaller compared to those observed in the \qwenbasefamily family. Similar trends are observed on AMC in Appendix~\ref{app:amc_results}.
    }
    \label{fig:reward_other_family}
\end{figure*}

\section{(Lack of) Generalization to Other Models}
\label{sec:others_dont}

Inspired by the unexpected effectiveness of spurious reward signals in improving the performance of \qwenmathfamily models, we study whether these rewards generalize to training other models.
We extend beyond the math-specialized \qwenmathfamily models to include general-purpose variants (\qwen, \qwensmall~\citep{Yang2024Qwen25TR}), and two additional model families: (a) the widely-used Llama3.1-8B(-Instruct) and Llama3.2-3B(-Instruct)~\citep{Dubey2024TheL3}, and (b) \olmo and \olmosft~\citep{OLMo20242O2}. \olmosft is instruction-tuned from \olmo; we include both to better understand the impact of SFT training on RLVR training. Moreover, we are optimistic that OLMo's open training data will enable future works to better study the origins of any reasoning behaviors (or lack thereof) we later observe. We train these 8 additional models with the same setup and rewards as in Section~\ref{sec:qwen_works}.
We report performance on MATH-500 in the main text of the paper and AMC in Appendix~\ref{app:amc_results} where the same trends are observed.
The AIME 2024 and 2025 benchmarks show similar but noisy trends (Appendix~\ref{app:aime_results}).

\noindent\textbf{Spurious rewards can benefit \qwenbasefamily models, but rarely help non-Qwen models.}
Figure~\ref{fig:reward_other_family} shows two trends. First, models within a family behave similarly: across \qwenbasefamily, all non-random rewards (even spurious incorrect) improve MATH-500, whereas OLMo stays flat under spurious rewards and gains mainly with ground-truth rewards.
We conjecture this family-level consistency reflects shared pretraining distributions that induce similar pre-RL behaviors.
Second, smaller models benefit less from spurious (e.g., random) rewards; we conjecture larger models preserve more pretraining priors that these rewards amplify.
Section~\ref{sec:analysis} further analyzes how pre-existing behaviors shape RLVR outcomes.
Importantly, these reward signals do not reliably transfer across families. Although spurious incorrect and other weak rewards consistently improve Qwen, each weak or spurious reward fails to help at least one other model and can be flat or even harmful. Overall, Qwen appears unusually robust to reward strength. Appendix~\ref{app:instruct_models} further shows that models already RL post-trained see minimal gains under nearly all rewards.

\noindent\textbf{Practical warning~\parbox{4mm}{\includegraphics[width=\linewidth]{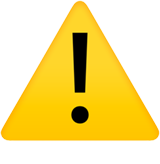}}: Proposed RLVR reward signals should be tested on diverse models!}
Many recent methods on RLVR for reasoning draw their conclusions primarily or exclusively from gains shown on Qwen models (non-exhaustively, \cite{zuo2025ttrl, zhao2025learning, wang2024openr, xie2025logic, hu2025open, zhang2025right}). As a case study, we experimented with recent work on (1) test time training \citep{zuo2025ttrl} and (2) one-shot RL~\citep{wang2025reinforcement} (setup details in Appendix~\ref{app:oneshot_ttrl_setup}). 
We find these methods exhibit a pattern similar to our results above: the proposed training signals yield strong improvements in the \qwenmathfamily or \qwenbasefamily models, matching the performance of the ground truth rewards (Figure~\ref{fig:ttrl_results} in Appendix~\ref{app:ttrl_and_one_shot_rl}). 
However, these same signals often fail to yield performance gains on other model families. Our findings suggest that existing Qwen-centric RLVR research should possibly be further validated on non-Qwen models.

\section{Understanding Training Signals from Random Rewards in GRPO}\label{sec:random_reward}

Prior work has shown that RLVR using majority-voted labels or format rewards can improve model performance in GRPO~\citep{zuo2025ttrl,zhao2025learning,wang2025reinforcement}. As a result, the effectiveness of majority-voted (even when incorrect\footnote{We hypothesize that incorrect rewards provide training signals because many remain close to ground truth and because they implicitly act as format rewards, incentivizing extractable answers.}) rewards and format rewards is less surprising. 
However, the effectiveness of random rewards remains puzzling, as they are constructed to contain no information. In this section, we present our hypotheses on how random rewards induce meaningful training signals in GRPO and validate them through ablation studies.



\begin{figure}[t]
    \centering
    \cblock{147}{112}{188} Random w/ Clipping Enabled \\
    \cblock{36}{174}{18} Random w/ Clipping Disabled \\
    \cblock{90}{168}{253} Random w/ Clipping Disabled (Mini-Batch Size $\uparrow$) \\
    \cblock{18}{58}{184} Random w/ Clipping Disabled (Rollout Batch Size $\downarrow$) \\
        \includegraphics[width=0.49\linewidth]{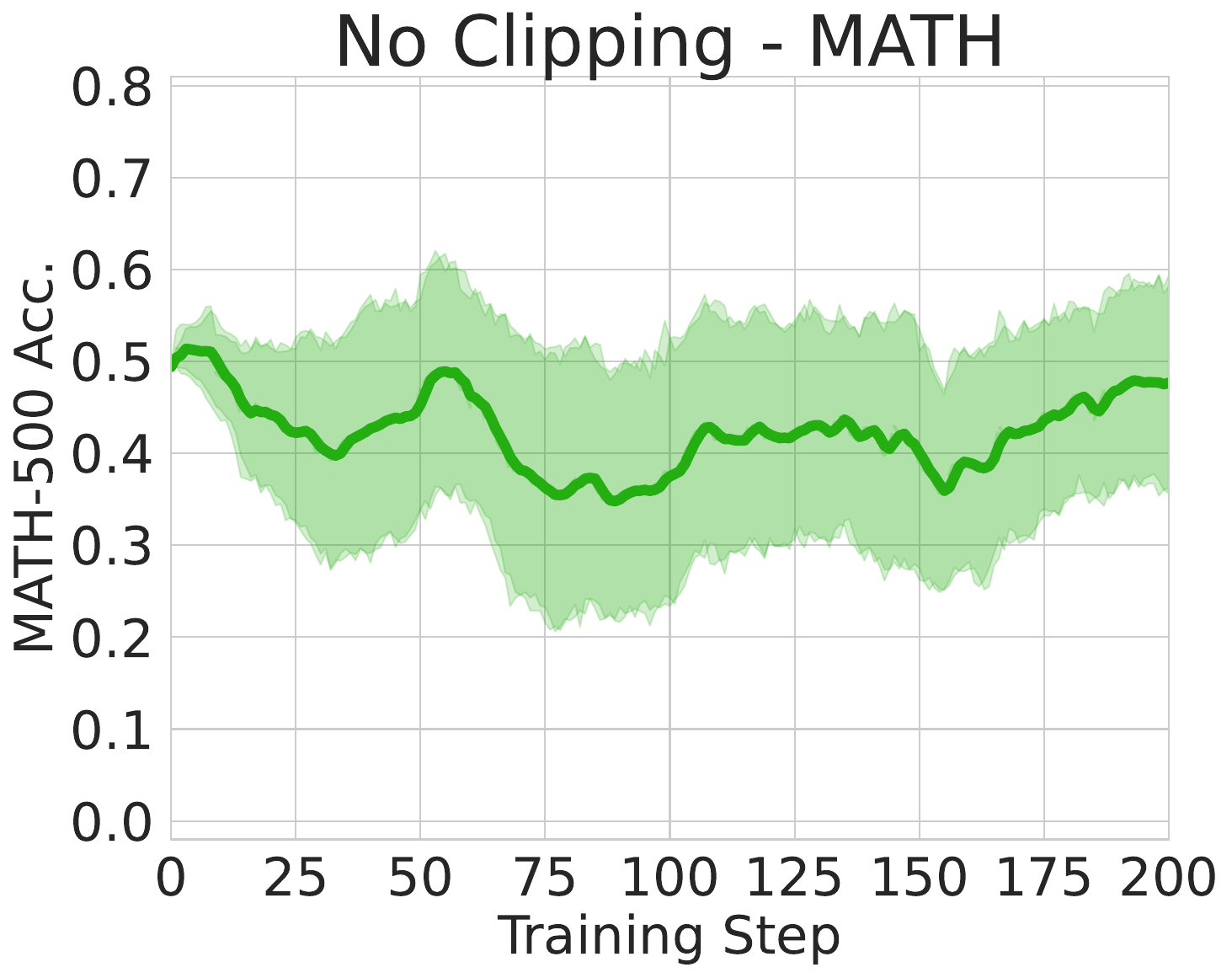}\hfill
        \includegraphics[width=0.49\linewidth]{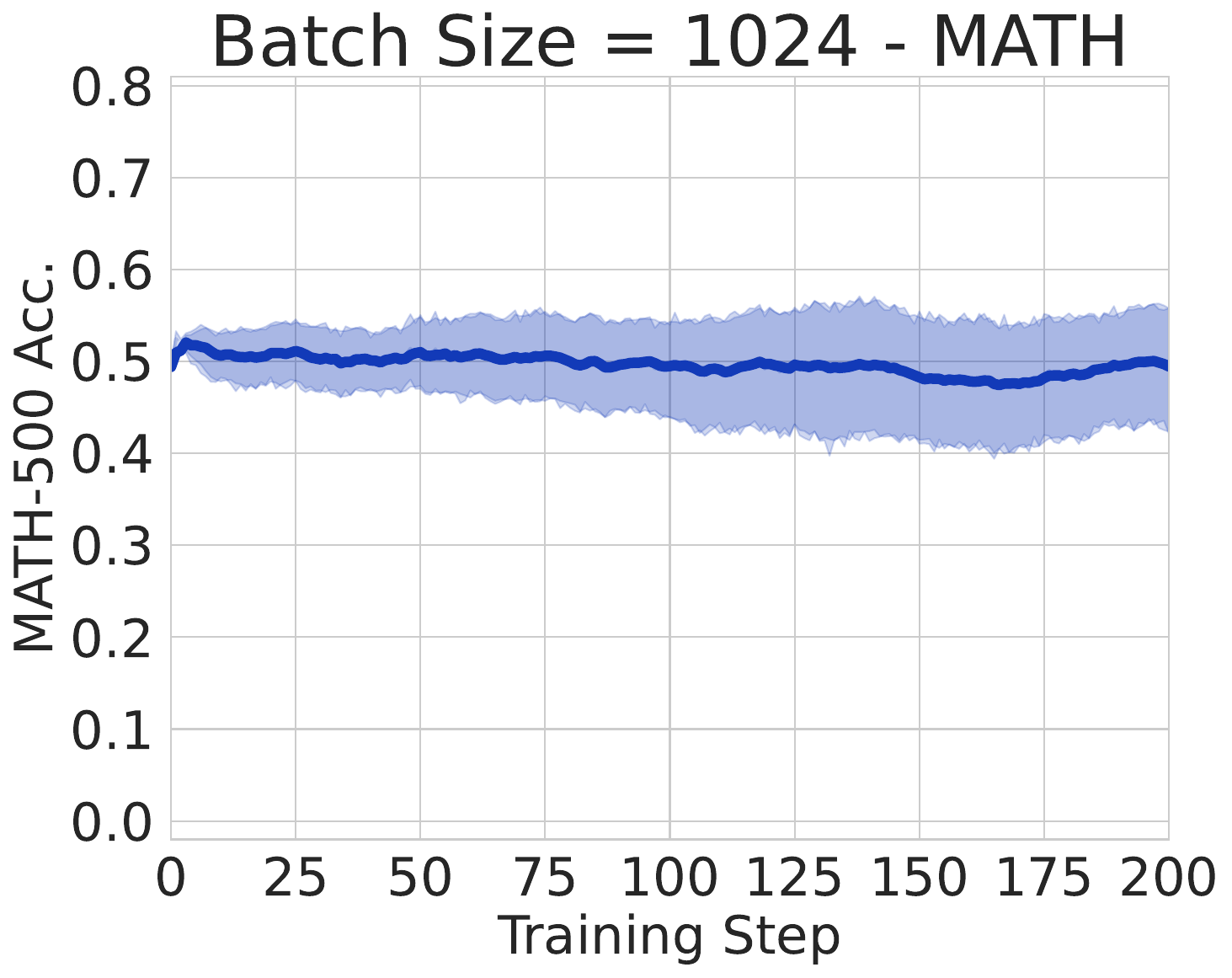}\\
        \includegraphics[width=0.49\linewidth]{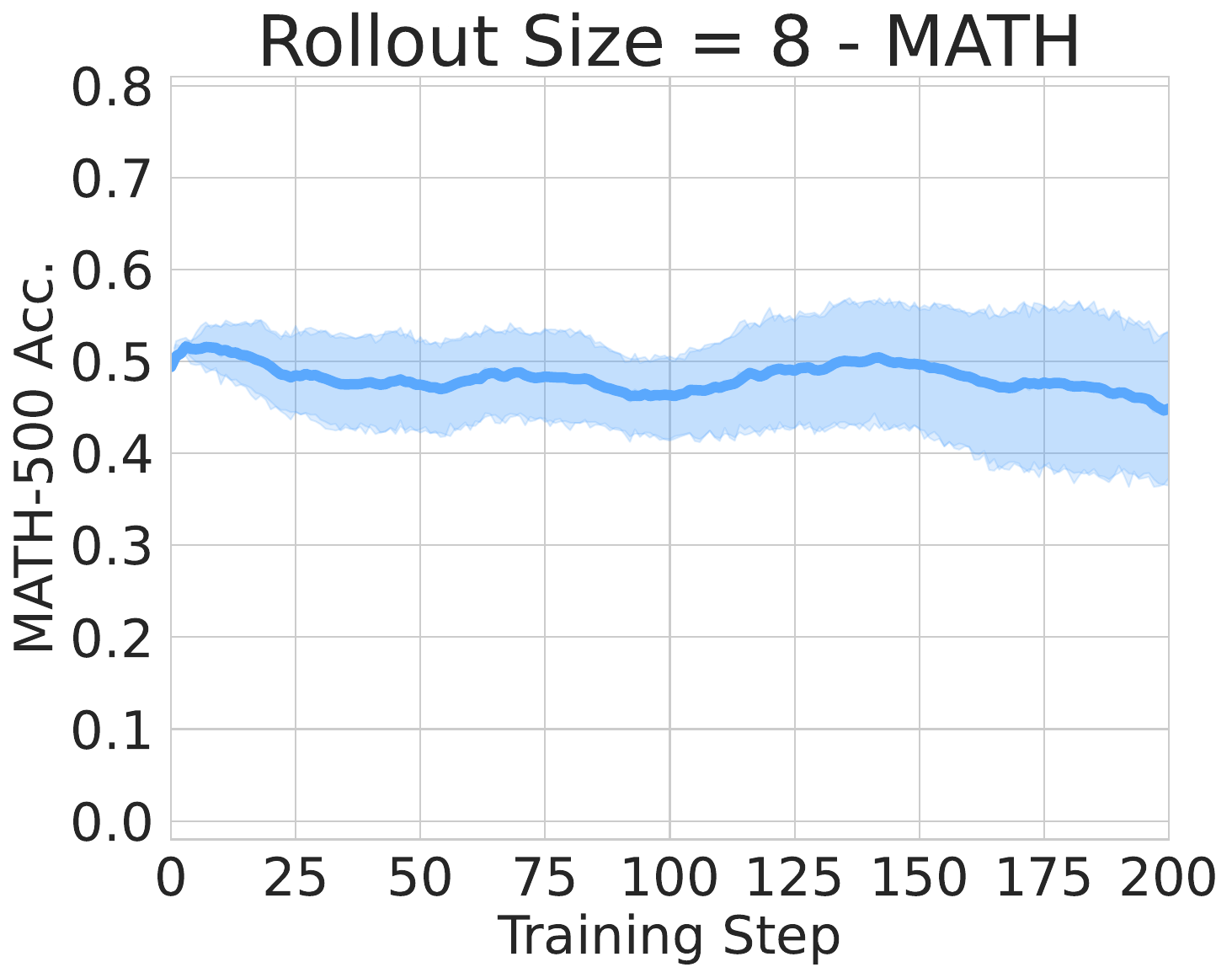}\hfill
        \includegraphics[width=0.49\linewidth]{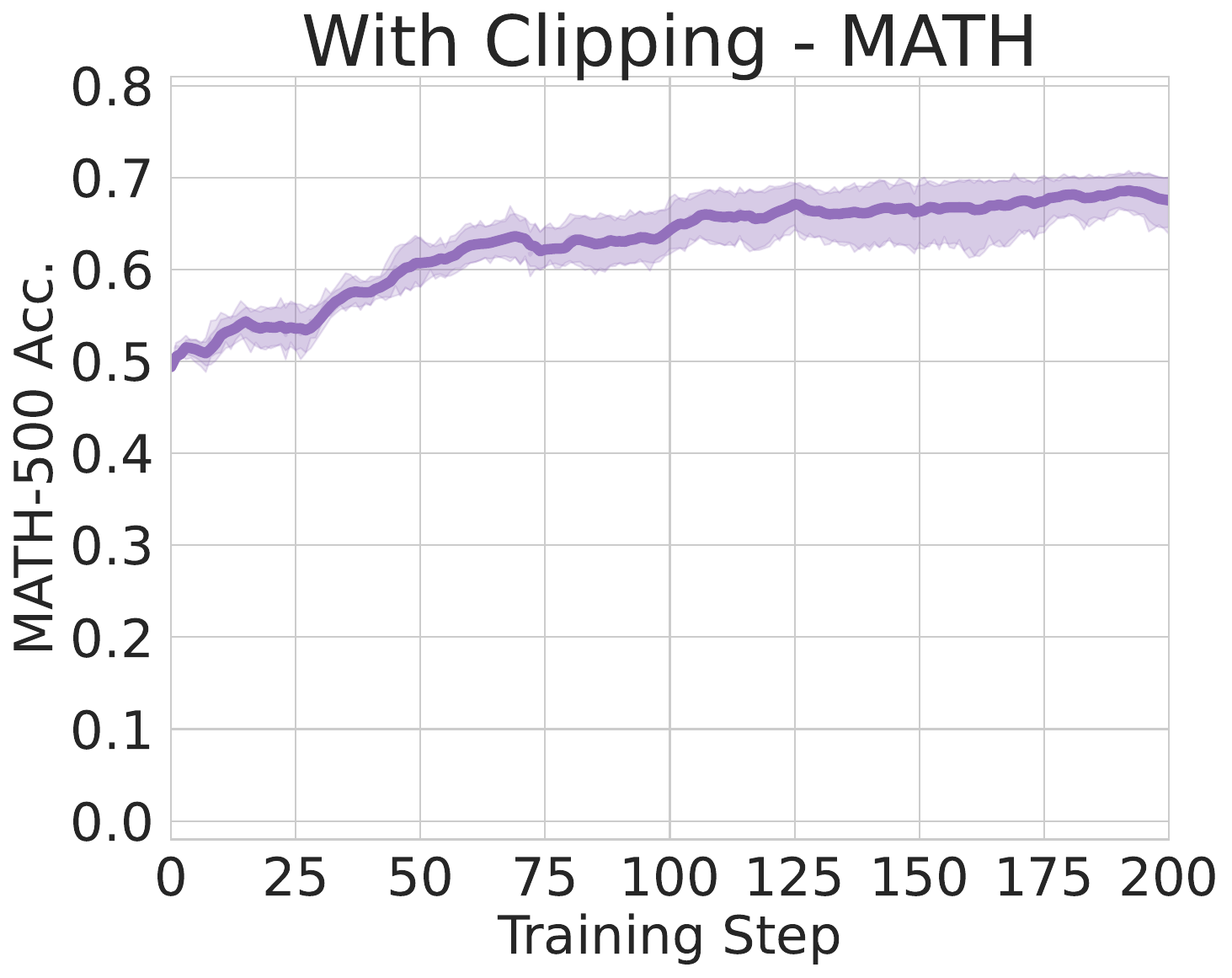}
    \caption{
    Average performance over multiple random seeds when ablating the clipping term in GRPO with random rewards for Qwen2.5-Math-7B.
    The first three panels correspond to \emph{no-clipping variants} (see text), while the last panel enables clipping.
    Random rewards yield consistent improvements only when clipping is present.
    }
\label{fig:clipping_ablation}
\end{figure}

\noindent\textbf{No training signal from random rewards when clipping is removed from GRPO.}
We analyze GRPO with the clipping term removed and show that random rewards alone do not yield effective training.
The \textcolor{red}{clipping term} appears in the GRPO objective\footnote{We omit the KL regularization term, which is disabled in our experiments.}:
\begin{equation}
\small
\label{eq:grpo-main-text}
\begin{aligned}
J(\theta)=
&\mathbb{E}_{x\sim\mathcal{D},\,y\sim\pi_{\text{old}}(\cdot\mid x)}
\left[
\sum_{t=1}^{|y|}
\min\Bigl(
\rho_t(y;\theta)\,\hat A(x,y),\;\right.\\
&\left.
\textcolor{red}{
\operatorname{clip}\bigl(\rho_t(y;\theta),1-\epsilon_c,1+\epsilon_c\bigr)
\hat A(x,y)}
\Bigr)
\right],
\end{aligned}
\end{equation}
where $x$ denotes the input, $y$ the generated sequence, $\pi_{\text{old}}$ the behavior policy, 
$\rho_t(y;\theta)=\pi_{\theta}(y_t\mid x,y_{<t})/\pi_{\text{old}}(y_t\mid x,y_{<t})$ the importance ratio at step $t$ (corresponding to the $t$-th generated token), and $\epsilon_c$ the clipping threshold. 
The advantage $\hat A(x,y)$ is computed from the reward $r(x,y)$ by centering it within each prompt, 
$\hat A(x,y)=r(x,y)-\mathbb{E}_{y'\sim\pi_{\text{old}}(\cdot\mid x)}[r(x,y')]$.
When clipping (the red term in Eq.~\ref{eq:grpo-main-text}) is removed and rewards are sampled independently of the responses, the expected training objective is zero, yielding no meaningful learning signal.
To demonstrate this, we consider three separate \textbf{no-clipping variants} of GRPO: disabling clipping in the implementation, increasing the mini-batch size to match the rollout size, or reducing the rollout size so that only a single gradient update is performed per rollout.\footnote{The latter two variants enforce $\pi_\theta=\pi_{\text{old}}$ during each rollout, yielding $\rho_t=1$ and thus eliminating the effect of clipping by construction.}
As shown in the first three panels of Figure~\ref{fig:clipping_ablation} (no-clipping variants), random rewards fail to yield consistent performance. In contrast, enabling clipping (bottom-right) leads to stable and consistent improvements.

\noindent\textbf{GRPO clipping bias can induce random reward training signals.}
We analyze the effect of clipping in GRPO under random rewards, with the full derivation provided in Appendix~\ref{app:random_reward}.
At a high level, clipping introduces a bias in the gradient update that depends on how the current policy’s token probabilities deviate from those of the behavior policy.
Factoring out reward-independent scalar terms, the expected gradient induced by random rewards can be written (up to a positive constant) as
\begin{equation}
\small
\mathbb{E}[\nabla_\theta J(\theta)]
\propto
\mathbb{E}_{x,y}\!\left[
\begin{cases}
\;\;\nabla_\theta \log \pi_{\theta,x}(y_t), & R_\theta < 1-\epsilon_c,\\
\;0, & |R_\theta-1|\le \epsilon_c,\\
-\nabla_\theta \log \pi_{\theta,x}(y_t), & R_\theta > 1+\epsilon_c,
\end{cases}
\right],
\end{equation}
where $R_\theta=\pi_{\theta,x}(y_t)/\pi_{\text{old},x}(y_t)$.
This form shows that clipping increases the probability of tokens that are already likely under the policy, while suppressing overly large deviations, even when rewards are uninformative. We illustrate this effect with the following example:

\begin{tcolorbox}[
  colback=gray!5,
  colframe=gray!50,
  boxrule=0.5pt,
  arc=2pt,
  left=6pt,
  right=6pt,
  top=6pt,
  bottom=6pt
]
\vspace{-5pt}
\noindent\textbf{Example.}
Let clipping threshold $\epsilon_c=0.2$ and consider two tokens with different probabilities under the policy. The same relative clipping threshold therefore induces asymmetric absolute clipping ranges within $[0,1]$ for low- and high-probability tokens.
If $\pi_{\text{old},x}(y_t)=0.85$ (frequently sampled), the policy probability would need to exceed $\pi_{\text{old},x}(y_t) \cdot (1+\epsilon_c) = 1.02$ to trigger upper clipping. Since $\pi_{\theta,x}(y_t) \leq 1$, this threshold can never be reached. Thus, this token receives nonnegative gradient bias, reinforcing its high probability.
In contrast, if $\pi_{\text{old},x}(y_t)=0.02$ (rarely sampled), upper clipping is triggered when $\pi_{\theta,x}(y_t) > 0.024$. Small increases in $\pi_{\theta,x}(y_t)$ can easily exceed this threshold, causing this token to receive negative gradient bias that suppresses its probability.
Thus, clipping bias asymmetrically suppresses low-probability tokens and reinforces high-probability ones.
\vspace{-5pt}
\end{tcolorbox}

In Appendix~\ref{app:random_reward}, we show that this effect persists across different random reward distributions, indicating its stability.
Our findings echo the intuition of \citet{yu2025dapo}, who show that clipping bias reduces exploration and increases exploitation in RLVR with ground-truth labels.

\section{Case Study: Code Reasoning as a Representative Pre-Existing Behavior}
\label{sec:analysis}

Previously, we showed that different models can exhibit substantially different outcomes when trained with the same weak or spurious reward signals.
Here, we propose a general hypothesis for these discrepancies and examine it through a focused case study of a representative pre-existing behavior observed in Qwen models.
Our central hypothesis is that differences in RLVR outcomes stem from differences in the reasoning strategies acquired during pretraining.
Some strategies may be readily amplified by RLVR, while others may be difficult to elicit or absent altogether, leading to divergent training dynamics across model families.

We study one such strategy---using code to support mathematical reasoning---which Qwen-Math models employ effectively, while other model families do so to a lesser extent (\S\ref{sec:analysis:qualitative}).
By tracing the prevalence of code reasoning over the course of RLVR training, we find evidence consistent with this hypothesis (\S\ref{sec:code_freq_acc}).
We note that we use code reasoning as a controlled and observable example to study how RLVR interacts with pre-existing internal strategies, rather than as an exhaustive characterization of model behavior.





\subsection{Different Models Exhibit Pre-existing Reasoning Strategy Discrepancies}
\label{sec:analysis:qualitative}

We analyze reasoning traces from different models on MATH-500 and observe clear discrepancies in pre-existing strategies.
\qwenmath frequently generates Python code in its reasoning (65.0\% of responses), despite lacking code execution; we refer to this behavior as \emph{code reasoning}.

\vspace{-8pt}
\begin{figure}[ht]
\centering
\begin{tcolorbox}[
 colframe=MidnightBlue,
 colback=blue!5,
 coltitle=white,
 title=\textbf{MATH Question:},
 fonttitle=\bfseries
]
What is the distance, in units, between the points $(2, -6)$ and $(-4, 3)$? Express your answer in simplest radical form.
\end{tcolorbox}
\begin{tcolorbox}[
 colframe=DodgerBlue4, 
 colback=white,
 coltitle=white,
 title={\textbf{Qwen2.5-Math-7B Solution (correct):}},
 fonttitle=\bfseries
]
To find the distance between two points $(x_1, y_1)$ and $(x_2, y_2)$ in a Cartesian plane...

Let's break this down step-by-step and compute the result using Python.

\begin{lstlisting}[language=Python]
import math
...
# Calculate the distance using the distance formula
distance = math.sqrt(dx**2 + dy**2)
print(distance)
\end{lstlisting}

output:
10.816653826391969

...

Thus, the final answer is:
\fbox{$3\sqrt{13}$}
\end{tcolorbox}\vspace{-2mm}
\caption{Example of code reasoning (see Figure~\ref{fig:qualitative_original_correct} for the complete response). Note that both the code and the code execution result are autoregressively generated by \qwenmath. \textbf{No external code interpreter was provided to the model.}}
\vspace{-5pt}
\label{fig:qualitative_original_correct_truncated}
\end{figure}

Crucially, code reasoning in \qwenmath is strongly correlated with correctness.
As shown in Table~\ref{tab:code_freq_initial}, responses containing code achieve substantially higher accuracy than those without.
\qwenmathsmall exhibits similar behavior, while other models do not.
We categorize these models as either \emph{No-Code} (e.g., Llama, \qwensmall, \olmo), which do not generate code, or \emph{Bad-Code} (e.g., \olmosft, \qwen), which frequently generate code but with degraded performance.
Thus, effective code reasoning appears to be a distinctive pre-existing capability of the \qwenmathfamily prior to RLVR.

We provide further analysis that hints at the origins of code reasoning of \qwenmath models in Appendix~\ref{app:qualitative}.
Note that code reasoning is not used as a complete explanation: other behaviors can also be elicited easily and often correlate with performance---we briefly discuss another such behavior, generation without repetition, in Appendix~\ref{app:analysis_repetition}. 
In addition, we find the initial performance of \qwenmath is sensitive to the prompts used for evaluation; we supplement a discussion on this impact in Appendix~\ref{app:prompt_effects}.

\subsection{RLVR with Spurious Rewards Upweight Pre-existing Reasoning Strategies}\label{sec:code_freq_acc}
Motivated by our observations above, we traced changes in the reasoning behavior of models throughout the course of RLVR training across two dimensions: (1) \textbf{Accuracy}: the average accuracy of the model on MATH-500, and (2) \textbf{Code reasoning frequency}: the percentage of model responses containing the string ``\texttt{python}''. 
We find that rewards that we employed in the paper, including spurious rewards such as the random and incorrect rewards, gain much of the accuracy on the \qwenmathfamily and Qwen2.5 through eliciting the correct reasoning strategy.

\begin{table}[]
    \centering
    \captionof{table}{We report the fraction of MATH-500 responses containing Python code before RL training, along with accuracy for code-based and natural-language-only responses. \qwenmathfamily models achieve higher accuracy when using code, whereas other models do not benefit from code reasoning. The remaining models (\qwensmall, \olmo, \llama, \llamasmall) never generate code.}
    \resizebox{\linewidth}{!}{
    \begin{tabular}{@{\hskip 3mm}l@{\hskip 3mm}@{\hskip 3mm}r@{\hskip 3mm}r@{\hskip 3mm}r@{\hskip 3mm}r}
    \toprule
        \multirow{2}{*}{\textbf{Model}} & \textbf{Qwen2.5-} & \textbf{Qwen2.5-} & \textbf{Qwen2.5-} & \textbf{OLMo2-} \\
         & \textbf{Math-7B} & \textbf{Math-1.5B} & \textbf{7B} & \textbf{7B-SFT} \\
        \midrule
        \textbf{Code Frequency} & 65.0 & 53.6 & 92.2 & 98.0   \\
        \textbf{Acc. w/ Code}   & 60.9 & 52.6 & 39.9 & 21.0 \\
        \textbf{Acc. w/ Lang}   & 35.0 & 17.2 & 61.5 & 40.0 \\
    \bottomrule
    \end{tabular}
    }
    \label{tab:code_freq_initial}
\end{table}

\begin{figure}[t]
    \centering
    \cblock{94}{149}{78} Ground Truth
    \cblock{70}{129}{188} Majority Vote
    \cblock{221}{162}{88} Format\\
    \cblock{211}{98}{83} Incorrect
    \cblock{147}{112}{188} Random
    \begin{subfigure}[t]{\linewidth}
        \centering
        \includegraphics[width=0.49\linewidth]{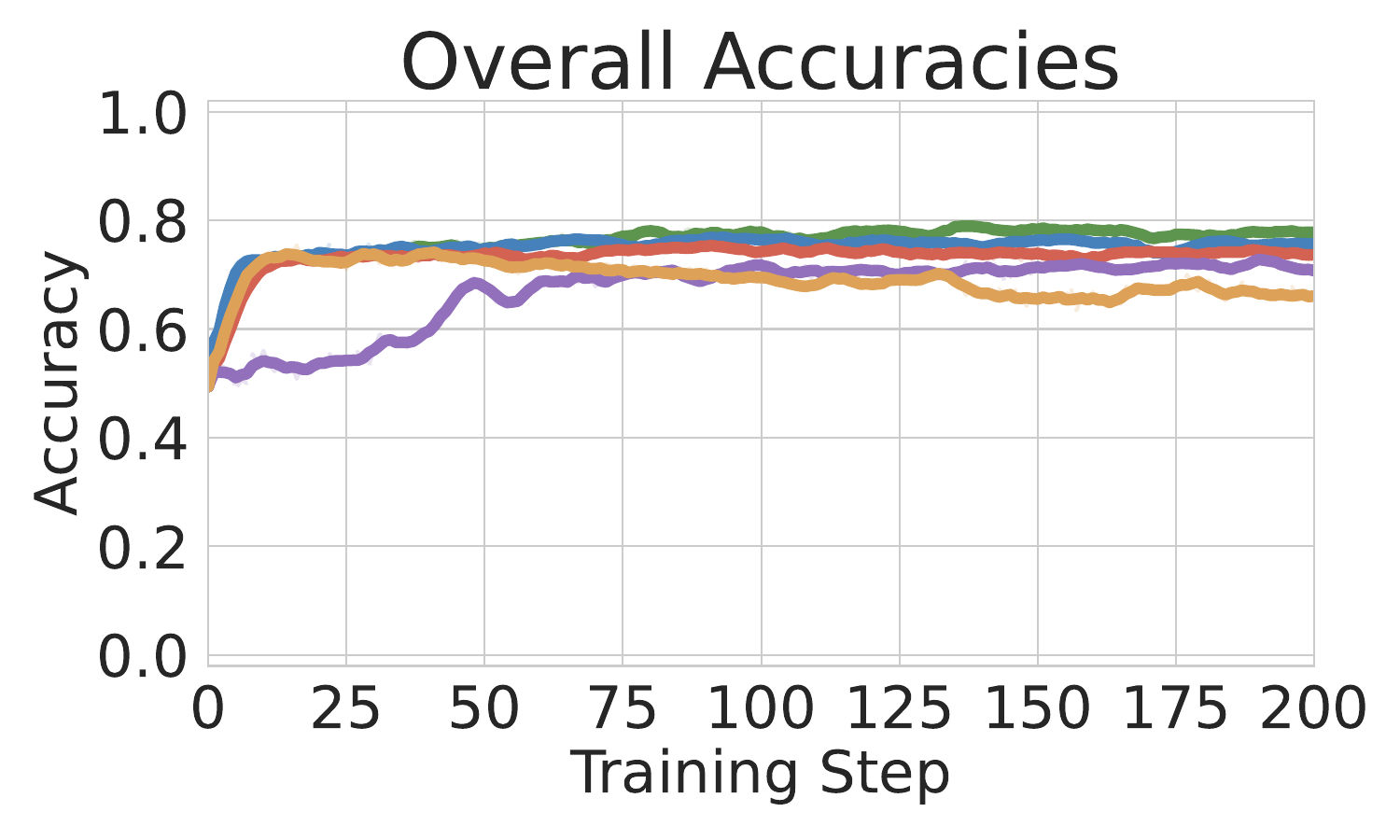}\hfill
        \includegraphics[width=0.49\linewidth]{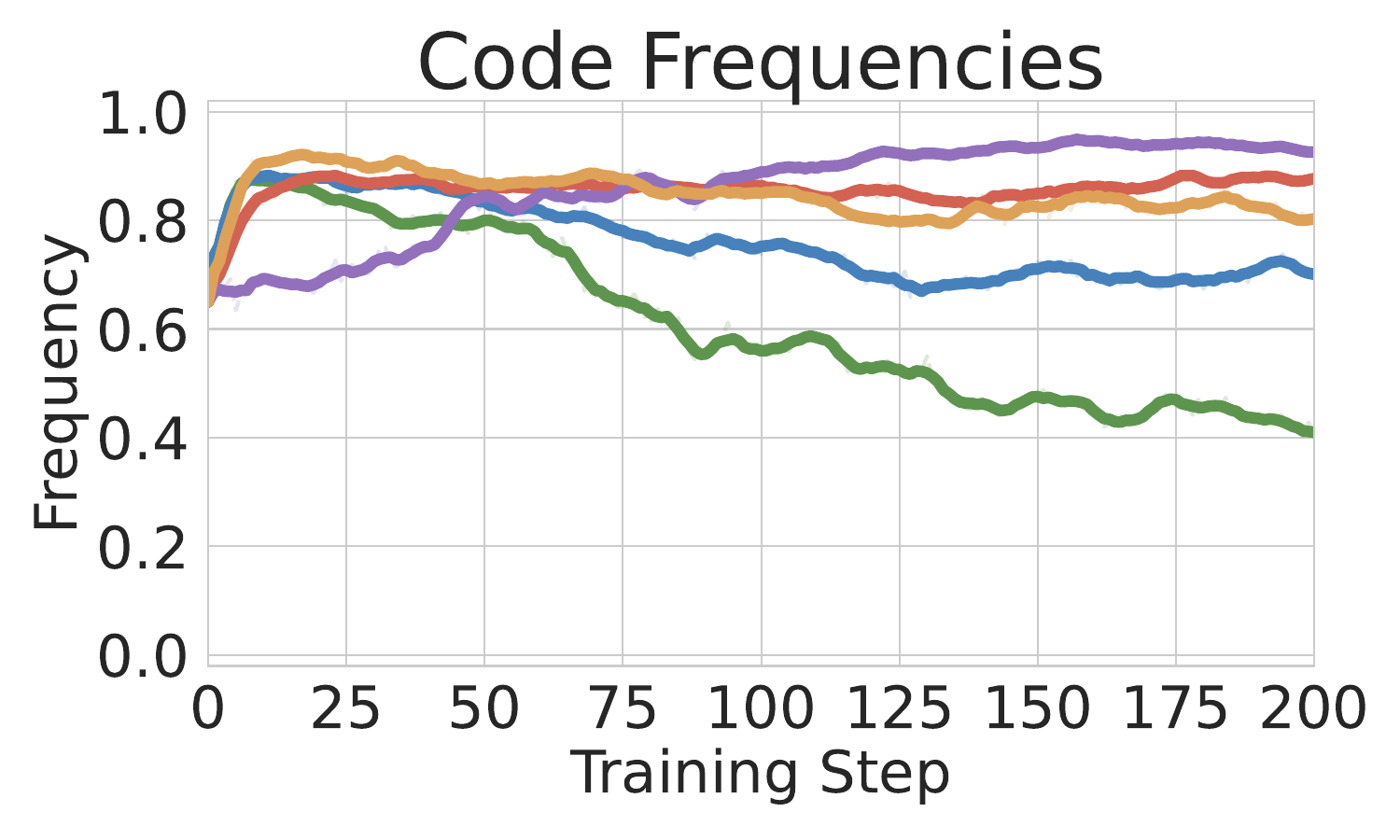}
    \caption{\qwenmath}
    \label{fig:code_freq_qwen_math}
    \end{subfigure}%
    \hfill
    \begin{subfigure}[t]{\linewidth}
        \centering
        \includegraphics[width=0.49\linewidth]{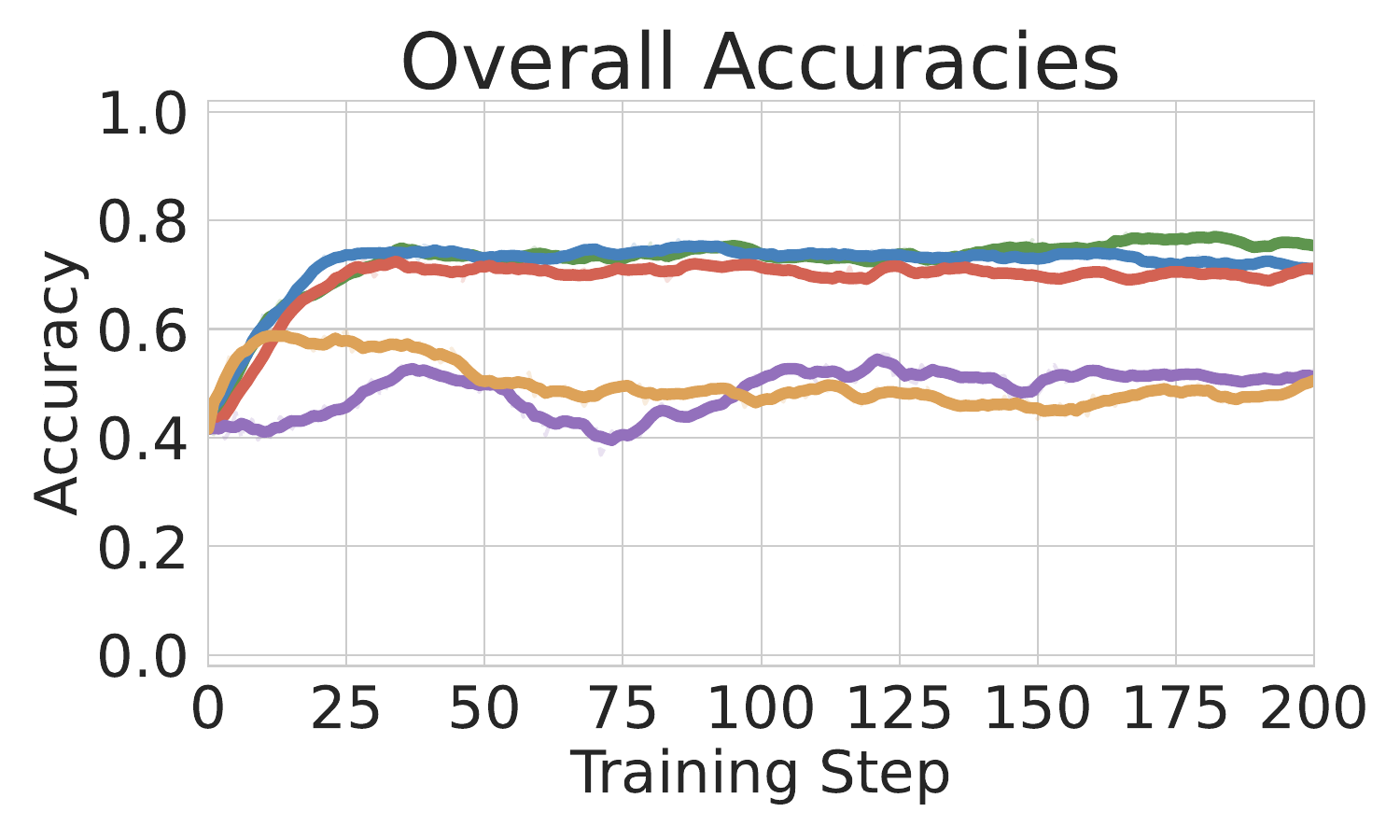} \hfill
        \includegraphics[width=0.49\linewidth]{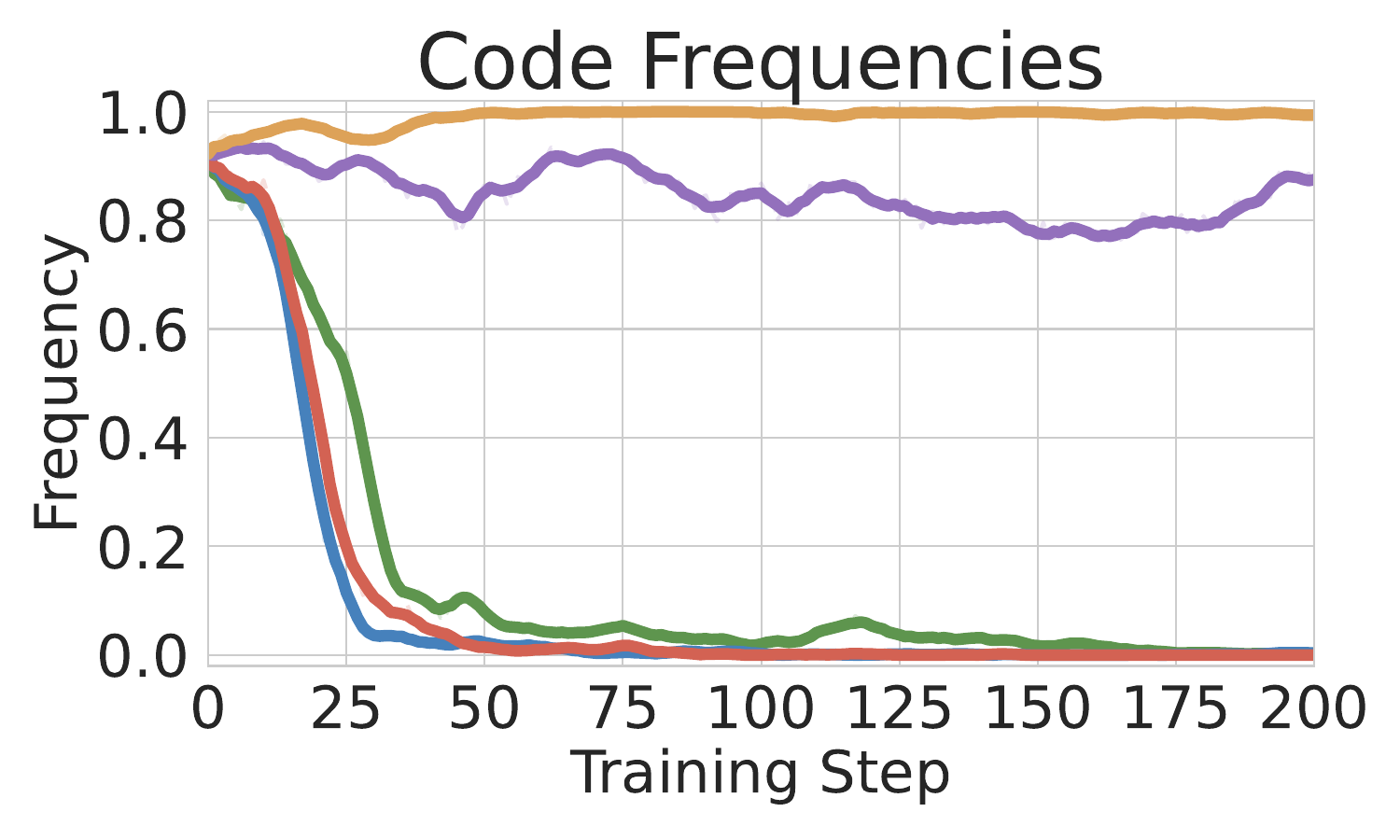}
    \caption{\qwen}
    \label{fig:code_freq_qwen}
    \end{subfigure}%
    \hfill
    \begin{subfigure}[t]{\linewidth}
        \centering
        \includegraphics[width=0.49\linewidth]{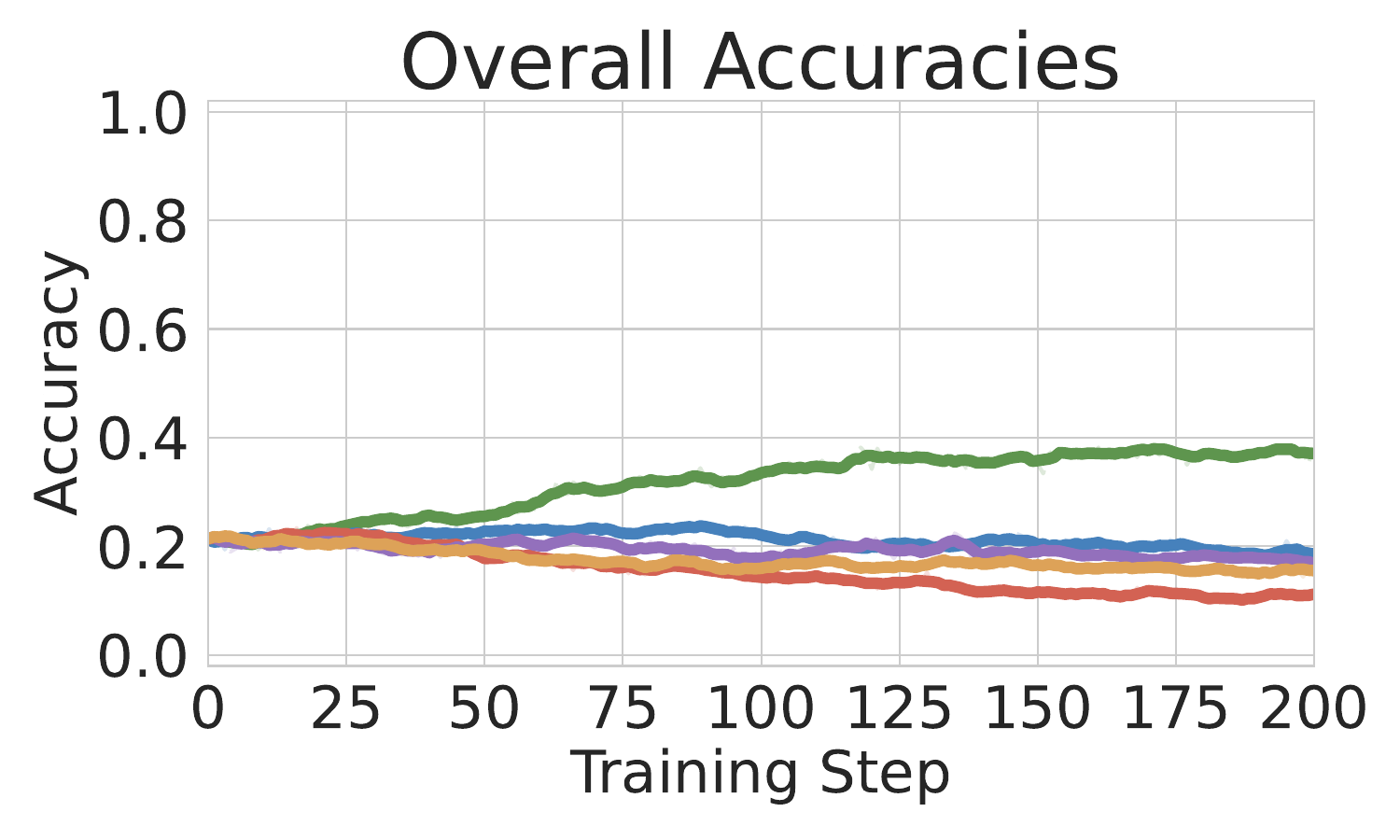}\hfill
        \includegraphics[width=0.49\linewidth]{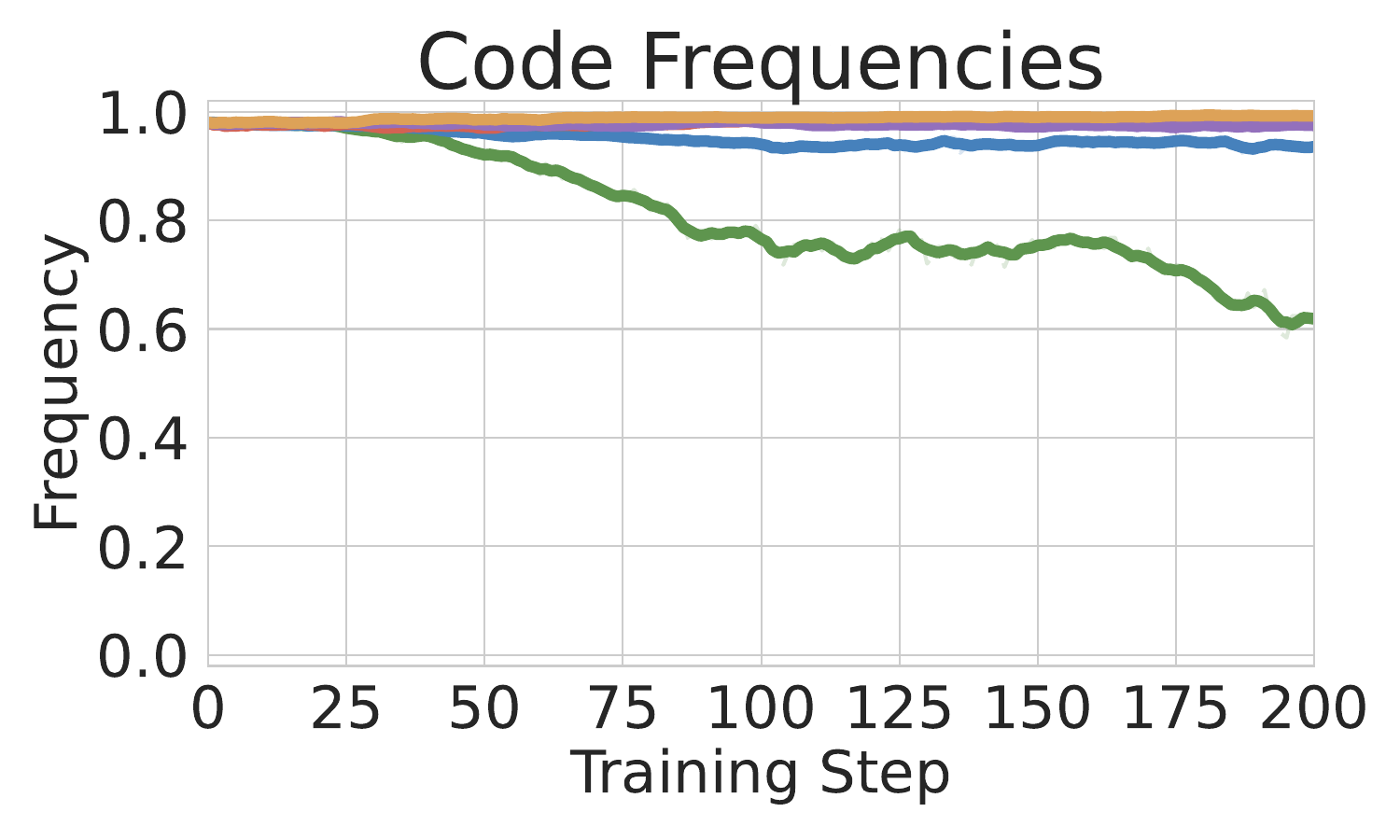}
    \caption{\olmosft}
    \label{fig:code_freq_olmosft}
    \end{subfigure}
    \caption{We track MATH-500 accuracy (left) and the proportion of answers containing Python code (right) for a \qwenmath model trained with different reward signals. Under weak (format) or spurious (random, incorrect) rewards, both accuracy and code frequency increase. In contrast, training with ground-truth rewards improves accuracy while code frequency eventually declines, suggesting a different improvement mechanism.
    }
    \label{fig:analysis_code_freq}
\end{figure}

\noindent\textbf{Performance correlates with code reasoning frequency.}
As shown in Figure~\ref{fig:analysis_code_freq}, prior to RLVR training, \qwenmath uses code reasoning in 65.0\% of MATH-500 solutions.
Under RLVR with weak or spurious rewards, code frequency rapidly increases (to $\sim$90\% within 15 steps and up to 95.6\% for random rewards), closely tracking accuracy gains.
With ground-truth rewards, code frequency also rises initially but later declines as natural-language reasoning improves, suggesting learning beyond reliance on code.
For \emph{Bad-Code} models, performance gains instead correlate with reduced code usage.
Additional analysis of these strategy shifts is provided in Appendix~\ref{app:strategy_switch}, indicating that gains from spurious rewards largely arise when models transition from text-based to code-based reasoning.

\subsection{Intervening Explicitly on Code Reasoning Frequency}
\label{sec:analysis:intervene}
We have shown observationally that code reasoning frequency increases during RLVR and correlates with improved test performance.
Here, we study the effect of explicitly inducing or suppressing code reasoning.
We present results for inducing additional code reasoning in the main text, while experiments suppressing code reasoning are deferred to Appendix~\ref{app:compound_results}.
Across both directions, these interventions produce outcomes consistent with our hypothesis.

\noindent\textbf{Inducing code reasoning improves \qwenmathfamily models but degrades others.}
We induce code reasoning via (1) prompting and (2) RLVR.
With prompting (forcing responses to begin with \textsc{``Let's solve this using Python.''}), MATH-500 accuracy improves by 24.2\%, 15.0\%, and 10.0\% for \qwenmathsmall, \qwenmath, and \qwensmall, respectively (Table~\ref{tab:model-performance}), while performance degrades for other models such as Llama and OLMo2—consistent with their lack of effective code reasoning (Section~\ref{sec:analysis:qualitative}).

\begin{figure}[t!]
    \centering
    \cblock{98}{62}{121} \small{Qwen-Math-7B}
    \cblock{134}{117}{171} \small{Qwen-Math-1.5B}
    \cblock{83}{97}{54} \small{Qwen-7B} \\
    \cblock{143}{163}{100} \small{Qwen-1.5B}
    \cblock{165}{94}{138} \small{Olmo2-7B-SFT}
    \cblock{200}{107}{144} \small{Olmo2-7B}\\
    \cblock{121}{106}{70} \small{Llama3.1-8B}
    \cblock{195}{166}{92} \small{Llama3.2-3B}\\
    \cblock{75}{100}{130} \small{Llama3.1-8B-Instruct}
    \cblock{107}{142}{185} \small{Llama3.2-3B-Instruct}
    \includegraphics[width=0.8\linewidth]{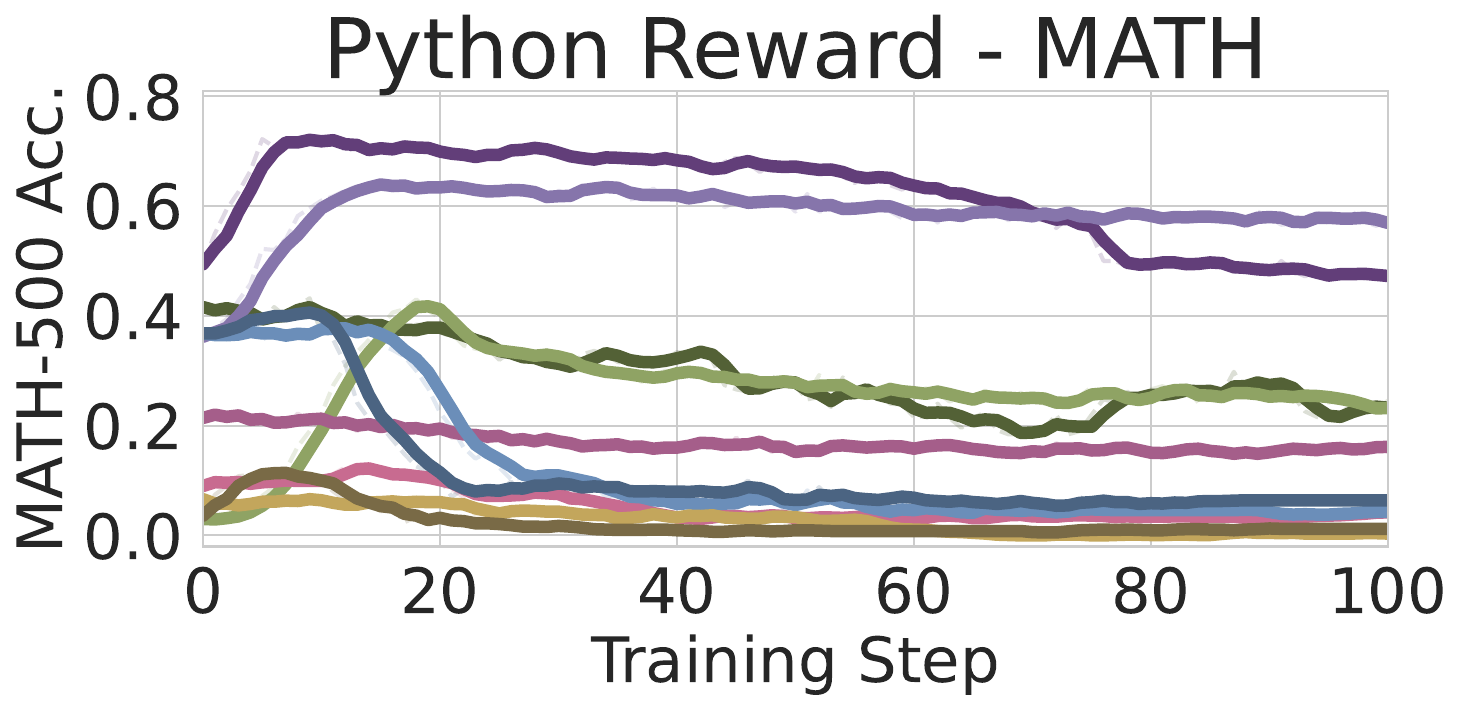}\vspace{-2mm}
    \caption{Performance when using our Python reward to explicitly encourage the model to perform code reasoning. This improves performance \emph{only} in \qwenmathfamily. 
    \qwenmath starts to generate > 99\% code reasoning in 20 training steps. }\vspace{-2mm}
    \label{fig:python_results}
\end{figure}

To induce code reasoning with RLVR, we assign a positive reward if and only if a response contains the string ``\texttt{python}'' (a \emph{Python reward}).
This rapidly increases code usage in \qwenmath to over 99\% within 20 training steps (Figure~\ref{fig:python_results}) and yields performance gains primarily in the \qwenmathfamily.
For these models, RLVR-based induction matches or exceeds the gains achieved through prompting, while other models show limited improvement.

\begin{table}[t]
\caption{Model performance on MATH-500 after augmenting the prompt to incentivize code reasoning. In this experiment, we force the model's first generated sentence to be ``Let's solve this using Python.'' When applied to \qwenmathfamily models, which have strong code reasoning priors, our ``code-forcing'' prompting strategy results in significantly increased test accuracy.}
    \label{tab:model-performance}
    \centering
    \resizebox{\linewidth}{!}{
    \begin{tabular}{@{\hskip 3mm}l@{\hskip 3mm}c@{\hskip 3mm}c@{\hskip 3mm}c@{\hskip 3mm}c@{\hskip 3mm}}
    \toprule
    \textbf{Model} & \textbf{Original} & \textbf{Prompting} & \textbf{Abs. Diff.} \\
    \midrule
    \qwenmathsmall & 36.2\% & 60.4\% & +24.2\% \\
    \qwenmath      & 49.4\% & 64.4\% & +15.0\% \\
    \qwensmall     &  3.0\% & 13.0\% & +10.0\% \\
    \qwen          & 41.6\% & 22.2\% & –19.4\% \\
    \llamasmall    & 36.8\% &  8.2\% & –28.6\% \\
    \llama         & 36.8\% & 15.2\% & –21.6\% \\
    \olmo          & 9.0\% &  7.8\% & –1.2\% \\
    \olmosft       & 21.4\% & 18.6\% & –2.8\% \\
    \bottomrule
    \end{tabular}}\\
\end{table}

\section{Related Work}

\noindent\textbf{Unsupervised Reinforcement Learning.}
Several approaches have explored unsupervised reinforcement learning. \citet{prasad2024self} introduced Self-Consistency Preference Optimization (ScPO), which trains models to prefer consistent answers over inconsistent ones on unsupervised problems. Similarly, Test-Time Reinforcement Learning (TTRL) \citep{zuo2025ttrl} leverages majority voting across sampled outputs to estimate pseudo-rewards, demonstrating significant performance improvements on mathematical reasoning tasks. 
EMPO~\citep{zhang2025right} adapts PPO~\citep{schulman2017proximal} or GRPO~\citep{shao2024deepseekmath} for unsupervised RLVR by calculating the rewards based on minimizing the entropy of queries in the semantic space.
These approaches suggest that internal model consistency can serve as an effective proxy for correctness, though they do not systematically investigate how different kinds of training rewards affect various model families during RLVR.

\noindent\textbf{Reinforcement Learning for Language Models.}
The development of language model capabilities has been significantly advanced through reinforcement learning approaches. RLHF has become a standard technique for aligning models with human preferences~\citep{ouyang2022training, bai2022constitutional}, while RLVR has proven effective for tasks with deterministic answers \citep{deepseekteam2024deepseek, gao2024designing, wen2025lightr1,lambert2024tulu3,deepscaler2025}. 
These methods traditionally rely on accurate supervision signals, either through human feedback or verifiable rewards. Recent work has explored reducing the dependence on human annotations through AI feedback mechanisms~\citep{bai2022constitutional} and through training dynamics analysis~\citep{zhao2025echo}, which supports the finding that RL primarily amplifies behaviors or capabilities already buried in the pretrained models~\citep{liu2025understanding,yue2025does}.

\section{Conclusion}
Our findings have three main implications: base model pretraining significantly affects RLVR outcomes; even corrupted or spurious supervision can enhance reasoning when it triggers useful existing behaviors; and effects observed in one model family may not generalize to others. Our work highlights the importance of 
(1) testing across multiple models with differing pretraining distributions, and
(2) testing across multiple different baselines, such as format and random rewards, when evaluating reinforcement learning techniques.

This work does not advocate the use of spurious rewards as a training strategy for improving model capabilities. Instead, spurious rewards are introduced solely for controlled analysis, as their misuse in deployed systems could lead to unreliable or undesirable behavior. Overall, the anticipated impact of this work is to improve the rigor and robustness of reinforcement learning research rather than to enable new applications.

\section*{Acknowledgments}
We thank Hamish Ivison and Trung Vu for insightful discussions. 
This research was developed in part with funding from the Defense Advanced Research Projects Agency's (DARPA) SciFy program (Agreement No. HR00112520300). The views expressed are those of the author and do not reflect the official policy or position of the Department of Defense or the U.S.~Government.
This work was supported by the Singapore National Research Foundation and the National AI Group in the Singapore Ministry of Digital Development and Information under the AI Visiting Professorship Programme (award number AIVP-2024-001), and by the AI2050 program at Schmidt Sciences. This work was supported by NSF grants 2112471 and 2134012.
SG is supported by the NSF GRFP. SSD acknowledges the support of NSF IIS-2110170, NSF DMS-2134106, NSF CCF-2212261, NSF IIS-2143493, NSF CCF-2019844, NSF IIS-2229881, and the Sloan Research Fellowship.

\bibliographystyle{icml2026}
\bibliography{neurips2025}


\newpage
\appendix
\onecolumn

\section*{Appendix}
\startcontents[sections]
\printcontents[sections]{l}{1}{\setcounter{tocdepth}{2}}

\newpage

\section{Experimental Setup}\label{app:setup}

\subsection{Preliminaries on GRPO} \label{sec:preliminaries}
Denote the input prompt as $x$ and the corresponding model rollouts as $y$. Denote the $t$-th token in the $i$-th rollout $y$ as $y_t$, where $1\leq t \leq |y|$.
GRPO loss, as introduced by \citep{shao2024deepseekmath}, contains the following components: For each prompt $x$, we compute the group-wise mean $\bar{r}_x$ and standard deviation $\sigma_x$. The normalized \emph{group-relative advantage} is 
\( \hat{A}(x,y) = \frac{r(x,y) - \bar{r}_x}{\sigma_x}. \)
Let $\pi_{\text{old}}$ be the behavior policy that produced the trajectories and $\pi_{\text{ref}}$ be a frozen reference policy (for example, the initial supervised model). We denote the token-level importance ratio by 
\(\rho_t(y;\theta) = \frac{\pi_{\theta}(y_t|x,y_{<t})}{\pi_\text{old}(y_t|x,y_{<t})} \), where $t$ is the token index.
With PPO-style clipping threshold $\epsilon_c$, KL-penalty weight $\lambda$,
the surrogate objective maximized is

\begin{equation}
\small
\label{eq:grpo}
\begin{aligned}
J(\theta)=
&\mathbb{E}_{x\sim\mathcal{D},y\sim\pi_\text{old}(\cdot|x)}\!
\left[
\sum_{t=1}^{|y|}
\min\Bigl(
\rho_t(y;\theta)\,\hat A(x,y),
\operatorname{clip}\bigl(\rho_t(y;\theta),1-\epsilon_c,1+\epsilon_c\bigr)
\,\hat A(x,y)
\Bigr)
\right]\\
&-\lambda\,\mathbb{E}_{x\sim\mathcal D}
\Bigl[\mathrm{KL}\bigl(\pi_{\theta}\|\pi_{\text{ref}}\bigr)\Bigr].
\end{aligned}
\end{equation}

The KL regularization is typically adopted to ensure our model does not deviate too far from the frozen reference model. However, since we focus primarily on verifiable rewards and investigate the signals produced by these rewards, we assume that distribution shift is of lesser concern following \citep{liu2025understanding}. Moreover, KL regularization adds confounding factors to our analysis. Recent work has also shown that removing the KL term leads to better performance \citep{hu2025open}. Therefore, we set $\lambda = 0$ in our work.

\subsection{Datasets and Models}\label{app:eval-setup}

We conduct our experiments on three canonical mathematical reasoning benchmarks:
\begin{itemize}[itemsep=1pt,topsep=0pt,leftmargin=12pt] 
\item \textbf{MATH-500} \citep{hendrycks2021measuring,lightman2023let}: A standardized subset of the MATH dataset focusing on advanced mathematical reasoning, including problems from algebra, calculus, and number theory.
\item \textbf{AMC} \citep{li2024numinamath}: The American Mathematics Competition dataset contains diverse mathematical problems ranging from algebra to combinatorics and geometry. For this benchmark, we report model's average performance over 8 trials ($\mathrm{avg@8}$).
\item \textbf{AIME} \citep{li2024numinamath}: The American Invitational Mathematics Examination dataset consists of challenging high-school level mathematical problems requiring multi-step reasoning. We include AIME questions from 2024 and 2025 for evaluation. For this benchmark, we report model's average performance over 8 trials ($\mathrm{avg@8}$).
\end{itemize}

For our initial experiments, we primarily utilize the Qwen2.5-Math-7B model, a 7 billion parameter language model specifically tuned for mathematical reasoning tasks. We select this model (1) for its strong baseline performance on mathematical problems while remaining computationally efficient for multiple experimental iterations and (2) because it is frequently used in prior work~\citep{zuo2025ttrl, wang2025reinforcement}.

\subsection{Training Configuration}\label{app:training-setup}
Unless otherwise specified, we train each model on 8 GPUs with a constant learning rate of 5e-7, a mini batch size (number of rollouts seen before a gradient update) of 128, and a rollout batch size (number of prompts we rollout at the same time) of 64. For each prompt, we collect 16 rollouts to compute advantages for GRPO update. We use a sampling temperature $\tau=1$.  We do not apply KL divergence loss or entropy loss in our training.

\subsection{Decoding Configuration}
We use temperature of 0.0 for pass at 1, and temperature of 0.6 for pass at k during decoding time. All other hyperparameters are the default value following the \citet{zuo2025ttrl} codebase. 

\subsection{Computation Resource}
Each RLVR run in our experiment takes approximately 24 hours on 8 A100s.

\subsection{Visualization}
In all plots, we smooth the metric of interest across ten training steps, and overlay the raw curves as transparent dashed lines. We do not smooth the step 0 performance, which corresponds to the performance of the base model before training. We report the smoothed value at the final training step as the final performance.

\subsection{Experimental Setups for TTRL and One-Shot RL}\label{app:oneshot_ttrl_setup}
\paragraph{TTRL setup.}
We follow all hyperparameters from the original TTRL paper~\citep{zuo2025ttrl}, and train all models using their publicly released implementation. Due to compute constraints, we do not extensively sweep hyperparameters on the new base models we consider. Note that the TTRL accuracies are not directly comparable to our majority-vote reward results, because (a) TTRL trains on (unlabeled) test prompts while we train on a much larger set of distinct train prompts, and (b) TTRL updates labels during training based on the online policy's majority vote, while we assign labels once offline.

\paragraph{Differences in our One-Shot RL setup from \cite{wang2025reinforcement}.}
We note that for the one-shot RL settings in our paper, we do not apply entropy loss for more consistent setups with other experiments and better stability. However, \cite{wang2025reinforcement} shows that the entropy loss may further improve the performance of one-shot RL and post-saturation generalization. We use the same training example $\pi_1$ (defined in \cite{wang2025reinforcement}) for one-shot RLVR experiments on all models as in \cite{wang2025reinforcement}. Their work discusses that different models should possibly select different examples for more performant one-shot RLVR training, but use $\pi_1$ in all experiments due to high cost of trying different possible 1-shot examples. In particular, we note that this setup difference may cause a noticeable performance difference for Qwen2.5-7B model compared to the entropy loss setting used in \cite{wang2025reinforcement} (71.2\% with entropy loss vs. 54.8\% without).

\section{The Curious Case of Random Rewards}\label{app:random_reward}
\begin{figure}[t]
\centering\hfill
\begin{minipage}[t]{0.45\linewidth}
\vspace{0pt}
  \includegraphics[width=\linewidth]{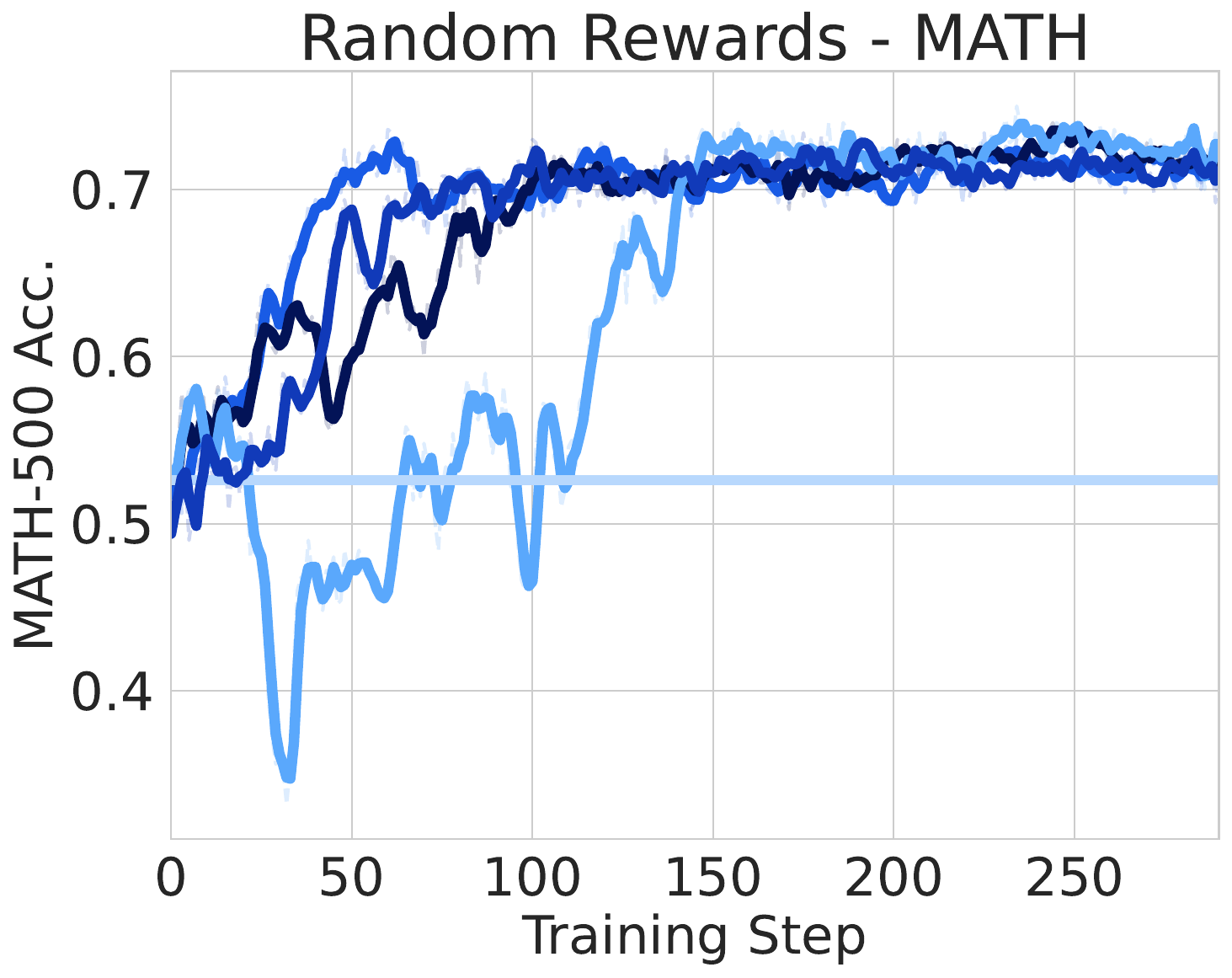}
\end{minipage}\hfill
\begin{minipage}[t]{0.45\linewidth}
\vspace{0pt}
  \caption{We train \qwenmath using GRPO with random rewards of different probability $\gamma \in \{0.7, 0.5, 0.3, 0.001, 0\}$. We note that with the exception of $\gamma=0$, which leads the model to not perform any learning, all other probability configuration successfully leads the model to achieve a significant performance gain after some period of random walk. Furthermore, once in the high-accuracy regime, the training procedure maintain model's performance.\vspace{1mm}\\
  \cblock{3}{19}{87} Random $\gamma = 0.7$\\
  \cblock{17}{58}{185} Random $\gamma = 0.5$\\
  \cblock{25}{91}{229} Random $\gamma = 0.3$\\
  \cblock{90}{168}{252} Random $\gamma = 0.001$\\
  \cblock{184}{216}{253} Random $\gamma = 0$}
  \label{fig:random_thresholds}
\end{minipage}\hfill
\end{figure}

In this section, we study a special case of our spurious rewards, random rewards, where the reward is randomly assigned independently of the model rollouts. We first show that random rewards work across several non-zero reward probabilities (Figure~\ref{fig:random_thresholds}), confirming that this is a stable observation. Next, we provide detailed gradient derivation in this special case and discuss our hypothesis on one of the potential sources of training signals with random rewards---the clipping bias in GRPO update.

\paragraph{Random rewards with varying probabilities consistently improve performance.} 
We train \qwenmath using GRPO with random rewards assigned by $\mathtt{Bernoulli}(\gamma)$ variables, where $\gamma \in \{0.7, 0.5, 0.3, 0.001, 0\}$. Each response receives reward 1 with probability $\gamma$ and 0 otherwise.
We find
that all non-zero probability configurations successfully lead to significant performance gains after an initial period of exploration, yielding comparable performance to ground truth rewards (Figure~\ref{fig:random_thresholds}). $\gamma = 0$ yields no improvement as expected, since constant rewards provide no learning signal. The convergence speed varies with $\gamma$, but all configurations eventually reach similar high-performance regimes, with accuracy improvements of 15-20 percentage points on MATH-500.

\paragraph{Recap of notations.} Recall that we denote the input prompt as $x$ and the corresponding model rollouts as $\{y^{(1)}, \cdots, y^{(G)}\}$, where $G\in \mathbb{N^+}$ is the rollout size. In addition, we denote the $t$-th token in the $i$-th rollout $y^{(i)}$ as $y^{(i)}_t$, where $t \in \mathbb{N^+}$ and $1\leq t \leq |y^{(i)}|$. 
The normalized group-relative advantage in GRPO is 
\( \hat{A}(x,y) = \frac{r(x,y) - \bar{r}_x}{\sigma_x} \), where $\bar{r}_x$ is the group-wise mean and $\sigma_x$ is the group-wise standard deviation.
The token-level importance ratio\footnote{For simplicity, we denote $\pi(y_t|x,y_{<t})$ as $\pi_x(y_t)$, and similarly $\pi_\theta(y_t|x,y_{<t})$ as $\pi_{\theta,x}(y_t)$, $\pi_\text{old}(y_t|x,y_{<t})$ as $\pi_{\text{old},x}(y_t)$.}  is
\(\rho_t(y;\theta) = \frac{\pi_{\theta,x}(y_t)}{\pi_{\text{old},x}(y_t)} \), where $t$ is the token index.

\subsection{Conjecture: Clipping Bias Brings Training Signals under Random Rewards}
\label{app:clipping_bias}

\subsubsection{Expected Advantage of Random Rewards} \label{app:random_zero_adv}
We investigate the expected advantage when the reward signal is assigned by a random $\mathtt{Bernoulli(\gamma)}$ variable.
For a prompt $x$ and rollouts $\{y^{(i)}\}_{i=1}^G$, the reward of each $y^{(i)}$ is $r(x,y^{(i)})\sim\mathtt{Bernoulli(\gamma)}$, so the expected average reward of $x$ is $\gamma$.
Assuming $\hat A_i = 0$ when $\sigma_x = 0$, the sum of the normalized advantages over $G$ rollouts $\sum_{i=1}^G\hat A(x,y^{(i)}):=\sum_{i=1}^G\frac{r(x,y^{(i)}) - \bar{r}_x}{\sigma_x} = \frac{\sum_{i=1}^G r(x,y^{(i)}) - G \cdot \bar{r}_x}{\sigma_x} = 0$ by construction.
Furthermore, for i.i.d.~rewards that are independent of the provided samples, such as the random rewards used in our experiments, $\mathbb{E}(\sum_{i=1}^G\hat A(x,y^{(i)})) = G\cdot\mathbb{E}(\hat{A}) = 0 \implies \mathbb{E}(\hat{A})=0$.

The clipping term in GRPO loss (Equation \ref{eq:grpo}) prevents excessive deviation from the previous policy, stabilizing training. 
Recent work suggests this term, which operates on the ratio $\rho_t(y; \theta) = \pi_{\theta,x}(y_t)/\pi_{\text{old},x}(y_t)$, introduces bias toward exploitation in the case of ground truth reward \citep{yu2025dapo}. 
Here, we analyze the bias in the settings of the random rewards, where each rollout receives an advantage that is independent of the rollout.

We note that the loss is no longer differentiable everywhere with clipping. In practical implementations, e.g., PyTorch, the gradients will be automatically set to 0 when the value is clipped. For simplicity, we discuss the gradient of the loss function by assuming 0 gradient at the non-differentiable points.

\subsubsection{Gradient Derivation of Clipping Bias under Random Rewards}\label{app:clipping_gradient}
In this section, we derive the training signals induced by the clipping factor, which we name as a ``clipping bias''. Specifically, we define the clipping bias as the difference in the expected gradients after adding clipping to the GRPO objective\footnote{For simplicity, we denote $\mathbb{E}_{x\sim\mathcal D,y\sim \pi_{\text{old}(\cdot|x)}}$ as $\mathbb{E}_{x,y}$}:
$$\mathtt{Bias}(\nabla_\theta J(\theta)) = \mathbb{E}_{x,y}[\nabla_\theta J(\theta)]- \mathbb{E}_{x,y}[\nabla_\theta L^{\text{unclipped}}(\theta)].$$

As we derived above, $\mathbb{E}(\hat{A}(x,y_j))=0$.
Assuming a simple loss with no clipping, the expectation of the policy gradient without the clipping term is also trivially zero given that $\hat A $ is independent of other variables: 
\begin{align*}
&\mathbb{E}_{x,y}[\nabla_\theta L^{\text{unclipped}}(\theta)] \\ =& 
\mathbb{E}_{x,y}\!
\left[
\frac1{|y|}\sum_{t=1}^{|y|}
\nabla_\theta \rho_t(y;\theta)\,\hat A(x,y)
\right] \\
=& 
\mathbb{E}(\hat A)\mathbb{E}_{x,y}\!
\left[
\frac1{|y|}\sum_{t=1}^{|y|}
\nabla_\theta \rho_t(y;\theta)
\right] \\ =& 0.
\end{align*}
Therefore, the clipping bias under random reward is 
$$\mathtt{Bias}(\nabla_\theta J(\theta)) = \mathbb{E}[\nabla_\theta J(\theta)]-\mathbb{E}[\nabla_\theta L^{\text{unclipped}}(\theta)]=\mathbb{E}[\nabla_\theta J(\theta)].$$

Next, we compute the exact form of this clipping bias, i.e., the gradient in GRPO with random rewards. Recall from Eq.~\ref{eq:grpo}, the surrogate objective that we aim to maximize is:
$$ J(\theta) = \min\Bigl( \rho_t(y;\theta)\,\hat A(x,y), \operatorname{clip}\bigl(\rho_t(y;\theta),1-\epsilon_c,1+\epsilon_c\bigr) \,\hat A(x,y) \Bigr).$$

For simplicity, we further denote $R_\theta = \rho_t(y;\theta)=\frac{\pi_{\theta,x}(y_t)}{\pi_{\text{old},x}(y_t)}$ to be the token-level importance ratio, 
$C = \operatorname{clip}(R_\theta, 1-\epsilon_c, 1+\epsilon_c)$ to be the clipping term, 
and $\hat A(x,y)$ as $\hat A$.
So we have
$ J(\theta) = \min\bigl( R_\theta\cdot\hat A, C\cdot\hat A \bigr)$.
We now analyze the gradient of GRPO loss with respect to the policy model parameters $\theta$
based on the sign of $\hat A$.

As discussed in Section~\ref{app:clipping_bias}, the loss function is not differentiable everywhere. Therefore, we discuss the gradients with respect to the differentiable regions below and set the gradients to 0 at the non-differentiable points at the end.

\paragraph{Case 1: $\hat A \ge 0$.}
When $\hat A \ge 0$, we can directly take $\hat A$ out of the $\min$ function. 
Thus, 
$$ J(\theta) = \min\bigl( R_\theta\cdot\hat A, C\cdot\hat A \bigr) = \hat A \cdot \min(R_\theta, C).$$
Since $C$ is defined as $C = \operatorname{clip}(R_\theta, 1-\epsilon_c, 1+\epsilon_c)$, 
we have:
\begin{itemize}
    \item If $R_\theta < 1-\epsilon_c$, then $C = 1-\epsilon_c$. Thus, $\min(R_\theta, C) = R_\theta$.
    \item If $1-\epsilon_c \le R_\theta \le 1+\epsilon_c$, then $C = R_\theta$. Thus, $\min(R_\theta, C) = R_\theta$.
    \item If $R_\theta > 1+\epsilon_c$, then $C = 1+\epsilon_c$. Thus, $\min(R_\theta, C) = 1+\epsilon_c$.
\end{itemize}
Combining these, when $\hat A \geq 0$, the objective is:
$$ L^{+}(\theta) = \hat A \cdot 
\begin{cases} 
R_\theta, & \text{if } R_\theta < 1+\epsilon_c, \\ 
1+\epsilon_c, & \text{if } R_\theta > 1+\epsilon_c. 
\end{cases} $$
The gradient with respect to $\theta$ for this case is:
$$ \nabla_\theta L^{+}(\theta) = \hat A \cdot 
\begin{cases} 
\nabla_\theta R_\theta, & \text{if } R_\theta < 1+\epsilon_c, \\ 
0, & \text{if } R_\theta > 1+\epsilon_c.
\end{cases} $$

\paragraph{Case 2: $\hat A < 0$.}
When $\hat A < 0$, multiplying by $\hat A$ flips the $\min$ function.
So, 
$$ L^{-}(\theta) = \min(R_\theta\cdot\hat A, C\cdot\hat A) = \hat A \cdot \max(R_\theta, C).$$

Applying a similar analysis, when $\hat A < 0$, 
the gradient with respect to $\theta$ for this case is:
$$ \nabla_\theta L^{-}(\theta) = \hat A \cdot 
\begin{cases} 
0, & \text{if } R_\theta < 1-\epsilon_c, \\ 
\nabla_\theta R_\theta, & \text{if } R_\theta > 1-\epsilon_c.
\end{cases} $$

\paragraph{Combining the cases.} Together, the gradient is:
$$ \nabla_\theta J(\theta) = \hat A \cdot \begin{cases} 
\nabla_\theta R_\theta, & \text{if }  \hat A \geq 0 \text{ and }R_\theta < 1+\epsilon_c,\\ 
0,               & \text{if }  \hat A \geq 0 \text{ and }R_\theta > 1+\epsilon_c, \\
0,               & \text{if }  \hat A < 0 \text{ and }R_\theta < 1-\epsilon_c,\\ 
\nabla_\theta R_\theta, & \text{if }  \hat A < 0 \text{ and }R_\theta > 1-\epsilon_c.
\end{cases} $$

Then, the clipping bias, $\mathtt{Bias}(\nabla_\theta J(\theta))$, is
\begin{align*}
\small
\centering
& \mathtt{Bias}(\nabla_\theta J(\theta))\\=&\mathbb{E}_{\hat A, x,y} \left[ \nabla_\theta J(\theta) \right] \\ 
    =& P(\hat A\geq 0) \cdot \mathbb{E}_{\hat A>0, x,y}[\nabla_\theta L^{+}(\theta)] + P(\hat A<0) \cdot \mathbb{E}_{\hat A<0, x,y}[\nabla_\theta L^{-}(\theta)] + 0 \\
    =& \mathbb{E}_{x,y} \left[ \begin{cases} 
        P(\hat A\geq 0) \cdot \mathbb{\mathbb{E}}_{\hat A\geq 0}[\hat A] \cdot \nabla_\theta R_\theta, & \text{if } R_\theta < 1+\epsilon_c, \\
        0,  & \text{if } R_\theta > 1+\epsilon_c.
    \end{cases} \right]\\ 
    &+ \mathbb{E}_{x,y} \left[ \begin{cases} 
        0,  & \text{if } R_\theta < 1-\epsilon_c, \\
        P(\hat A<0) \cdot \mathbb{\mathbb{E}}_{\hat A<0}[\hat A] \cdot \nabla_\theta R_\theta, & \text{if } R_\theta > 1-\epsilon_c.
    \end{cases} \right]\\
    =& \mathbb{E}_{x,y} \left[ \begin{cases} 
        P(\hat A\geq 0) \cdot \mathbb{\mathbb{E}}_{\hat A\geq 0}[\hat A] \cdot \nabla_\theta R_\theta, & \text{if } R_\theta < 1-\epsilon_c, \\
        (P(\hat A\geq 0) \cdot \mathbb{\mathbb{E}}_{\hat A\geq 0}[\hat A]  
        + P(\hat A<0) \cdot \mathbb{\mathbb{E}}_{\hat A<0}[\hat A]) \cdot \nabla_\theta R_\theta, & 
        \text{if } 1-\epsilon_c < R_\theta < 1+\epsilon_c,
        \\
        P(\hat A<0) \cdot \mathbb{\mathbb{E}}_{\hat A<0}[\hat A] \cdot \nabla_\theta R_\theta, & \text{if } R_\theta > 1+\epsilon_c. 
    \end{cases} \right]
\end{align*}

By definition, $\hat A$ is a normalized distribution and $\mathbb{E}(\hat{A})=0$ from Appendix \ref{app:random_zero_adv}, so 
$$\mathbb{E}(\hat{A}) = P(\hat A\geq 0) \, \mathbb{E}_{\hat A\geq 0}(\hat A) + P(\hat A<0) \, \mathbb{E}_{\hat A<0}(\hat A)=0,\text{ and}$$
$$P(\hat A\geq 0) \, \mathbb{E}_{\hat A\geq 0}(\hat A)=-P(\hat A<0) \, \mathbb{E}_{\hat A<0}(\hat A) \ge 0.$$
We denote $\mu = P(\hat A\geq 0) \, \mathbb{E}_{\hat A\geq 0}(\hat A)=-P(\hat A<0) \, \mathbb{E}_{\hat A<0}(\hat A)$, which by definition is positive.
And we substitute $R_\theta =\frac{\pi_{\theta,x}(y_t)}{\pi_{\text{old},x}(y_t)}$ in the conditions. Finally, we set the gradients on non-differentiable points to 0 and have
\begin{align*}
\small
\mathtt{Bias}(\nabla_\theta J(\theta))= \mu\cdot\mathbb{E}_{x,y} \left[ \begin{cases} 
        \nabla_\theta R_\theta, & \text{if } 
        \pi_{\theta,x}(y_t) < \pi_{\text{old},x}(y_t)\cdot(1-\epsilon_c), \\
        \multirow{2}{*}{0,}  & \text{if } \pi_{\text{old},x}(y_t)\cdot(1-\epsilon_c) \leq \pi_{\theta,x}(y_t) \\
            &\quad \leq \pi_{\text{old},x}(y_t)\cdot (1+\epsilon_c), \\
        -\nabla_\theta R_\theta, & \text{if } 
        \pi_{\theta,x}(y_t) > \pi_{\text{old},x}(y_t)\cdot (1+\epsilon_c).
    \end{cases} \right]
\end{align*}

Recall that $R_\theta = \rho_t(y;\theta)=\frac{\pi_{\theta,x}(y_t)}{\pi_{\text{old},x}(y_t)}$ and both $\pi_{\theta,x}(y_t)$ and $\pi_{\text{old},x}(y_t)$ in range $[0, 1]$. 
Therefore, we observe that there is a \textit{positive gradient} $\big($gradient that increases $\pi_{\theta,x}(y_t)\big)$ if $R_\theta < 1-\epsilon_c$, and a \textit{negative gradient} $\big($gradient that decreases $\pi_{\theta,x}(y_t)$$\big)$ if $R_\theta > 1+\epsilon_c$, which means the clipping bias discourages the model from leaving the clipping region. 
In the rest of the paper, for simplicity we refer to the positive gradient case as ``positive gradient bias'' and the other way as ``negative gradient bias''.

\subsubsection{Clipping Creates Asymmetric Updates Towards Model Prior Knowledge}\label{app:clipping_hypothesis}

In this section, we provide further evidence for our hypothesis that the clipping bias increases the likelihood of high-probability rollouts, similar to \citet{yu2025dapo}'s analysis on GRPO with ground-truth labels.
We showcase this trend with a simple example.
Consider a case where a token has a high-probability $\pi_{\text{old},x}(y_t)=0.85$ and an $\epsilon_c =0.2$ that we adopt in our experiments.
Then, the upper threshold from the bias formula becomes $\pi_{\text{old},x}(y_t) \cdot(1+\epsilon_c) = 1.02$.
Since the probability output by the policy model cannot exceed 1, the upper clipping threshold of $1.02$ is never reached. Thus, the gradient bias is nonnegative for this token, 
leading to a net positive gradient bias on the policy model, which leads to increase in probability on this token.

On the other hand, for a low-probability token where $\pi_{\text{old},x}(y_t)=0.02$, the policy model receives a negative gradient bias when $\pi_{\theta,x}(y_t) > \pi_{\text{old},x}(y_t) \cdot(1+\epsilon_c)=0.024$; and positive gradient bias when $\pi_{\theta,x}(y_t) <\pi_{\text{old},x}(y_t) \cdot(1-\epsilon_c)= 0.016$.
The low threshold for negative gradient bias makes penalties more likely than in the high-probability case.

The clipping range width scales linearly with the original probability: higher-probability tokens have wider ranges and face fewer penalties. This asymmetric treatment prevents low-probability samples from receiving substantial upweighting during training, causing the model to concentrate probability mass on its existing distribution.

\paragraph{Empirical validation.}
We empirically validate our conjecture by 
disabling the clipping term in the GRPO loss, and observe differences in the training dynamics.
Specifically, we focus on \emph{Token Probability} $\pi_{\theta}$ and \emph{Code Frequency} as defined below.

Recall that $\pi_{\text{old}}$ is the old policy from the previous training step
and the $\pi_{\theta}$ the current policy,
let $G$ be the number of rollouts generated per prompt; let $B$ be the mini batch size, which is the number of prompts on which a single gradient update is performed; and let $M$ be the number of gradient updates from the old policy, $\pi_{\text{old}}$, to the new policy model, $\pi_\theta$. Note that $G\cdot B\cdot M$ is the total number of rollouts generated by $\pi_{\text{old}}$. The average token probability of $\pi_\theta$ over all rollouts $\{y^{(k)}\}$ at any step can be expressed as:

$$\pi_{\theta,x}(y)=\frac{1}{M}\sum_{i=1}^{M}\frac{1}{B}\sum_{j=1}^{B}\frac{1}{G}\sum_{k=1}^G\pi_{\theta_i,x_j}(y^{(k)}).$$

\begin{figure*}[t!]
    \centering
    \cblock{36}{174}{18} Random w/o Clipping
    \cblock{147}{112}{188} Random w/ Clipping
    
    \begin{subfigure}[t]{0.49\textwidth}
        \centering
        \includegraphics[width=\linewidth]{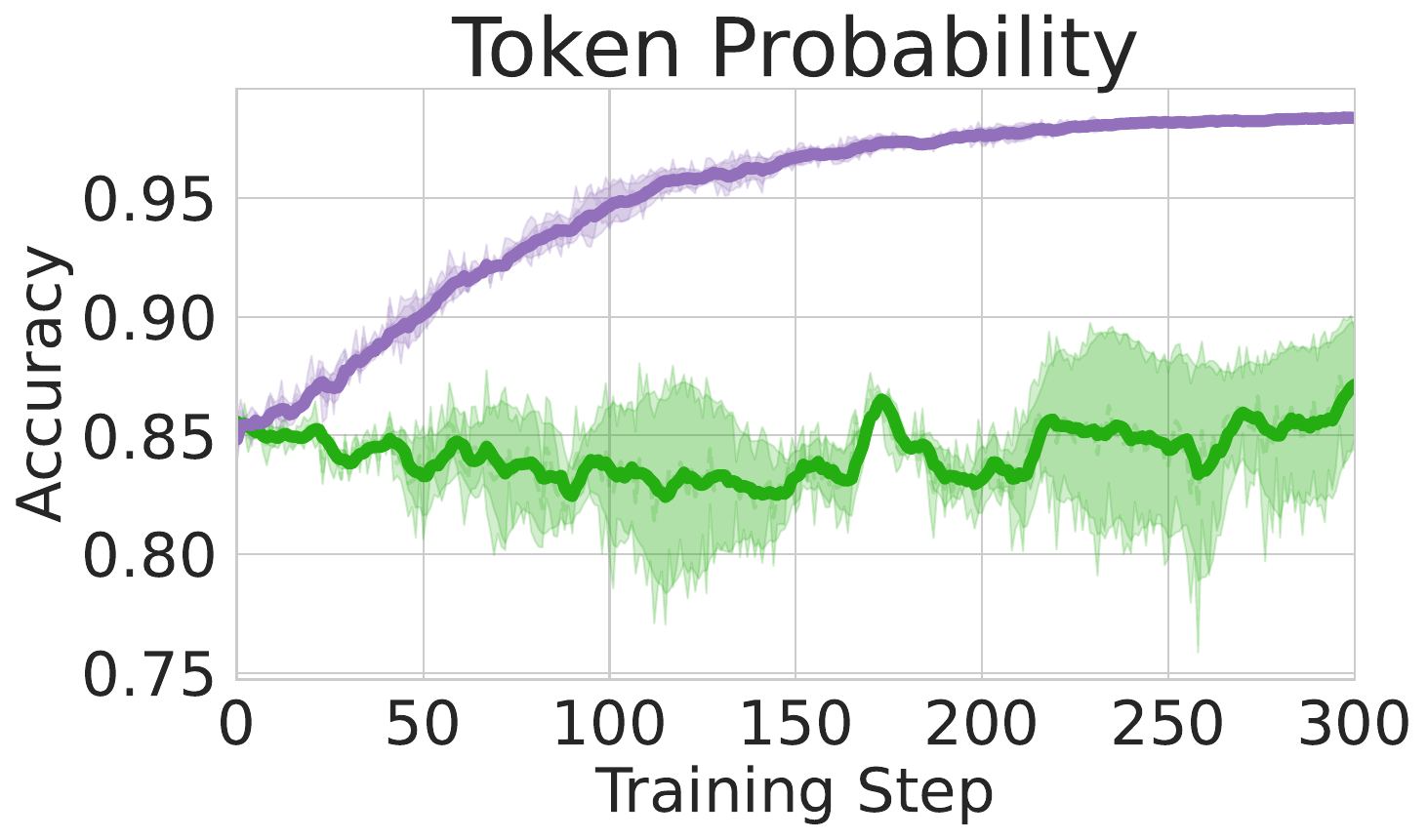}
    \caption{Token Probability}
    \label{fig:noclip_action_probs_actionprob}
    \end{subfigure}%
    \hfill
    \begin{subfigure}[t]{0.49\textwidth}
        \centering
        \includegraphics[width=\linewidth]{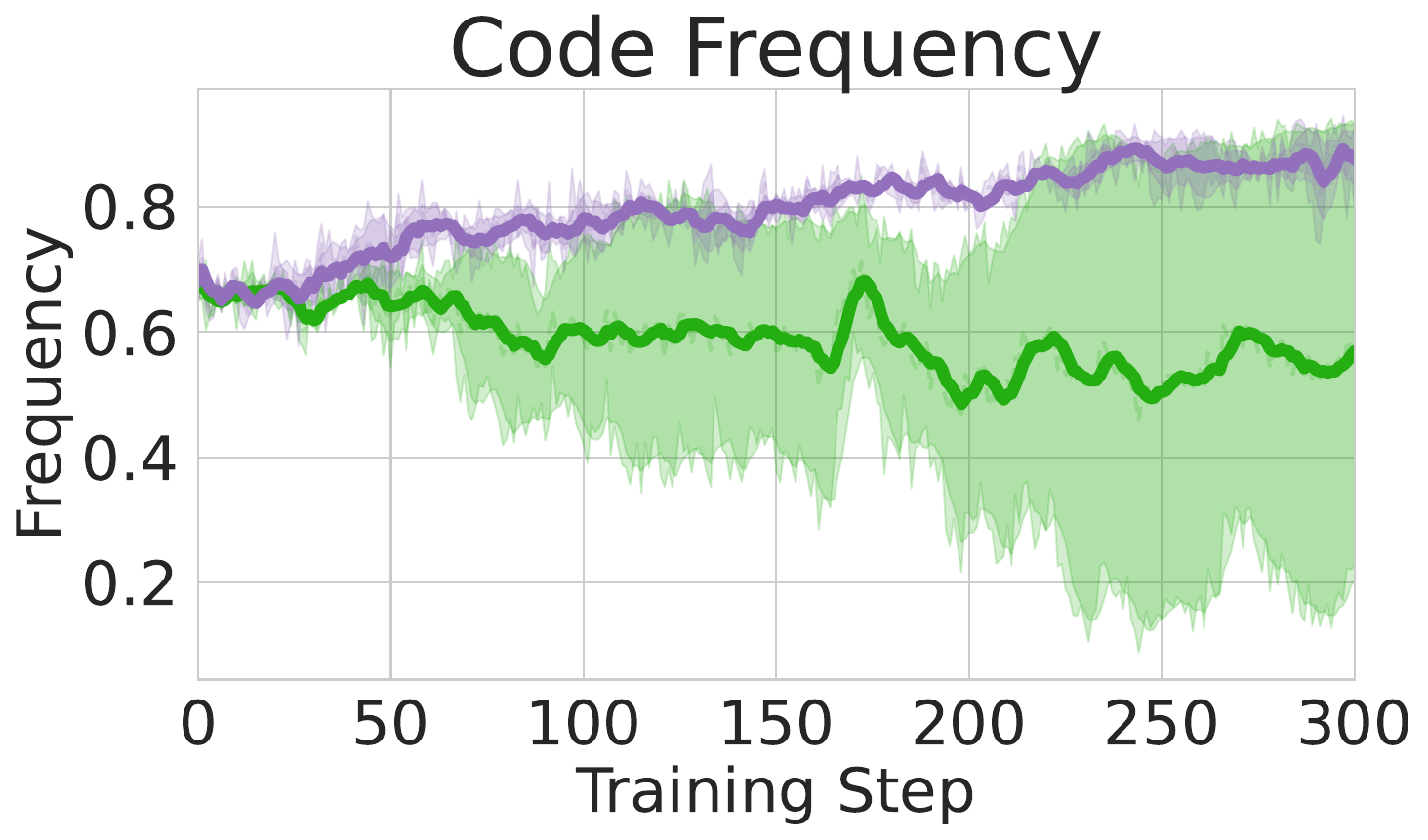}
    \caption{Code Frequency}
    \label{fig:noclip_action_probs_codefreq}
    \end{subfigure}
    
    \caption{Token probability 
    and code frequency of random rewards with (purple) and without (green) clipping on \qwenmath. 
    } 
    \label{fig:noclip_action_probs}
\end{figure*}

Figure~\ref{fig:noclip_action_probs} (a) shows that the average token probability mass increases for the standard loss of GRPO with clipping (purple line), but stays relatively constant when clipping is disabled. In addition, we log the frequency of code reasoning throughout the RLVR training in Figure~\ref{fig:noclip_action_probs} (b).
We hypothesize that the increase in average token probability correlates with the increase in code reasoning behavior, which, as we conjectured in Section~\ref{sec:analysis}, enhances the performance of the model, as we empirically show in Figure \ref{fig:analysis_code_freq} of Section \ref{sec:code_freq_acc}.

\paragraph{Why only Qwen-Math models benefit.} The clipping bias mechanism operates on all models, but its effectiveness depends on what behaviors get amplified. The bias systematically favors a model's pre-existing high-probability behaviors, but only benefits performance if those behaviors correlate with correctness.
For \qwenmathfamily models, code reasoning occurs in 65\% of responses and correlates strongly with correctness (64\% accuracy vs.~29\% without code). 
We observe that, when RLVR with clipping bias increases the model's code reasoning frequency to >90\%, performance improves substantially.
\emph{No-Code models} (Llama, OLMo2-7B) generate no code reasoning, so clipping bias has no beneficial pattern to amplify.
On the other hand, \emph{Bad-Code models} (OLMo2-7B-SFT) generate code 98\% of the time but with 21\% accuracy vs.~40\% for natural language, indicating that good performance does not rely on code reasoning.
Overall, the clipping bias is universal, but its impact on performance depends entirely on whether the model's dominant pre-existing behaviors happen to be effective reasoning strategies.

\paragraph{Implications.} Our findings reveal that GRPO's clipping mechanism provides a meaningful training signal even from pure noisy rewards, by systematically favoring the model's pre-existing behavioral patterns. This suggests that the apparent ``reward signal'' in random reward training is actually an artifact of the optimization algorithm's bias toward exploiting learned priors rather than exploring new behaviors.
This mechanism explains why random rewards work for Qwen2.5-Math models (which have strong code reasoning priors) but fail for other model families lacking these pre-existing capabilities. The training algorithm amplifies whatever reasoning patterns already correlate with correctness, regardless of the reward signal's actual informativeness.

\subsection{High Stochasticity in RLVR Without Clipping}\label{app:clip_random}

\begin{figure}
    \centering
     \begin{subfigure}[t]{\textwidth}
        \centering
        \includegraphics[width=0.245\linewidth]{figures/figure13/noclip_seeds_avg.pdf}
        \includegraphics[width=0.245\linewidth]{figures/figure13/batchbig_seeds_9_avg.pdf}
        \includegraphics[width=0.245\linewidth]{figures/figure13/batchsmall_seeds_9_avg.pdf}
        \includegraphics[width=0.245\linewidth]{figures/figure13/random_seeds_9_avg.pdf}
        \caption{Average results over multiple random seeds for each setting.
        }
        \label{fig:noclip_seed_math}
    \end{subfigure}

    \begin{subfigure}[t]{0.245\textwidth}
        \centering
        \includegraphics[width=\linewidth]{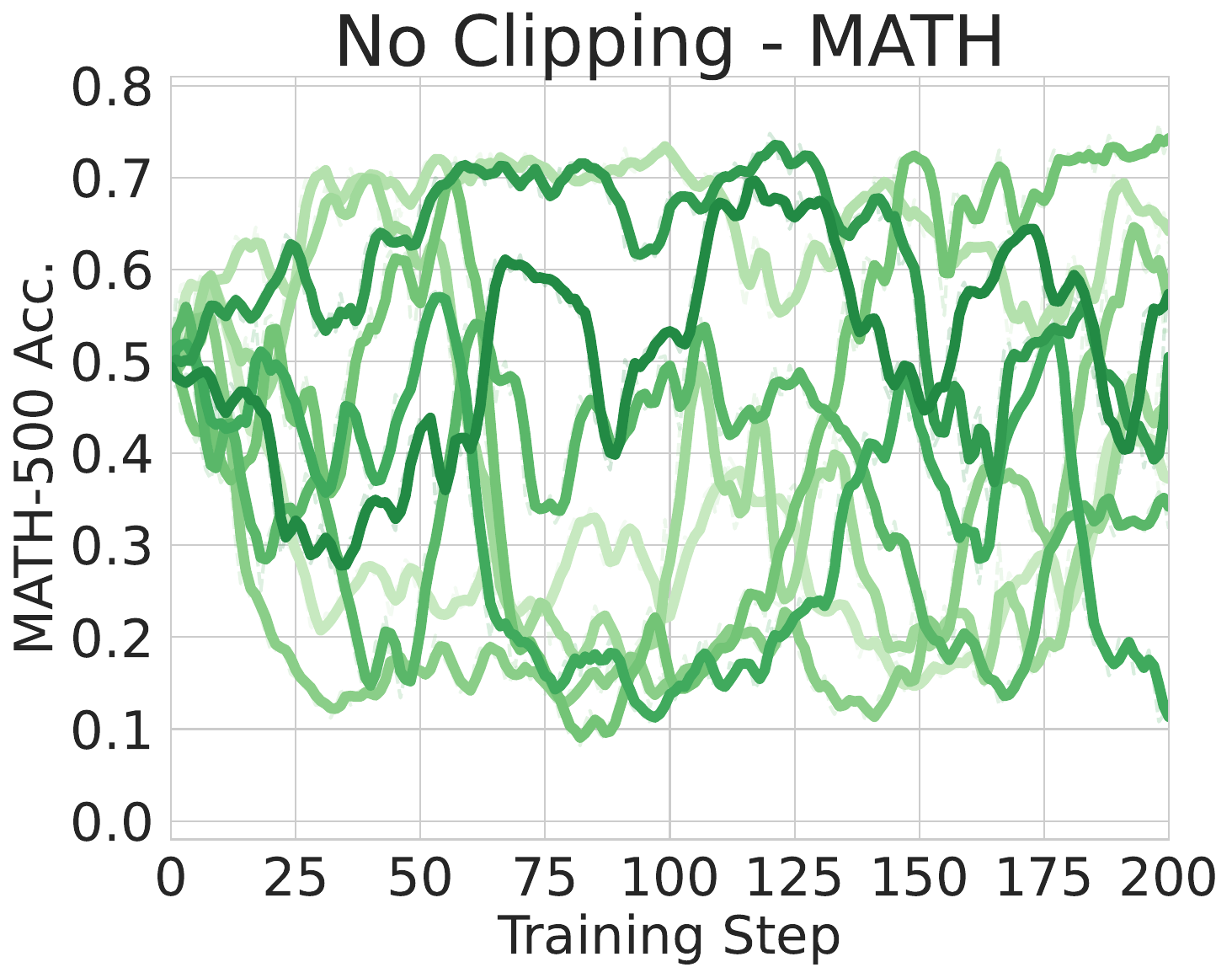}\\
        \includegraphics[width=\linewidth]{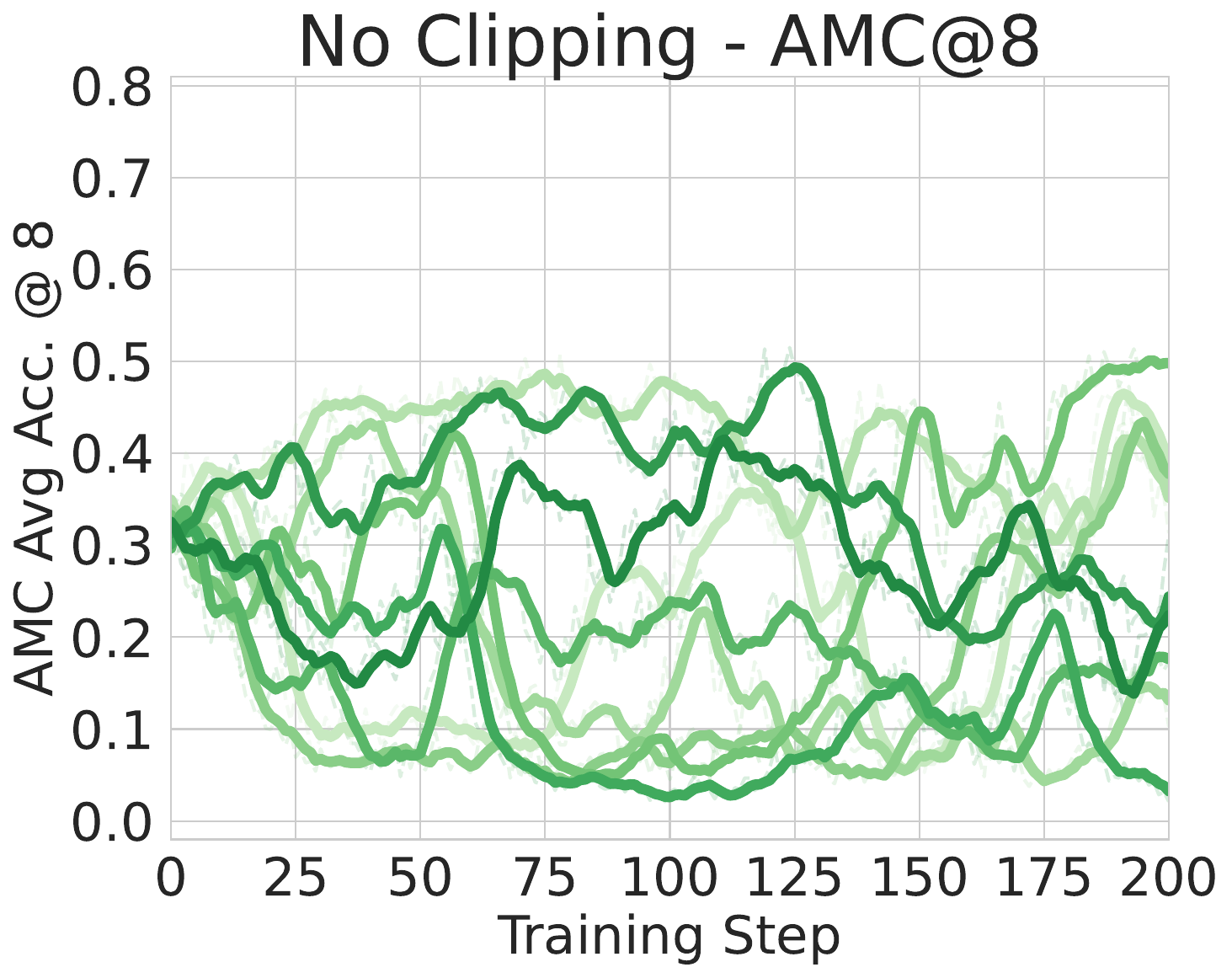}
        \caption{Turning off clipping with 8 gradient updates per rollout.}
        \label{fig:noclip_seed_math}
    \end{subfigure}%
    \hfill
    \begin{subfigure}[t]{0.245\textwidth}
        \centering
        \includegraphics[width=\linewidth]{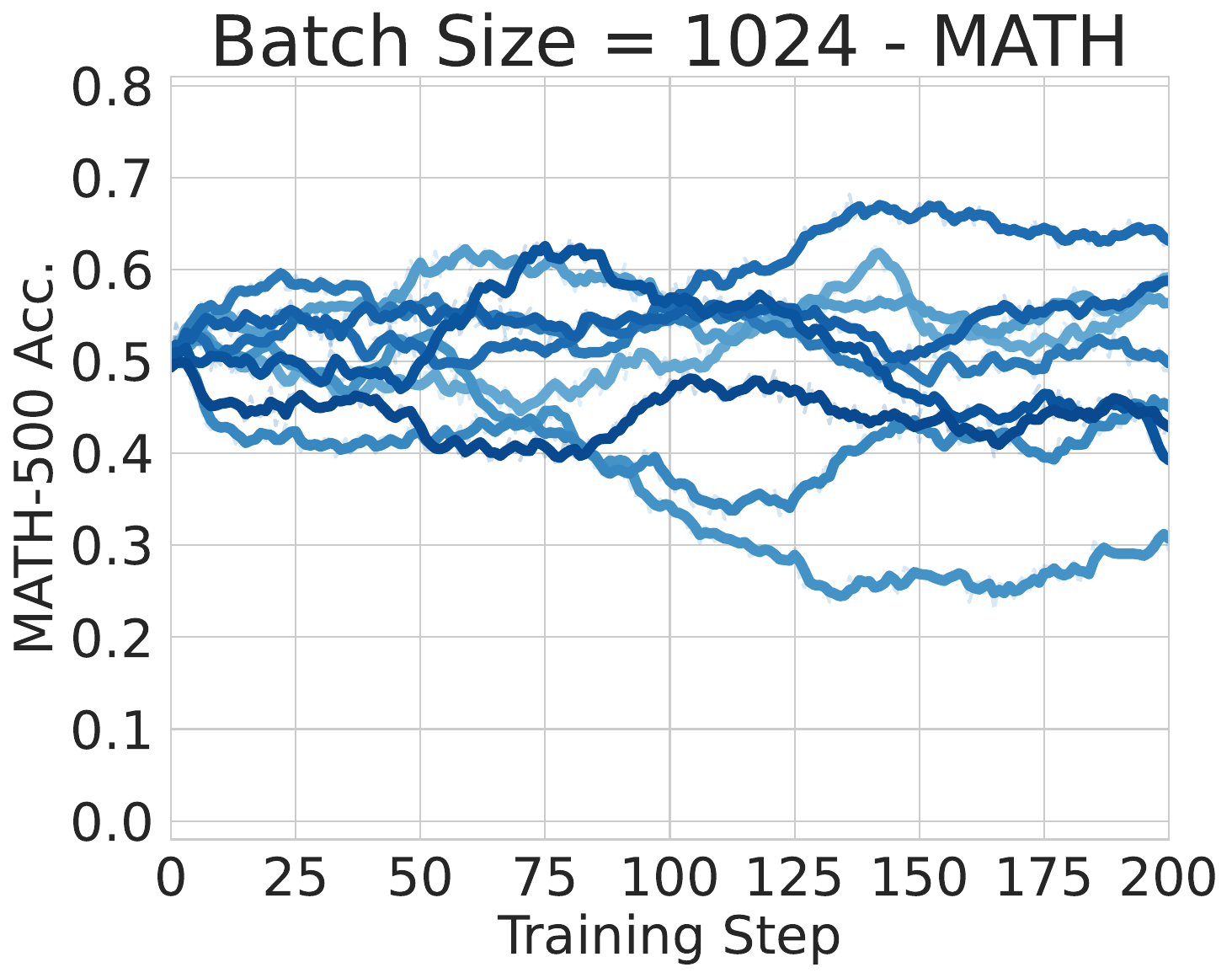}\\
        \includegraphics[width=\linewidth]{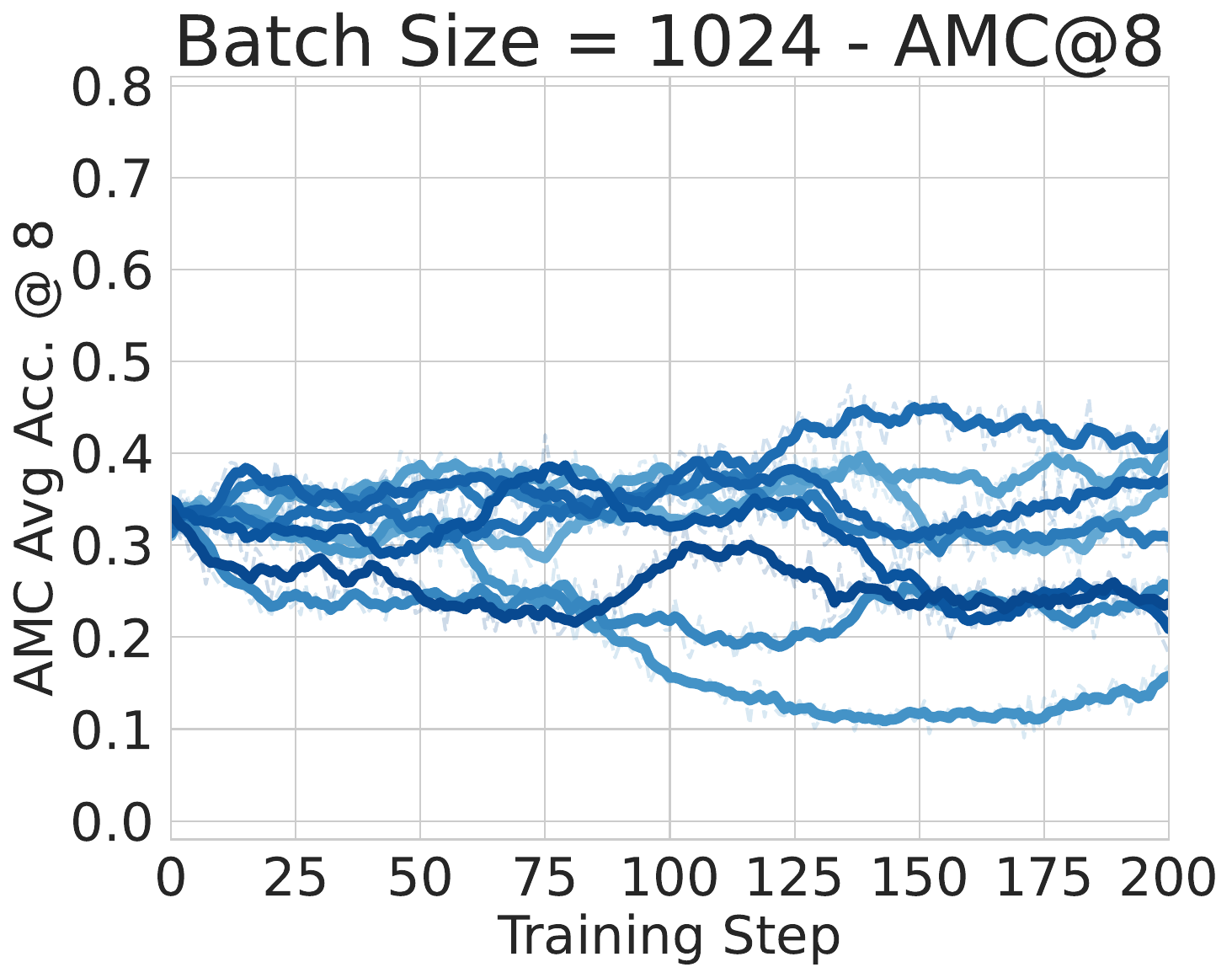}
        \caption{Increasing batch size to ensure only 1 gradient update per rollout.}
        \label{fig:noclip_seed_amc}
    \end{subfigure}%
    \hfill
    \begin{subfigure}[t]{0.245\textwidth}
        \centering
        \includegraphics[width=\linewidth]{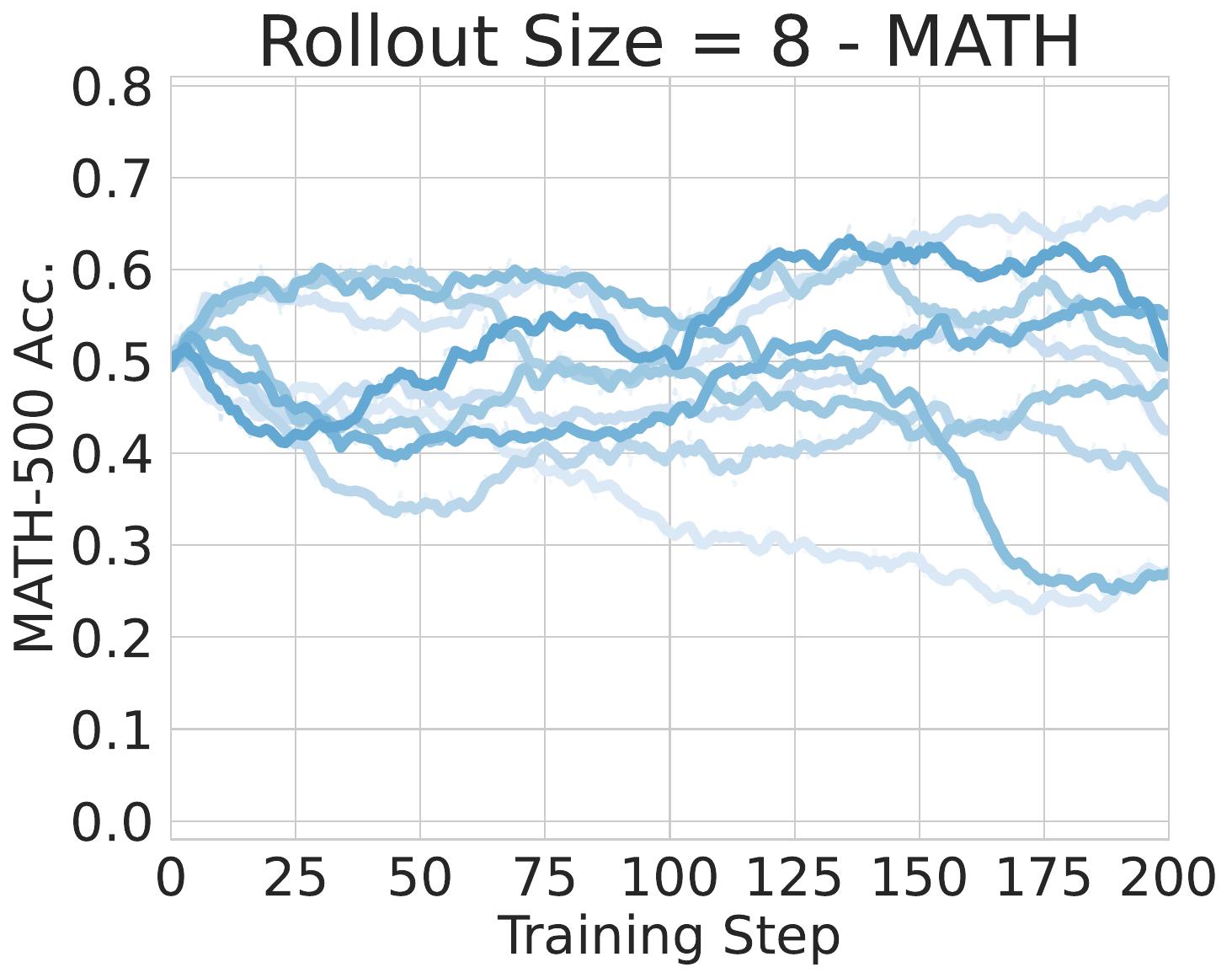}\\
        \includegraphics[width=\linewidth]{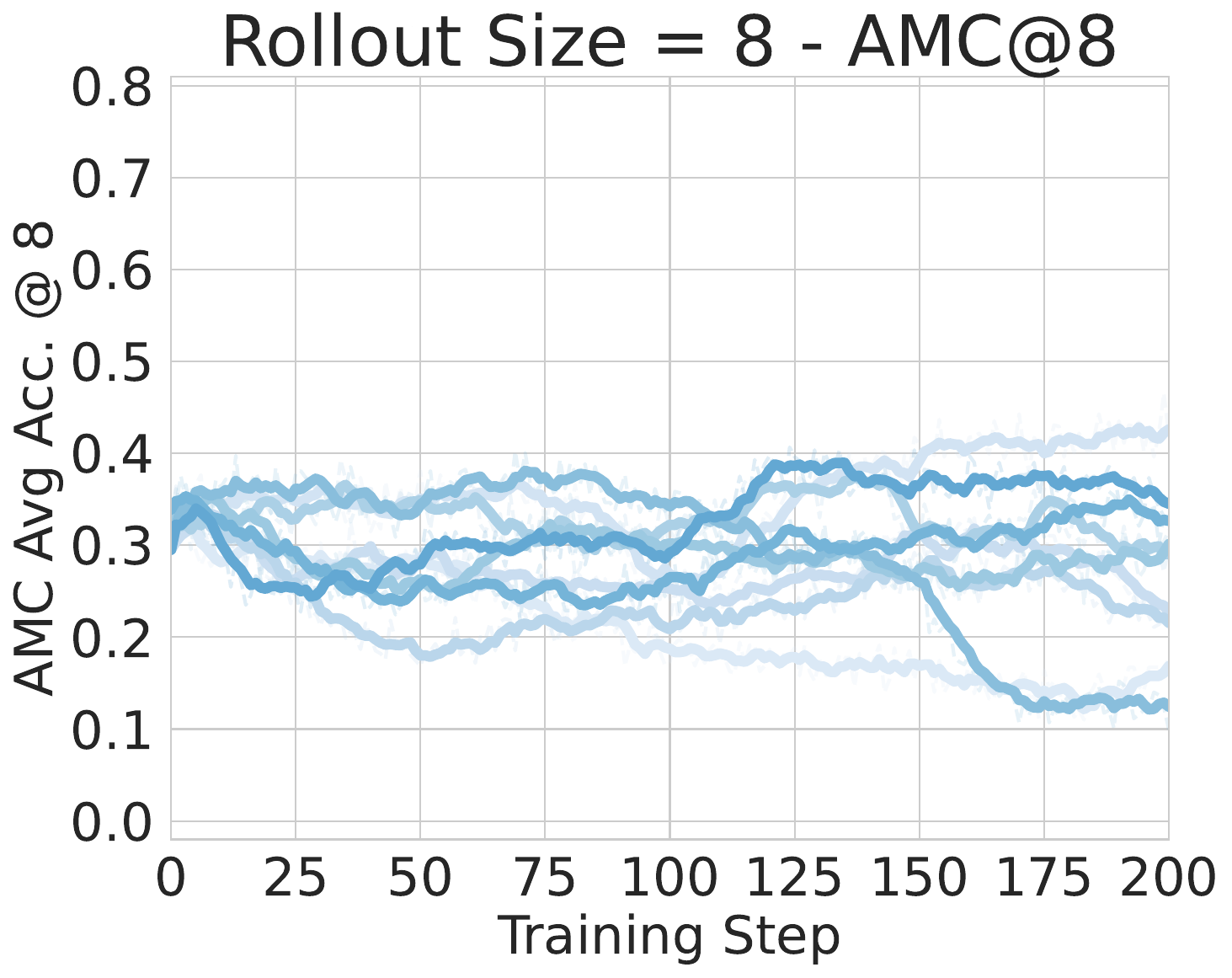}
        \caption{Decreasing rollout batch size to ensure only 1 gradient update per rollout.}
        \label{fig:noclip_seed_math}
    \end{subfigure}%
    \hfill
    \begin{subfigure}[t]{0.245\textwidth}
        \centering
        \includegraphics[width=\linewidth]{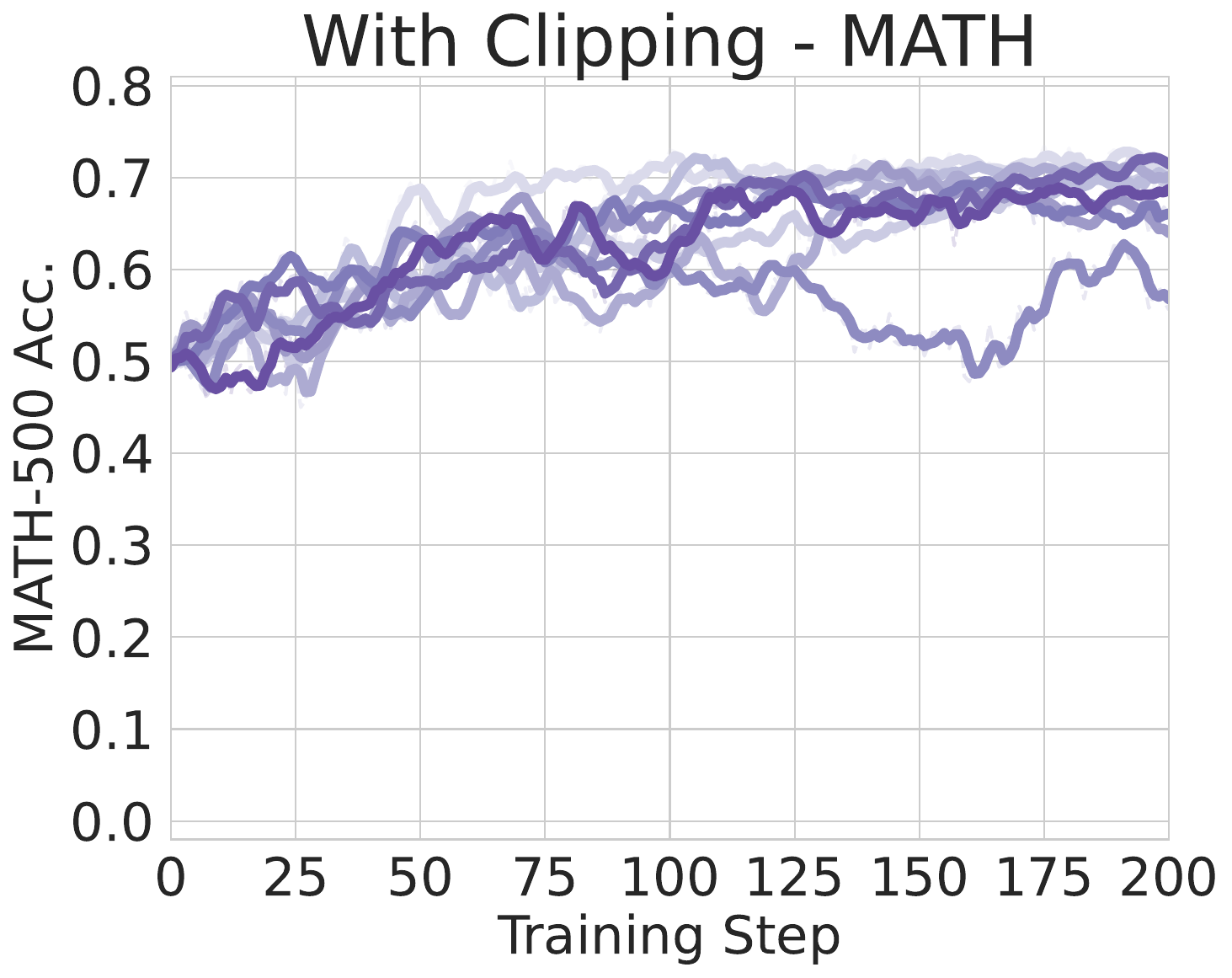}\\
        \includegraphics[width=\linewidth]{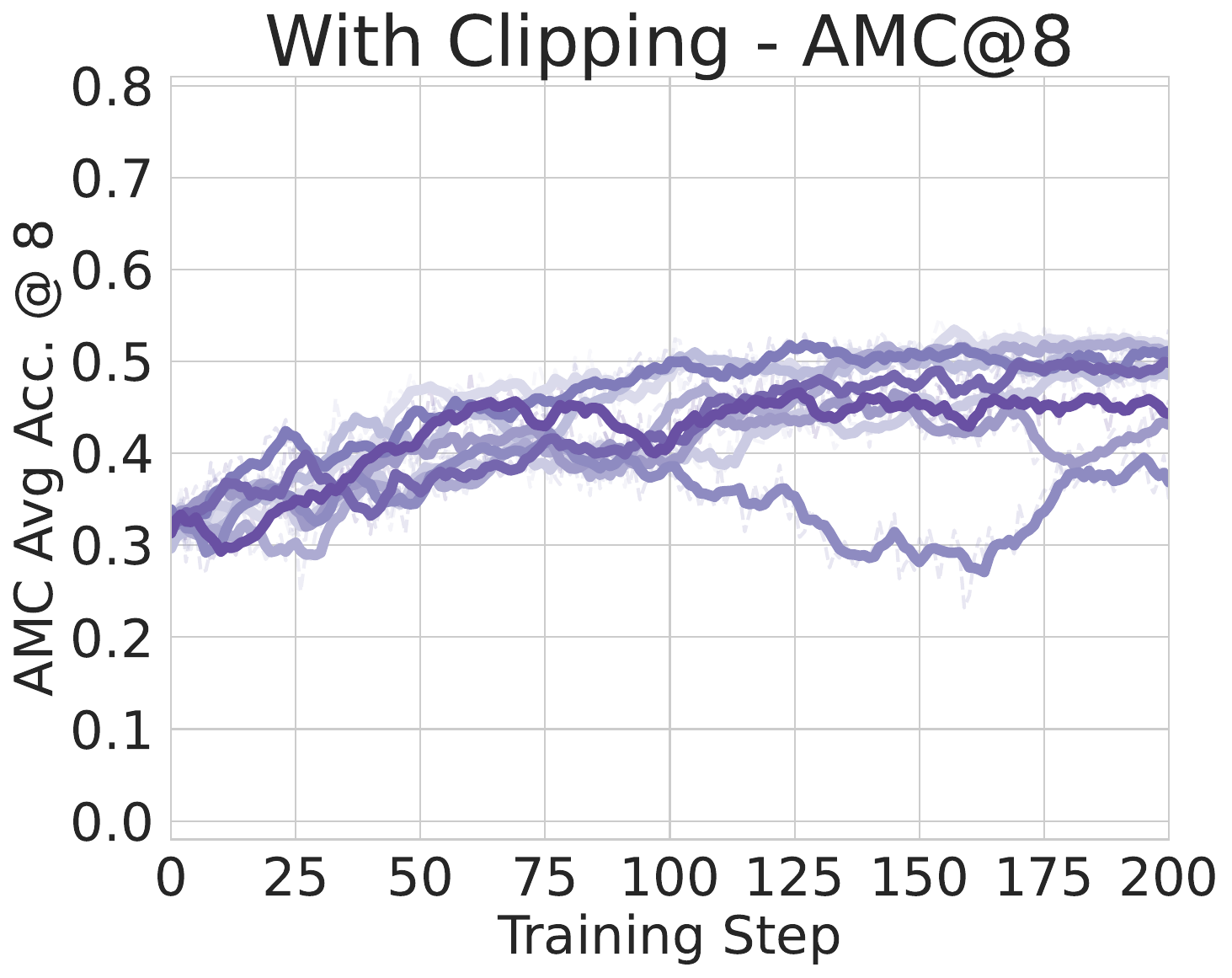}
        \caption{Default GRPO loss with random reward.}
        \label{fig:noclip_seed_amc}
    \end{subfigure}%
    \caption{RLVR performance on random rewards with disabled clipping across multiple different seeds. Models without clipping show no meaningful performance improvement on average, while models with clipping demonstrate consistent performance gains using random rewards. Disabling clipping in implementation produces high variance in results, occasionally yielding high accuracy scores. Experiments with adjusted batch sizes exhibit greater stability but involve 8 times fewer gradient updates than the disabled clipping experiments.}
    \label{fig:noclip_seed}
\end{figure}

In this section, we empirically show that training without clipping has stochastic training dynamics and can sometimes obtain performance improvement despite the expected gradient being zero.
Following the experimental setup in Section~\ref{sec:random_reward}, we disable the clipping effect through the following interventions: 
(i) removing clipping from the loss implementation, (ii) increasing mini-batch size to match rollout size, and (iii) decreasing the rollout size to match mini-batch size across multiple random seeds. We report the results in Figure~\ref{fig:noclip_seed} across different random seeds.

When clipping is disabled by adjusting batch size, model performance remains stable within a consistent range without converging toward either extreme of the accuracy spectrum. These runs involve 8 times fewer gradient updates due to the batch configuration.

In contrast, removing clipping from the implementation produces extreme stochasticity, occasionally resulting in high-performance convergence across multiple runs.
While the exact mechanism remains unclear, we hypothesize that inherent randomness in RLVR training contributes to these observations. The group size $G=16$ used in GRPO training creates high variance in gradient updates. Combined with the instability of unclipped training, this variance may accidentally reinforce certain reasoning patterns, leading to improved performance in some runs.
We leave a systematic analysis of this behavior to future work.

\section{Additional Results on AMC}\label{app:amc_results}

\begin{figure*}[t!]
    \centering
    \cblock{94}{149}{78} Ground Truth
    \cblock{70}{129}{188} Majority Vote
    \cblock{221}{162}{88} Format
    \cblock{211}{98}{83} Incorrect
    \cblock{147}{112}{188} Random
    \begin{subfigure}[t]{0.245\textwidth}
        \centering
        \includegraphics[width=\linewidth]{figures/figure3/rlvr_qwen_1.5b_MATH.pdf}\\
        \includegraphics[width=\linewidth]{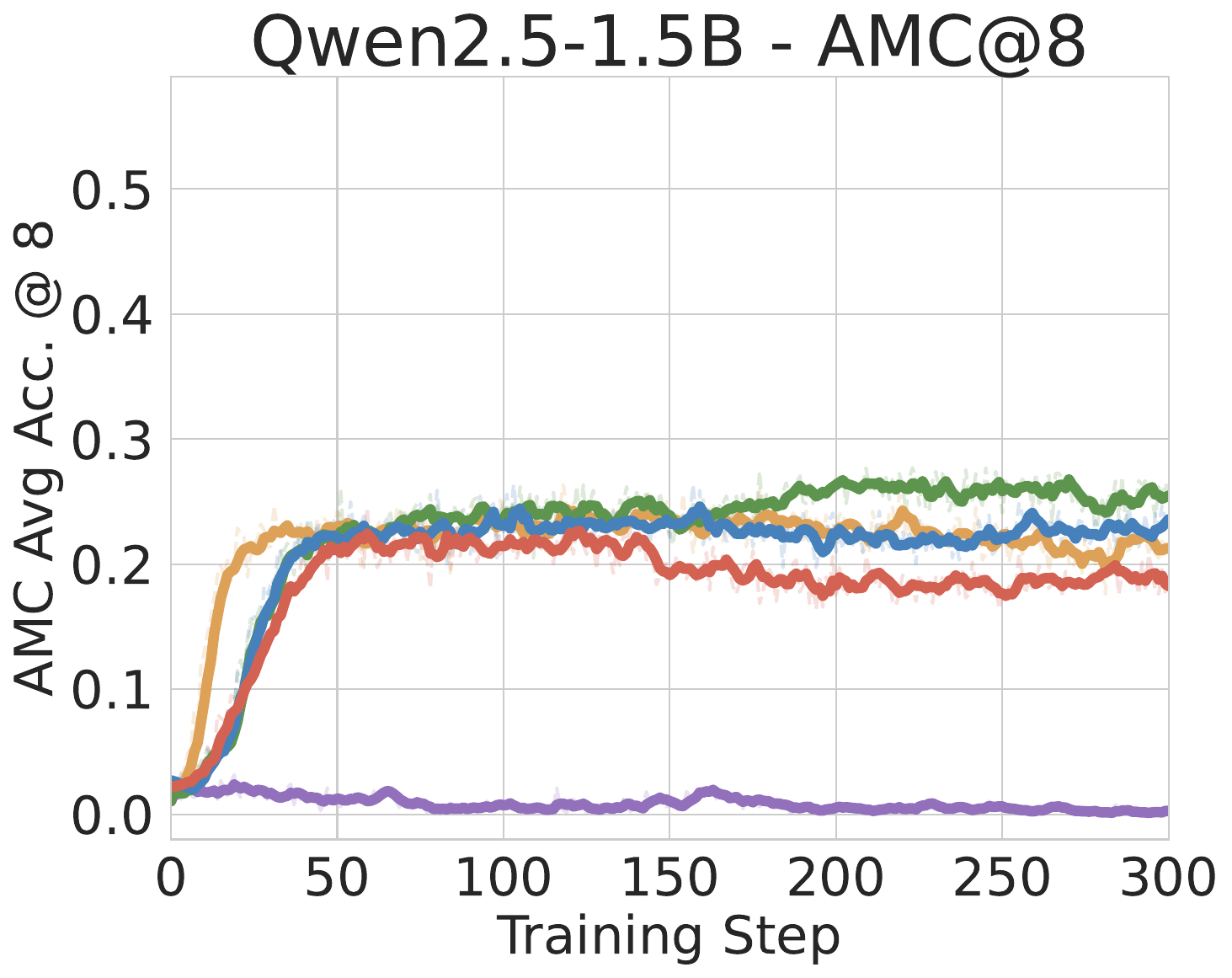}
        \caption{Qwen2.5-1.5B}
        \label{fig:qwen1.5b_results}
    \end{subfigure}%
    ~
    \begin{subfigure}[t]{0.245\textwidth}
        \centering
        \includegraphics[width=\linewidth]{figures/figure3/rlvr_qwen_7b_MATH.pdf}
        \includegraphics[width=\linewidth]{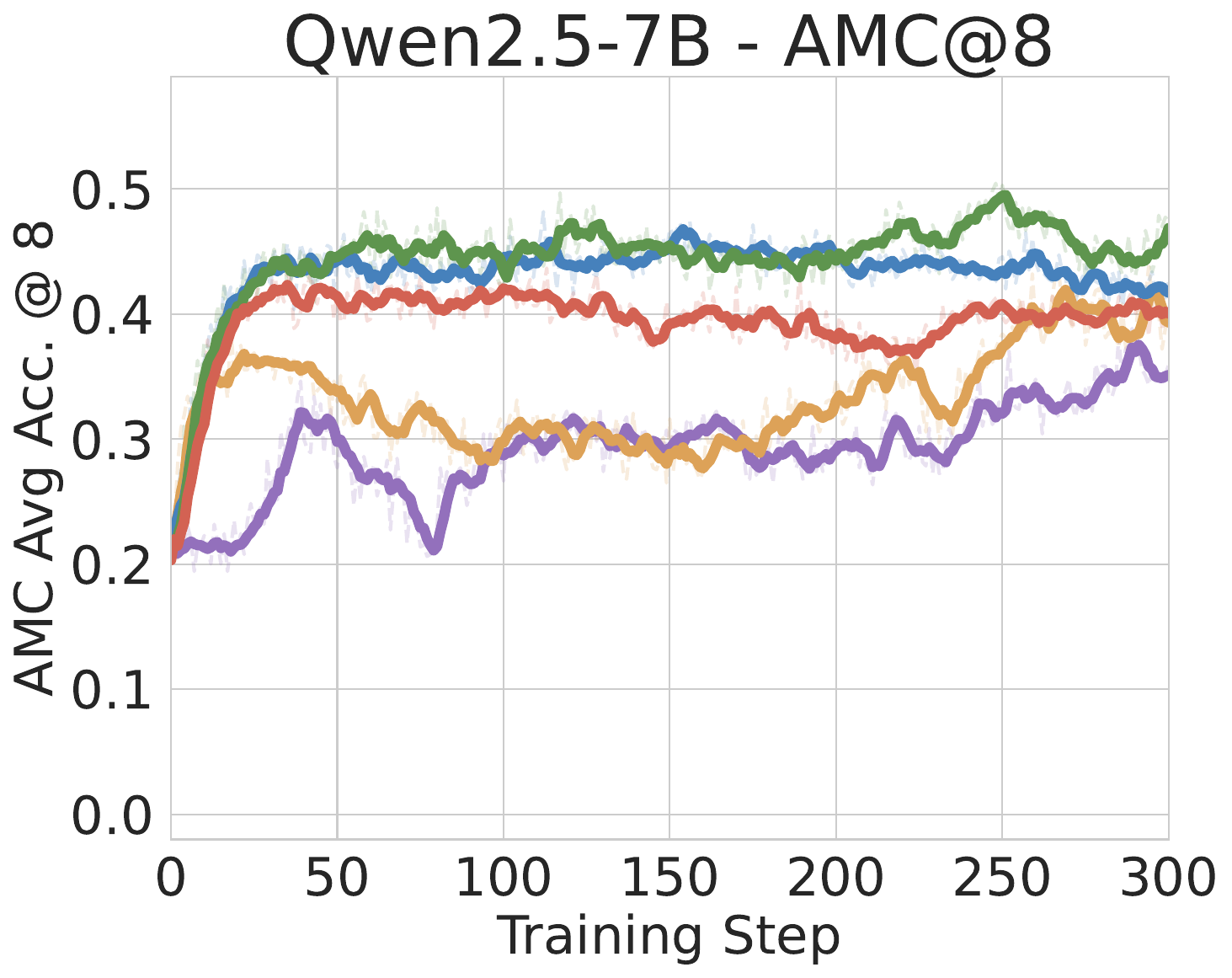}
        \caption{Qwen2.5-7B}
        \label{fig:qwen_results}
    \end{subfigure}%
    ~
    \begin{subfigure}[t]{0.245\textwidth}
        \centering
        \includegraphics[width=\linewidth]{figures/figure3/rlvr_olmo_base_MATH.pdf}
        \includegraphics[width=\linewidth]{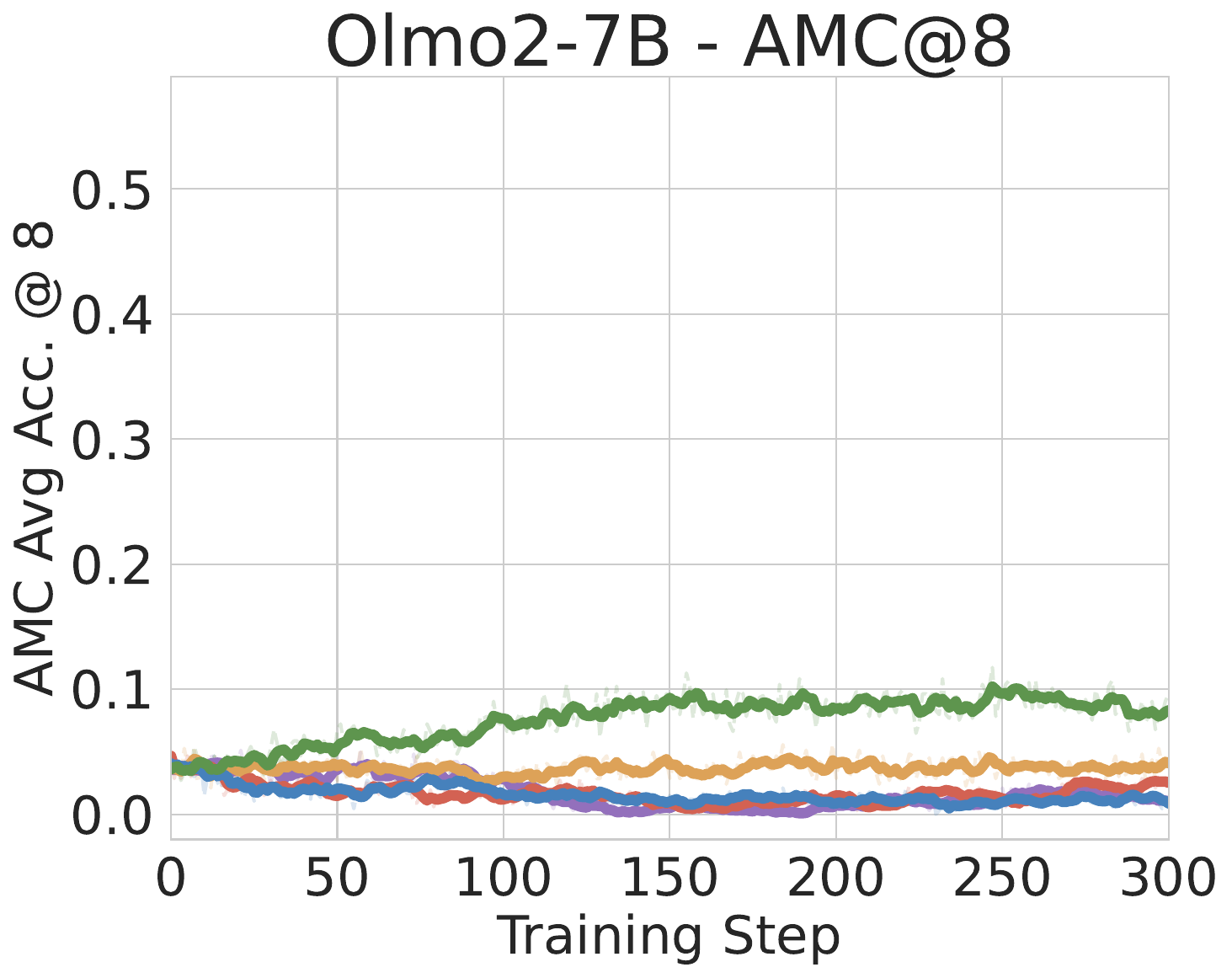}
    \caption{Olmo-2-1124-7B}
    \label{fig:olmo_results}
    \end{subfigure}%
    ~ 
    \begin{subfigure}[t]{0.245\textwidth}
        \centering
        \includegraphics[width=\linewidth]{figures/figure3/rlvr_olmo_sft_MATH.pdf}
        \includegraphics[width=\linewidth]{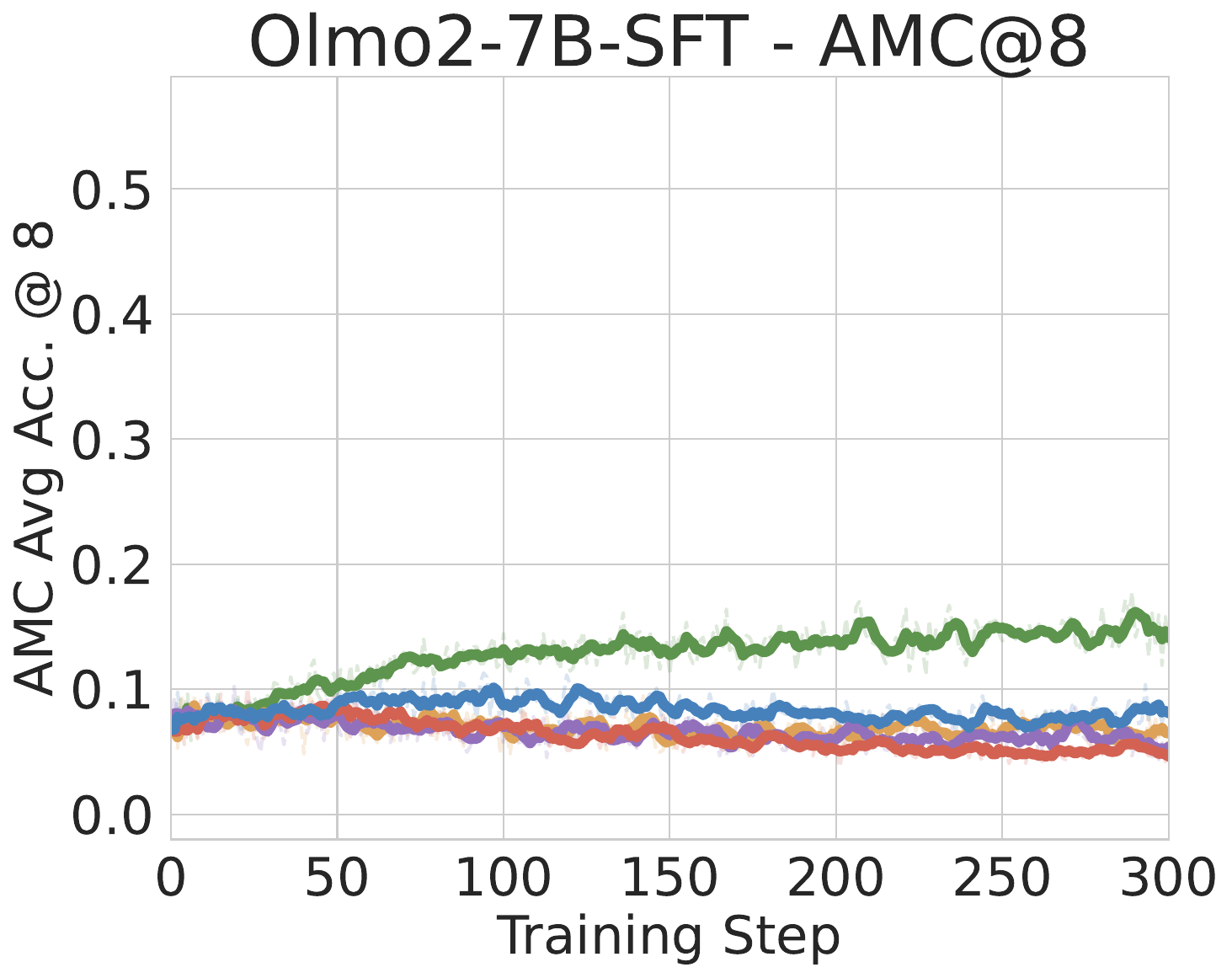}
    \caption{Olmo-2-1124-7B-SFT}
    \label{fig:olmo_sft_results}
    \end{subfigure}

    \begin{subfigure}[t]{0.245\textwidth}
        \centering
        \includegraphics[width=\linewidth]{figures/figure3/rlvr_llama_3b_base_MATH.pdf}
        \includegraphics[width=\linewidth]{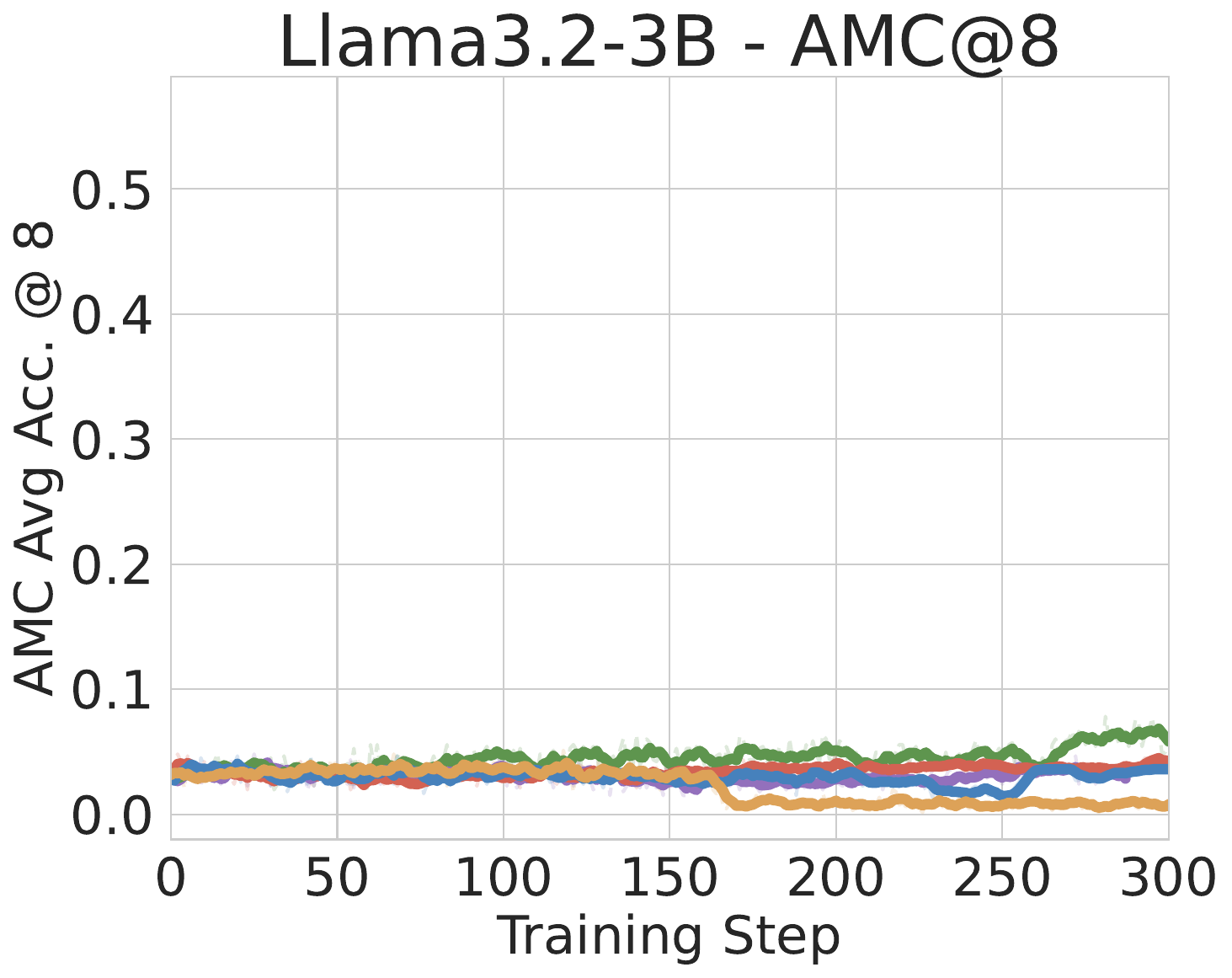}
    \caption{Llama3.2-3B}
    \label{fig:llama3_base_results}
    \end{subfigure}%
    ~
    \begin{subfigure}[t]{0.245\textwidth}
        \centering
        \includegraphics[width=\linewidth]{figures/figure3/rlvr_llama_8b_base_MATH.pdf}
        \includegraphics[width=\linewidth]{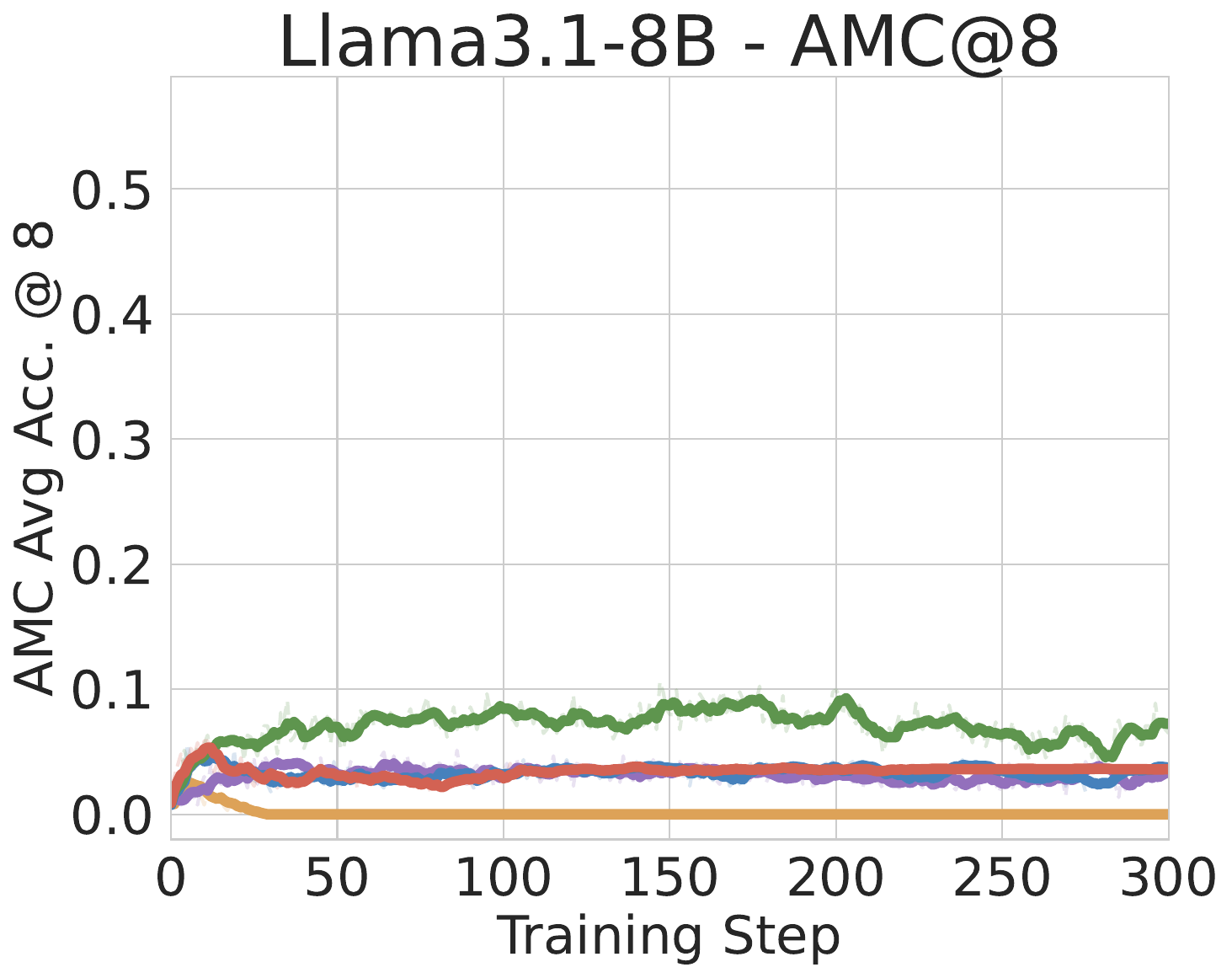}
    \caption{Llama3.1-8B}
    \label{fig:llama3_base_results}
    \end{subfigure}%
    ~
    \begin{subfigure}[t]{0.245\textwidth}
        \centering
        \includegraphics[width=\linewidth]{figures/figure3/rlvr_llama_3b_instruct_MATH.pdf}
        \includegraphics[width=\linewidth]{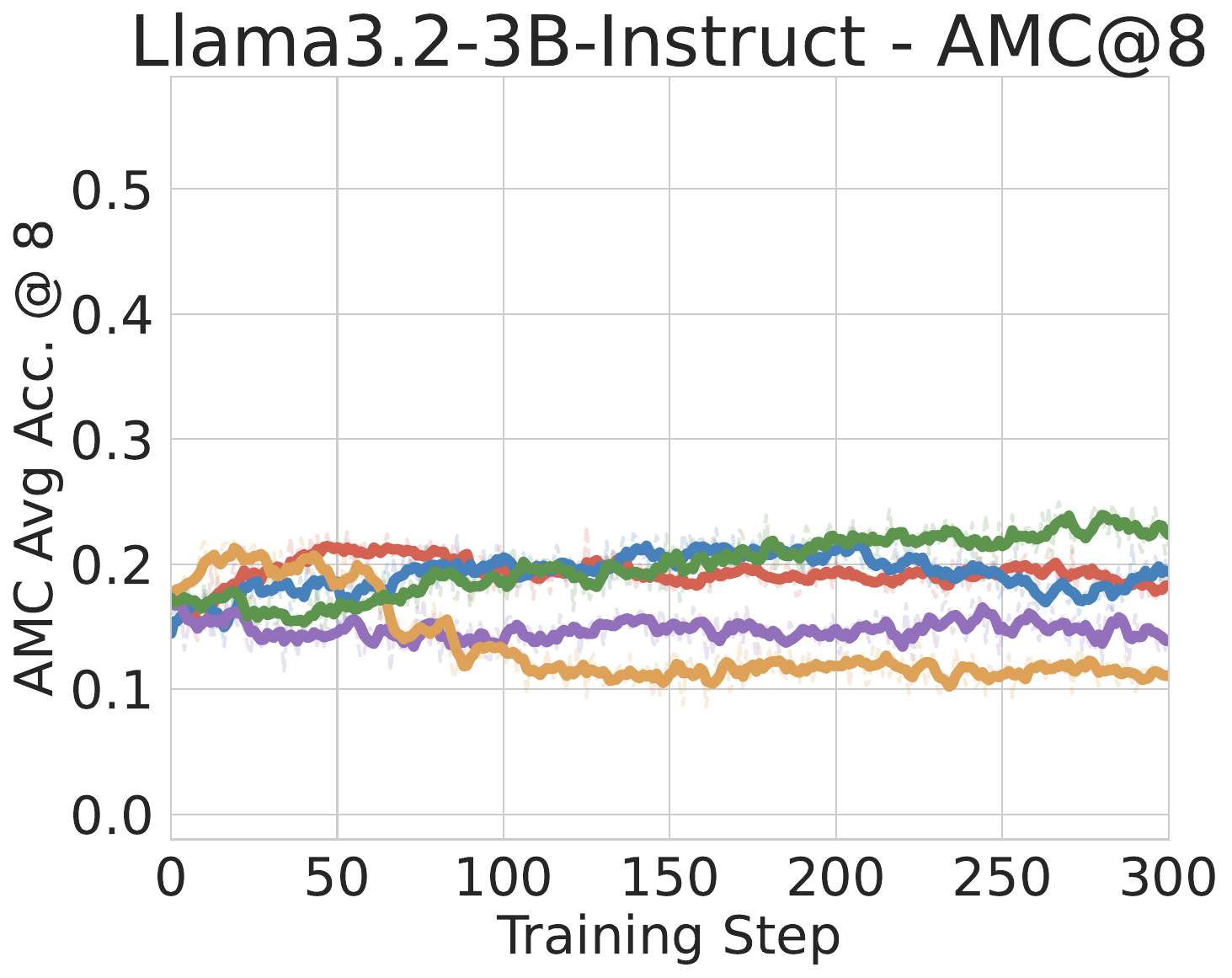}
    \caption{Llama3.2-3B-Instruct}
    \label{fig:llama3_results}
    \end{subfigure}%
    ~
    \begin{subfigure}[t]{0.245\textwidth}
        \centering
        \includegraphics[width=\linewidth]{figures/figure3/rlvr_llama_8b_instruct_MATH.pdf}
        \includegraphics[width=\linewidth]{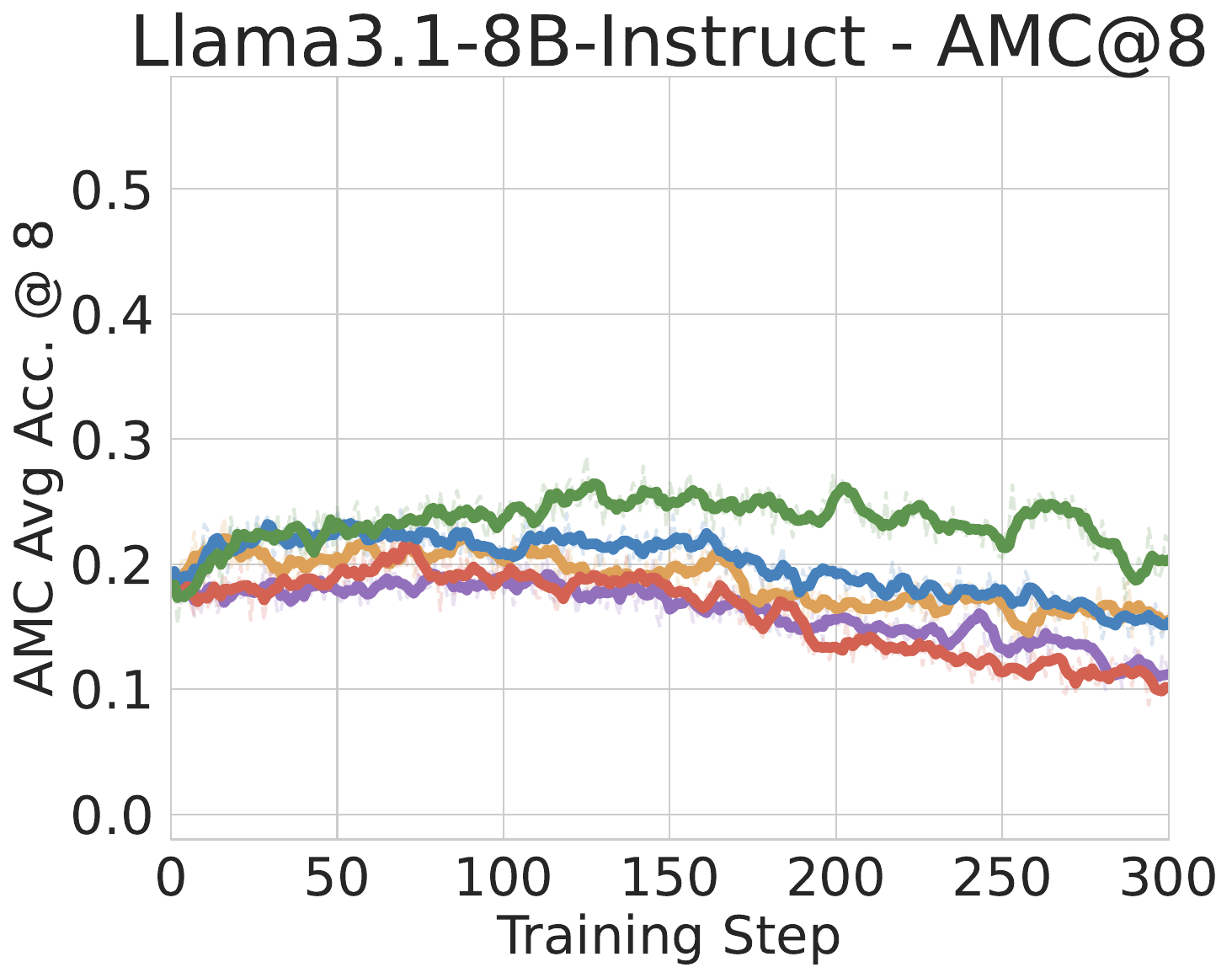}
    \caption{Llama3.1-8B-Instruct}
    \label{fig:llama8_results}
    \end{subfigure}
    
    \caption{Varying rewards across additional model classes. Spurious rewards remain effective on general-purpose \qwenbasefamily models but generally fail to yield any gains on other model families. The performance improvements on non-\qwenbasefamily models are substantially smaller compared to those observed in the \qwenbasefamily family. 
    } 
    \label{fig:app_reward_other_family}
\end{figure*}

We supplement additional AMC results for non-Qwen models in Figure~\ref{fig:app_reward_other_family}. The trends are consistent with the MATH-500 results that are shown in Section~\ref{sec:others_dont}.

\section{Additional Results on AIME 2024 and AIME 2025}\label{app:aime_results}
\begin{figure*}[h]
    \centering
    \cblock{94}{149}{78} Ground Truth
    \cblock{70}{129}{188} Majority Vote
    \cblock{211}{98}{83} Incorrect
    \cblock{221}{162}{88} Format
    \cblock{147}{112}{188} Random
    \begin{subfigure}[t]{0.49\textwidth}
        \centering
        \includegraphics[width=0.49\linewidth]{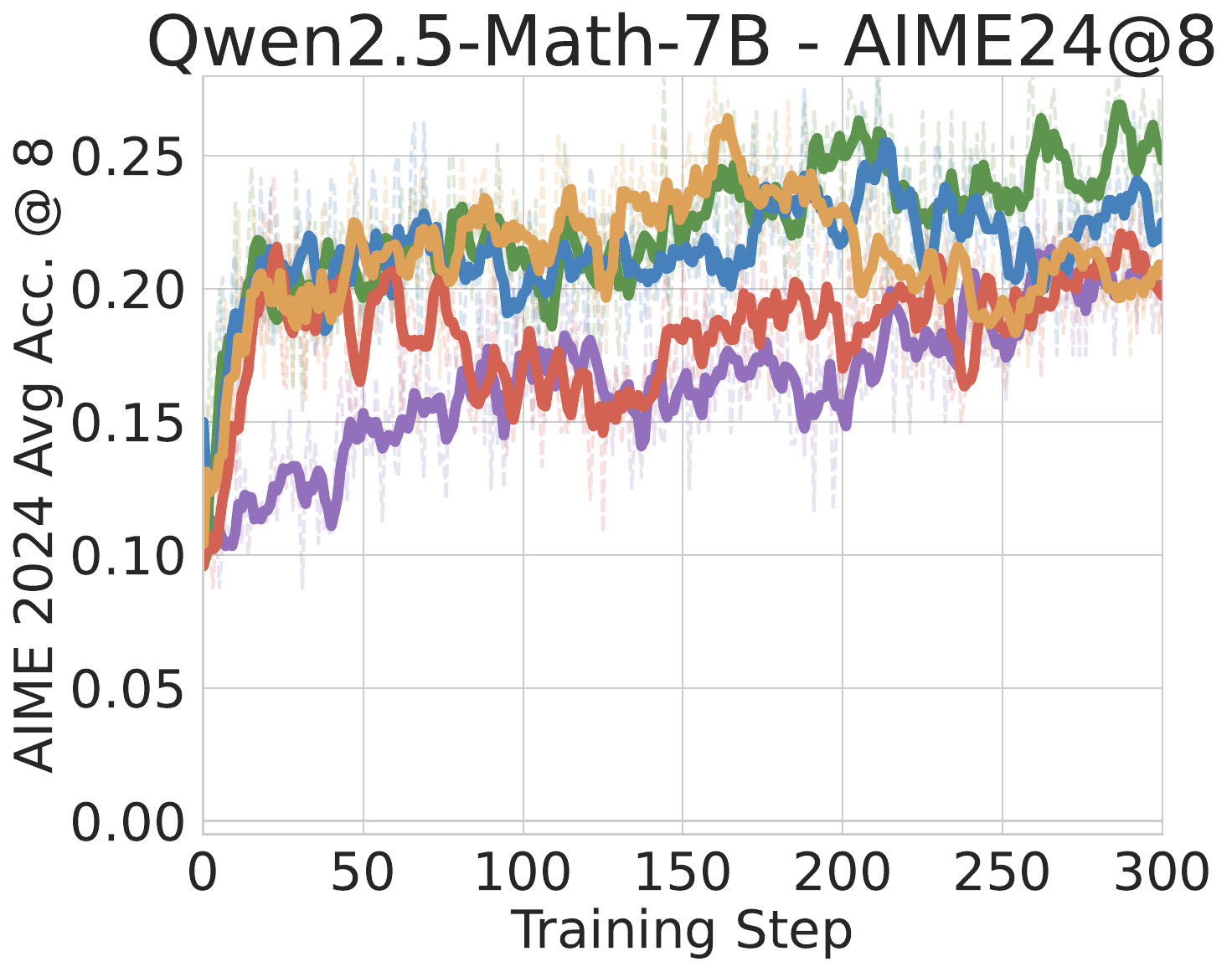}
        \includegraphics[width=0.49\linewidth]{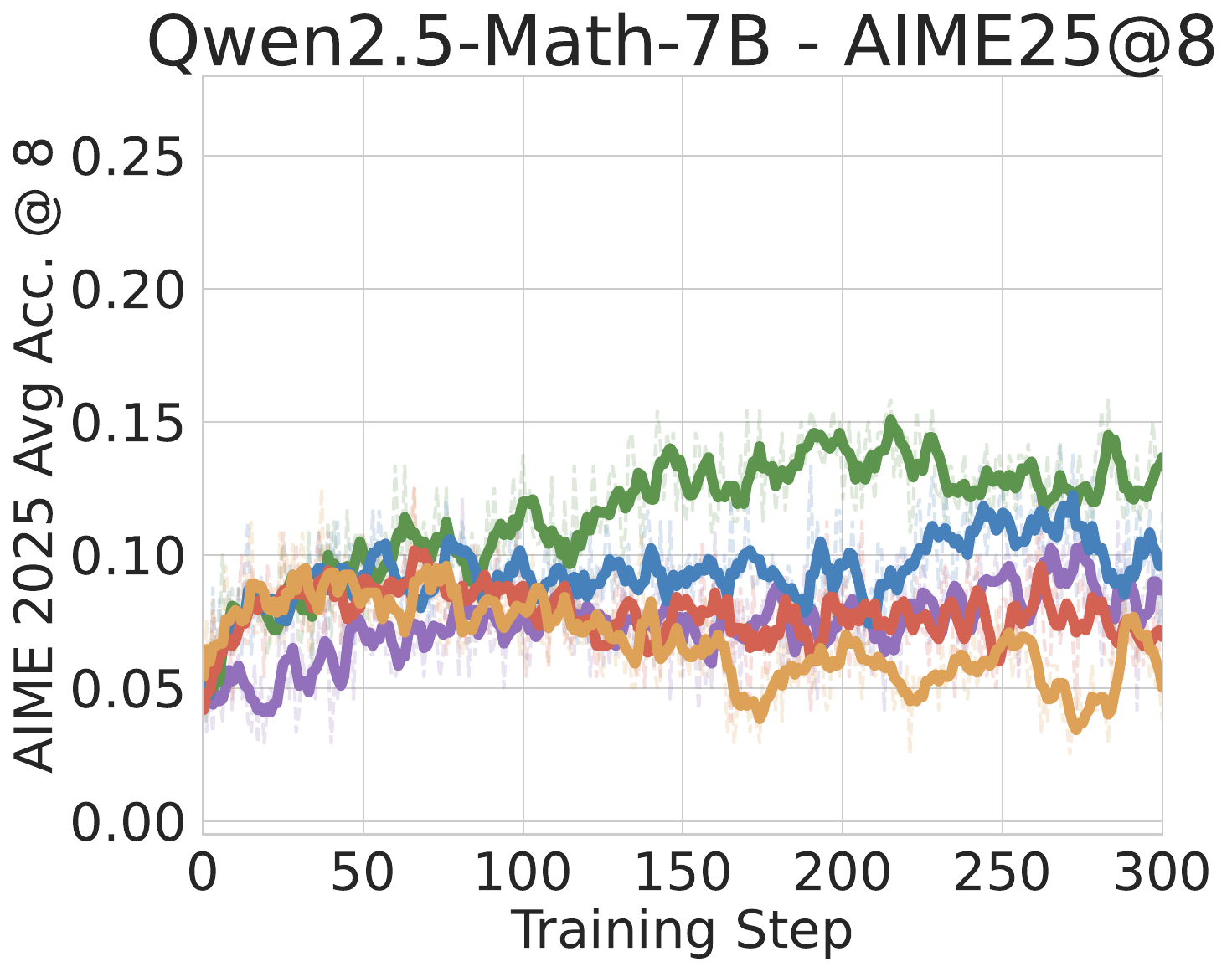}\vspace{-2mm}
        \caption{\qwenmath}
        \label{fig:qwen_7b_results_aime}
    \end{subfigure}%
    ~
    \begin{subfigure}[t]{0.49\textwidth}
        \centering
        \includegraphics[width=0.49\linewidth]{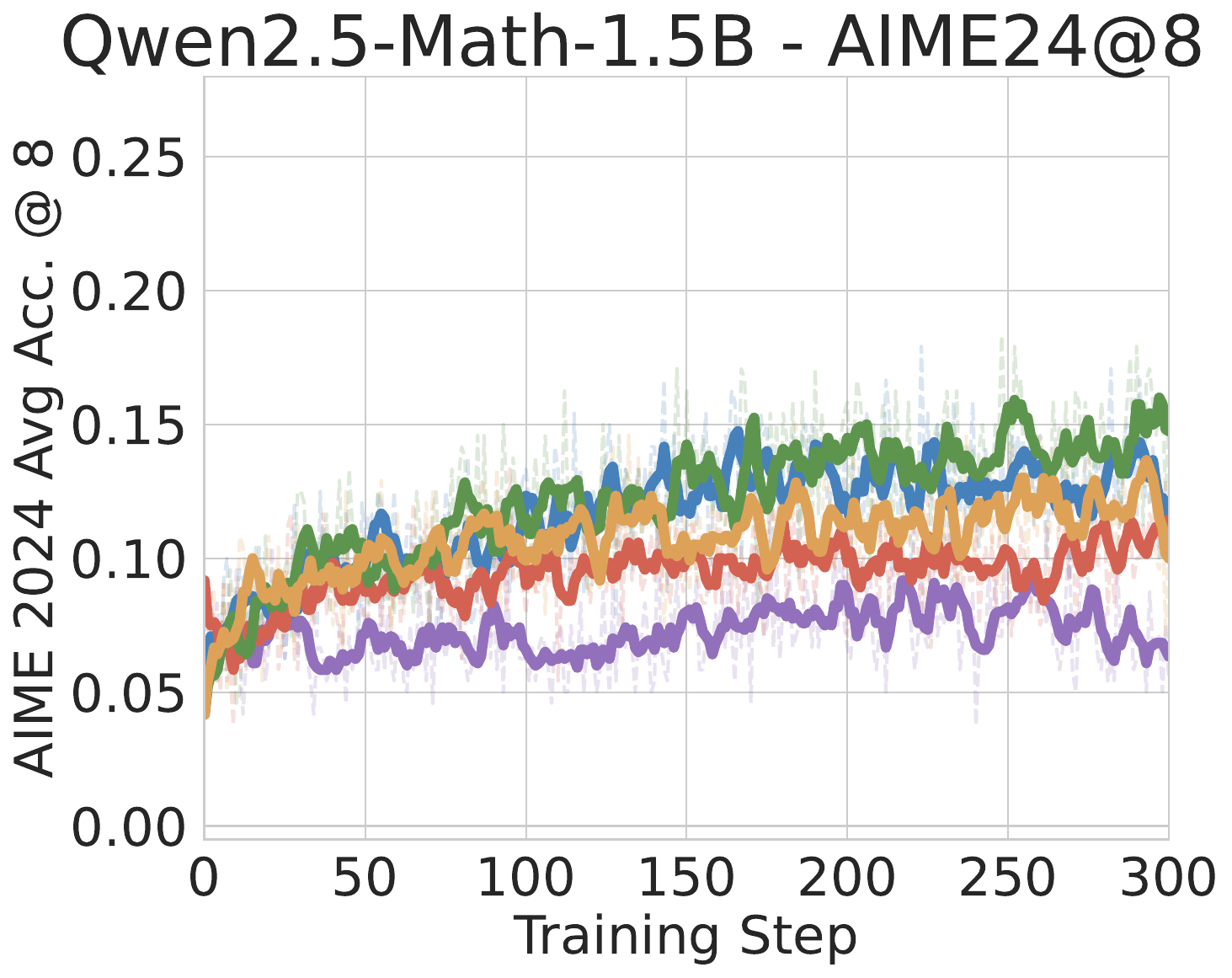}
        \includegraphics[width=0.49\linewidth]{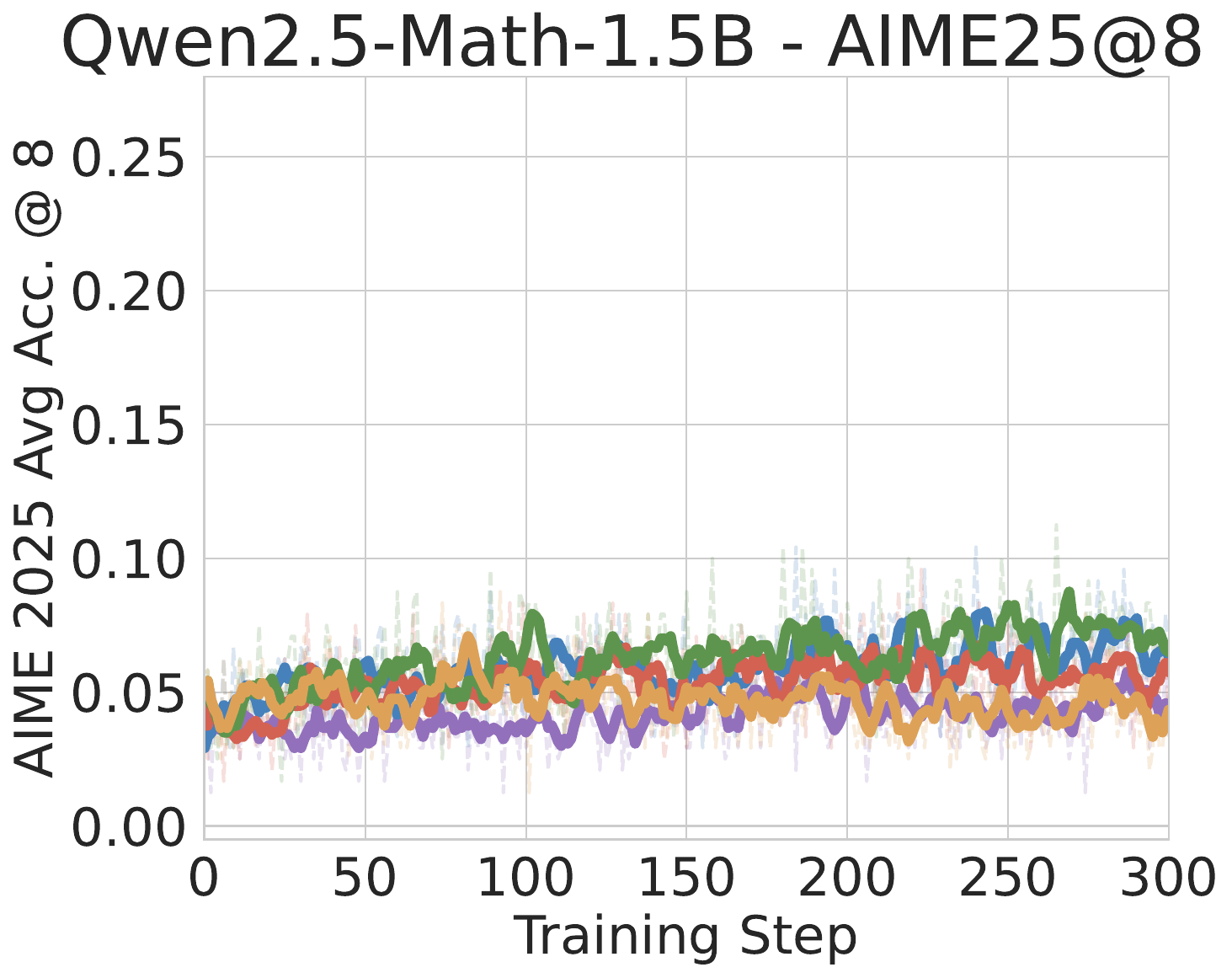}\vspace{-2mm}
        \caption{\qwenmathsmall}
        \label{fig:qwen_1b_results_aime}
    \end{subfigure}\vspace{-2mm}
    \caption{\qwenmathfamily Model performance on AIME 2024 and AIME 2025.
    }\label{fig:qwen_math_results_aime}
\end{figure*}

\begin{figure*}[t!]
    \centering
    \cblock{94}{149}{78} Ground Truth
    \cblock{70}{129}{188} Majority Vote
    \cblock{211}{98}{83} Incorrect
    \cblock{221}{162}{88} Format
    \cblock{147}{112}{188} Random
    \begin{subfigure}[t]{0.245\textwidth}
        \centering
        \includegraphics[width=\linewidth]{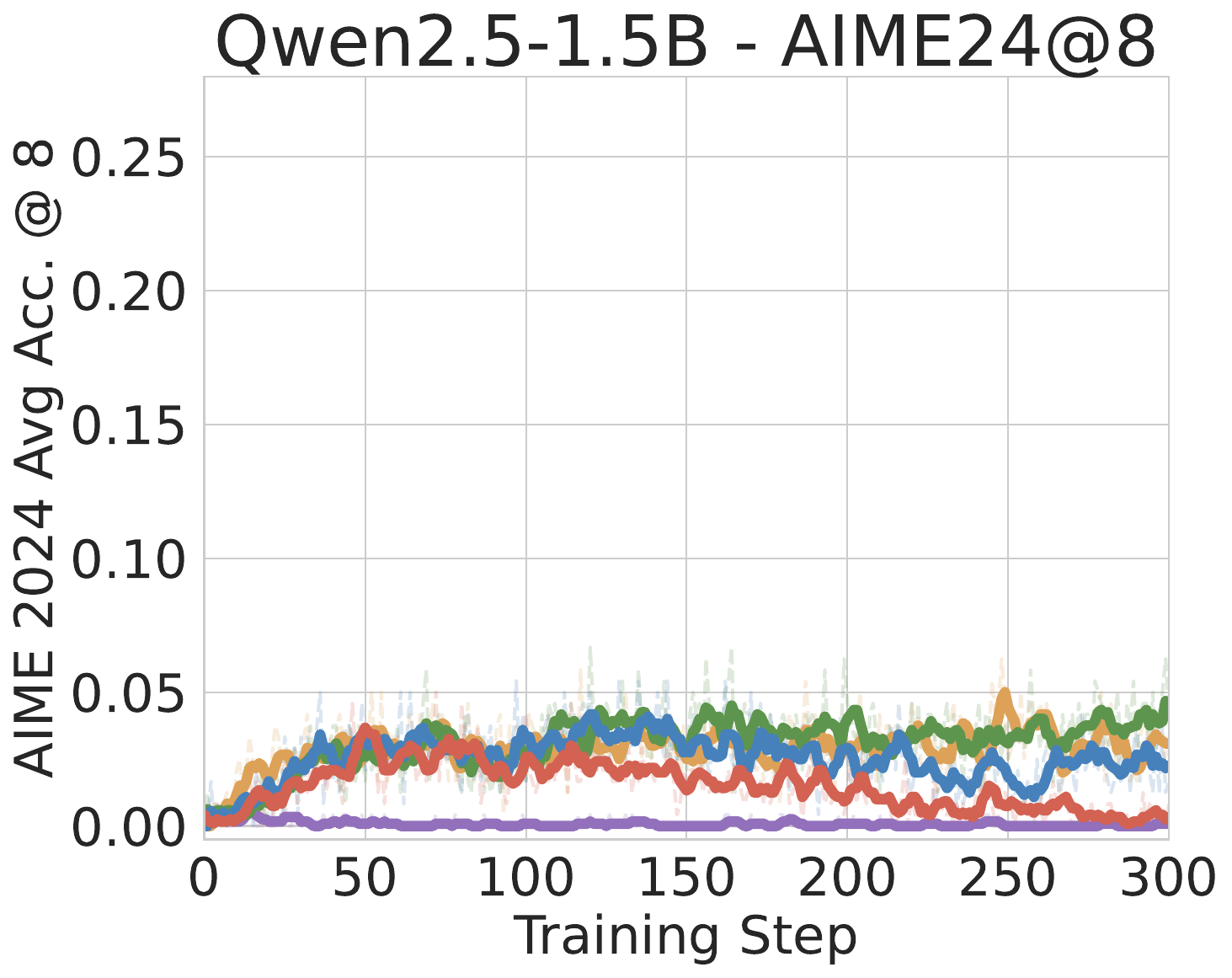}\\
        \includegraphics[width=\linewidth]{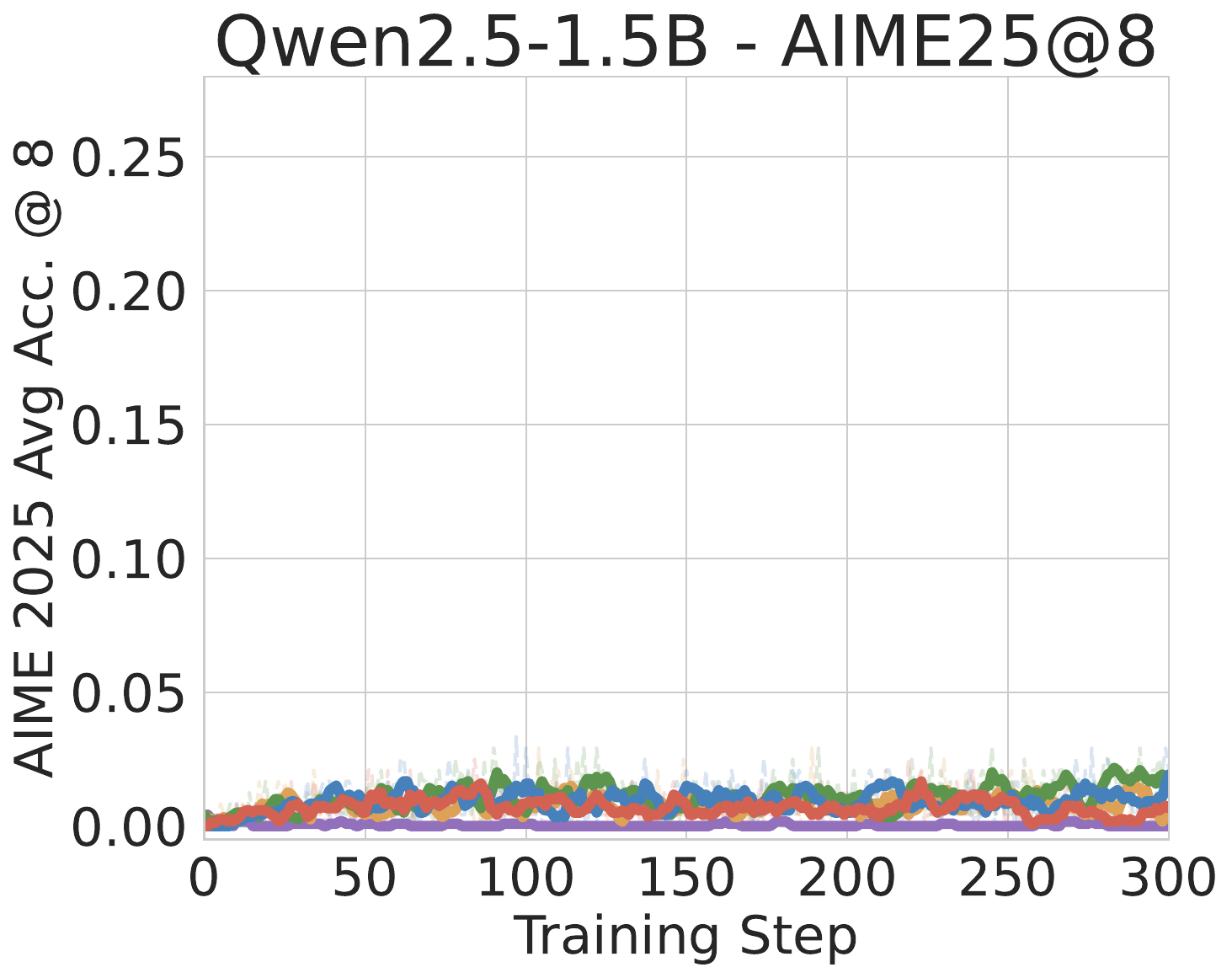}\vspace{-2mm}
        \caption{Qwen2.5-1.5B}
        \label{fig:qwen1.5b_results_aime}
    \end{subfigure}%
    ~
    \begin{subfigure}[t]{0.245\textwidth}
        \centering
        \includegraphics[width=\linewidth]{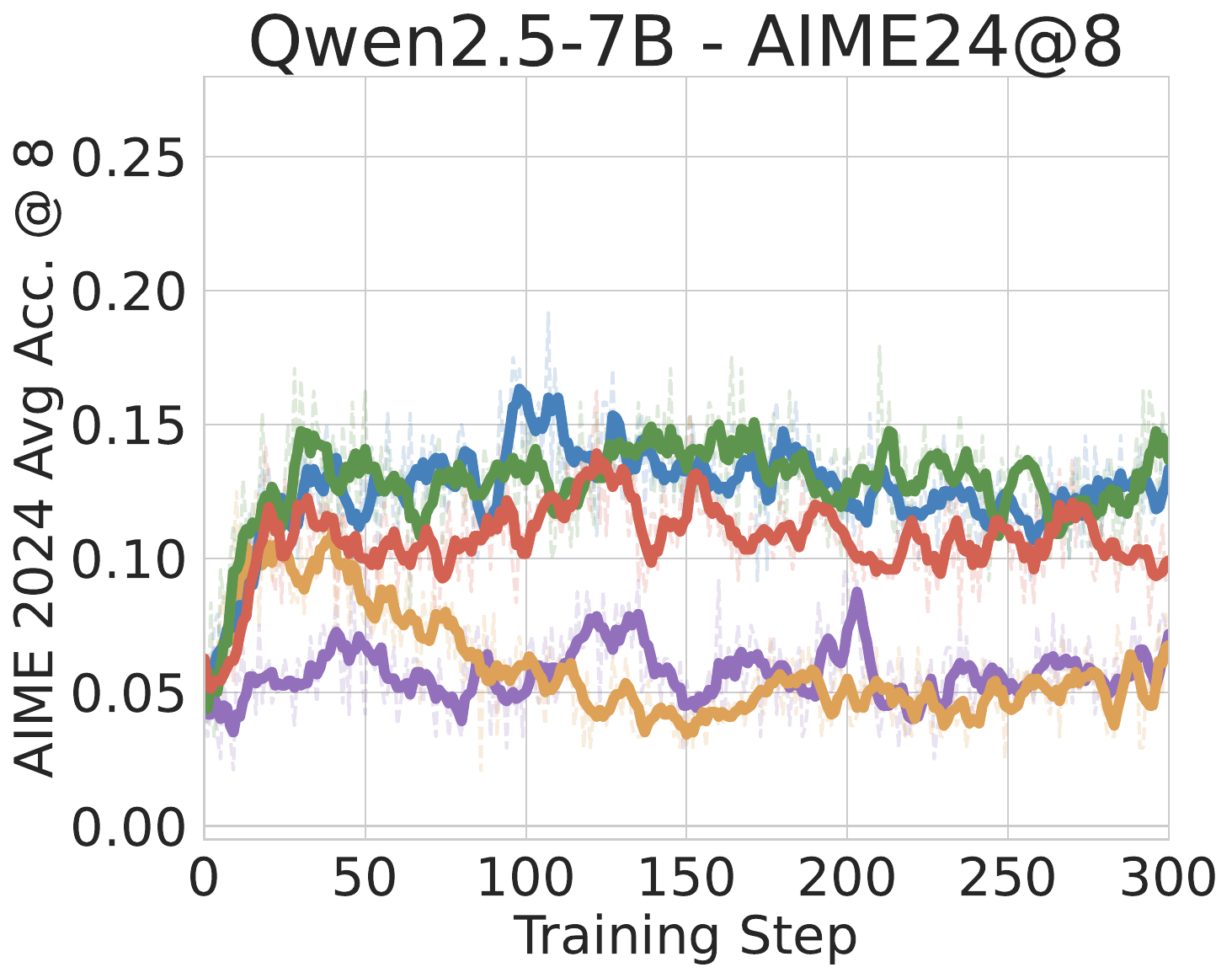}
        \includegraphics[width=\linewidth]{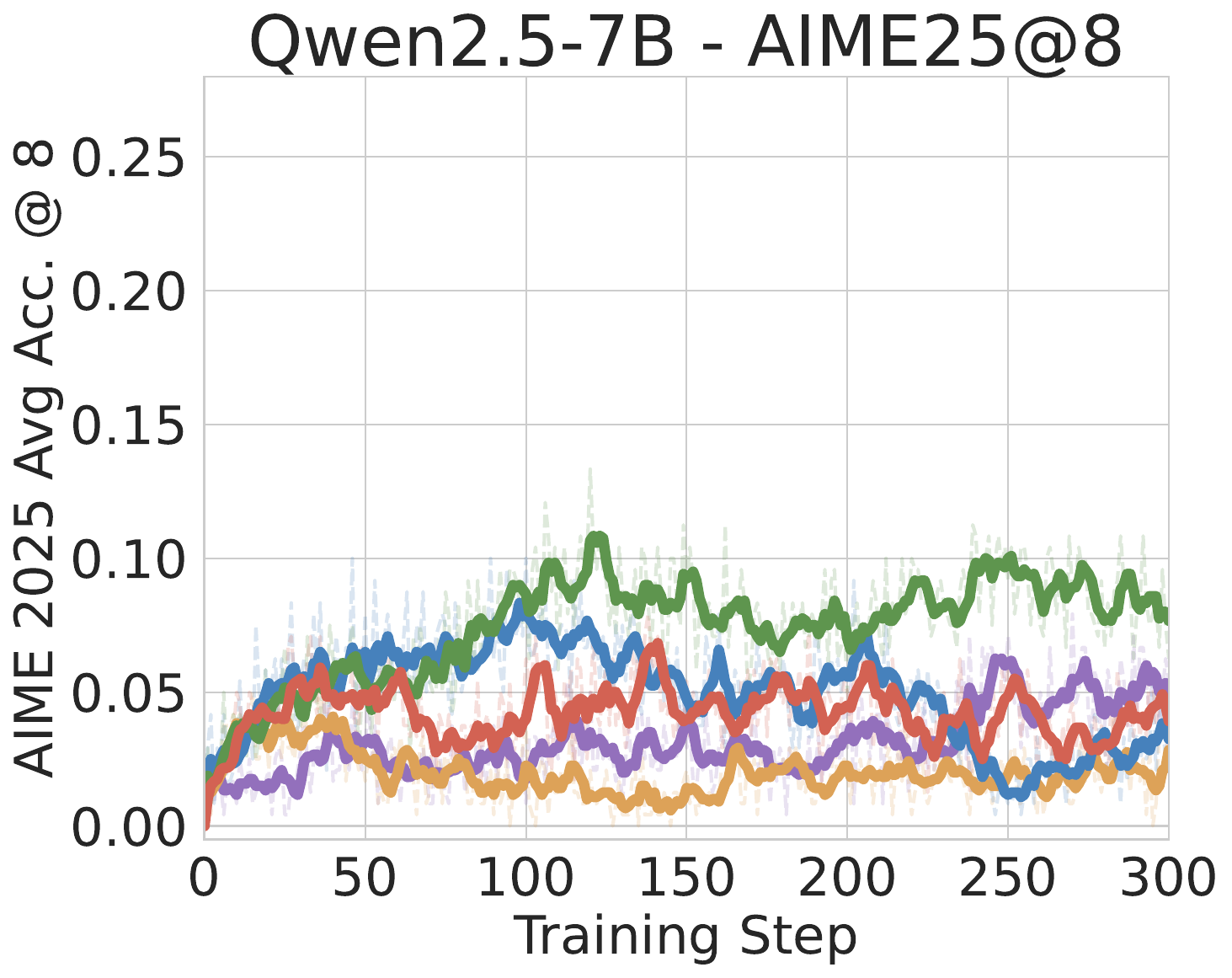}\vspace{-2mm}
        \caption{Qwen2.5-7B}
        \label{fig:qwen_results_aime}
    \end{subfigure}%
    ~
    \begin{subfigure}[t]{0.245\textwidth}
        \centering
        \includegraphics[width=\linewidth]{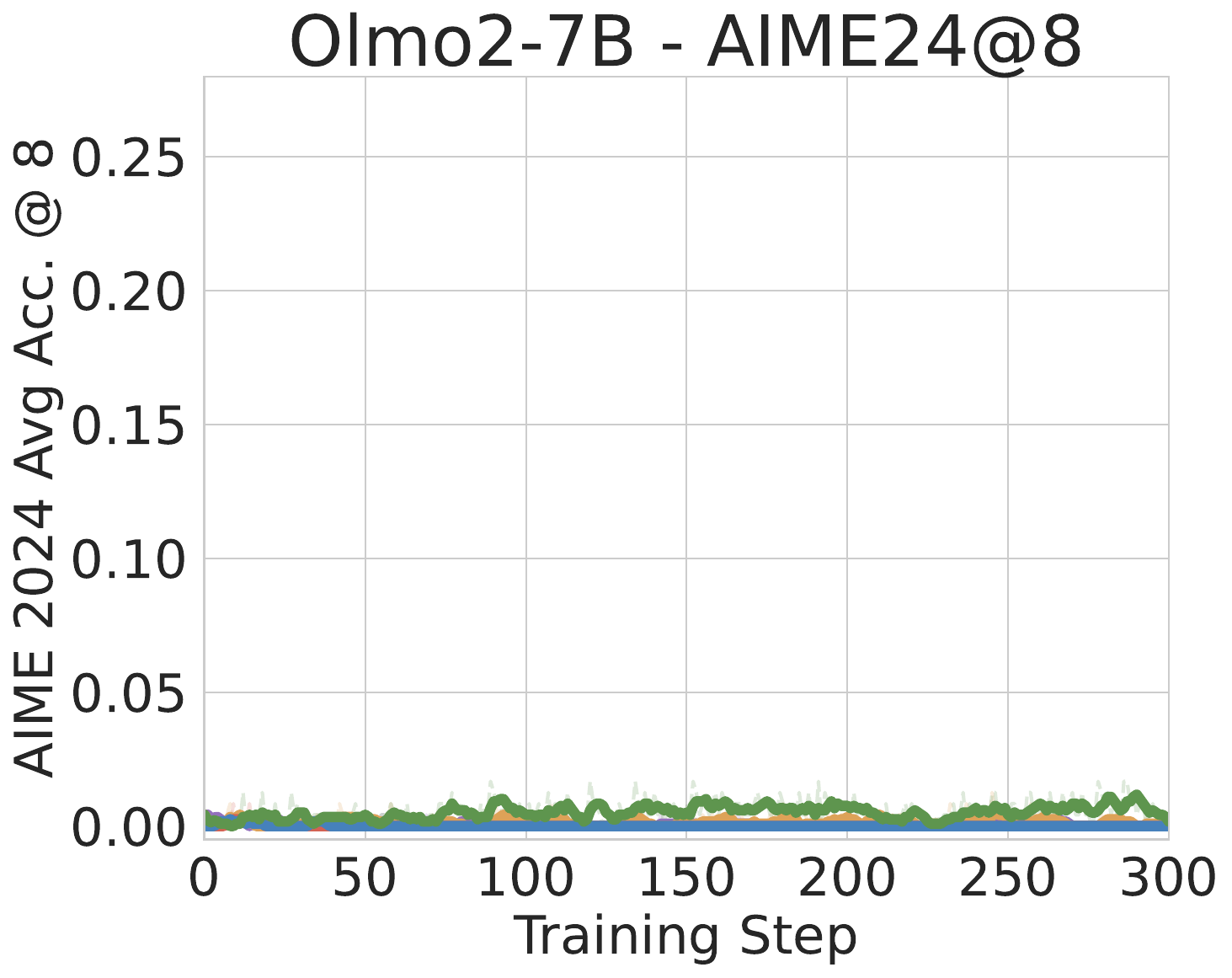}
        \includegraphics[width=\linewidth]{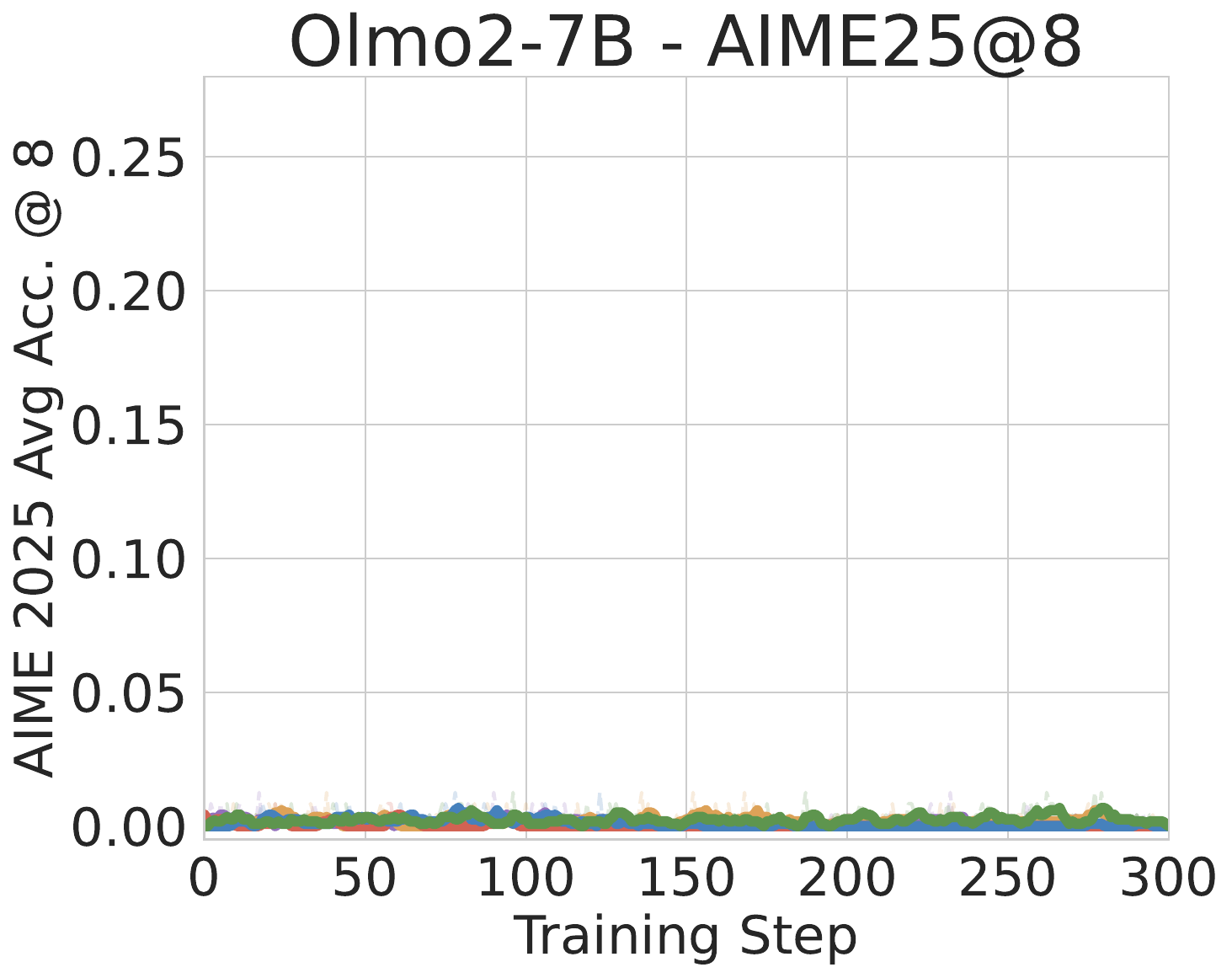}\vspace{-2mm}
    \caption{Olmo-2-1124-7B}
    \label{fig:olmo_results_aime}
    \end{subfigure}%
    ~ 
    \begin{subfigure}[t]{0.245\textwidth}
        \centering
        \includegraphics[width=\linewidth]{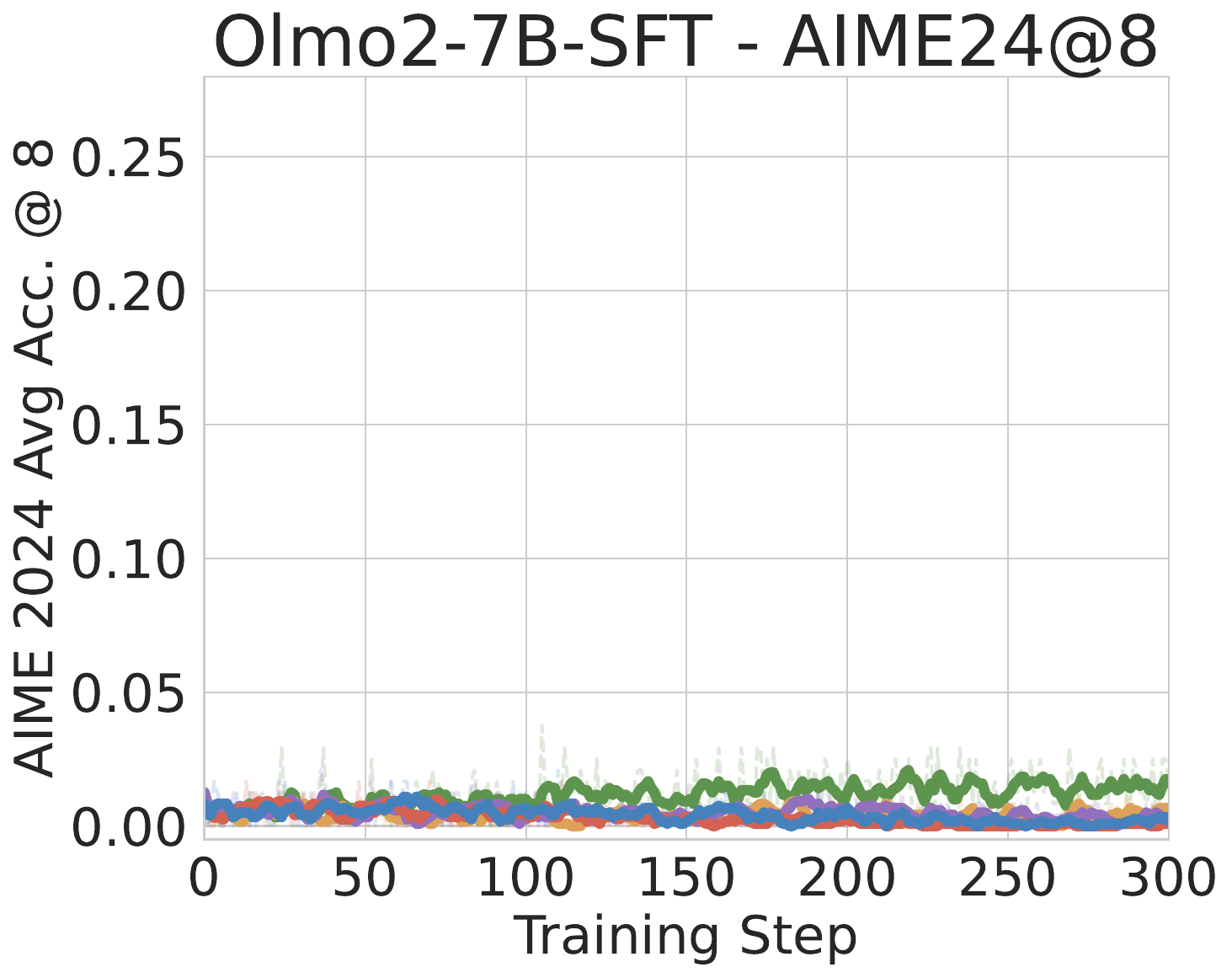}
        \includegraphics[width=\linewidth]{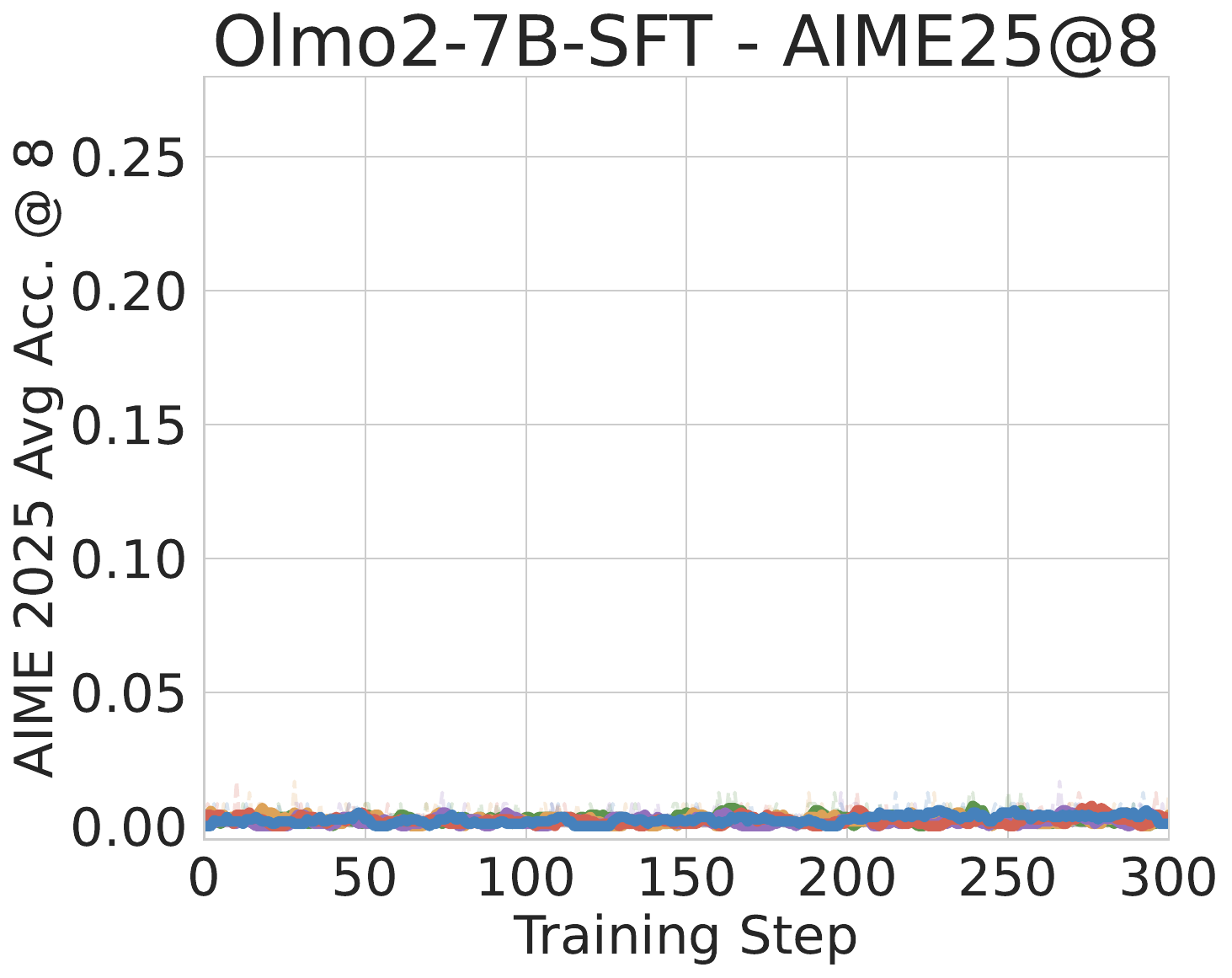}\vspace{-2mm}
    \caption{Olmo-2-1124-7B-SFT}
    \label{fig:olmo_sft_results_aime}
    \end{subfigure}\vspace{2mm}

    \begin{subfigure}[t]{0.245\textwidth}
        \centering
        \includegraphics[width=\linewidth]{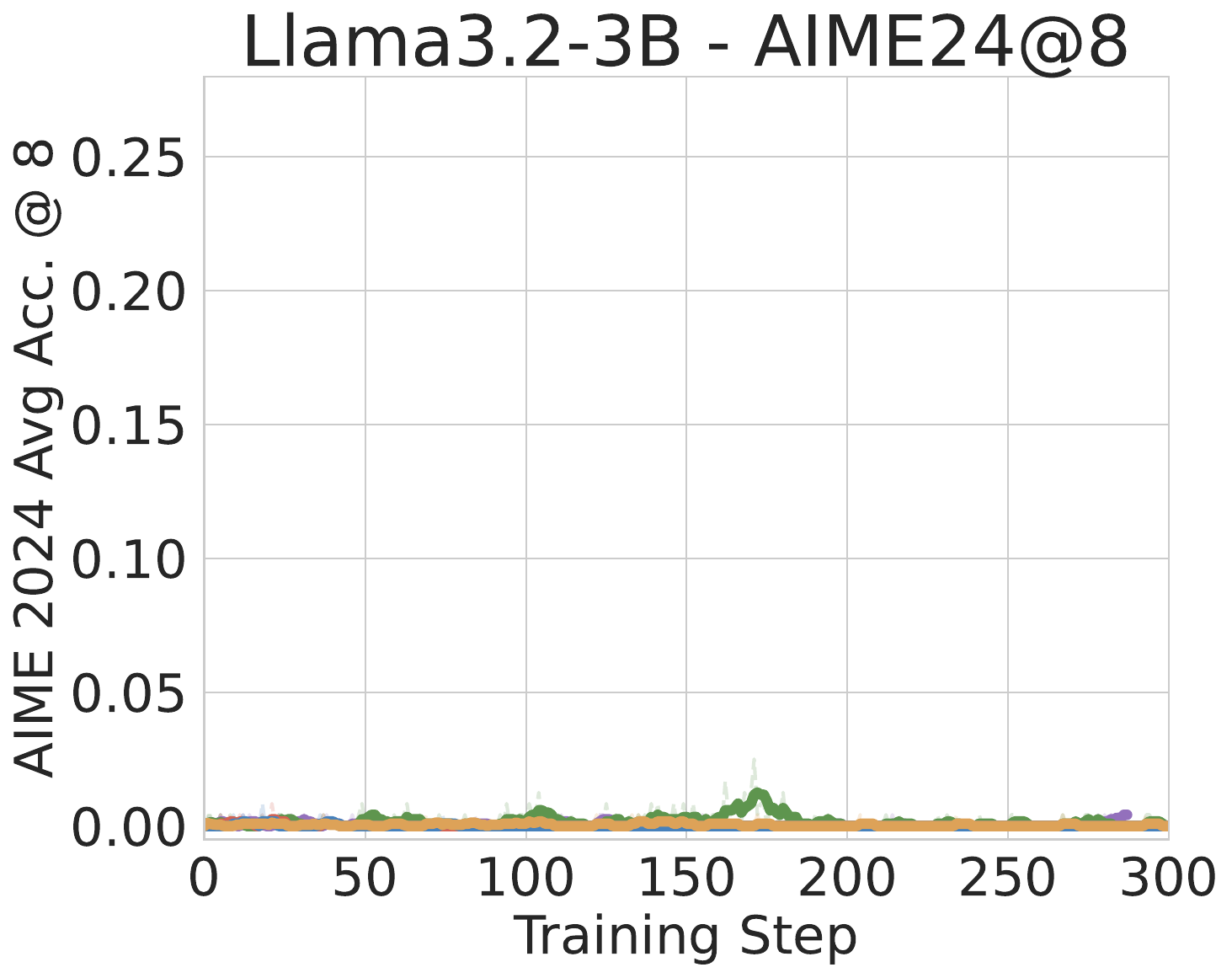}
        \includegraphics[width=\linewidth]{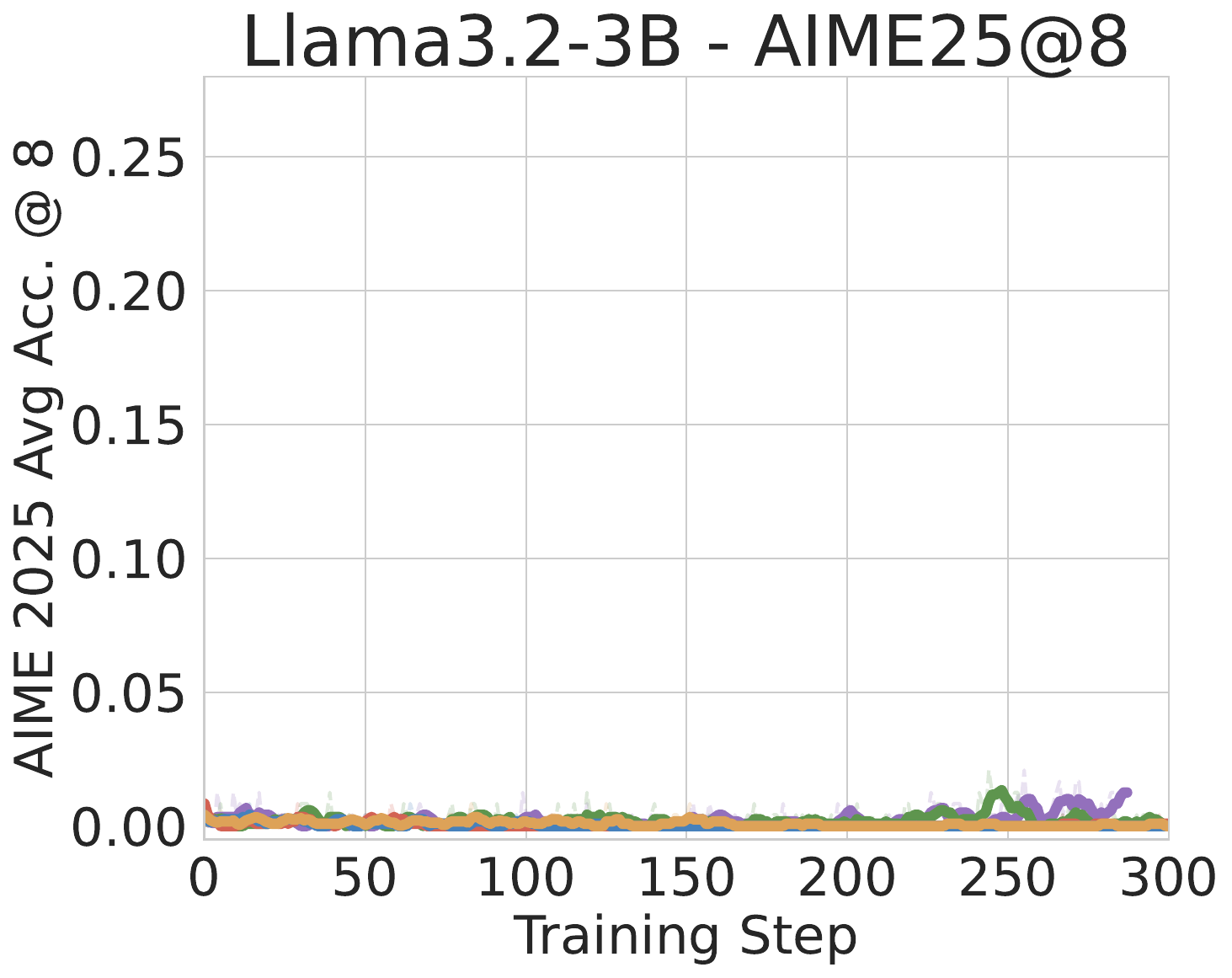}\vspace{-2mm}
    \caption{Llama3.2-3B}
    \label{fig:llama3_base_results_aime}
    \end{subfigure}%
    ~
    \begin{subfigure}[t]{0.245\textwidth}
        \centering
        \includegraphics[width=\linewidth]{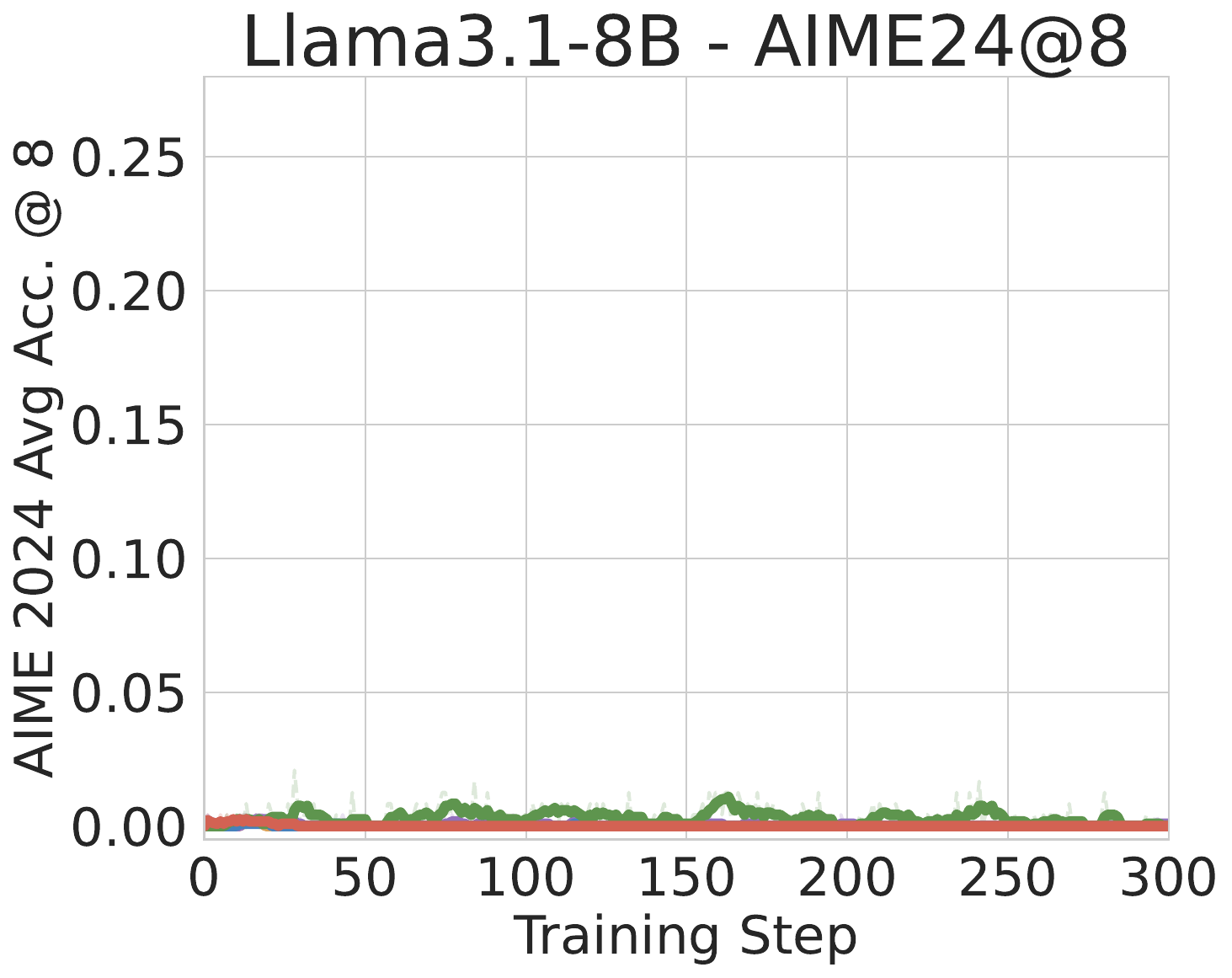}
        \includegraphics[width=\linewidth]{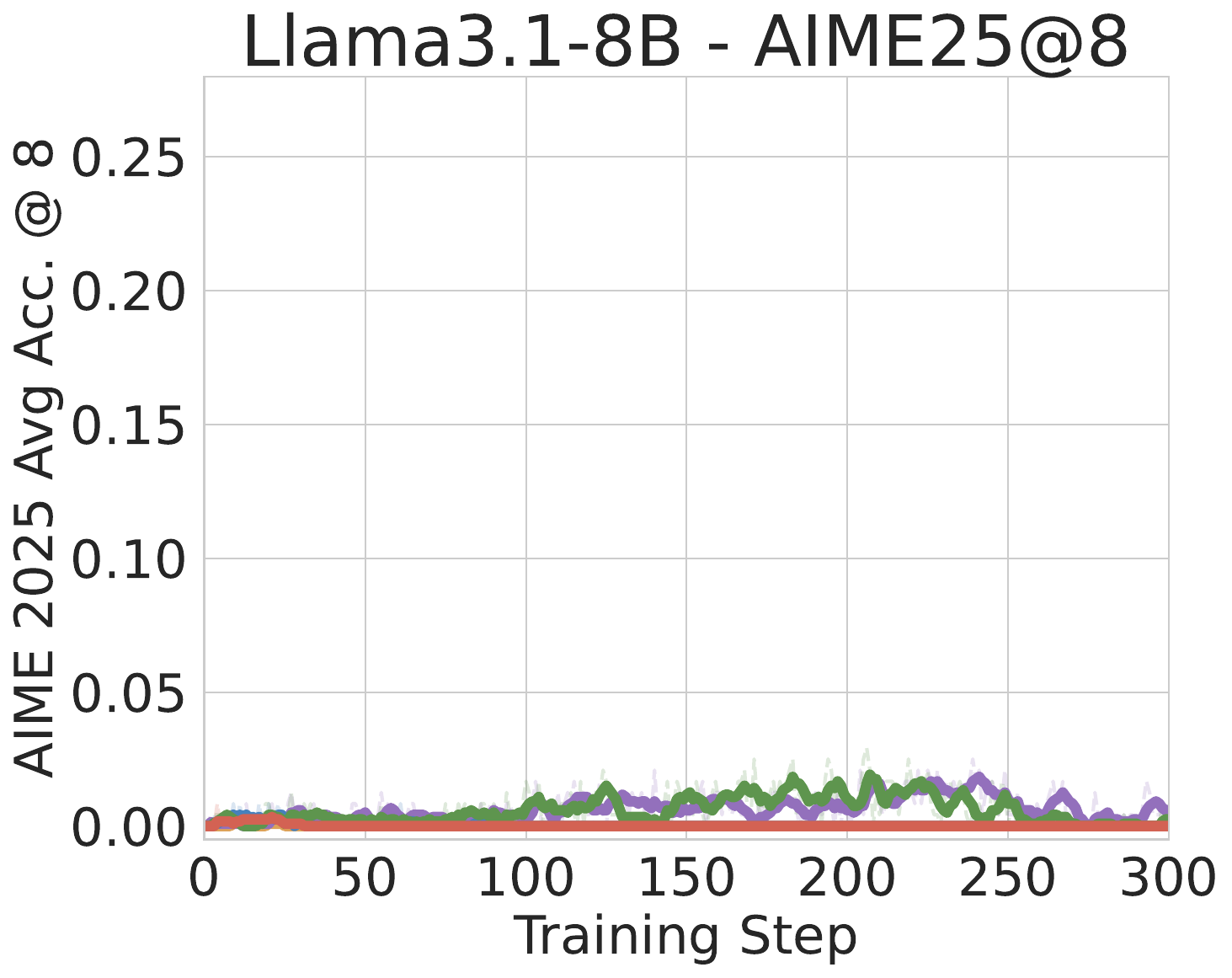}\vspace{-2mm}
    \caption{Llama3.1-8B}
    \label{fig:llama3_base_results_aime}
    \end{subfigure}%
    ~
    \begin{subfigure}[t]{0.245\textwidth}
        \centering
        \includegraphics[width=\linewidth]{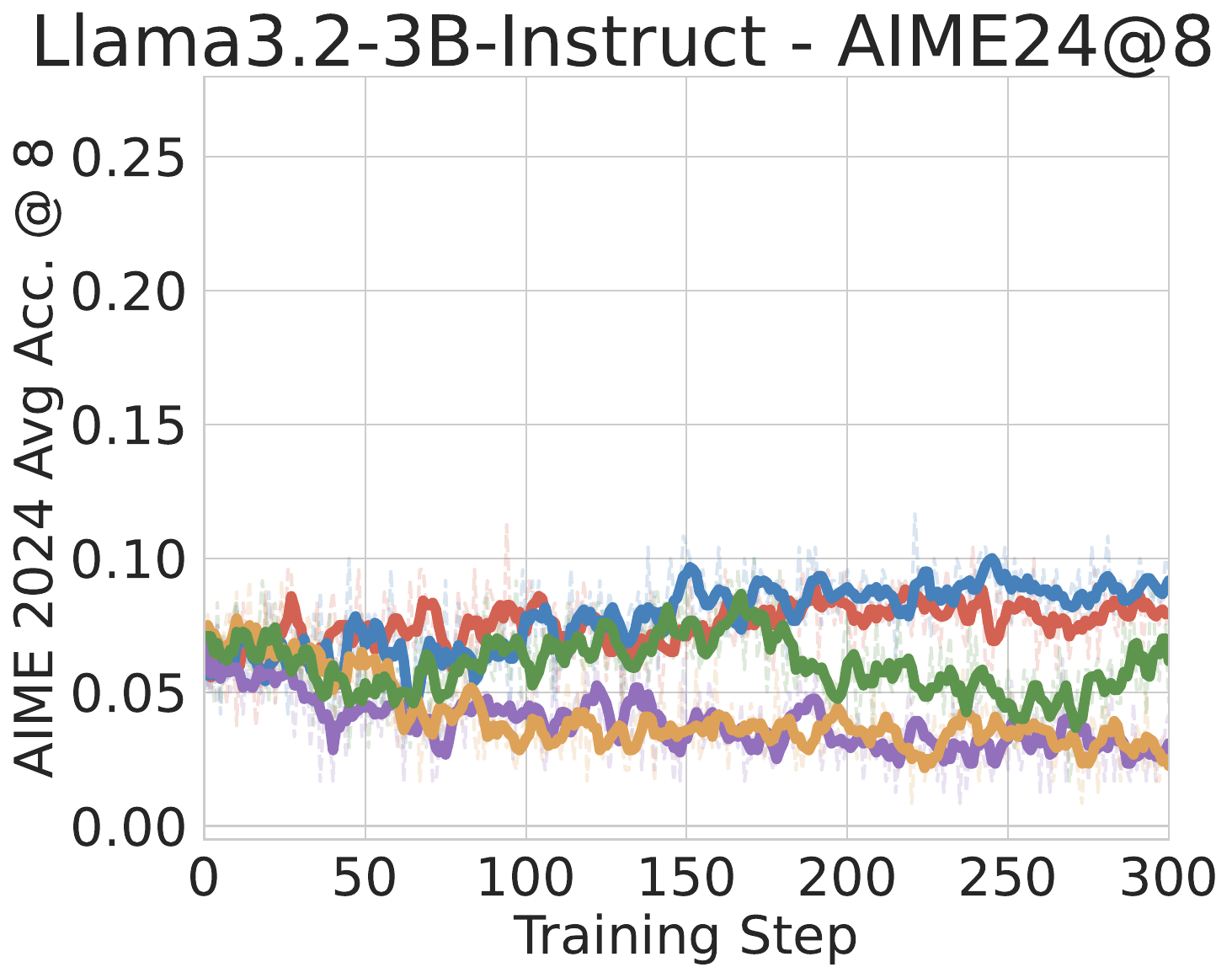}
        \includegraphics[width=\linewidth]{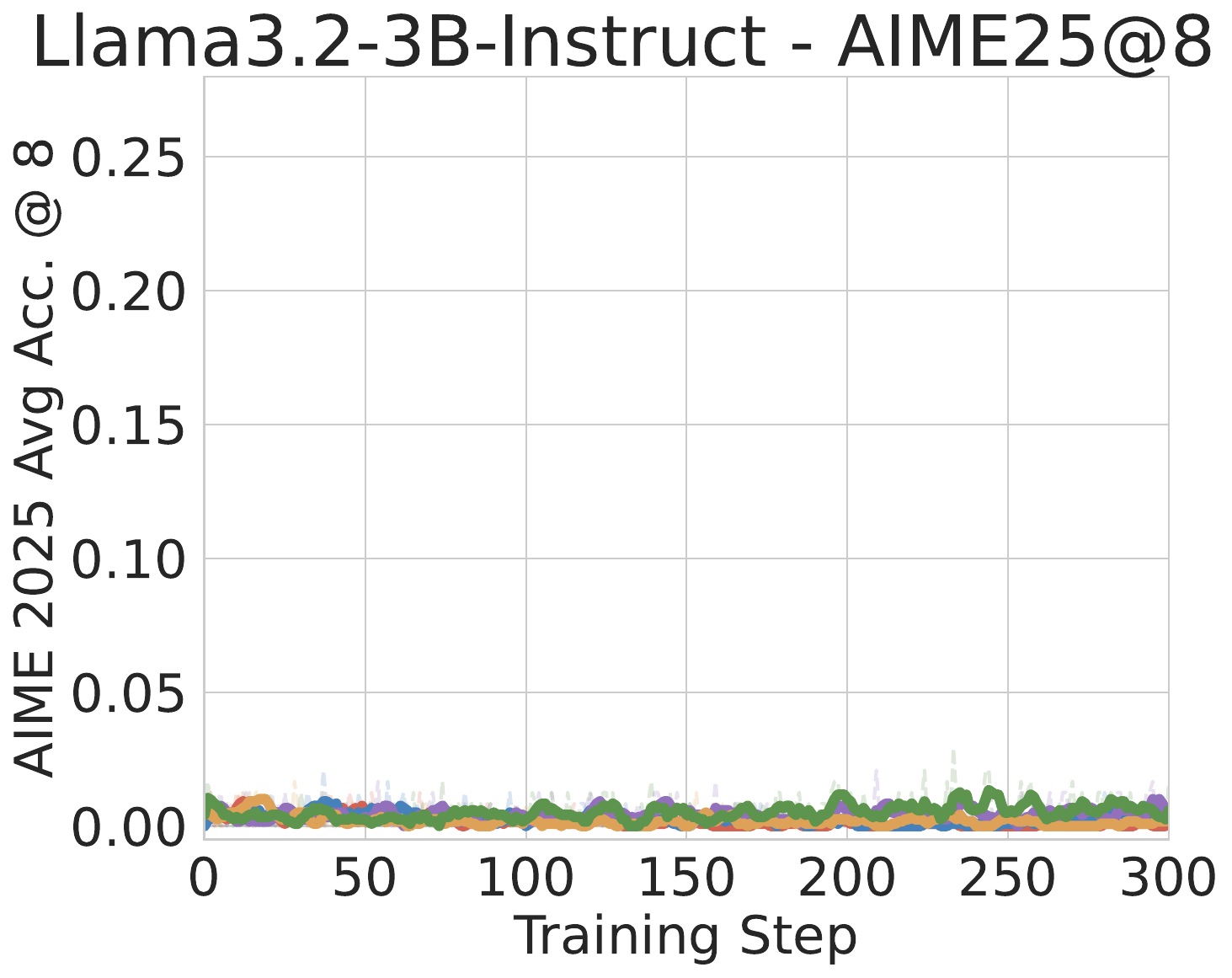}\vspace{-2mm}
    \caption{Llama3.2-3B-Instruct}
    \label{fig:llama3_results_aime}
    \end{subfigure}%
    ~
    \begin{subfigure}[t]{0.245\textwidth}
        \centering
        \includegraphics[width=\linewidth]{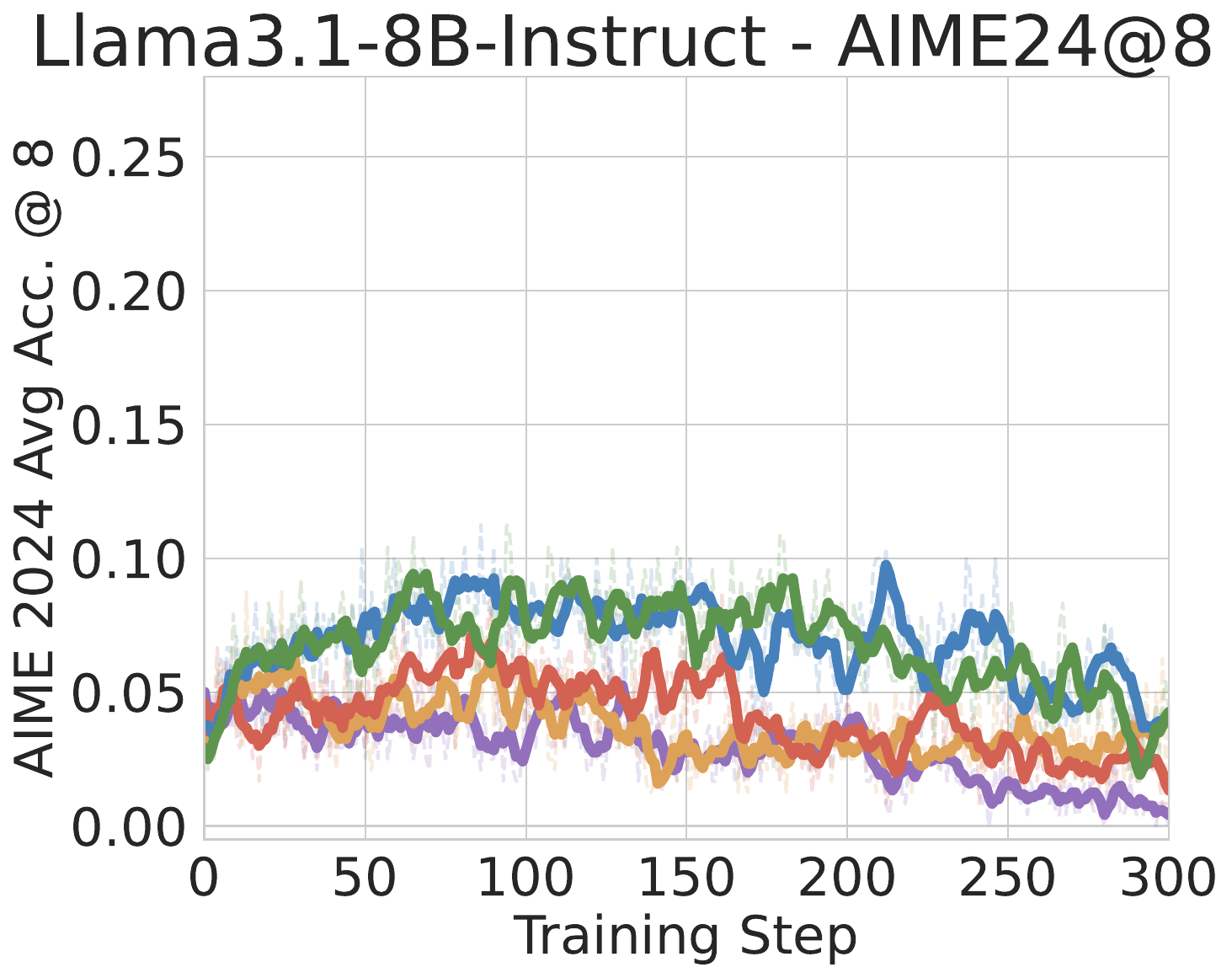}
        \includegraphics[width=\linewidth]{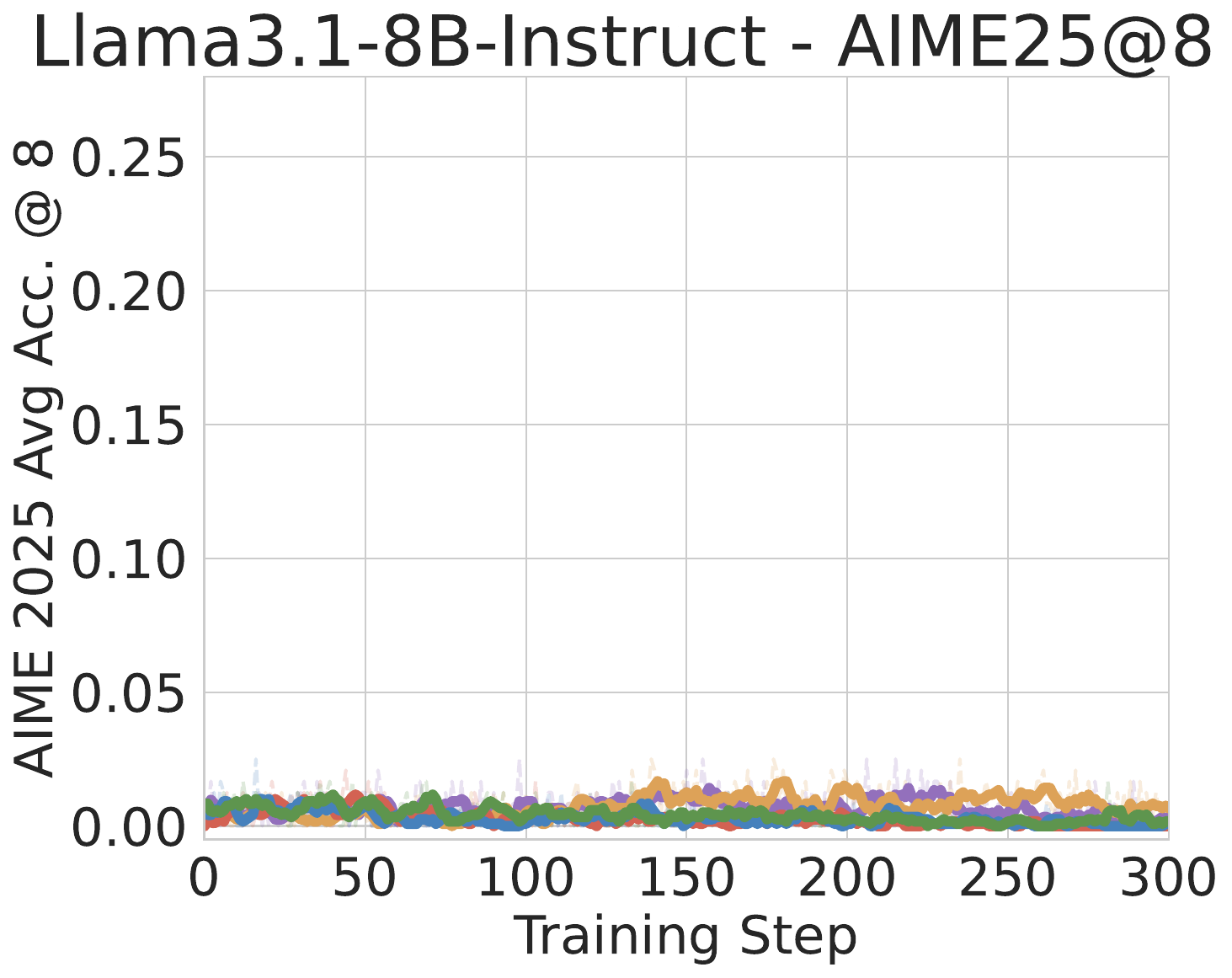}\vspace{-2mm}
    \caption{Llama3.1-8B-Instruct}
    \label{fig:llama8_results_aime}
    \end{subfigure}\vspace{-1mm}
    
    \caption{Varying rewards across additional model classes on the AIME 2024 and AIME 2025 benchmarks. Note that AIME 2025 was released after all the models' release dates, serving as a good resource to examine any dataset contamination phenomenon. Weak and spurious rewards (except for random) remain effective on general-purpose \qwenbasefamily models but generally fail to yield any gains on other model families. The performance improvements on non-\qwenbasefamily models are substantially smaller compared to those observed in the \qwenbasefamily family. Note that the AIME benchmarks contain 30 questions each, so small differences in accuracy (less than $\sim2$ pp.) may not be significant.
    } 
    \label{fig:rewards_other_family_aime}
\end{figure*}

We present additional results on the AIME benchmarks, which are challenging math Olympiad tests containing significantly harder problems than those in MATH-500 or AMC. We evaluate average@8 accuracy on AIME24 and AIME25~\citep{li2024numinamath}. AIME25 was created after the release date of all models considered in our study. Thus, evaluating performance on AIME25 allows us to control the risk that our models' have seen similar problems during web pretraining. We evaluate the trained models from Section~\ref{sec:qwen_works} and Section~\ref{sec:others_dont}. We show results on \qwenmathfamily models in Figure~\ref{fig:qwen_math_results_aime} and on the 8 additional models from Section~\ref{sec:others_dont} in Figure~\ref{fig:rewards_other_family_aime}.

\subsection{Spurious Rewards Yield Significant RLVR Gains on \qwenmathfamily}
As shown in Figure~\ref{fig:qwen_math_results_aime}, spurious rewards can consistently yield performance gains on Qwen-Math models on AIME24.
Intriguingly, we find that any AIME24 gains achievable from training Qwen models with spurious rewards largely vanish when evaluating on AIME 2025. We speculate that AIME25 contains questions that are more out-of-distribution to Qwen's pretrained knowledge; spurious rewards---which largely serve to elicit existing knowledge---hence no longer provide benefit.

\subsection{(Lack of) Generalization to Other Models}
Overall, results on other models are largely consistent with our earlier findings on AMC and MATH-500 (\S\ref{sec:others_dont}). Only \qwen, \qwensmall, and \llama exhibit any notable gains from any reward signals on AIME. For \qwenbasefamily models, weak and spurious rewards can yield gains; for example, format reward for \qwensmall and incorrect reward for \qwen. For \llama, only standard rewards (e.g., ground truth and majority vote) yield gains. 
As observed above, performance and gains from RLVR training are lower across the board for all models on AIME25.

\begin{figure}[t]
    \centering
    \parbox{7mm}{\includegraphics[width=\linewidth,height=1.5mm]{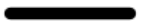}}\hspace{1mm}Compound\hspace{3mm}
    \parbox{7mm}{\includegraphics[width=\linewidth,height=1.5mm]{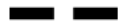}}\hspace{1mm}Original\\
    \cblock{98}{62}{121} Qwen-Math-7B  
    \cblock{134}{117}{171} Qwen-Math-1.5B  
    \cblock{83}{97}{54} Qwen-7B  
    \cblock{143}{163}{100} Qwen-1.5B  
    \cblock{165}{94}{138} Olmo2-7B-SFT
    \vspace{1mm}
    
    \begin{subfigure}[t]{0.33\textwidth}
        \centering
        \includegraphics[width=\linewidth]{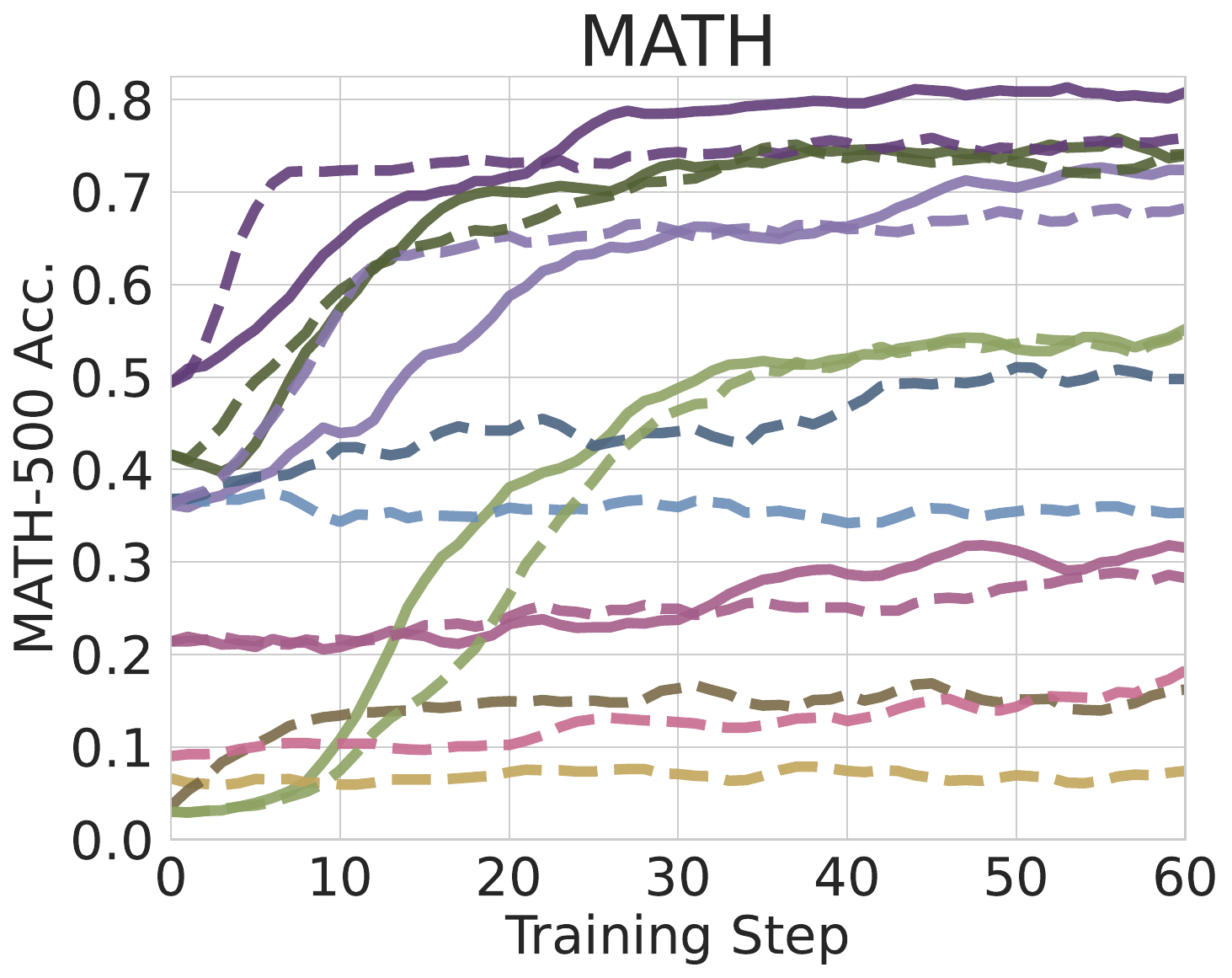}\\
        \includegraphics[width=\linewidth]{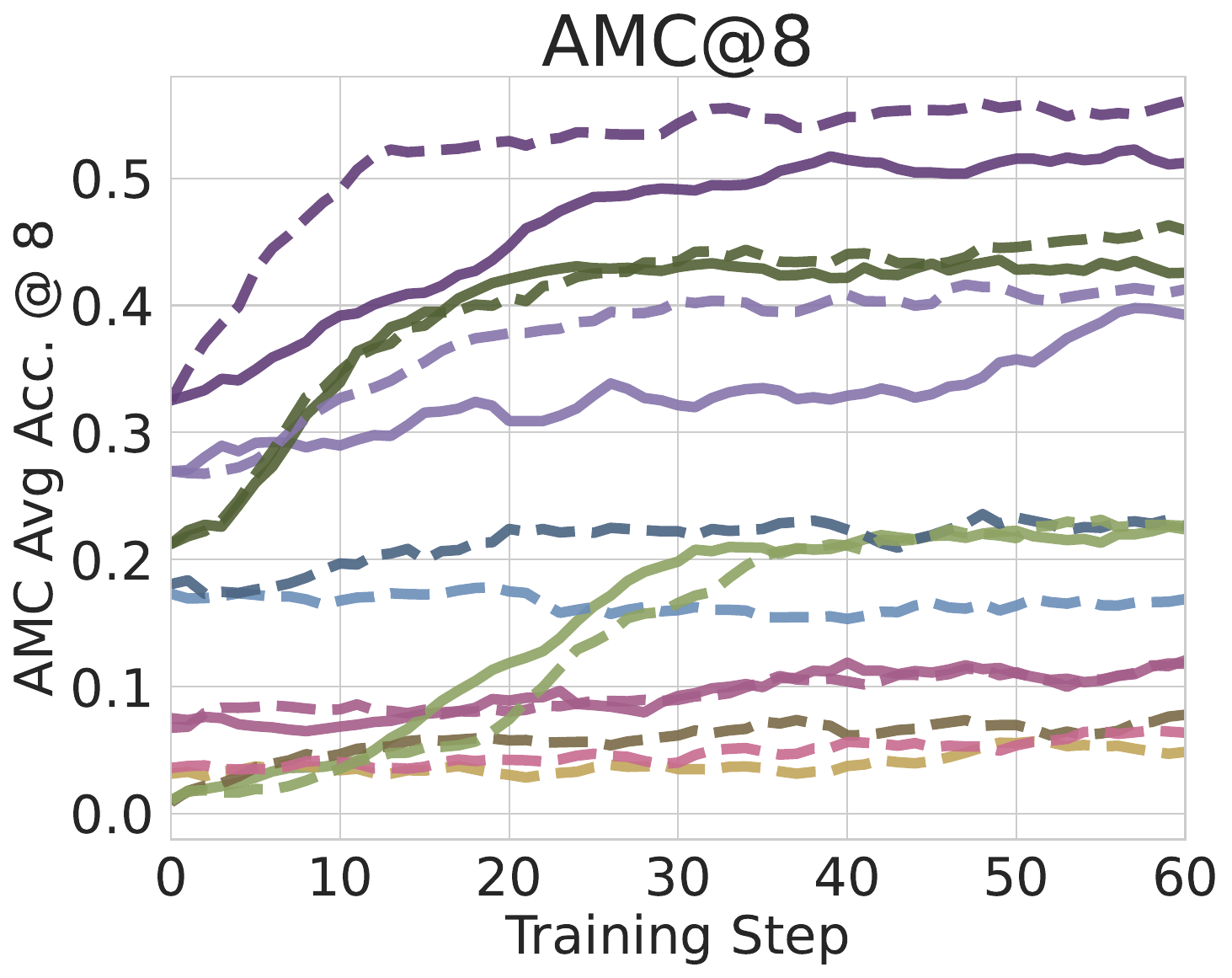}\\
        \includegraphics[width=\linewidth]{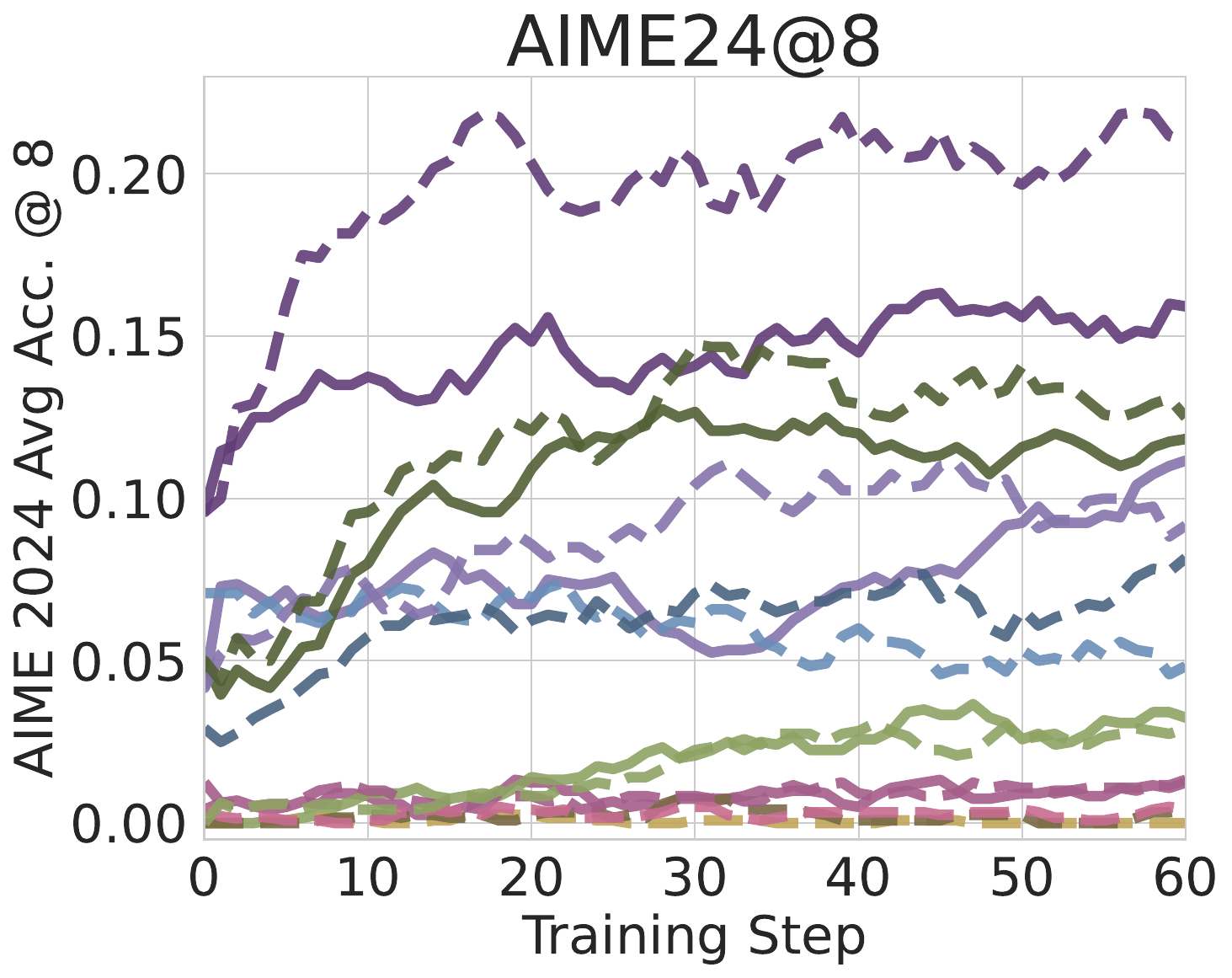}\\
        \includegraphics[width=\linewidth]{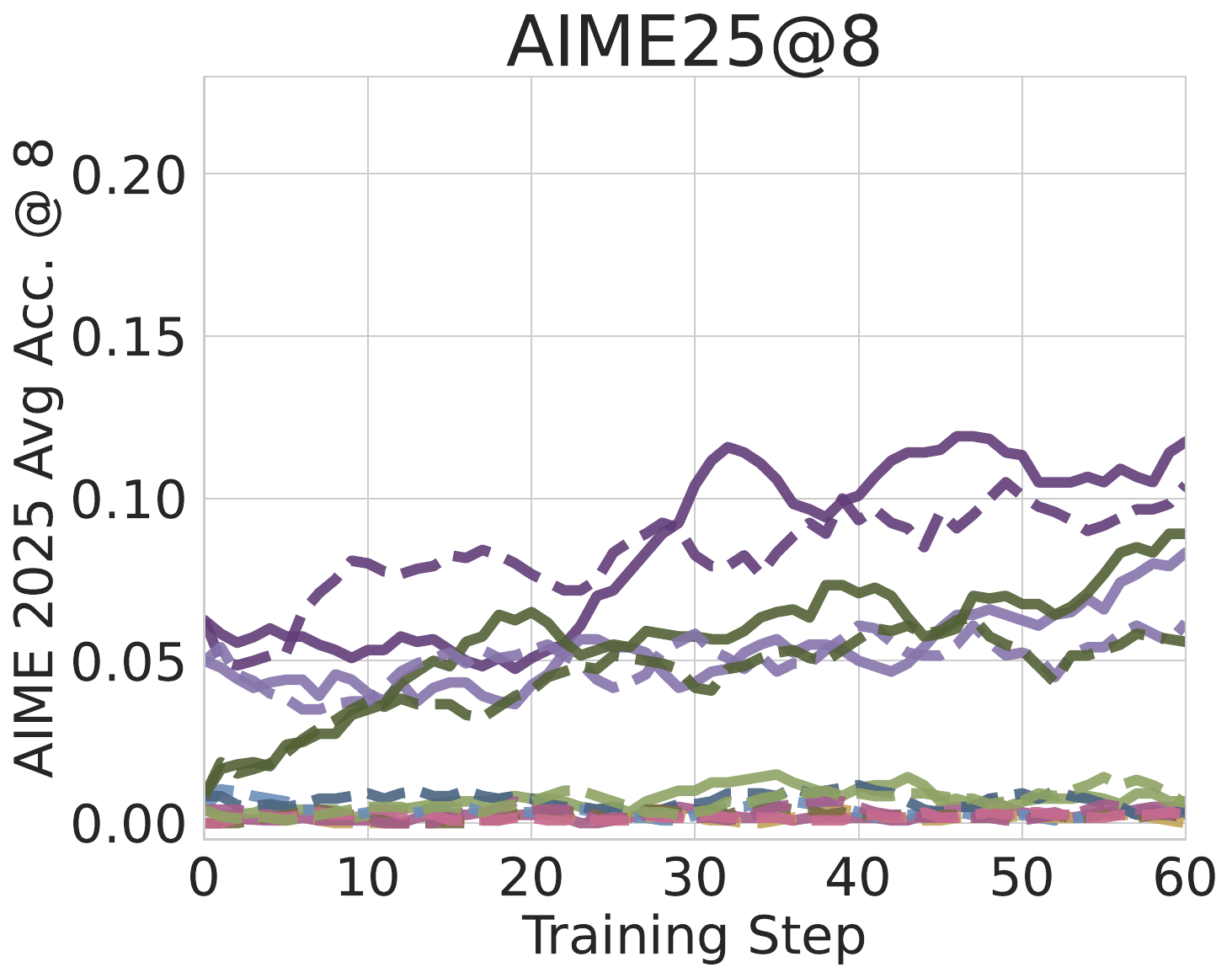}
        \vspace{-2mm}
    \caption{Ground Truth w/o Python}
    \label{fig:compound_reward_gt_extra}
    \end{subfigure}%
    ~
    \begin{subfigure}[t]{0.33\textwidth}
        \centering
        \includegraphics[width=\linewidth]{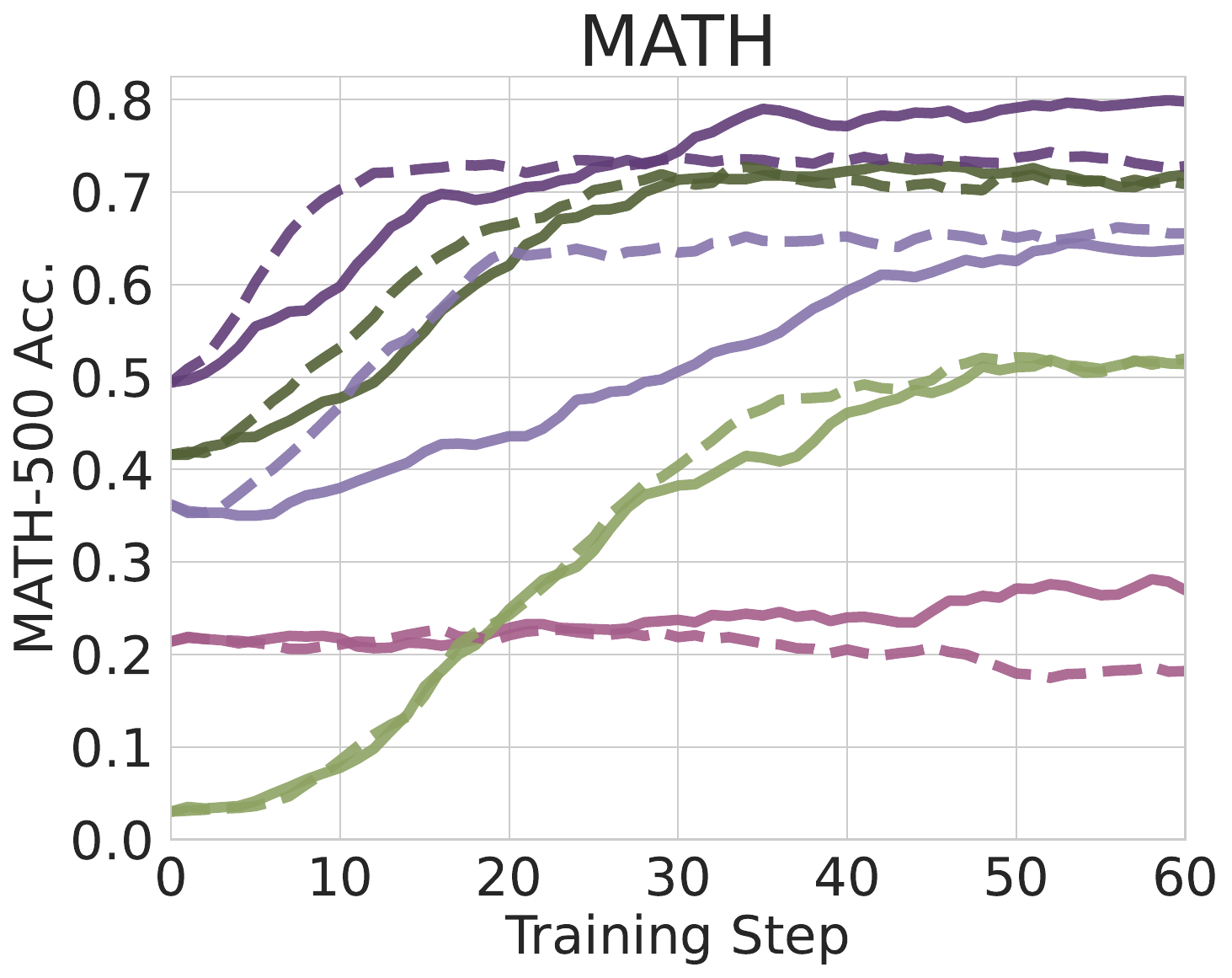}\\
        \includegraphics[width=\linewidth]{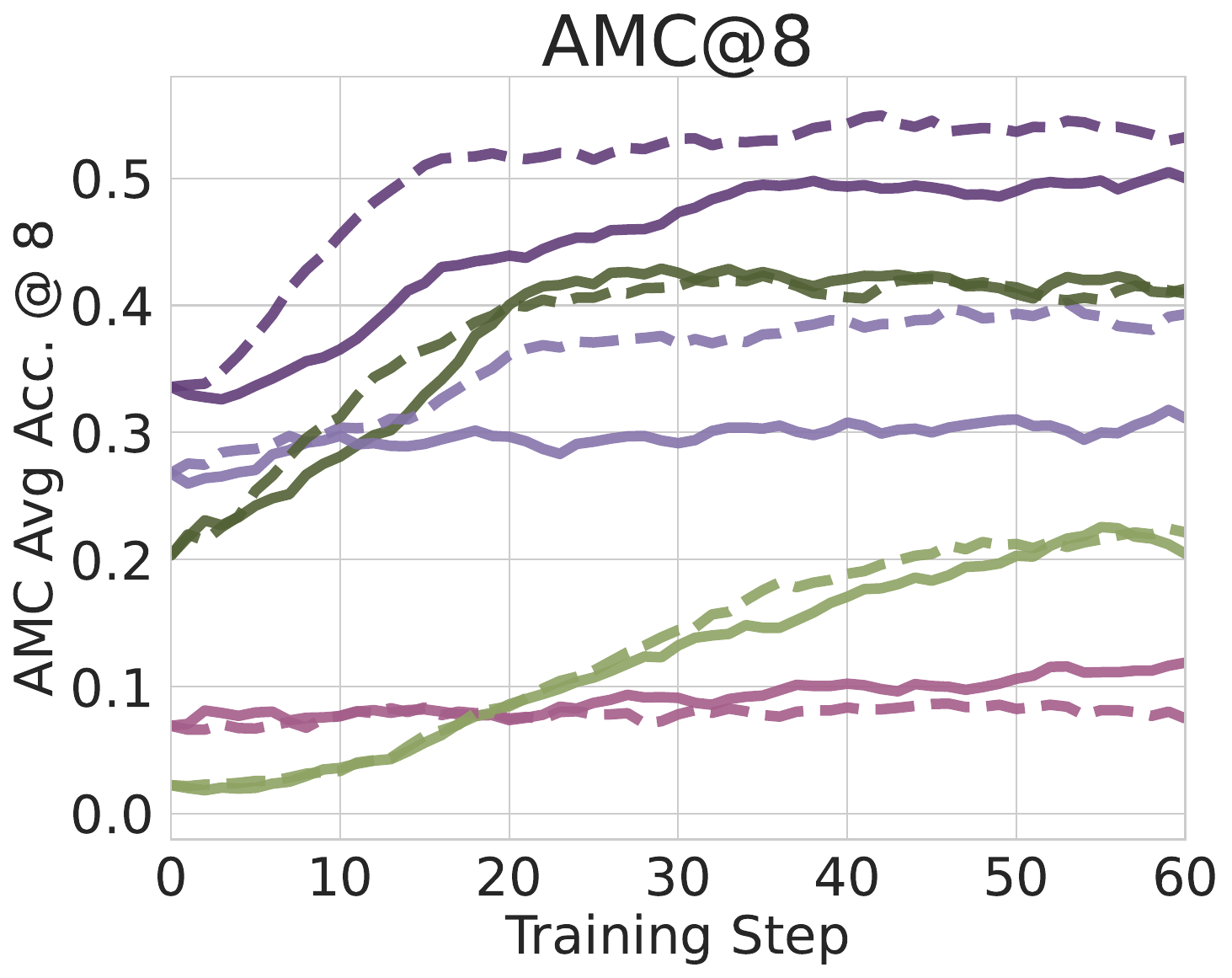}\\
        \includegraphics[width=\linewidth]{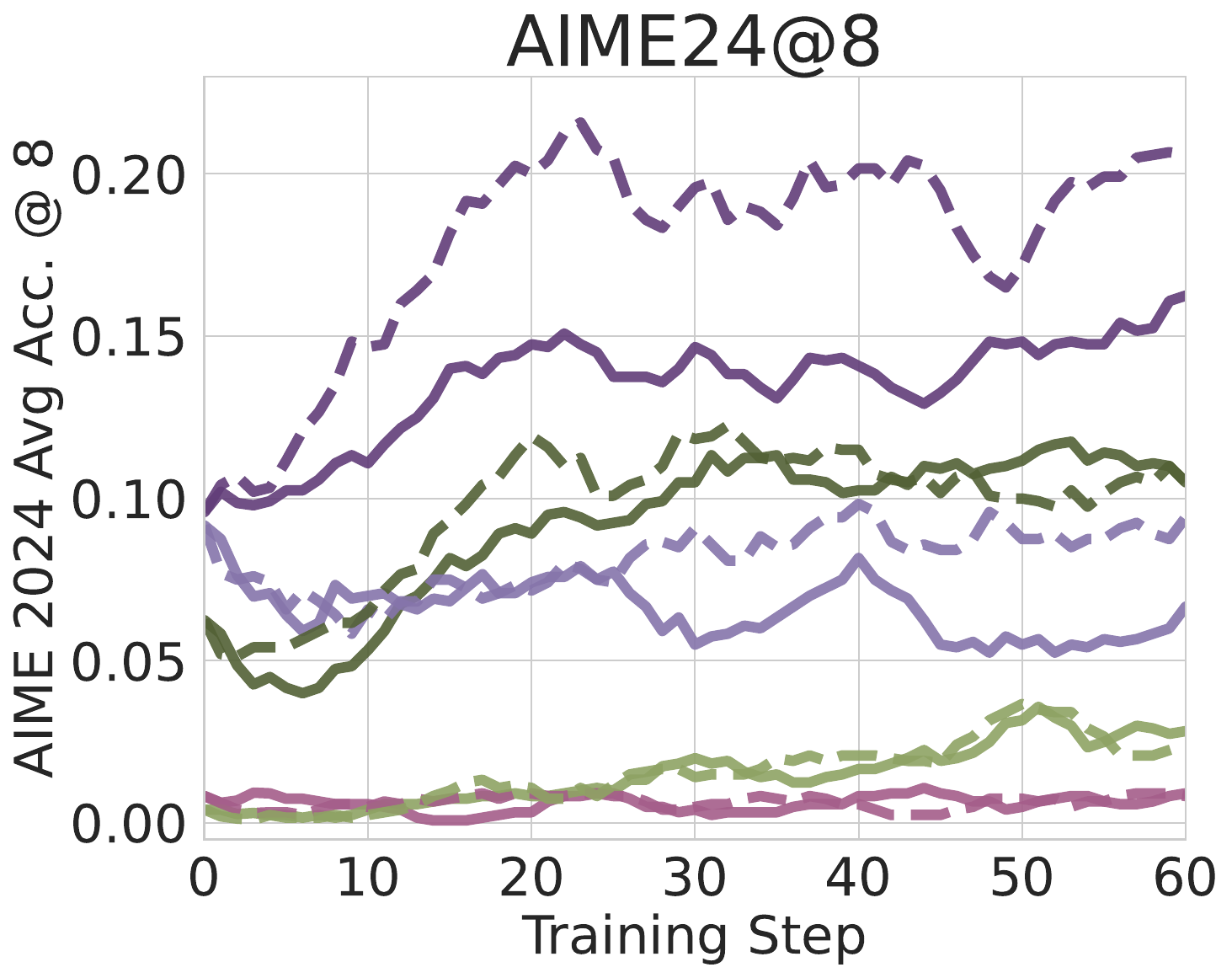}\\
        \includegraphics[width=\linewidth]{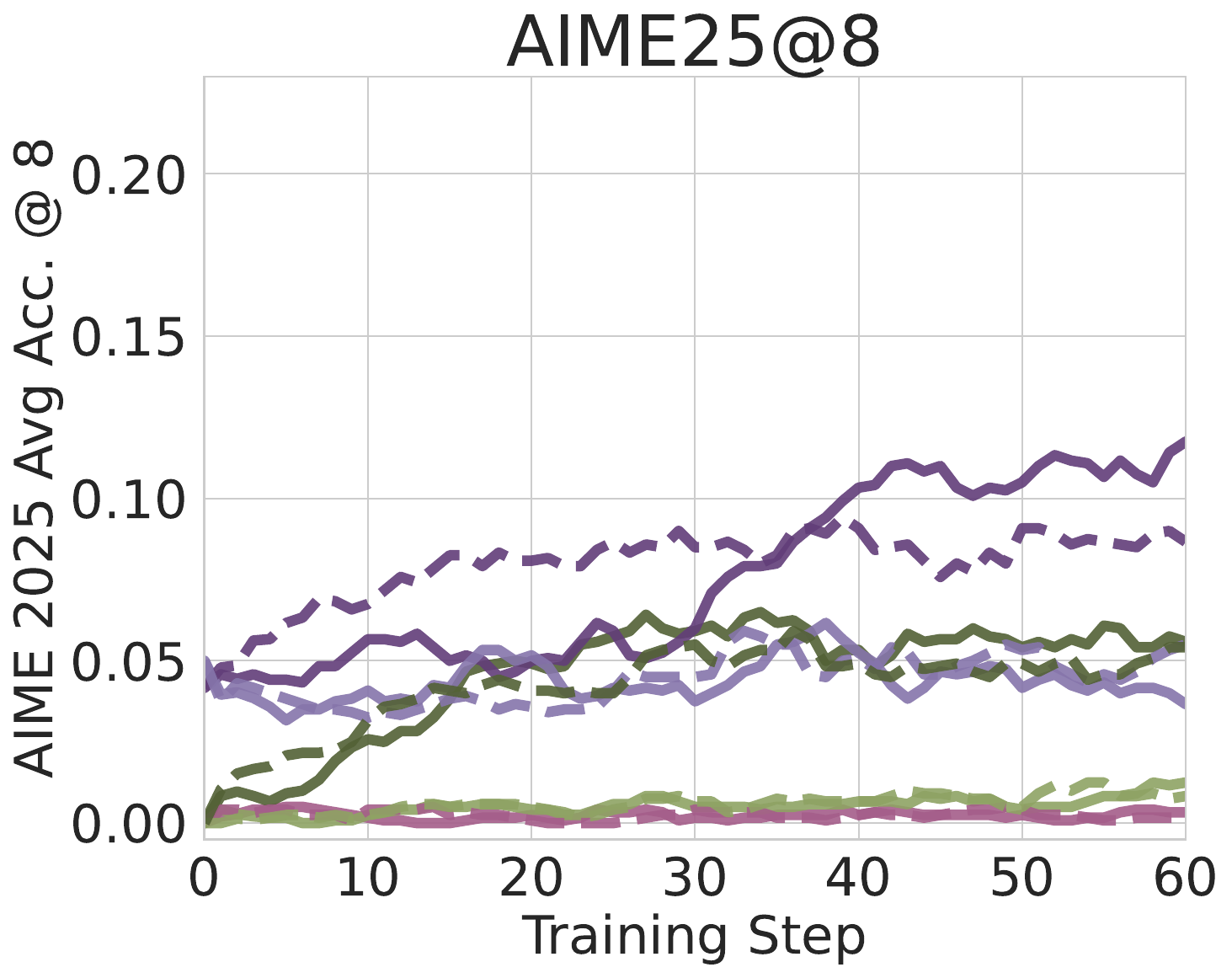}
        \vspace{-2mm}
    \caption{Incorrect w/o Python}
    \label{fig:compound_reward_incorrect_extra}
    \end{subfigure}%
    ~
    \begin{subfigure}[t]{0.33\textwidth}
        \centering
        \includegraphics[width=\linewidth]{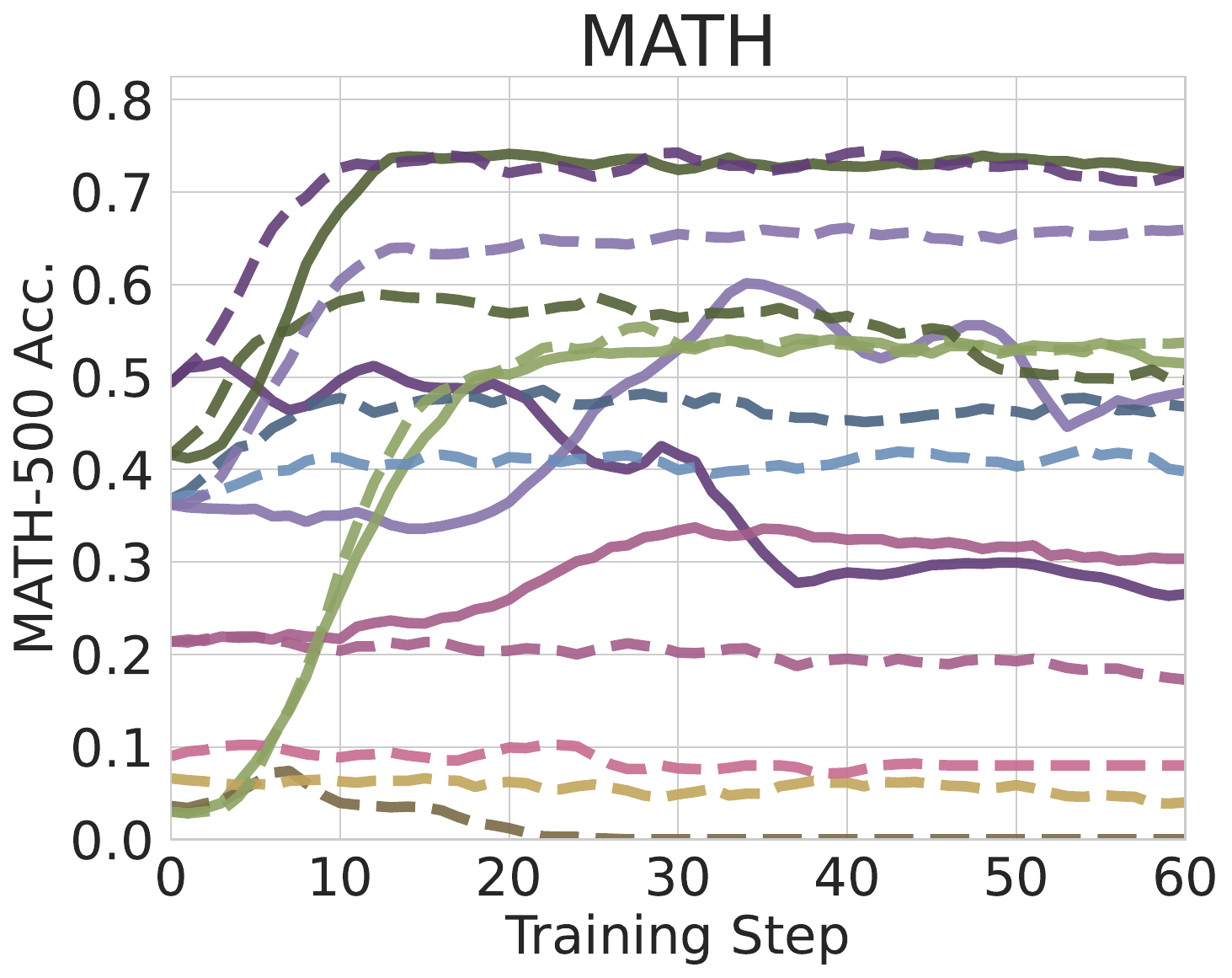}\\
        \includegraphics[width=\linewidth]{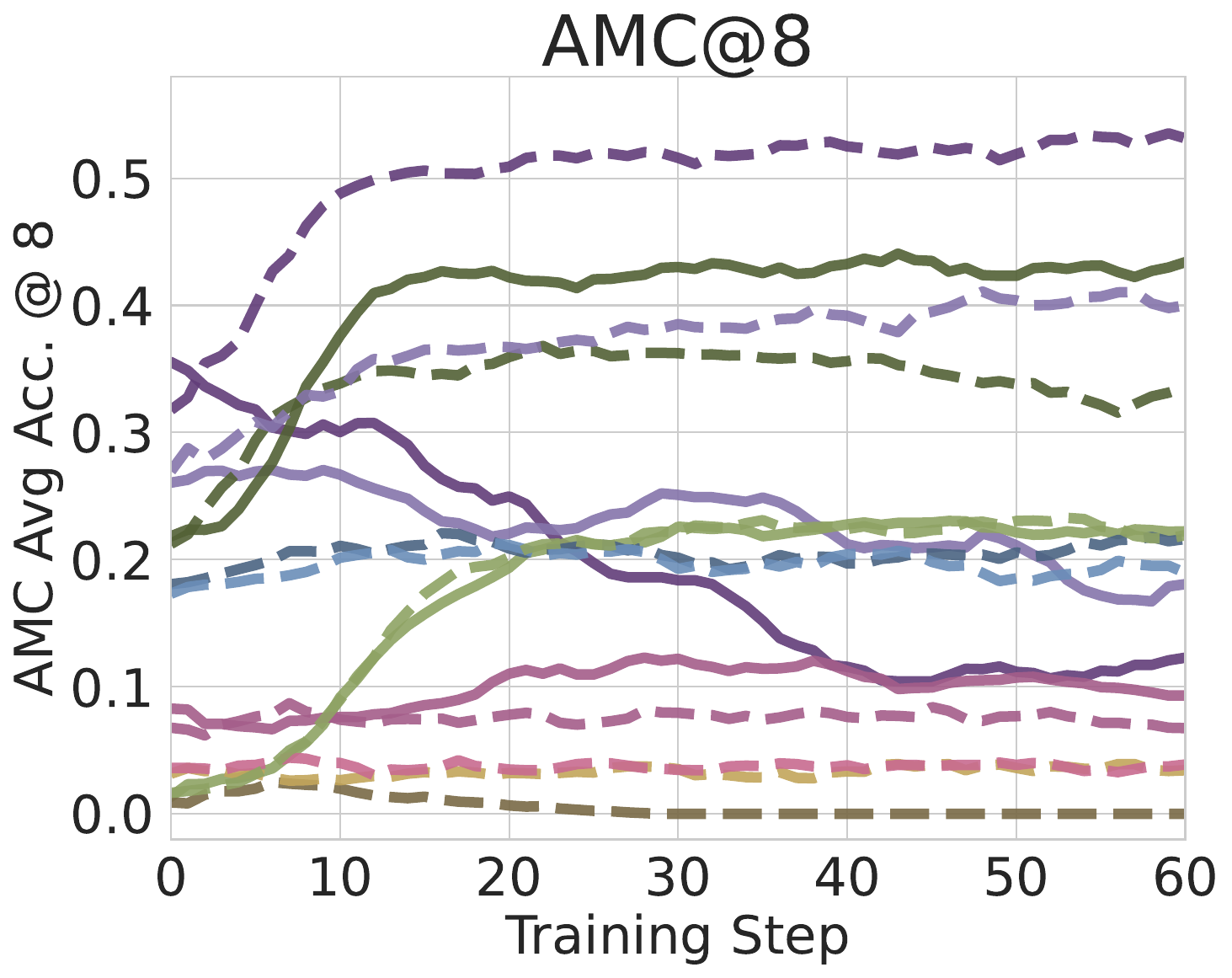}\\
        \includegraphics[width=\linewidth]{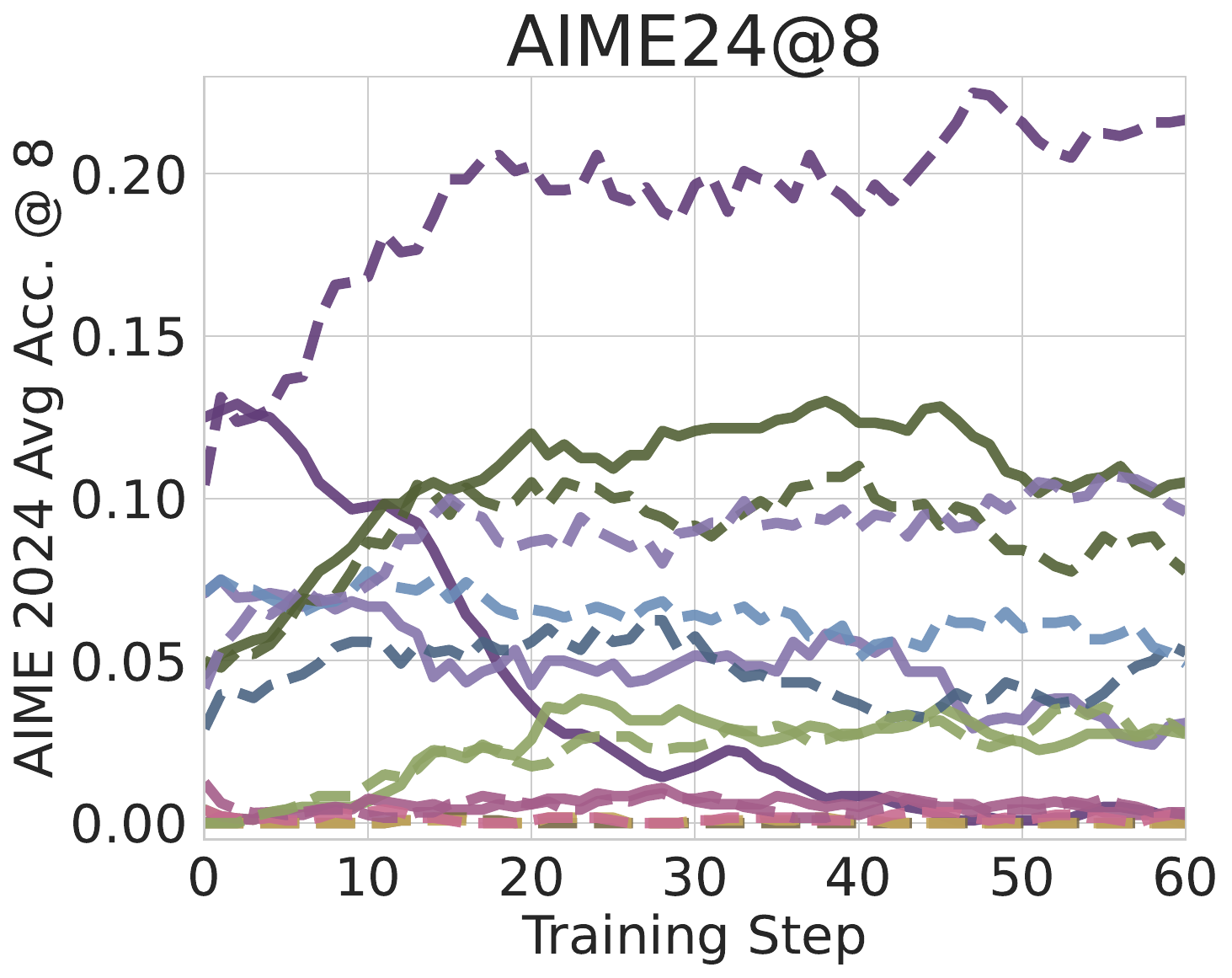}\\
        \includegraphics[width=\linewidth]{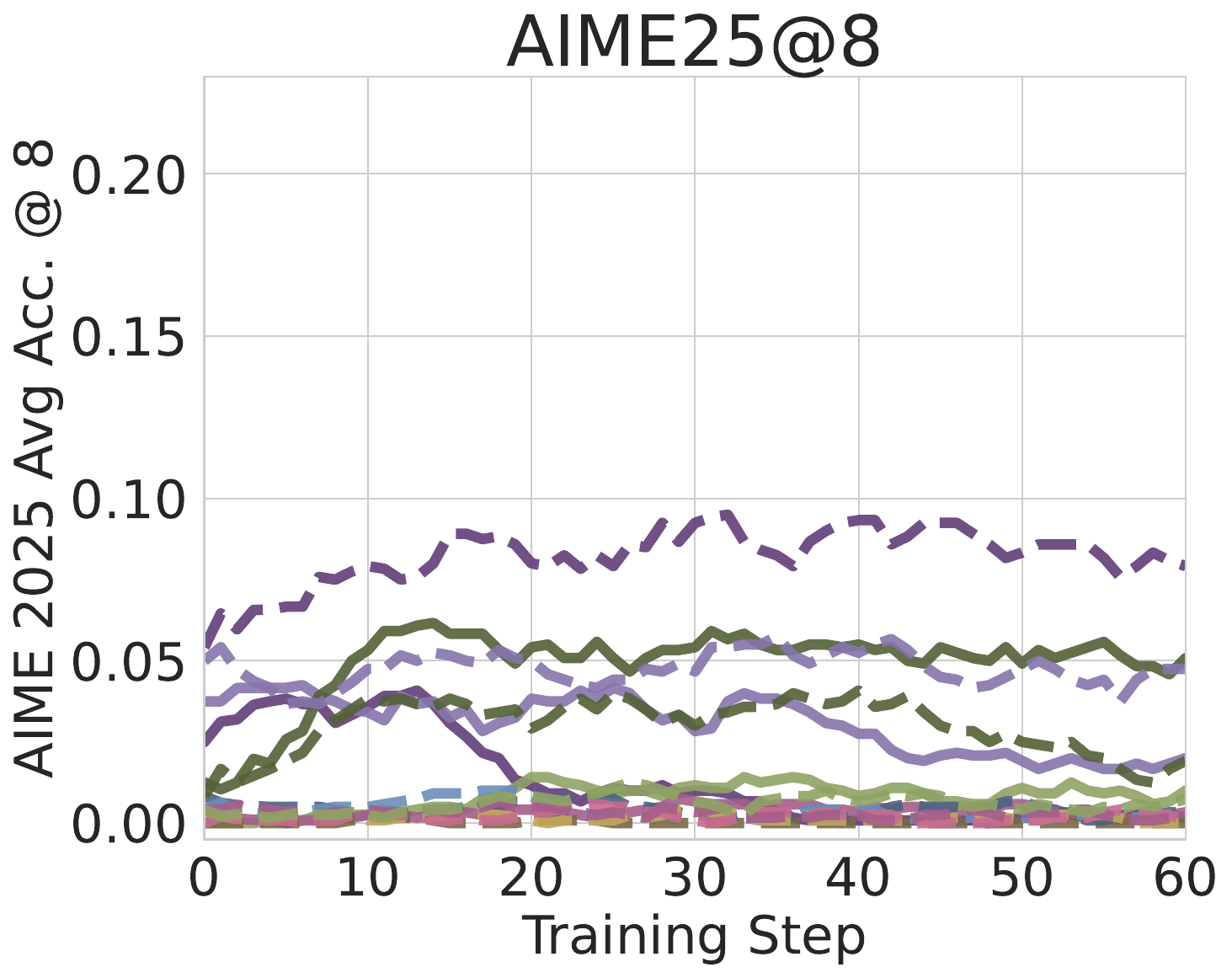}
        \vspace{-2mm}
    \caption{Format w/o Python}
    \label{fig:compound_reward_format_extra}
    \end{subfigure}%
    
    \caption{
    We present additional compound reward results on (1) smaller 1.5B Qwen models and (2) AMC and AIME benchmarks. See Section~\ref{sec:analysis:intervene} for full details of the setup. Our compound rewards intersect (a) our original rewards with a (b) \textit{no Python reward} that only rewards responses without Python code. Overall, our findings here are largely consistent with those from our main text. Note that AIME is a very small test set, so small differences in accuracy (less than $\sim2$ percentage points) may not be meaningful.
    }
    \label{fig:compound_reward_extra}\vspace{-2mm}
\end{figure}

\section{Examination of Existing Methods on Models Beyond Qwen}\label{app:ttrl_and_one_shot_rl}

\begin{figure}
    \centering
    \cblock{98}{62}{121}\qwenmath 
    \cblock{134}{117}{171}\qwenmathsmall  
    \cblock{83}{97}{54}\qwen  
    \cblock{143}{163}{100}\qwensmall  
    \cblock{165}{94}{138}\olmosft\\
    \cblock{200}{107}{144}\olmo 
    \cblock{121}{106}{70}\llamabase
    \cblock{195}{166}{92}\llamabasesmall  
    \cblock{75}{100}{130}\llama 
    \cblock{107}{142}{185}\llamasmall  \\
    \begin{subfigure}[t]{0.495\textwidth}
        \centering
        \includegraphics[width=0.49\linewidth]{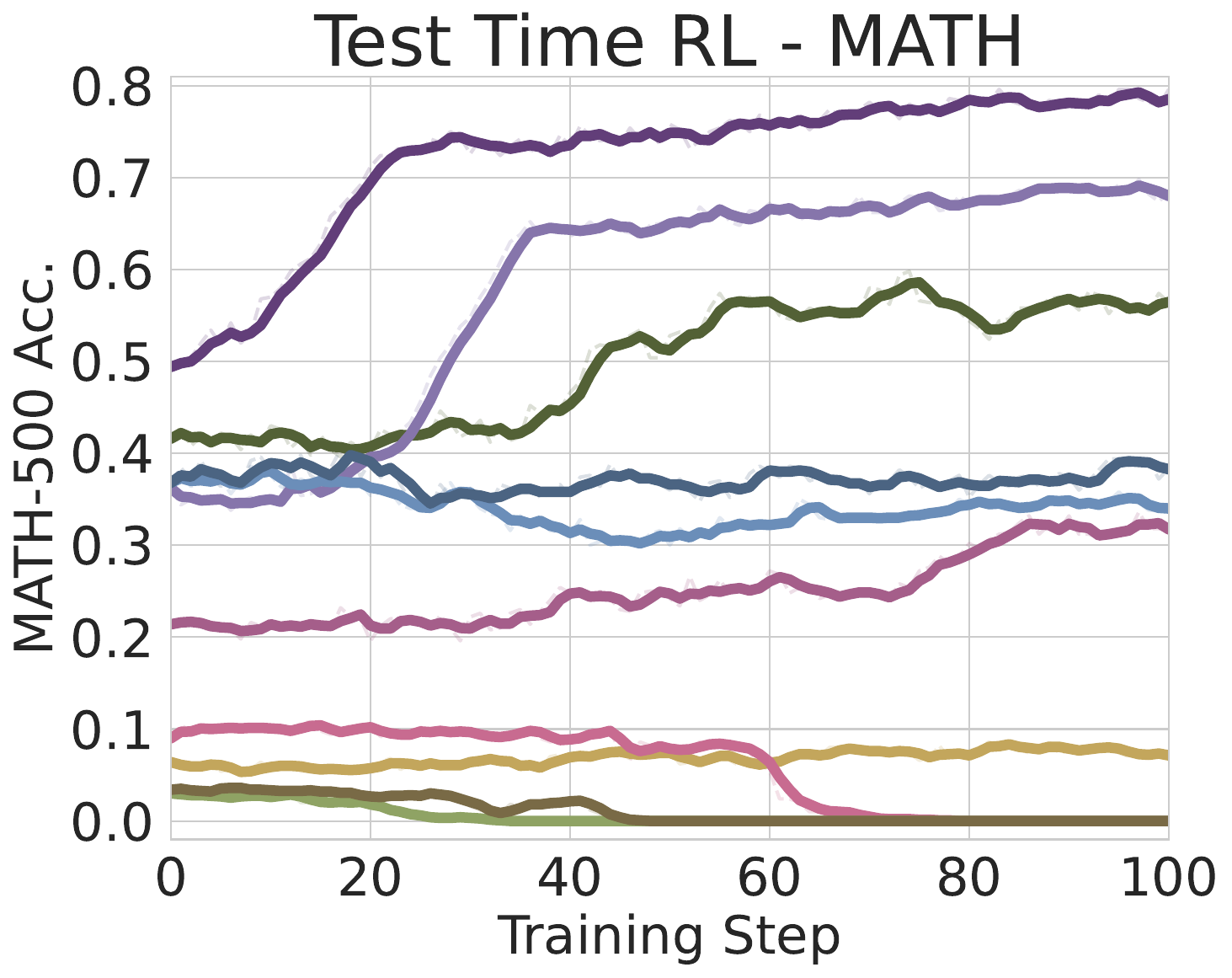}
        \includegraphics[width=0.49\linewidth]{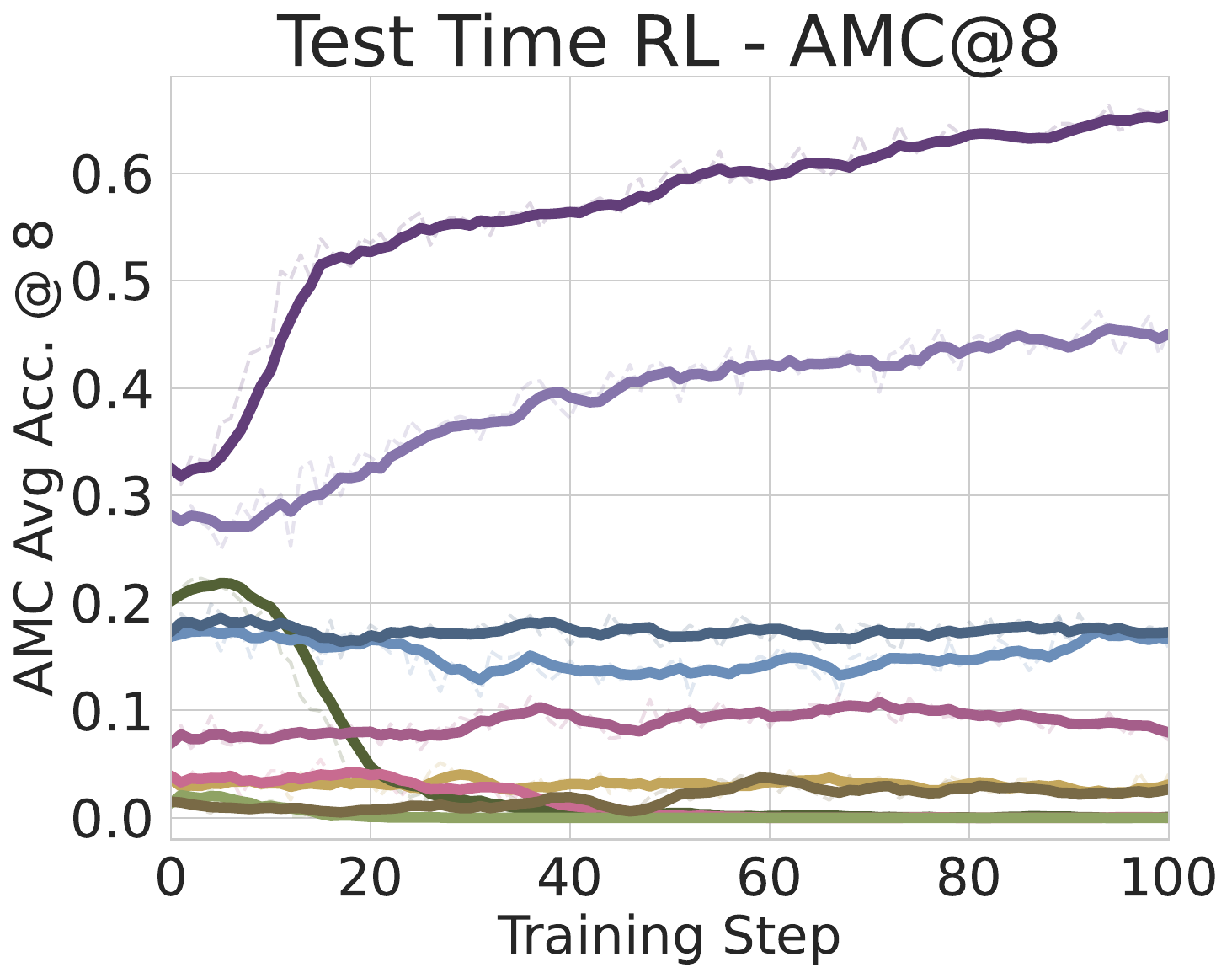}
        \caption{TTRL Results}
        \label{fig:llama3_results}
    \end{subfigure}%
    \hfill
    \begin{subfigure}[t]{0.495\textwidth}
        \centering
        \includegraphics[width=0.49\linewidth]{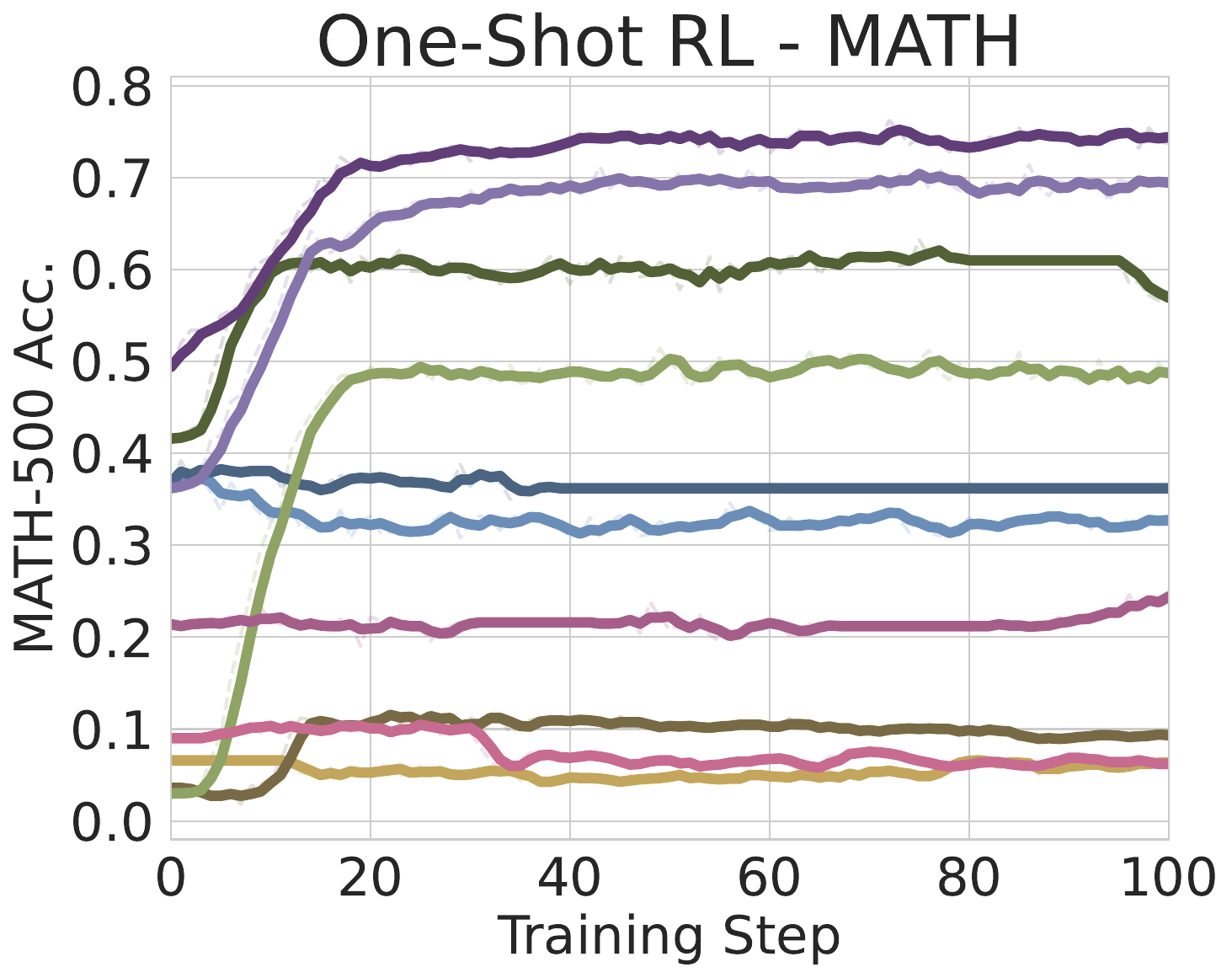}
        \includegraphics[width=0.49\linewidth]{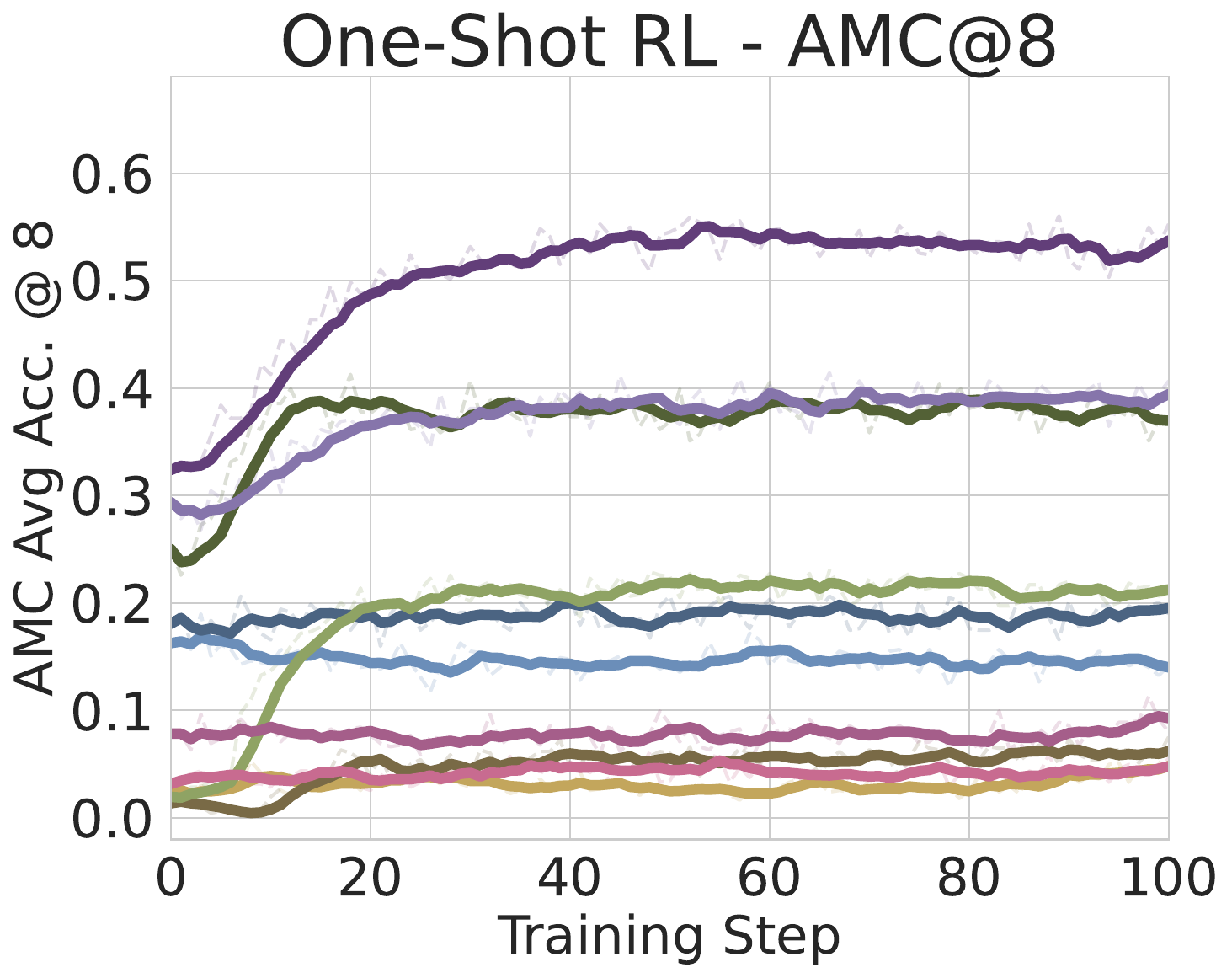}
        \caption{One-Shot RL Results}
        \label{fig:llama3_results}
    \end{subfigure}%
    \caption{We evaluate two recent weak supervision RL methods---TTRL~\citep{zuo2025ttrl} and One-Shot RL~\citep{wang2025reinforcement}---on diverse base models. We find that the proposed training rewards can consistently work on Qwen models.
    Yet with few exceptions, those same proposed signals often yield no gains on other model families, mirroring the limited generalization observed when training with our own spurious rewards. See Appendix~\ref{app:oneshot_ttrl_setup} for setup details. }
    \label{fig:ttrl_results}\vspace{-2mm}
\end{figure}

In this section, we examine the existing methods---TTRL and 1-shot RL---that apply weak supervision during RL, as shown in Figure~\ref{fig:ttrl_results}. 
We find both methods are significantly effective on Qwen models, but not others.

\section{Compound Rewards that Inhibit Code Reasoning}
\label{app:compound_results}

\begin{figure}[h]\vspace{-3mm}
    \centering
    \cblock{98}{62}{121} Qwen-Math-7B  
    \cblock{83}{97}{54} Qwen-7B  
    \cblock{165}{94}{138} Olmo2-7B-SFT  \\
    \parbox{7mm}{\includegraphics[width=\linewidth,height=1.5mm]{figures/line.png}}\hspace{1mm}Compound Reward\hspace{3mm}
    \parbox{7mm}{\includegraphics[width=\linewidth,height=1.5mm]{figures/dash.png}}\hspace{1mm}Original Reward
    \begin{subfigure}[t]{0.325\textwidth}
        \centering
        \includegraphics[width=\linewidth]{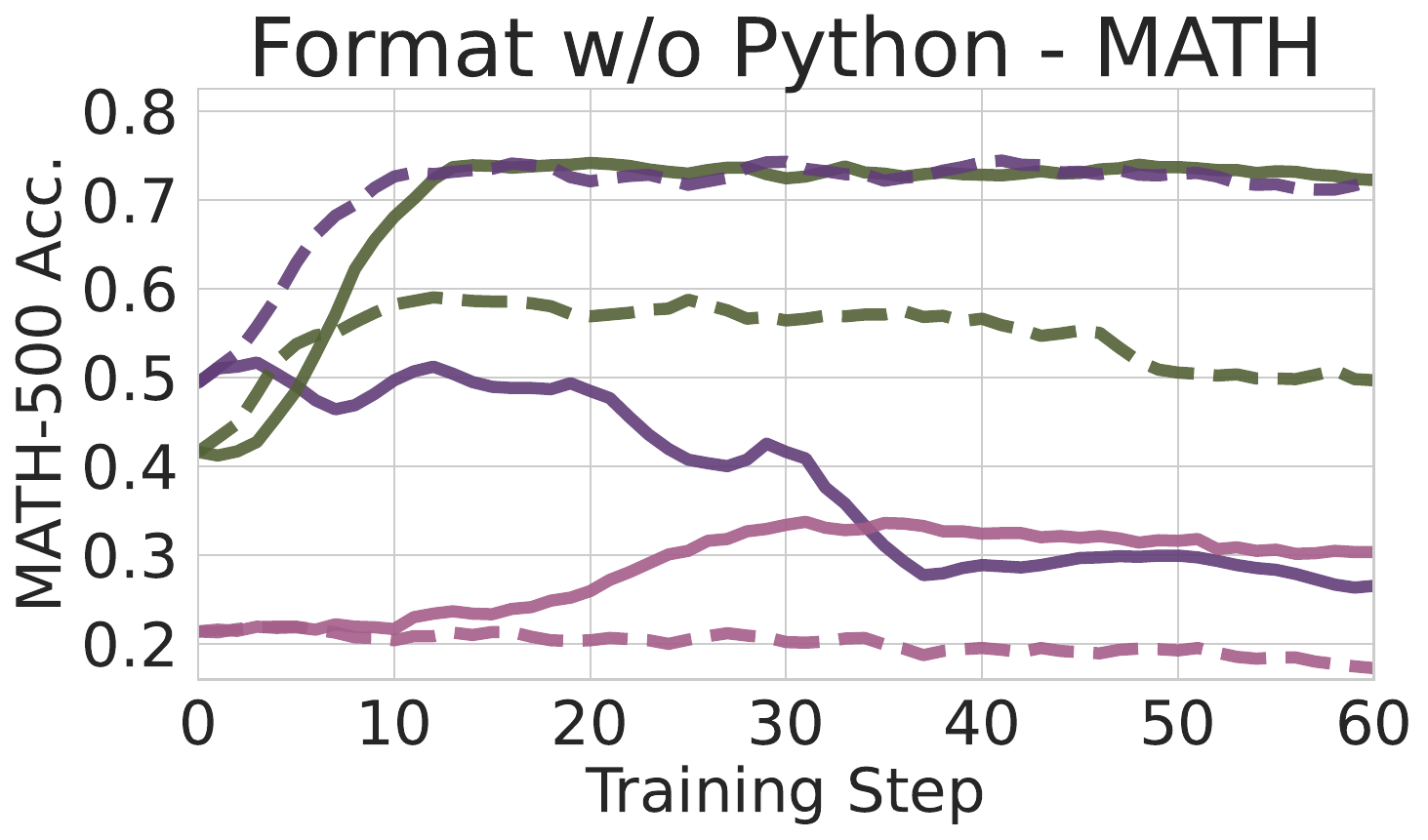}\\ 
        \includegraphics[width=\linewidth]{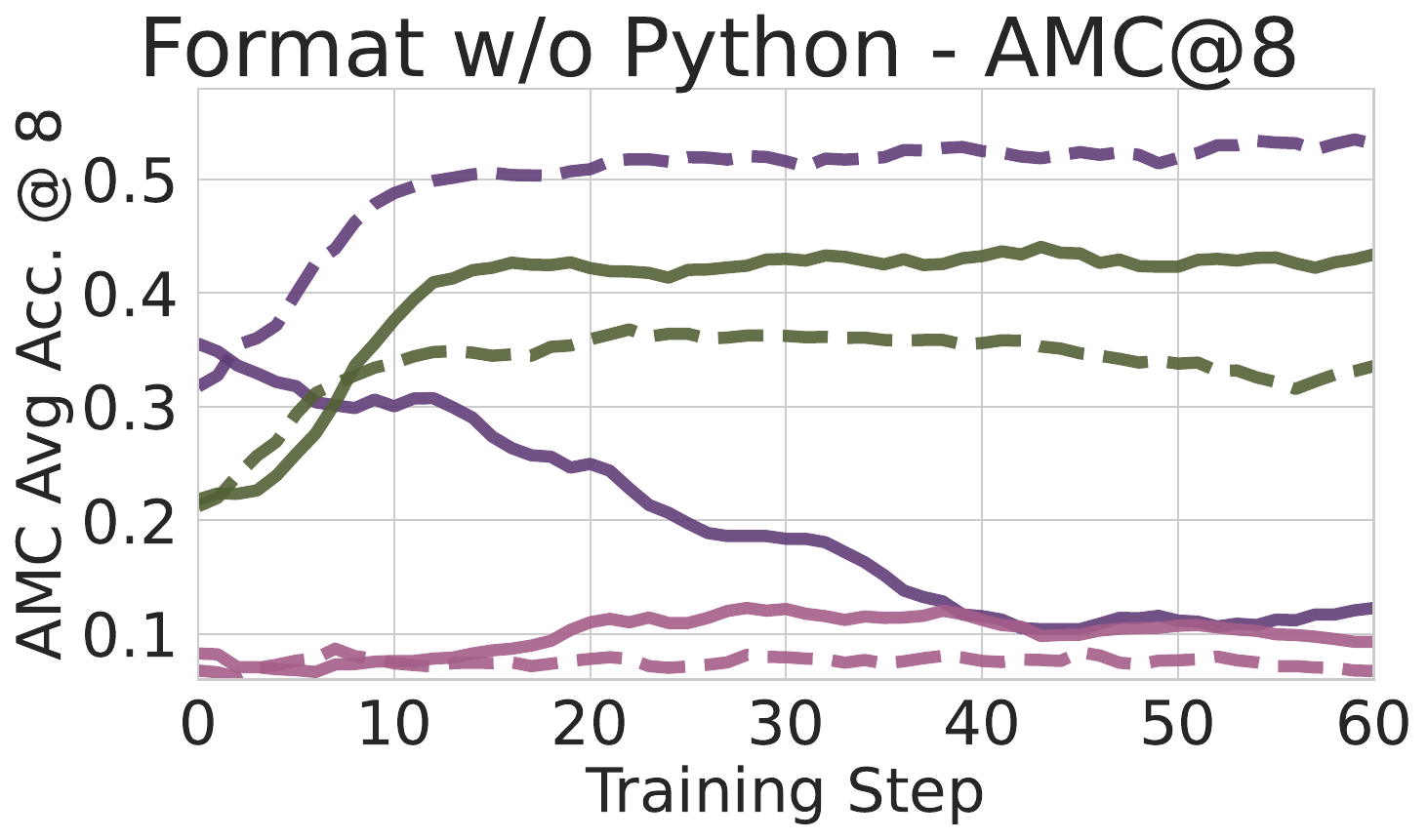}
    \caption{Format w/o Python}
    \end{subfigure}%
    ~
    \begin{subfigure}[t]{0.325\textwidth}
        \centering
        \includegraphics[width=\linewidth]{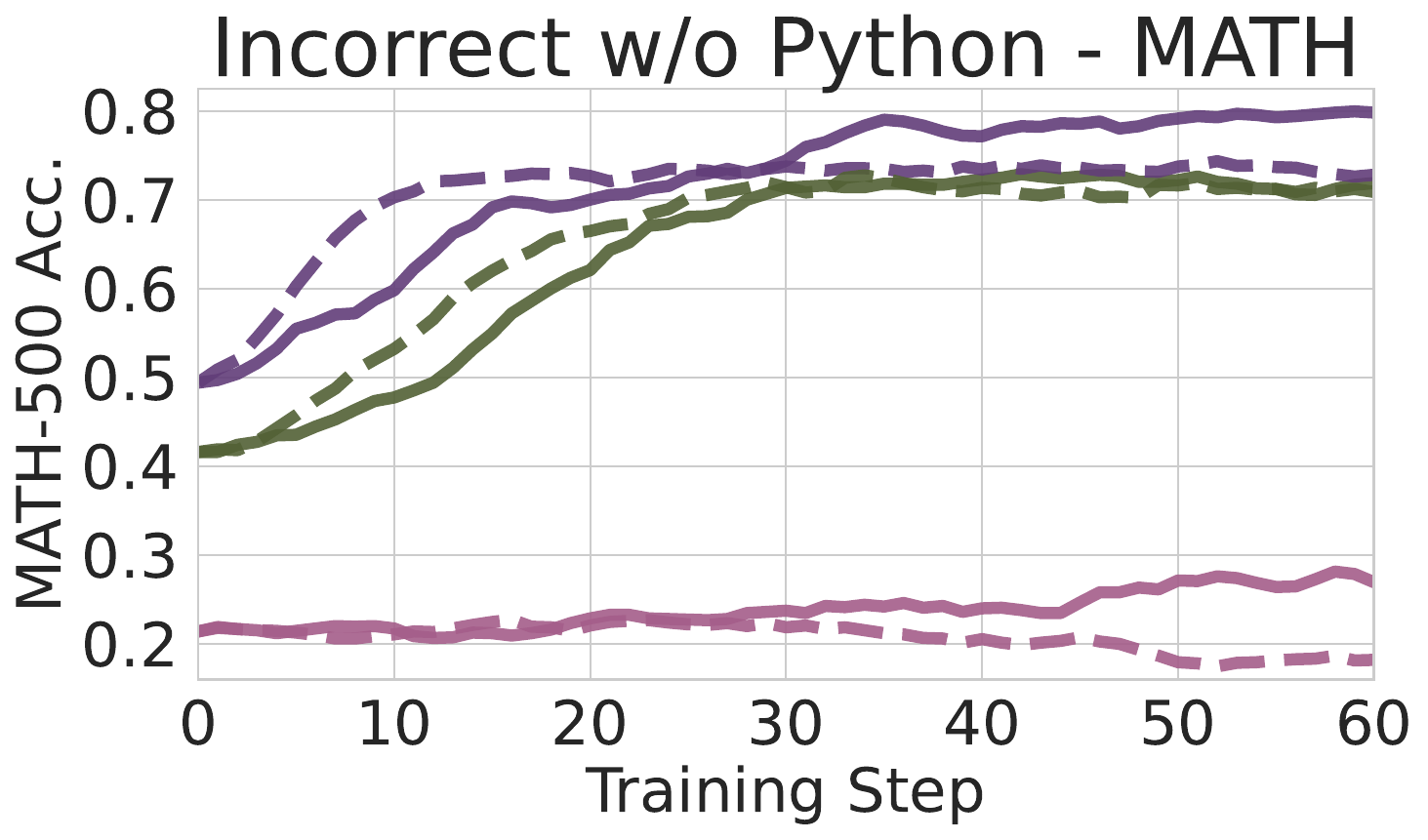}\\
        \includegraphics[width=\linewidth]{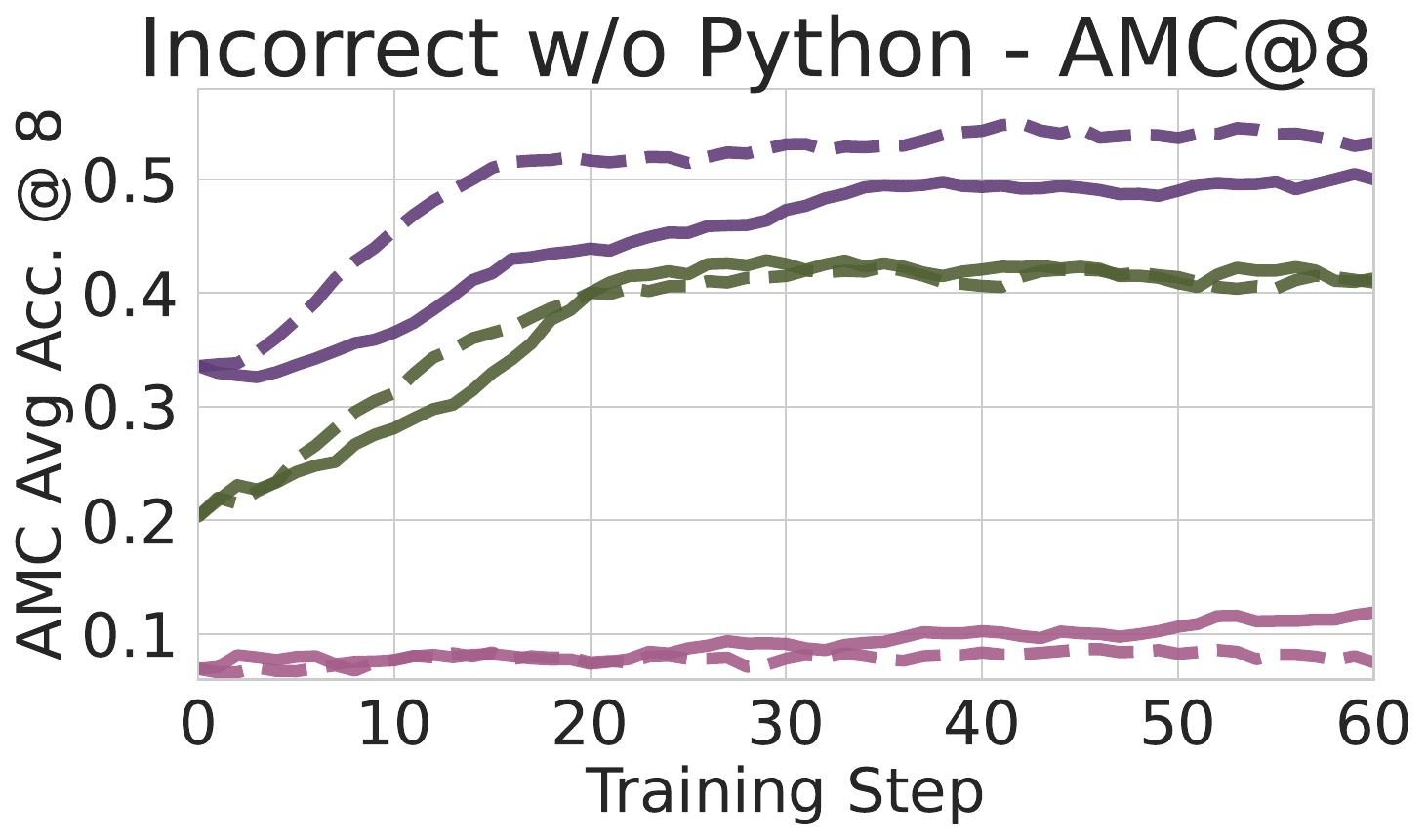}
    \caption{Incorrect w/o Python}
    \label{fig:compound_reward_incorrect}
    \end{subfigure}%
    ~
    \begin{subfigure}[t]{0.325\textwidth}
        \centering
        \includegraphics[width=\linewidth]{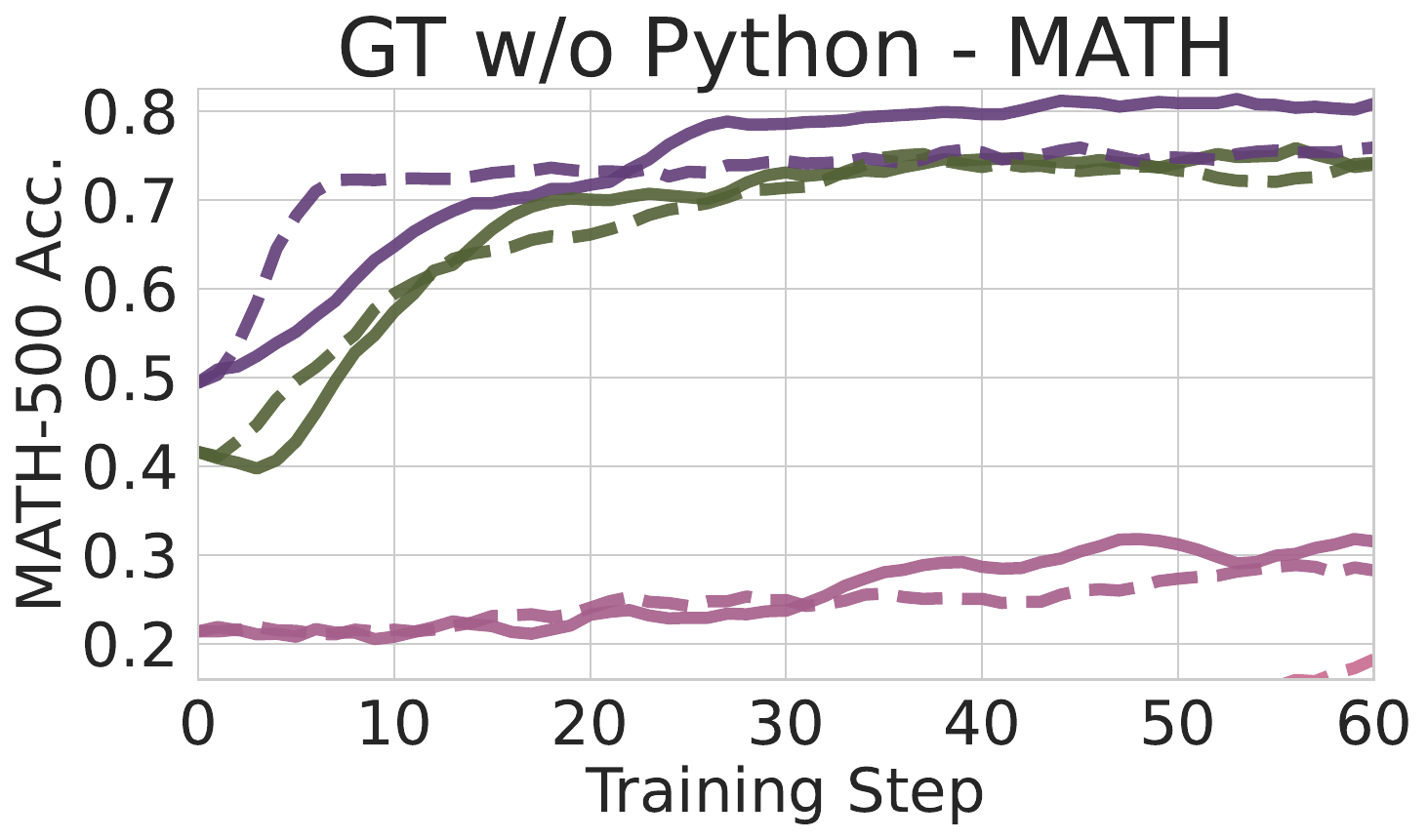}\\
        \includegraphics[width=\linewidth]{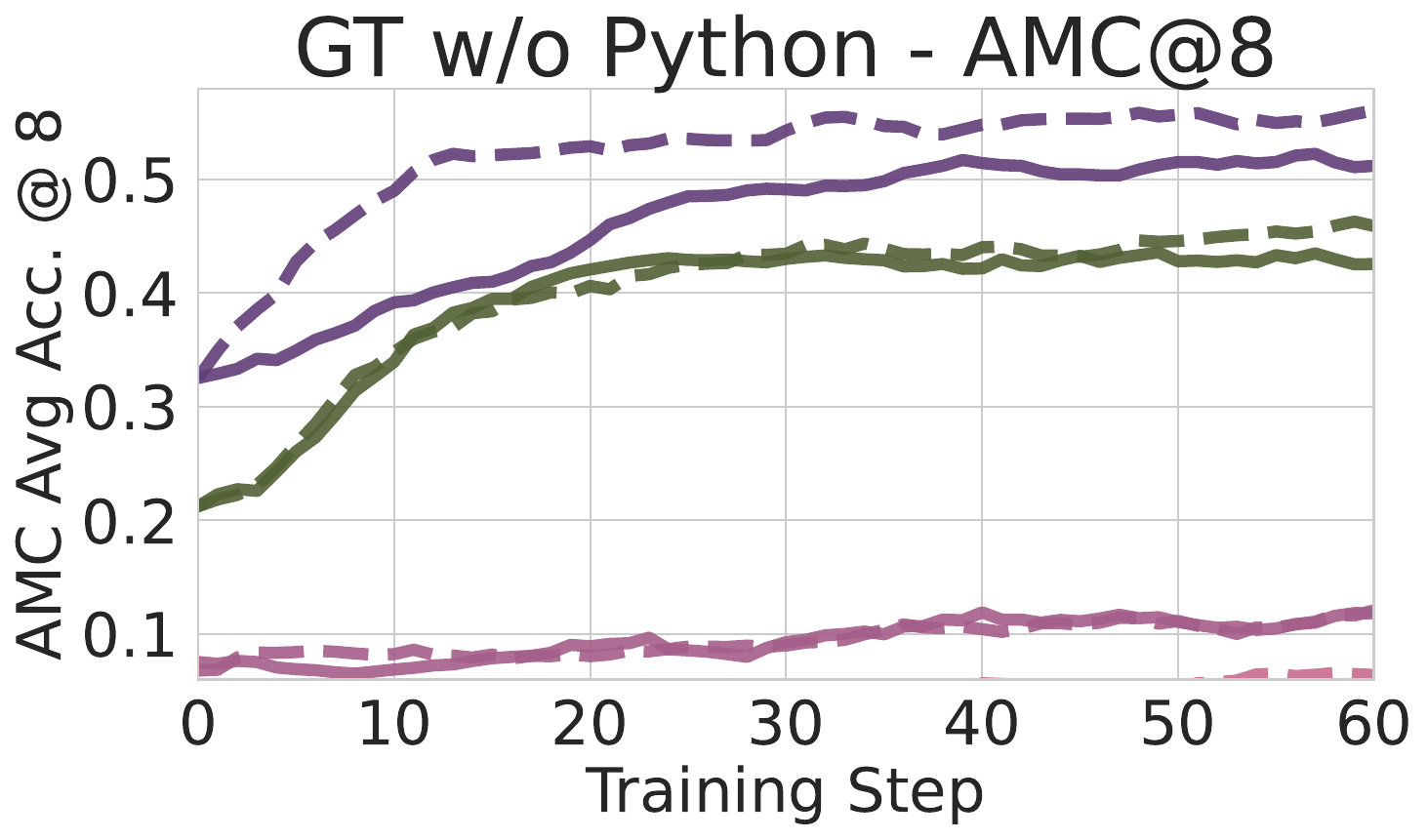}
    \caption{Ground Truth w/o Python}
    \label{fig:compound_reward_gt}
    \end{subfigure}
    \caption{RLVR with compound rewards that intersect (i) our original rewards with a (ii) \textit{no Python reward} that only rewards responses without Python code. 
    We defer corroborating results on AIME and more models to Appendix~\ref{app:compound_results}.
    }
    \label{fig:compound_reward}
\end{figure}

\paragraph{Inhibiting code reasoning during RLVR with spurious rewards can reduce gains on \qwenmath, but increase gains on other models.}
We hypothesized that code reasoning is part of the source of weak and spurious rewards gains.
Contrapositively, \textit{penalizing} code frequency could potentially reduce the gains of these rewards in \qwenmathfamily. To test this, we designed compound rewards that intersect each weak or spurious reward with a \textit{no Python reward}, so that a response is 
rewarded if and only if (1) it satisfies the original spurious reward condition and (2) it does not contain the string ``\texttt{python}''.
Consistent with our hypothesis, format rewards cease to improve \qwenmath when paired with the no code reward (Figure~\ref{fig:compound_reward_gt}). 
The spurious incorrect compound reward matches the original incorrect reward on MATH-500; on more difficult benchmarks such as AMC and AIME, gains also persist but are reduced (Figures~\ref{fig:compound_reward_incorrect},~\ref{fig:compound_reward_extra}). Thus, while ablating code reasoning does hurt the incorrect reward as predicted, we posit that other beneficial behaviors (e.g., reduced repetition, see Appendix~\ref{app:analysis_repetition}) may still be superficially elicited.
In addition, ground‑truth rewards still yield gains, suggesting benefits beyond eliciting code reasoning (consistent with the code‑frequency trends in Figure~\ref{fig:analysis_code_freq}).

Intriguingly, for \emph{Bad-Code} models \qwen and \olmosft, compound rewards often \textit{outperform} the originals. Most strikingly, \olmosft---which degrades under standalone format or incorrect rewards---gains +8.9 and +5.5 points, respectively, once the no-code reward is added.
We hypothesize that this is because \qwen and \olmosft exhibit weak code reasoning before RLVR training, so compound rewards explicitly downweight a behavior that is suboptimal for these models.

In Figure~\ref{fig:compound_reward_extra}, we present additional results for our compound rewards on (1) more models (\qwensmall, \qwenmathsmall) and (2) more benchmarks (AIME24, AIME25). Overall, results corroborate our analysis in Section~\ref{sec:analysis:intervene}; \qwenmathsmall follows the same overall trends as \qwenmath. Compound rewards also continue to benefit \qwensmall, where gains are comparable to the gains from using the original reward. This slightly contrasts our observations in \qwen, where compound rewards often yielded stronger gains. We conjecture that \qwensmall is stronger at code reasoning, which is consistent with our finding in Section~\ref{sec:analysis:intervene} that inducing code reasoning via prompting improves performance in in \qwensmall but not \qwen. We are unsure of the exact reason for this discrepancy between these two models (which have the same pretraining data) and leave it for future work.

\section{Beyond Code Reasoning: Another Beneficial Pattern that RLVR Can Easily Elicit}\label{app:analysis_repetition}
In the main paper, we show that RLVR with spurious rewards can improve \qwenmathfamily's performance by surfacing useful reasoning patterns learned during pretraining, and we use code reasoning as a standout example. We note that code reasoning is one distinctive reasoning representation, but not the only one. In this section, we briefly discuss another pattern---no repetition---that can also be easily elicited by RLVR in addition to code reasoning.

We observe that \qwenmathfamily models have a relatively higher tendency to produce repetitive outputs compared to Llama3 and Olmo2 models. We find that answers with code reasoning often do not have this issue. Therefore, we further study the effect of purely discouraging repetition in answers.
To study this, we design a new \textit{repetition reward} that returns a score of 0 when the answer contains any string repeated more than 10 times. Otherwise, it rewards the model with a full score of 1.

\begin{figure}[t]
    \centering
    \cblock{98}{62}{121}\qwenmath 
    \cblock{134}{117}{171}\qwenmathsmall  
    \cblock{83}{97}{54}\qwen  
    \cblock{143}{163}{100}\qwensmall  
    \cblock{165}{94}{138}\olmosft\\
    \cblock{200}{107}{144}\olmo 
    \cblock{121}{106}{70}\llamabase
    \cblock{195}{166}{92}\llamabasesmall  
    \cblock{75}{100}{130}\llama 
    \cblock{107}{142}{185}\llamasmall  \\
    \begin{subfigure}[t]{0.48\textwidth}
        \centering
        \includegraphics[width=\linewidth]{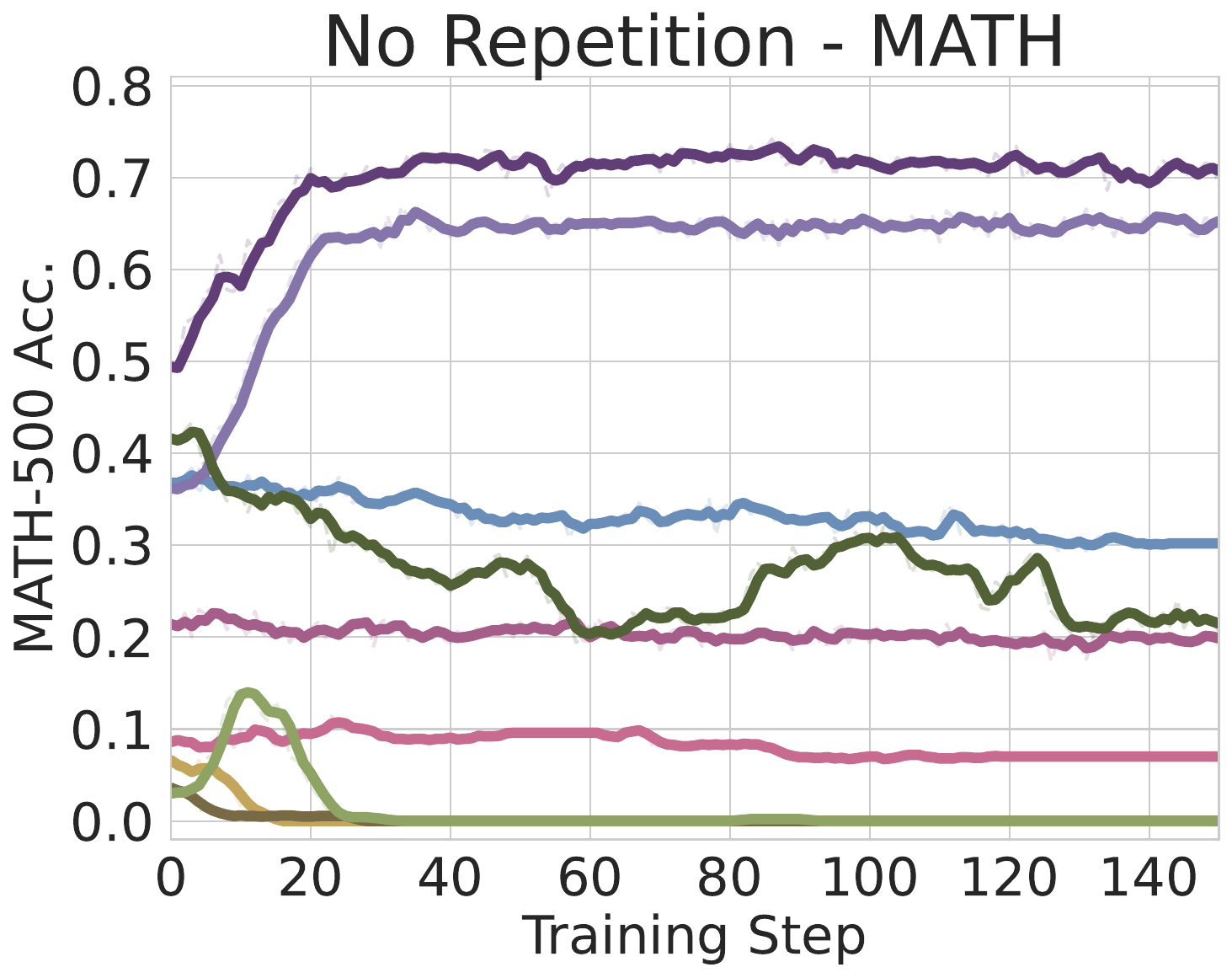}\vspace{-2mm}
    \caption{MATH-500}
    \label{fig:llama3_results}
    \end{subfigure}%
    \hfill
    \begin{subfigure}[t]{0.48\textwidth}
        \centering
        \includegraphics[width=\linewidth]{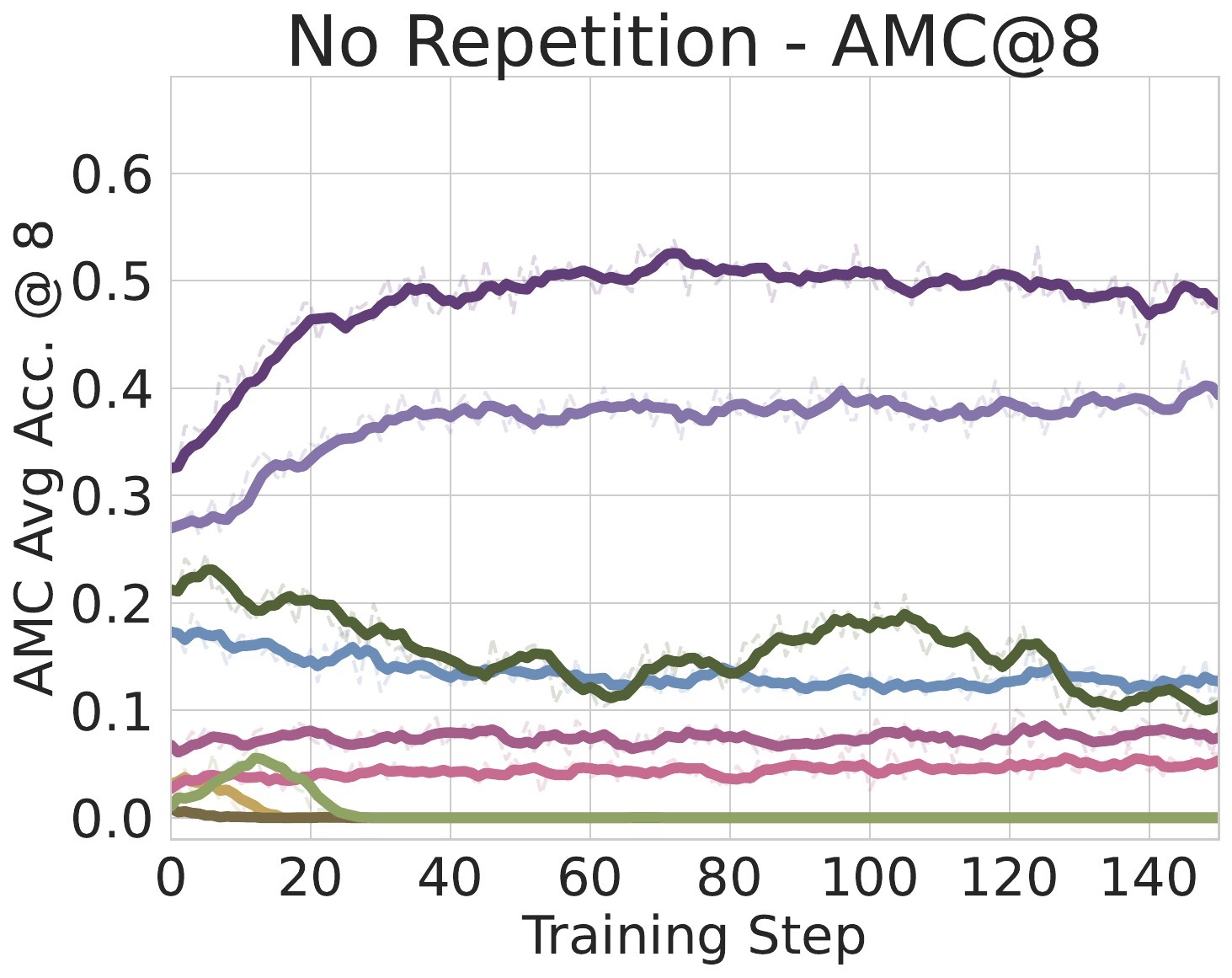}\vspace{-2mm}
    \caption{AMC Avg. @8}
    \label{fig:llama3_results}
    \end{subfigure}%
    \hfill
    \begin{subfigure}[t]{0.48\textwidth}
        \centering
        \includegraphics[width=\linewidth]{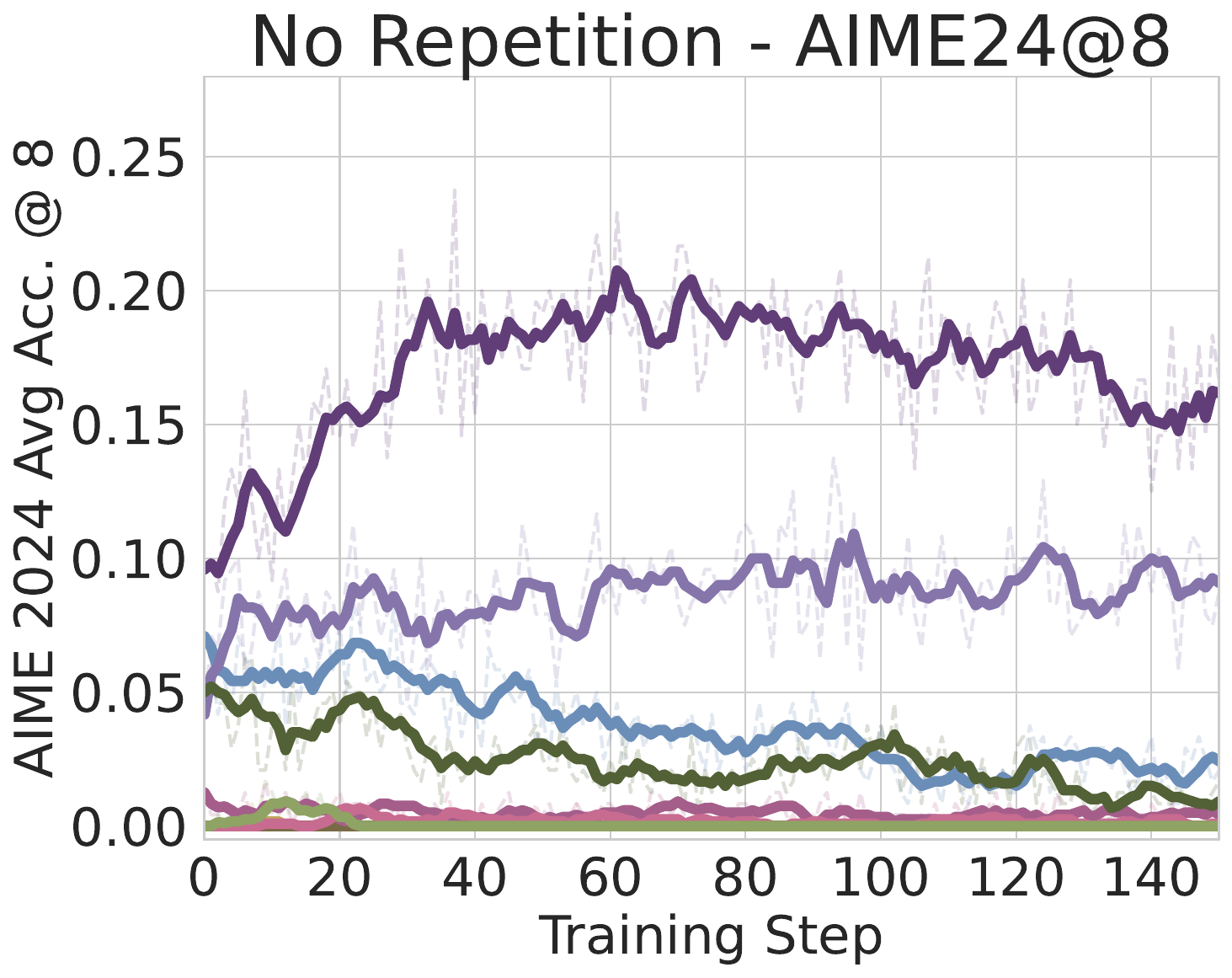}\vspace{-2mm}
    \caption{AIME 2024 Avg. @8}
    \label{fig:llama3_results}
    \end{subfigure}%
    \hfill
    \begin{subfigure}[t]{0.48\textwidth}
        \centering
        \includegraphics[width=\linewidth]{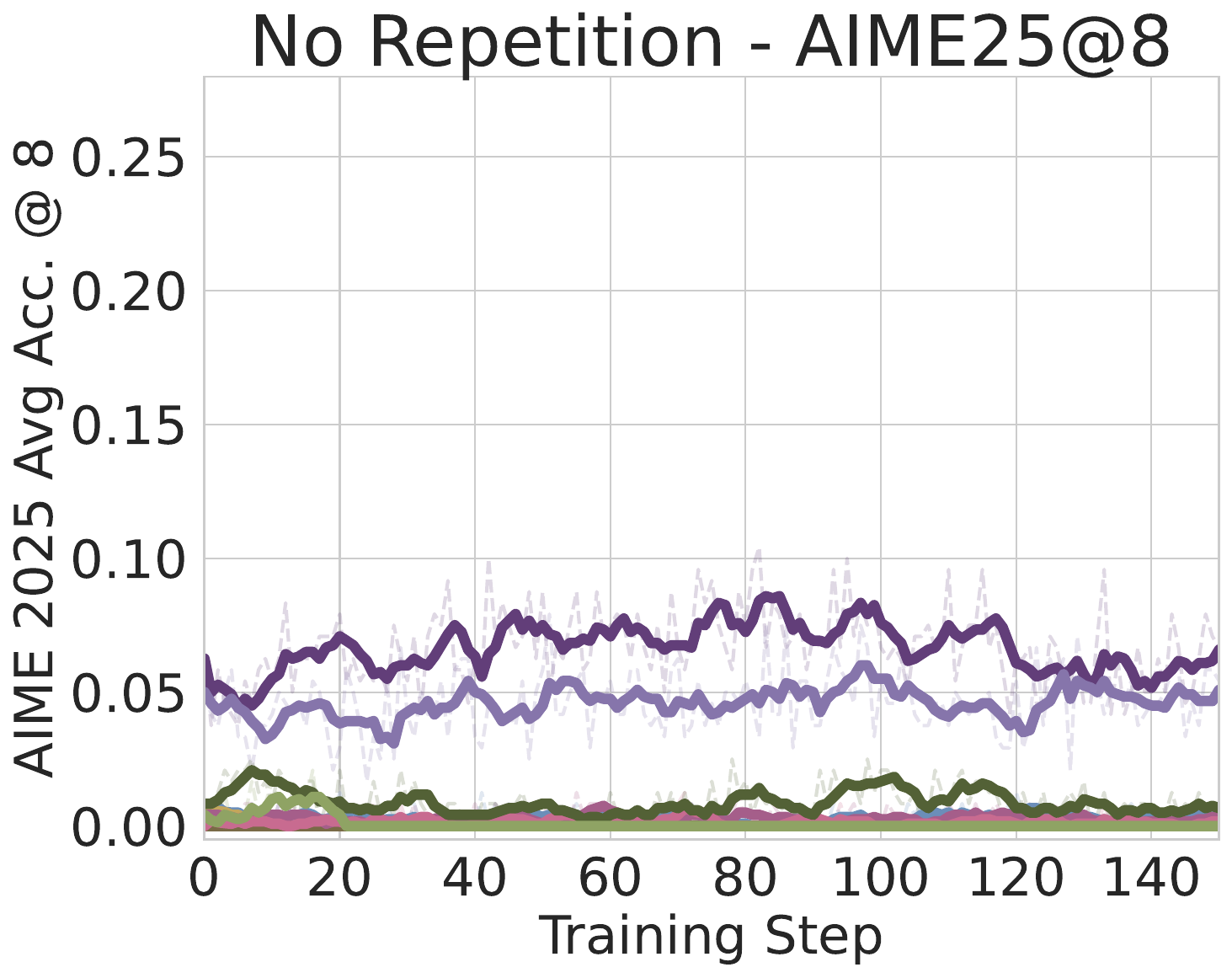}\vspace{-2mm}
    \caption{AIME 2025 Avg. @8}
    \label{fig:llama3_results}
    \end{subfigure}%
    \caption{
    We design a new type of reward, \emph{no-repetition reward}, which assign a score of 1 to responses that do not contain obvious repetition and 0 to responses that contain obvious string repetition. We find, no-repetition reward effectively improves the performance of \qwenmathfamily, while not others.
    }
    \label{fig:no_repetition}\vspace{-2mm}
\end{figure}

As shown in Figure~\ref{fig:no_repetition}, the RLVR no-repetition reward improves the performance of \qwenmath and \qwenmathsmall on MATH and AMC. We observe minimal or even negative improvement on other models.
Based on our findings, we hypothesize that various patterns exist whose presence correlates with answer correctness. These patterns, including code reasoning and no repetition, can be easily elicited by RLVR even when the rewards provide no information about the ground-truth answers. Still, the effectiveness of eliciting these patterns is heavily model-dependent.

\section{Switch of Reasoning Strategies during RLVR}\label{app:strategy_switch}

In this section, we present further analysis on the change of reasoning strategies during RLVR for \qwenmathfamily and \qwenbasefamily models.

\begin{figure}[t]
    \centering
    \begin{subfigure}[t]{0.198\textwidth}
        \centering
        \includegraphics[width=\linewidth]{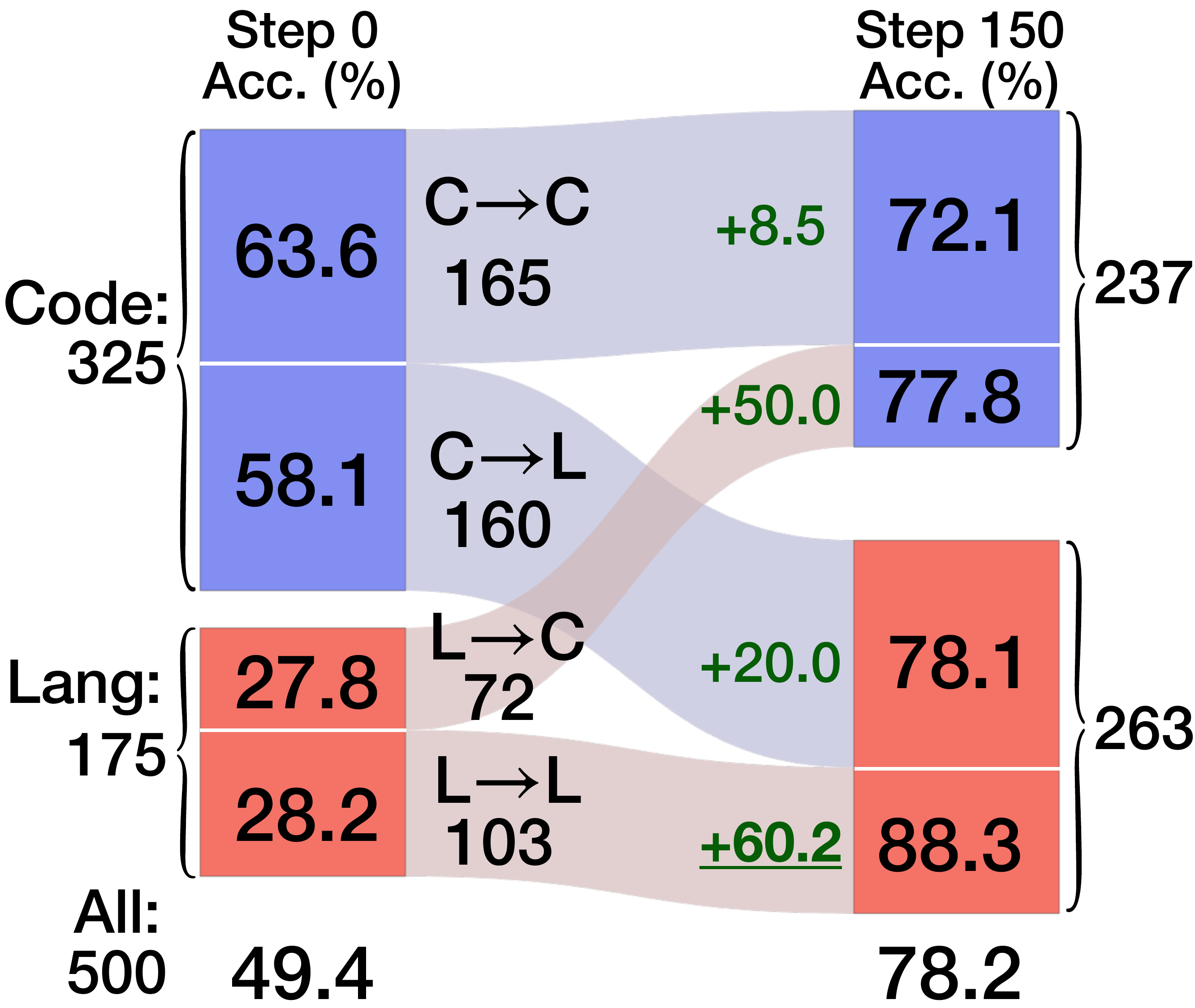}
    \caption{Ground Truth}
    \label{fig:qwen_math_1.5b_code_reward}
    \end{subfigure}%
    \hfill
    \begin{subfigure}[t]{0.198\textwidth}
        \centering
        \includegraphics[width=\linewidth]{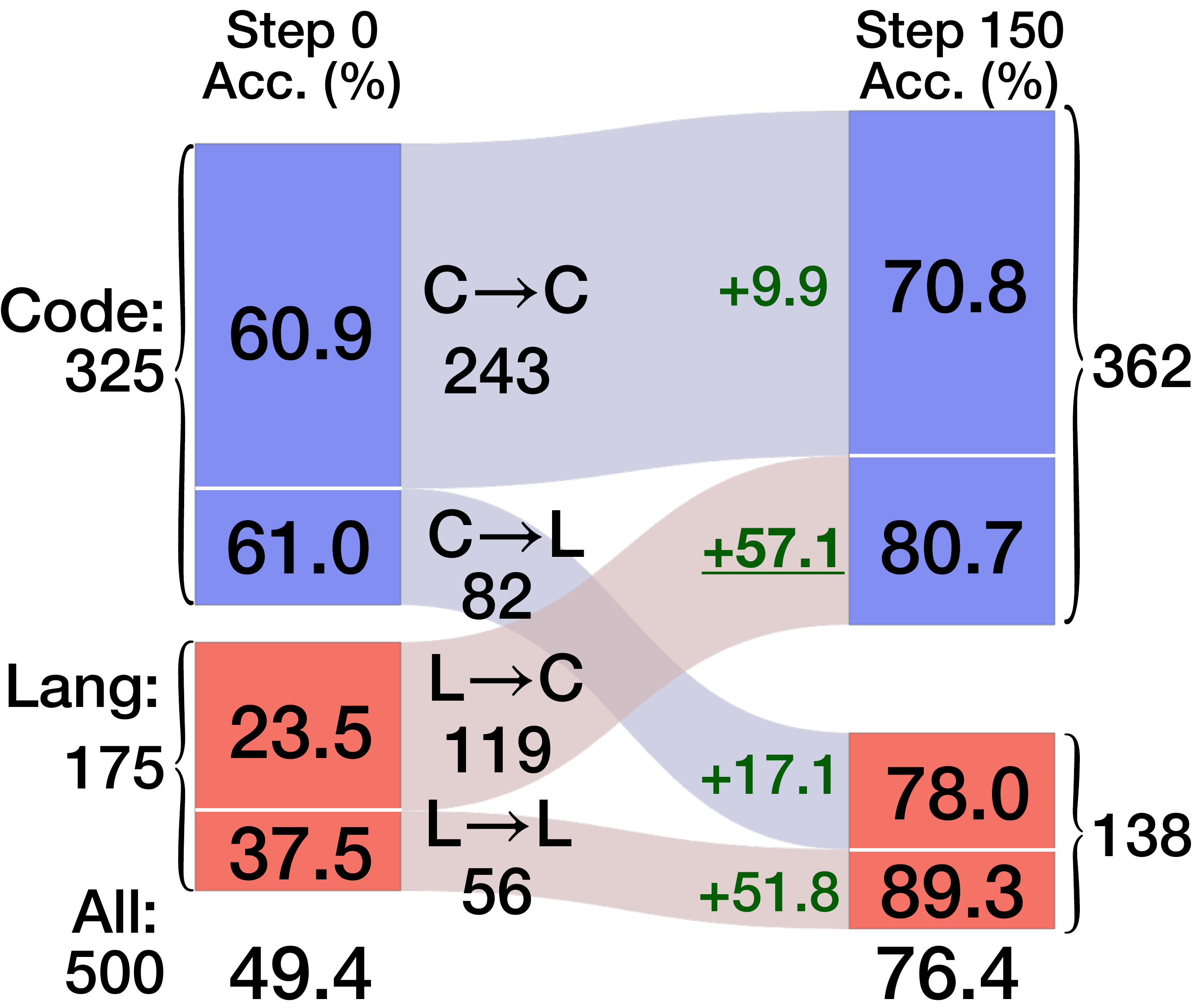}
    \caption{Majority Vote}
    \label{fig:qwen_math_7b_code_reward}
    \end{subfigure}%
    \hfill
    \begin{subfigure}[t]{0.198\textwidth}
        \centering
        \includegraphics[width=\linewidth]{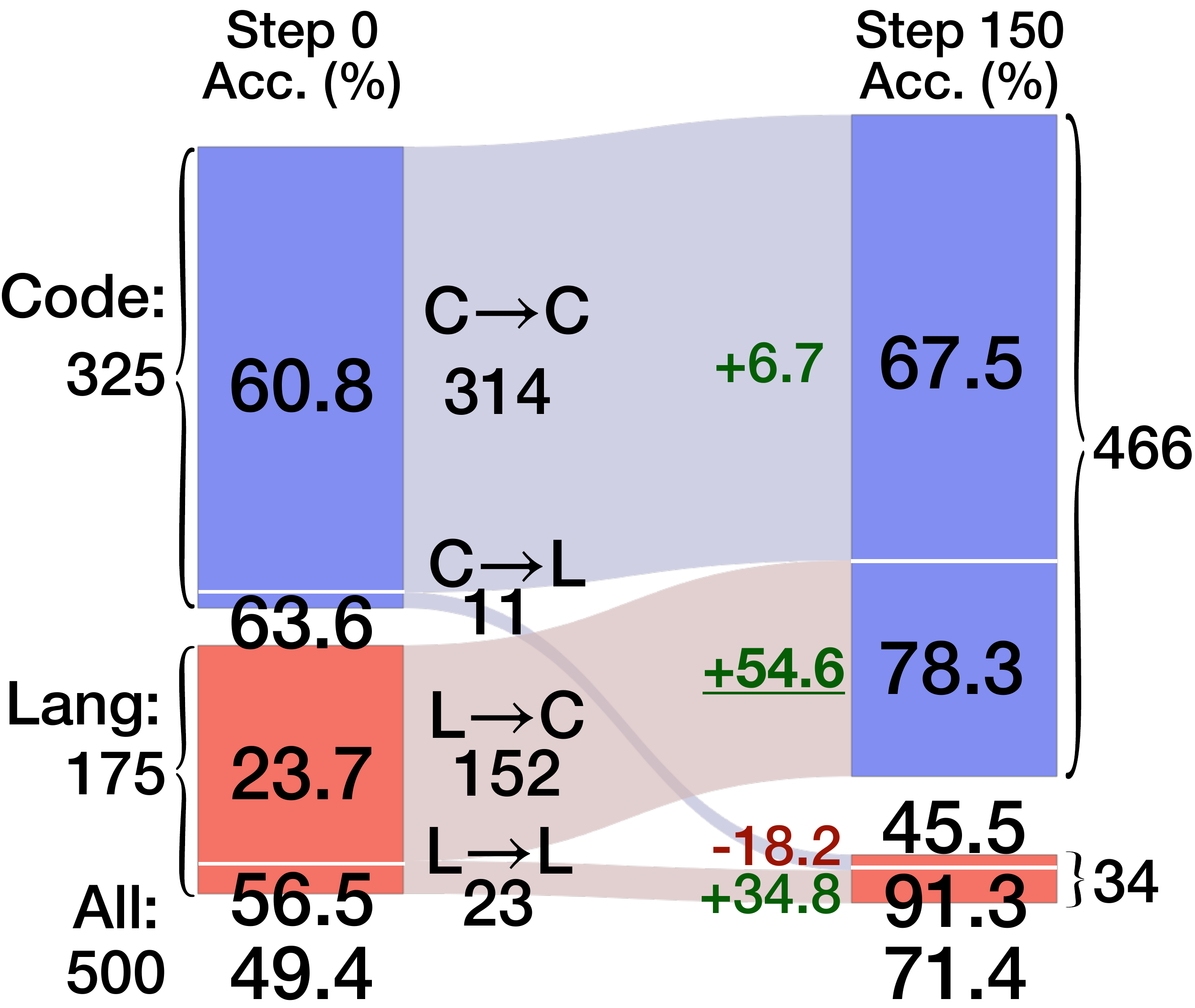}
    \caption{Format}
    \label{fig:qwen_7b_code_reward}
    \end{subfigure}%
    \hfill
    \begin{subfigure}[t]{0.198\textwidth}
        \centering
        \includegraphics[width=\linewidth]{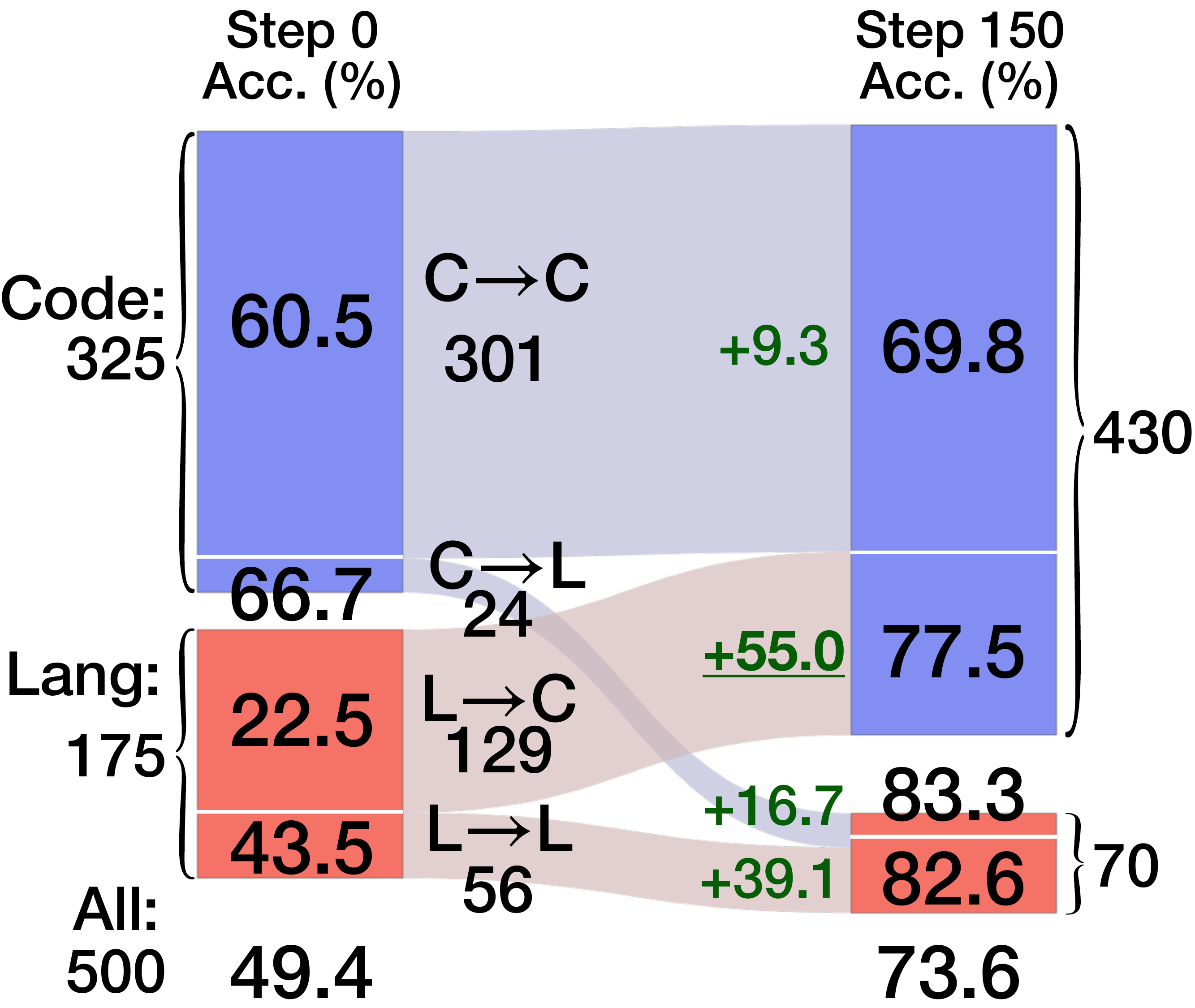}
    \caption{Incorrect}
    \label{fig:qwen_1.5b_code_reward}
    \end{subfigure}%
    \hfill
    \begin{subfigure}[t]{0.198\textwidth}
        \centering
        \includegraphics[width=\linewidth]{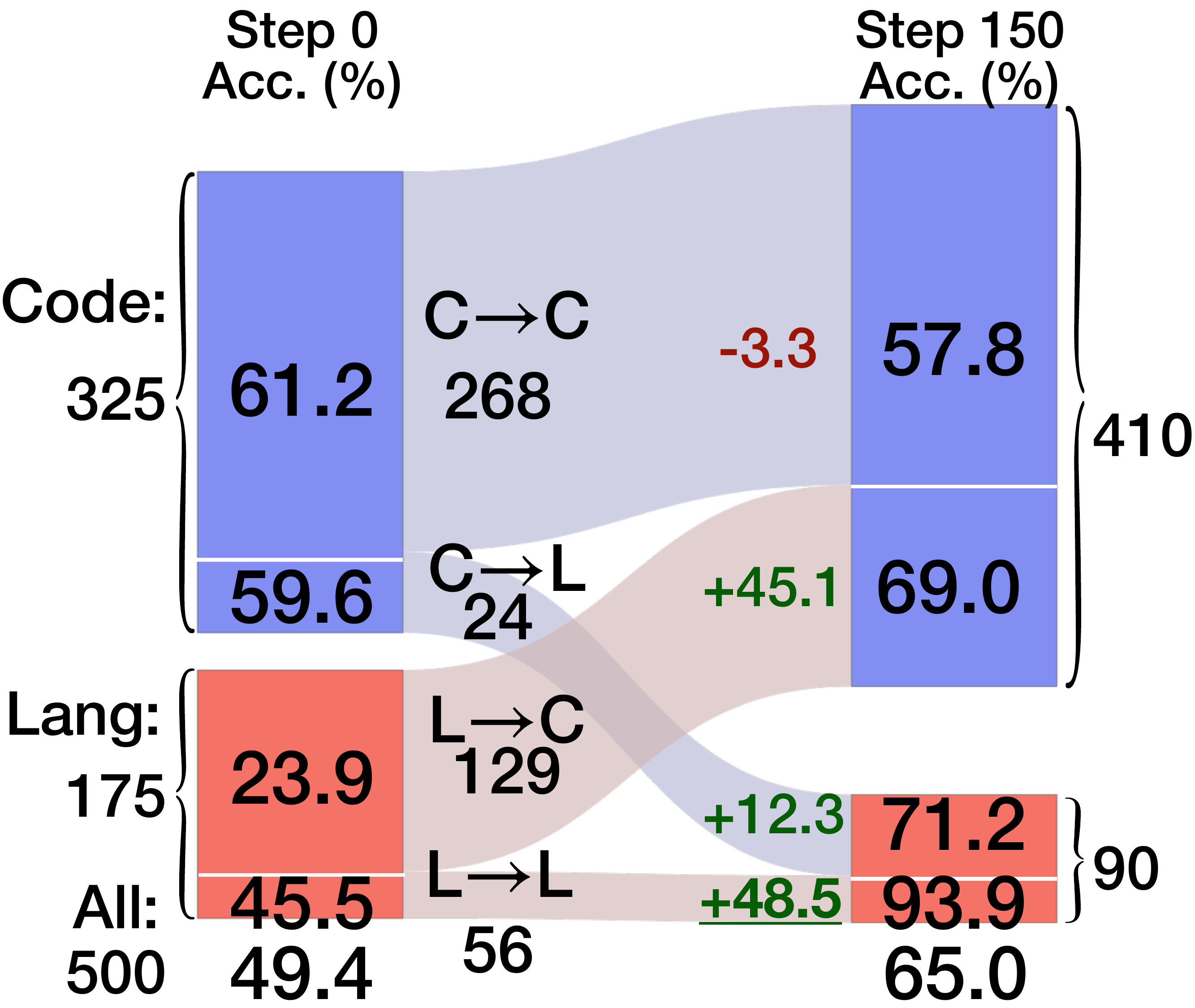}
    \caption{Random}
    \label{fig:olmo_sft_code_reward}
    \end{subfigure}
    \caption{Reasoning strategy switching and fine-grained performance of \qwenmath on the MATH-500 test set before and after RLVR with different training signals.
    \textcolor{blue}{Blue} labels are the problem for which the model uses code reasoning, while \textcolor{red}{red} labels indicate reasoning traces using only natural language.
    Accuracy of each disjoint subset of problems before and after RLVR is shown in the shaded ends, and the size of each subset is shown in the lightly shaded region along with the change in accuracy. 
    For all weak and spurious rewards, the model tends to use more code reasoning after RLVR. There is a small proportion of originally-code-reasoning problems being switched to language reasoning (Code$\rightarrow$Lang); a majority of originally-language-reasoning problems convert to code reasoning (Lang$\rightarrow$Code), on which we see the most significant performance increase after RLVR.
    }
    \label{fig:flow_rates}
\end{figure}

\paragraph{Reasoning strategy switches during RLVR.} 
\qwenmath's accuracy increases by an average of $23.5$ absolute points across different training signals.
To further break down this gain, we track the performance of models trained on each training signal across four disjoint subsets of the test prompts:
\textbf{(1) Code$\rightarrow$Code}: the model uses code reasoning both before and after RLVR; 
\textbf{(2) Code$\rightarrow$Lang}: the model initially uses code reasoning but switches to natural language reasoning;
\textbf{(3) Lang$\rightarrow$Code}: the model initially uses natural language reasoning but switches to code reasoning; and
\textbf{(4) Lang$\rightarrow$Lang}: the model uses natural language reasoning both before and after RLVR.
Specifically, we focus on two interconnected metrics: frequency and accuracy of each subset.
To systematically quantify the contribution of each subset to the performance gain, we define a \emph{Partial Contribution Score}, $C_d$, for any subset $d\subseteq\mathcal{D}$ of the entire test set $\mathcal{D}$, such that $C_d$ is the ratio between the net increase in the number of correctly answered problems in $d$ divided by the net increase in the number of correctly answered problems in $\mathcal{D}$:
\begin{equation*}
    C_d = \frac
{\sum_{x\in d}\mathbb{I}[\text{correct}(x_t)] - \mathbb{I}[\text{correct}(x_0)]}
{\sum_{x\in \mathcal{D}}\mathbb{I}[\text{correct}(x_t)] - \mathbb{I}[\text{correct}(x_0)]},\hspace{2mm}\text{where }x_0\text{ and }x_t\text{ are the initial and final answers.}
\end{equation*}

\paragraph{Frequency:} Figure~\ref{fig:flow_rates} shows \qwenmath's reasoning strategy switching pattern. For all weak and spurious rewards, the model uses more code reasoning after RLVR. While few originally-code-reasoning problems switch to language reasoning (Code$\rightarrow$Lang), most originally-language-reasoning problems convert to code reasoning (Lang$\rightarrow$Code). Ground truth reward does not follow this pattern.
For \emph{Bad-Code} models (\qwen and \olmosft), meaningful rewards steer models away from bad code reasoning. Code reasoning decreases with ground truth, majority vote, and incorrect rewards for \qwen, but only with ground truth and majority vote for \olmosft (Figures~\ref{fig:code_freq_qwen} and \ref{fig:code_freq_olmosft}).
For \emph{No-Code} models, RLVR fails to elicit meaningful changes in reasoning strategy, as this capability is likely not learned during pre-training.

\begin{table}[h]
    \centering
    \caption{Partial contribution to the overall performance gain averaged over rewards that successfully steered the model's reasoning strategy (Figure~\ref{fig:analysis_code_freq}).}
    \small
    \begin{tabular}{lrrr}
    \toprule
        \textbf{Model}       & \textbf{\qwenmath} & \textbf{\qwenmathsmall} & \textbf{\qwen} \\
        \midrule
        \textbf{Avg. Total Gain} & $\uparrow$ 23.5\% & $\uparrow$ 28.5\% & $\uparrow$ 30.6\% \\
        \midrule
        $\mathbf{C}_{\textbf{Code$\rightarrow$Code}}$ & 11.6\% &  2.8\% & 0.2\% \\
        $\mathbf{C}_{\textbf{Code$\rightarrow$Lang}}$ &  8.6\% &  2.0\% & \textbf{93.9\%} \\
        $\mathbf{C}_{\textbf{Lang$\rightarrow$Code}}$ & \textbf{58.3\%} & \textbf{78.7\%} & 0.0\% \\
        $\mathbf{C}_{\textbf{Lang$\rightarrow$Lang}}$ & 21.4\% & 16.5\% &  5.9\% \\
    \bottomrule
    \end{tabular}
    \label{tab:partial_contribution}
\end{table}

\textbf{Accuracy:} 
From Figure~\ref{fig:flow_rates}, there is a drastic increase in accuracy in the Lang$\rightarrow$Code subset after RLVR across all training signals. This is reflected in Table~\ref{tab:partial_contribution}, which shows that 58.3\% of the performance gain of \qwenmath is from this subset. Similarly in \qwenmathsmall, switching from natural language reasoning to code reasoning contribute to 78.7\% of the performance gain. 
For the \emph{Bad-Code} models, Code$\rightarrow$Lang contributes to 93.9\% of the performance gain of \qwen. This is intuitive, as the model has a higher langauge reasoning accuracy than code reasoning accuracy, RLVR training essentially encourages the model to use the reasoning strategy that it is better at. 
For \emph{No-Code} models, since there is no code reasoning before or after RLVR, all performance gains (or losses) are from the \textbf{Lang$\rightarrow$Lang} subset.
These results suggest that much of the accuracy gain from RLVR on these spurious rewards in \qwenmathfamily and Qwen2.5 is simply from eliciting the right reasoning strategy from the model.

\section{Spurious Prompts: \qwenmath's Unreasonably High Sensitivity to Prompts }\label{app:prompt_effects}

\begin{table}[t]
    \centering
    \scalebox{0.9}{\begin{tabular}{@{\hskip 1mm}c@{\hskip 0mm}c@{\hskip 0mm}c@{\hskip 1mm}}
    \toprule
        \textbf{Prompt Name} & \textbf{System Prompt} & \textbf{User Prompt} \\
        \midrule
        \textbf{Qwen Default} & Please reason step by step, and put                          &  \multirow{2}{*}{\{\}} \\
        \textbf{\citep{yang2024qwen2}}         & your final answer within \texttt{\textbackslash boxed\{\}}.  & \\
        \midrule
        \textbf{Math Problem} & You are a helpful Assistant. & Math Problem: \{\}\\
        \midrule
        \textbf{Simplerl-zoo} & \multirow{2}{*}{You are a helpful Assistant.} & \{\}\textbackslash nPlease reason step by step, and put \\
        \textbf{\citep{zeng2025simplerl}} & & your final answer within \texttt{\textbackslash boxed\{\}}.\\
        \midrule
        \multirow{11}{*}{\makecell{\textbf{Sober} \\ \textbf{\citep{hochlehnert2025soberlookprogresslanguage}}}}
         & 
         \multirow{11}{*}{\makecell{Please reason step by step, \\
         and put your final answer \\within \texttt{\textbackslash boxed\{\}}.
         \footnote{Note that the original paper did not include a system prompt. However, when applying chat templates to a conversation that lacks a system prompt (as occurs in the code from \citet{hochlehnert2025soberlookprogresslanguage}), the \qwenmathfamily models automatically prepend their default system prompt.}
         }}
         & Solve the following math problem   \\
        && efficiently and clearly. The last line of  \\
        && your response should be of the following \\
        && format: 'Therefore, the final answer is: \\
        && The last line of your response should be\\
        && of the following format: 'Therefore, the \\
        && final answer is: \texttt{\textbackslash boxed\{ANSWER\}}. \\
        && I hope it is correct'  (without quotes) where \\
        && ANSWER  is just the final number or \\
        && expression that solves the problem. Think \\
        && step by step before answering.\textbackslash n\textbackslash n\{\} \\
        \midrule
        \multirow{11}{*}{\textbf{Spurious Prompt}} & \multirow{11}{*}{\makecell{Lorem ipsum dolor sit amet,\\consectetuer adipiscing elit.}} & \multirow{11}{*}{\makecell{
        Ut purus elit, vestibulum ut, placerat\\
        ac, adipiscing vitae, felis. Curabitur\\
        dictum gravida mauris. Nam arcu libero, \\
        nonummy eget, consectetuer id, vulputate a,\\
        magna. Donec vehicula augue eu neque.\\
        Pellentesque habitant morbi tristique\\
        senectus et netus et malesuada fames ac\\
        turpis egestas. Mauris ut leo. Cras viverra\\
        metus rhoncus sem. Nulla et lectus vestibulum\\
        urna fringilla ultrices. Phasellus eu\\
        tellus sit amet tortor gravida placerat.\textbackslash n\textbackslash n\{\}}} \\
        &&\\
        &&\\
        &&\\
        &&\\
        &&\\
        &&\\
        &&\\
        &&\\
        &&\\
        \\
    \bottomrule
    \end{tabular}}
    \caption{Details of different prompts used in previous RLVR research. We explore the impact of existing prompts (Qwen Default, Simplerl-zoo, and Sober) and 2 our two proposed prompts---the first prompt, Math Prompt, indicates the evaluation domain without extra information about the format; the second prompt, Spurious Prompt, is a randomly picked LaTeX placeholder text generated by \textsc{lipsum}. Our motivation for introducing these 2 prompts is to study how concise domain knowledge or even random strings in context can boost the evaluation performance.
    \vspace{-2mm}}
    \label{tab:eval_prompting_ablation}
\end{table}

\begin{figure*}[t!]
    \centering
    \cblock{94}{149}{78} Ground Truth
    \cblock{221}{162}{88} Format
    \cblock{147}{112}{188} Random
    \begin{subfigure}[t]{\textwidth}
        \centering
        \includegraphics[width=0.245\linewidth]{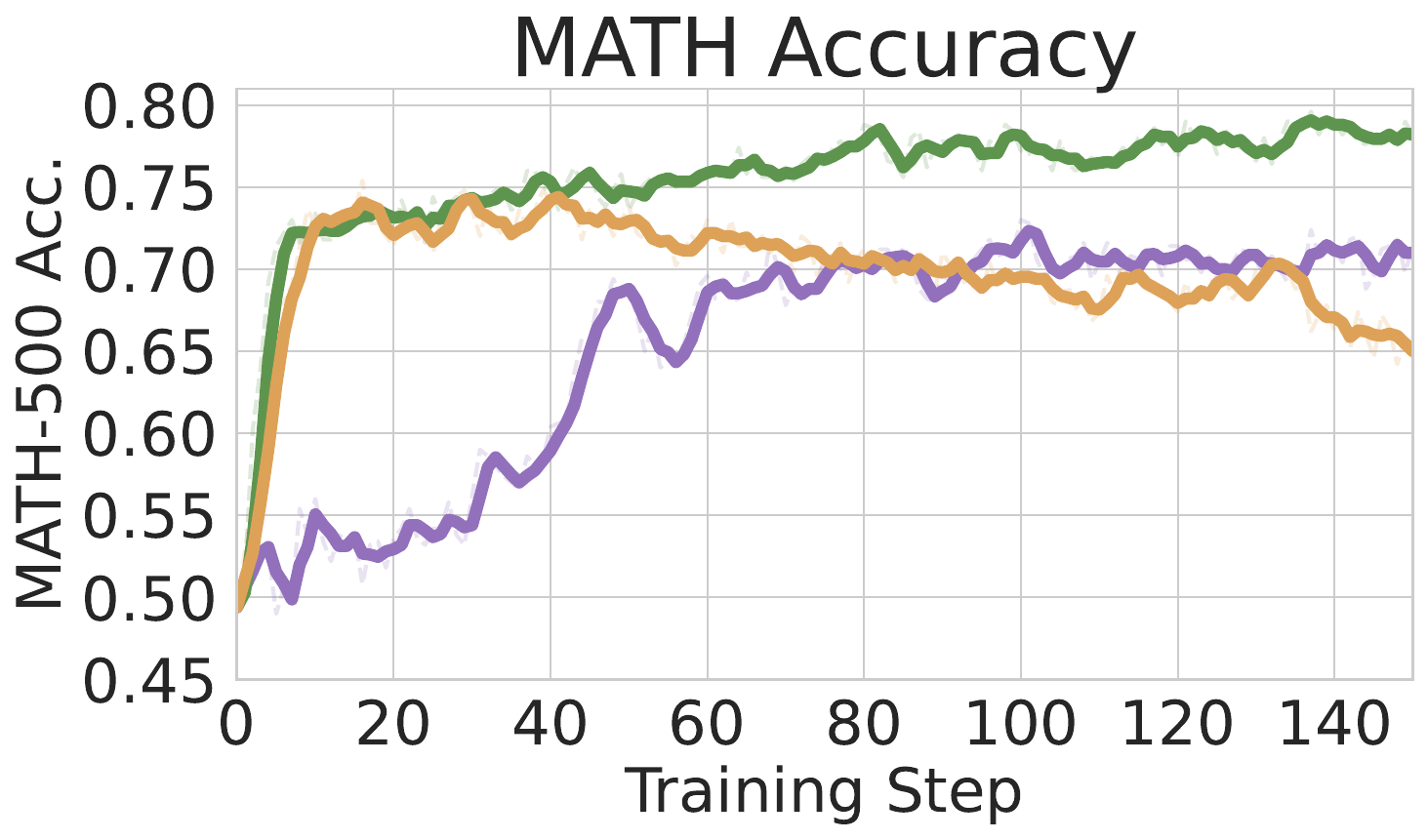}
        \includegraphics[width=0.245\linewidth]{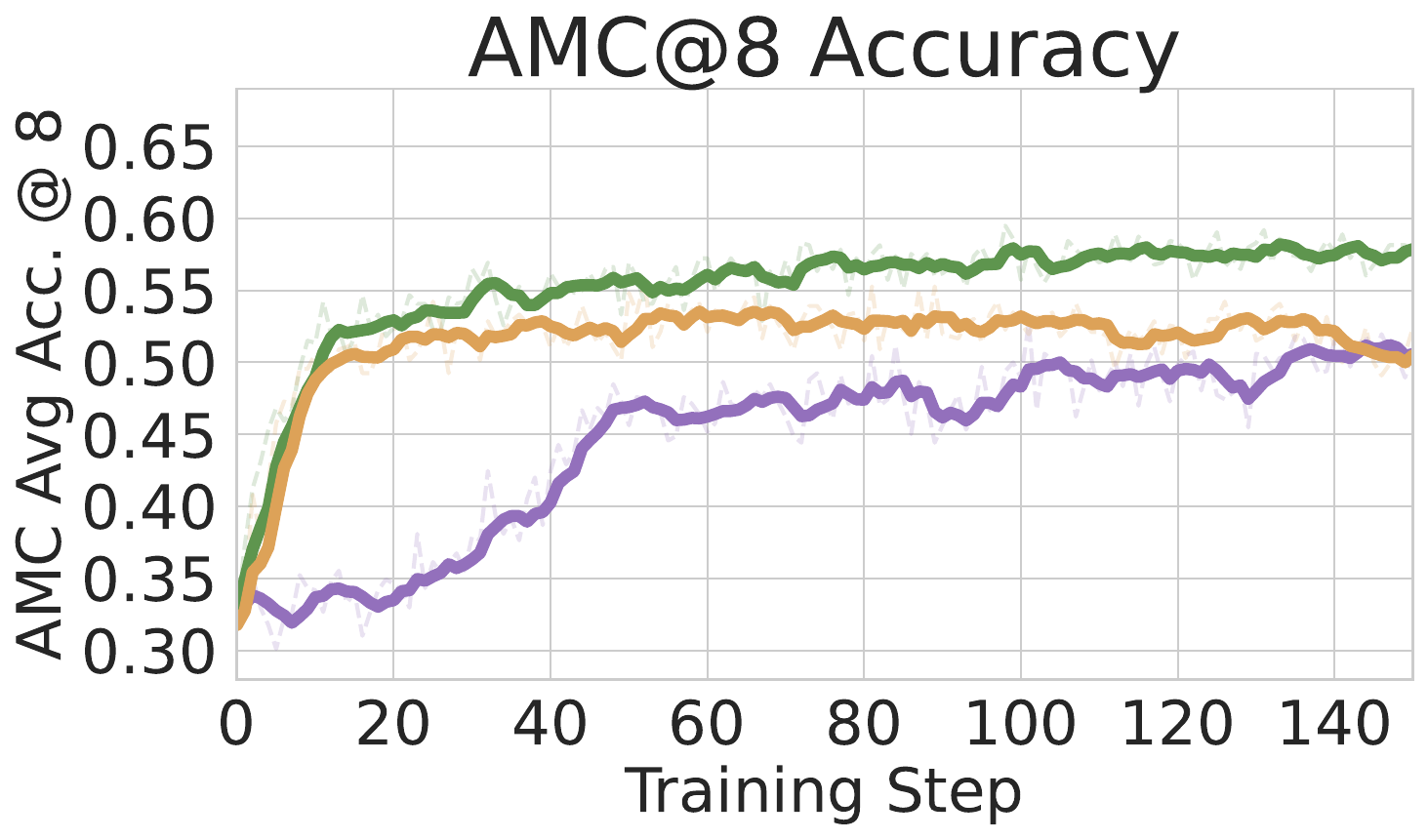}
        \includegraphics[width=0.245\linewidth]{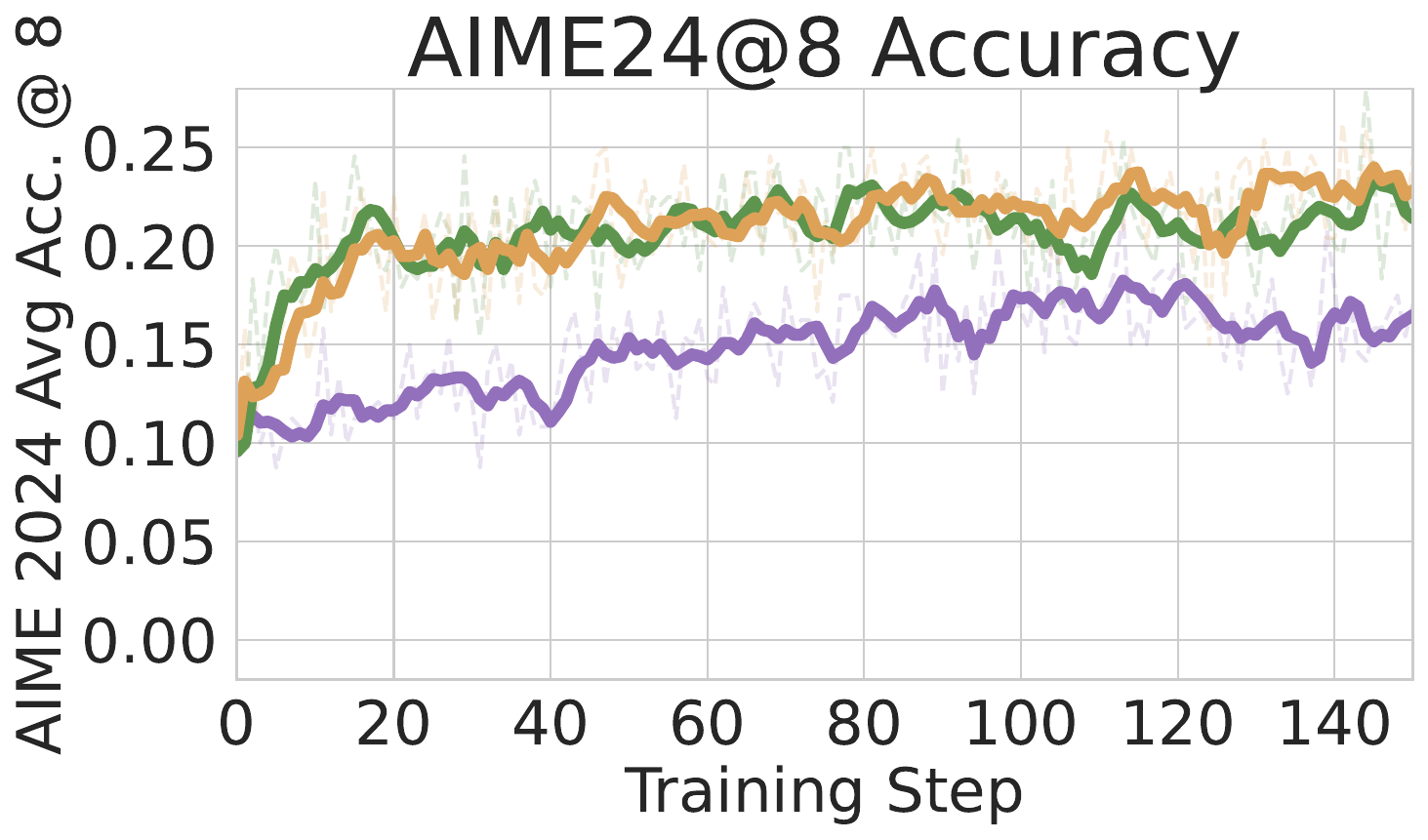}
        \includegraphics[width=0.245\linewidth]{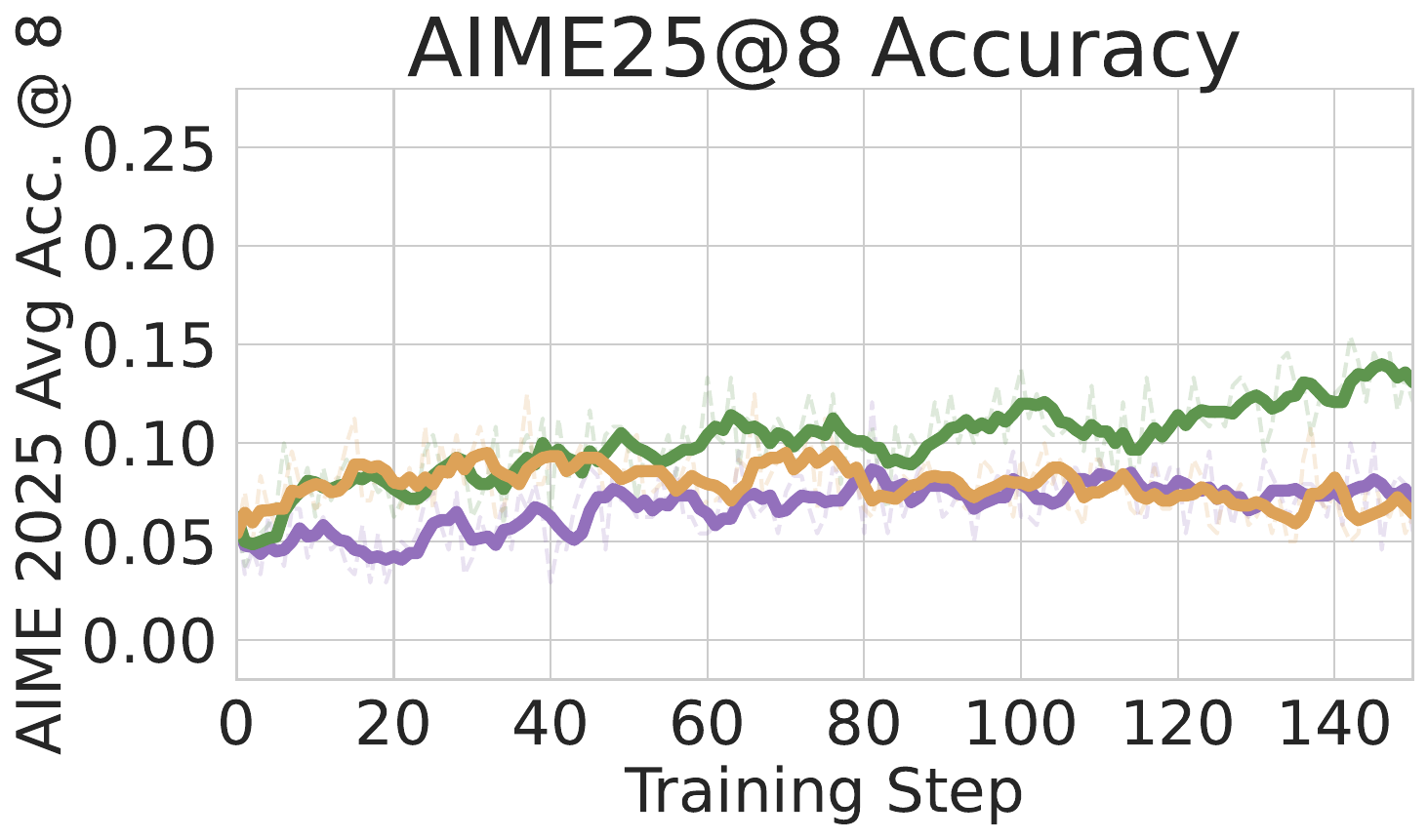}\vspace{-2mm}
        \caption{Default \qwenmathfamily Prompt~\citep{yang2024qwen2}\vspace{1mm}}
        \label{fig:default_prompt_results}
    \end{subfigure}\\
    
    \begin{subfigure}[t]{\textwidth}
        \centering
        \includegraphics[width=0.245\linewidth]{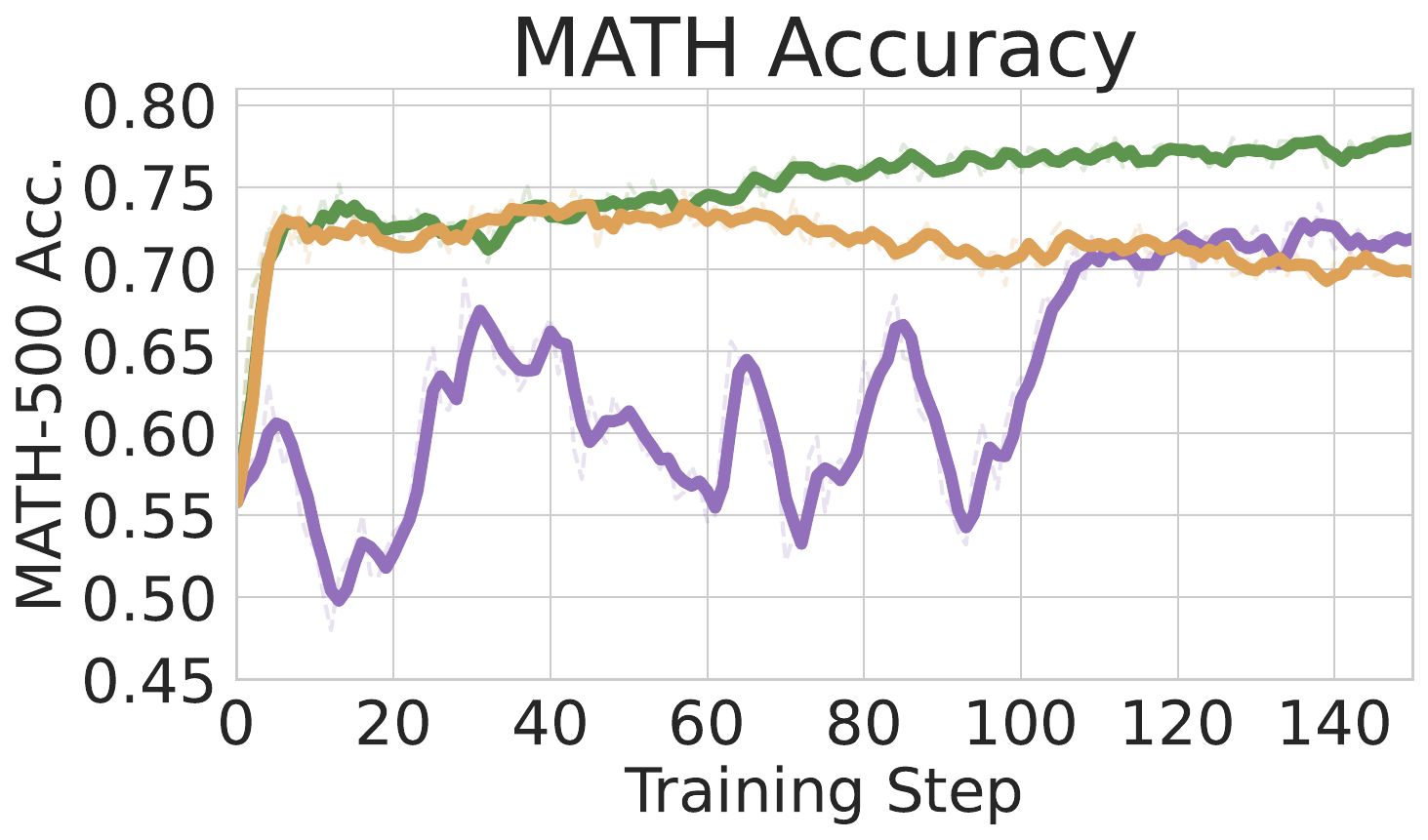}
        \includegraphics[width=0.245\linewidth]{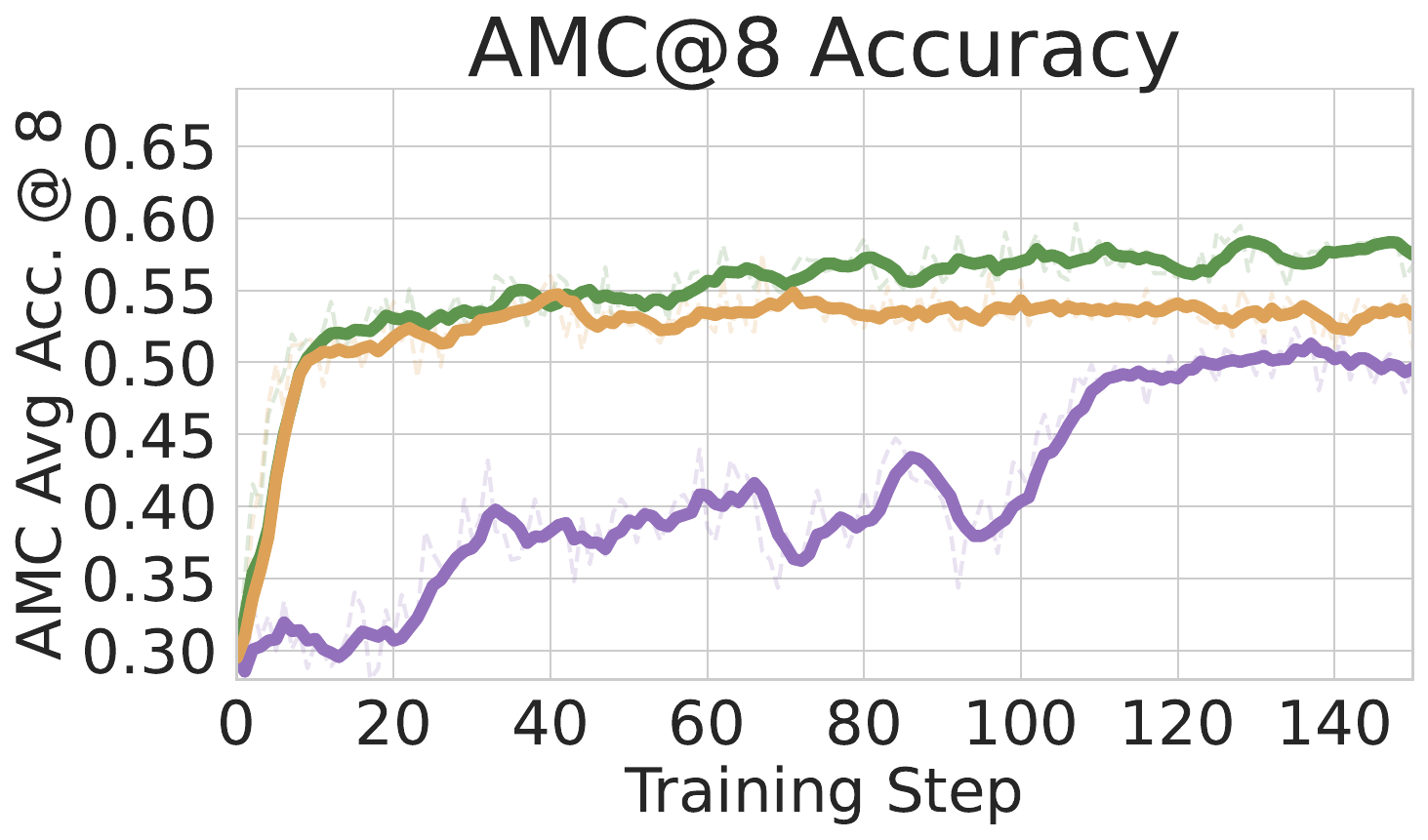}
        \includegraphics[width=0.245\linewidth]{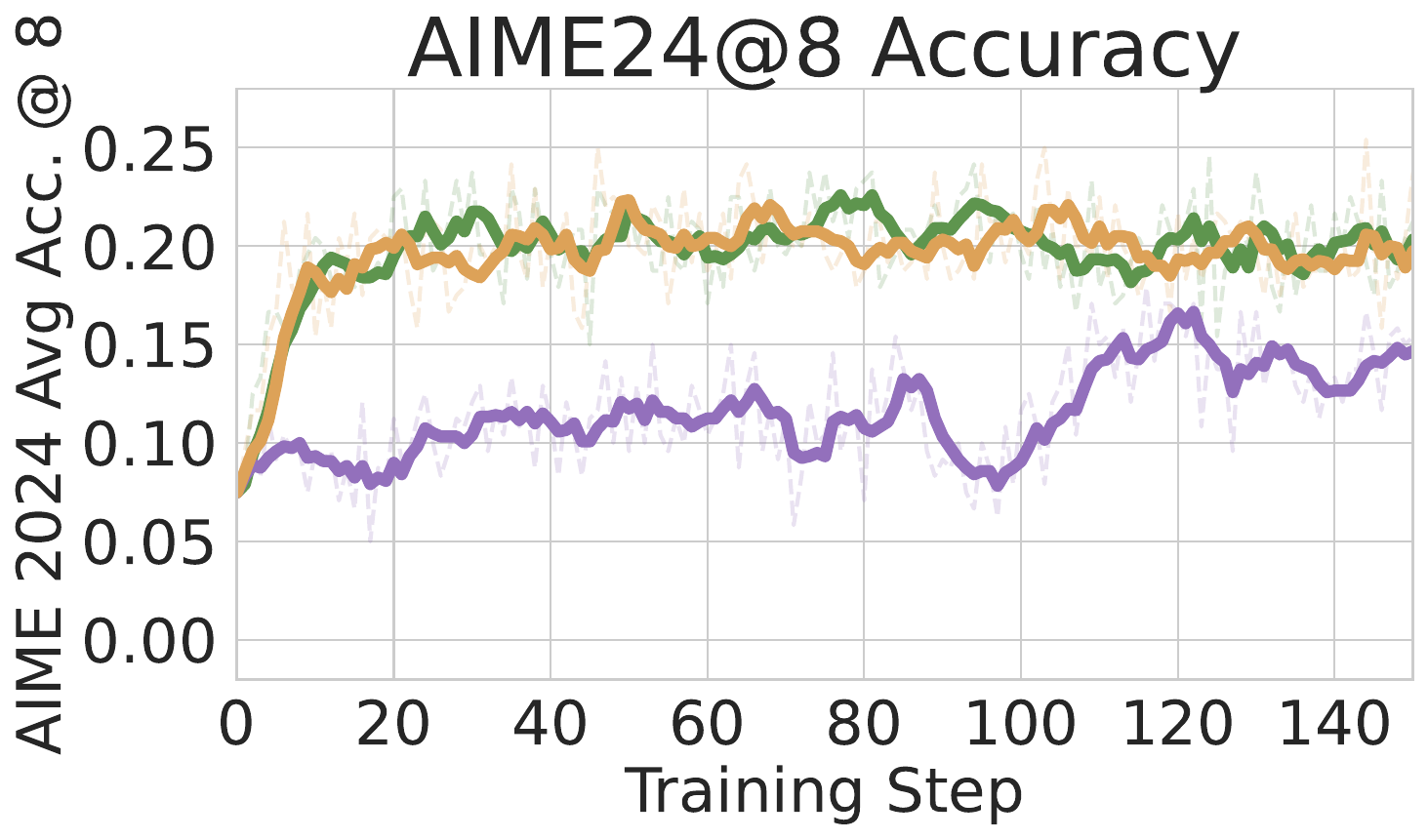}
        \includegraphics[width=0.245\linewidth]{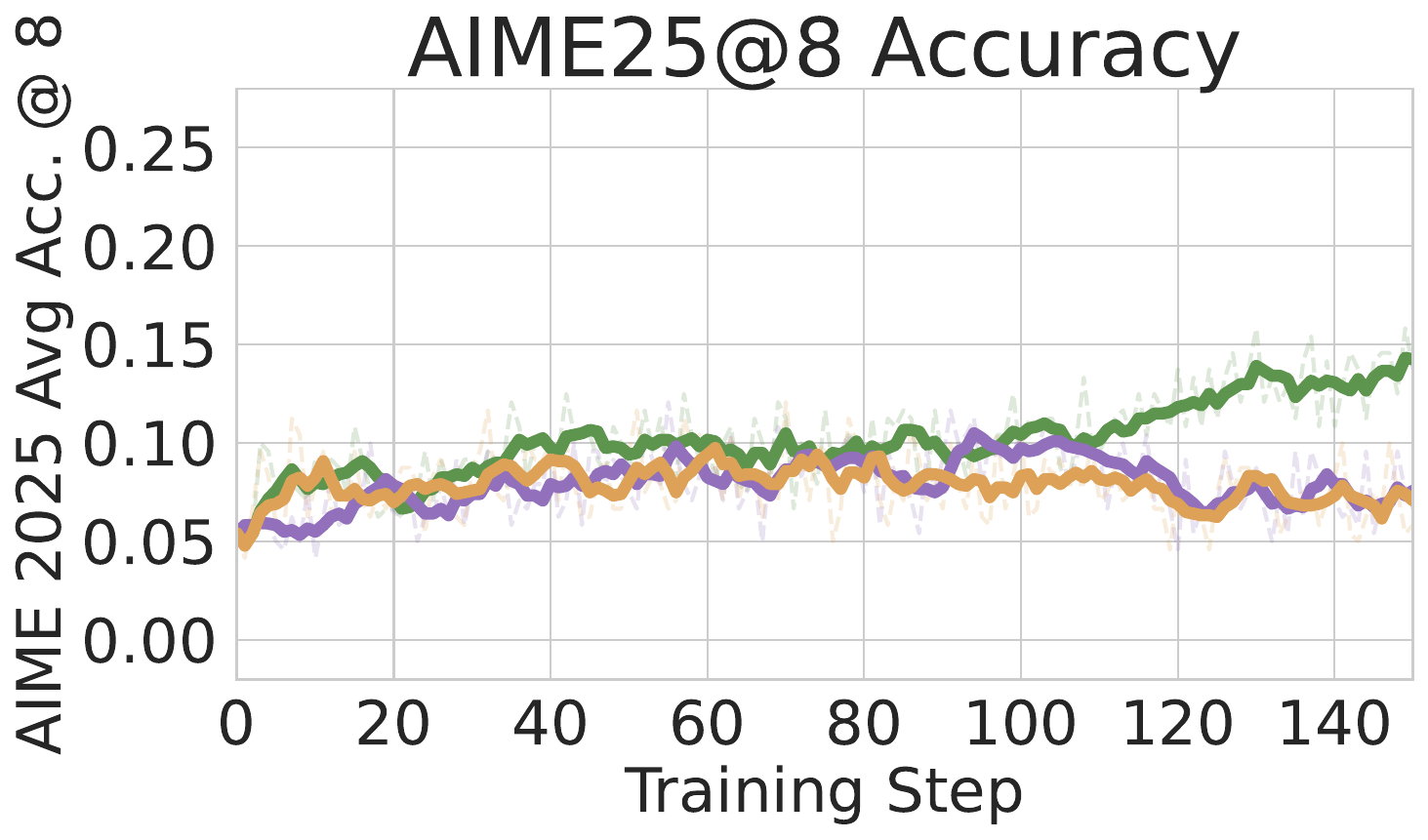}\vspace{-2mm}
        \caption{``\texttt{Math Problem:}'' Prompt\vspace{1mm}}
        \label{fig:mathproblem_prompt_results}
    \end{subfigure}\\
    
    \begin{subfigure}[t]{\textwidth}
        \centering
        \includegraphics[width=0.245\linewidth]{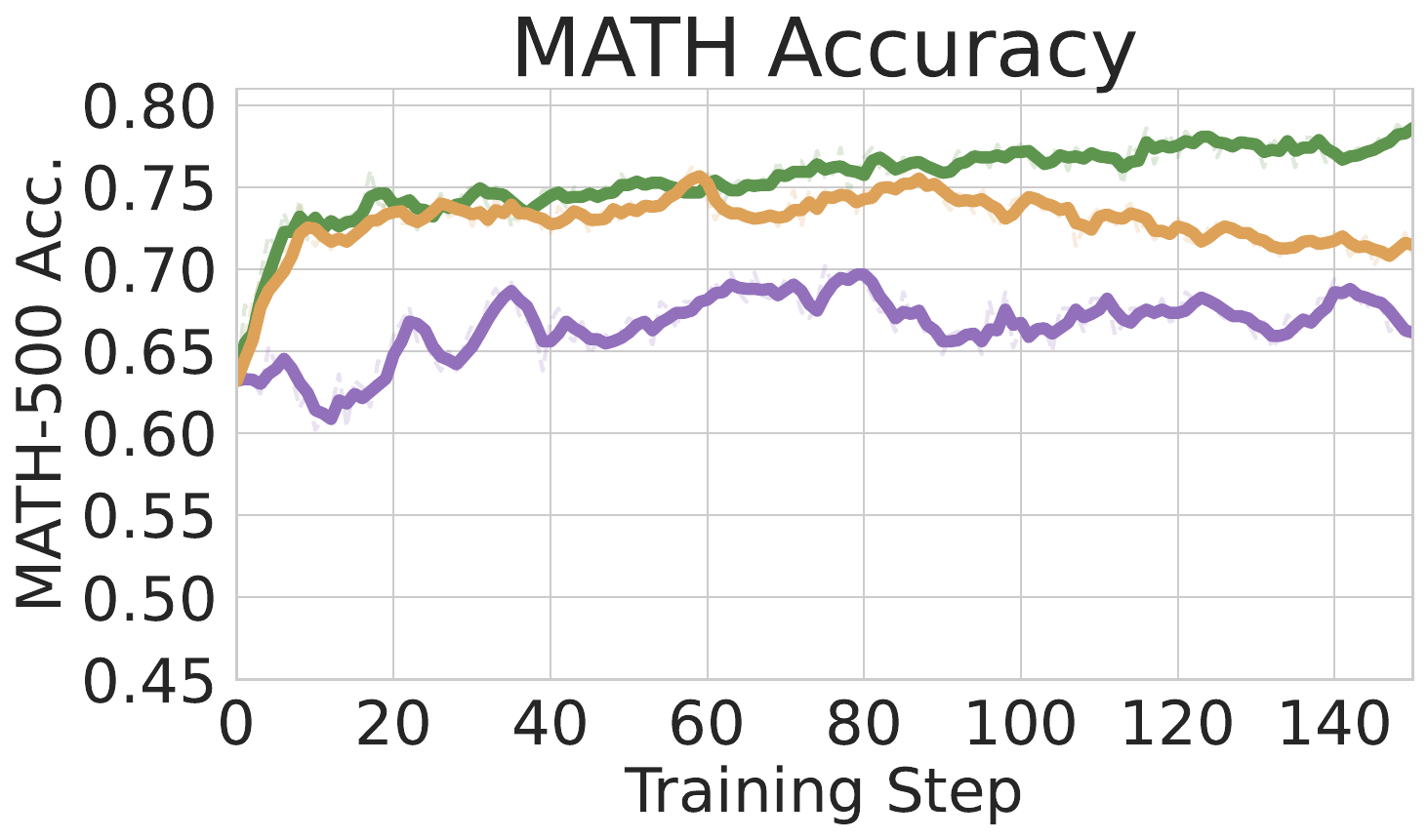}
        \includegraphics[width=0.245\linewidth]{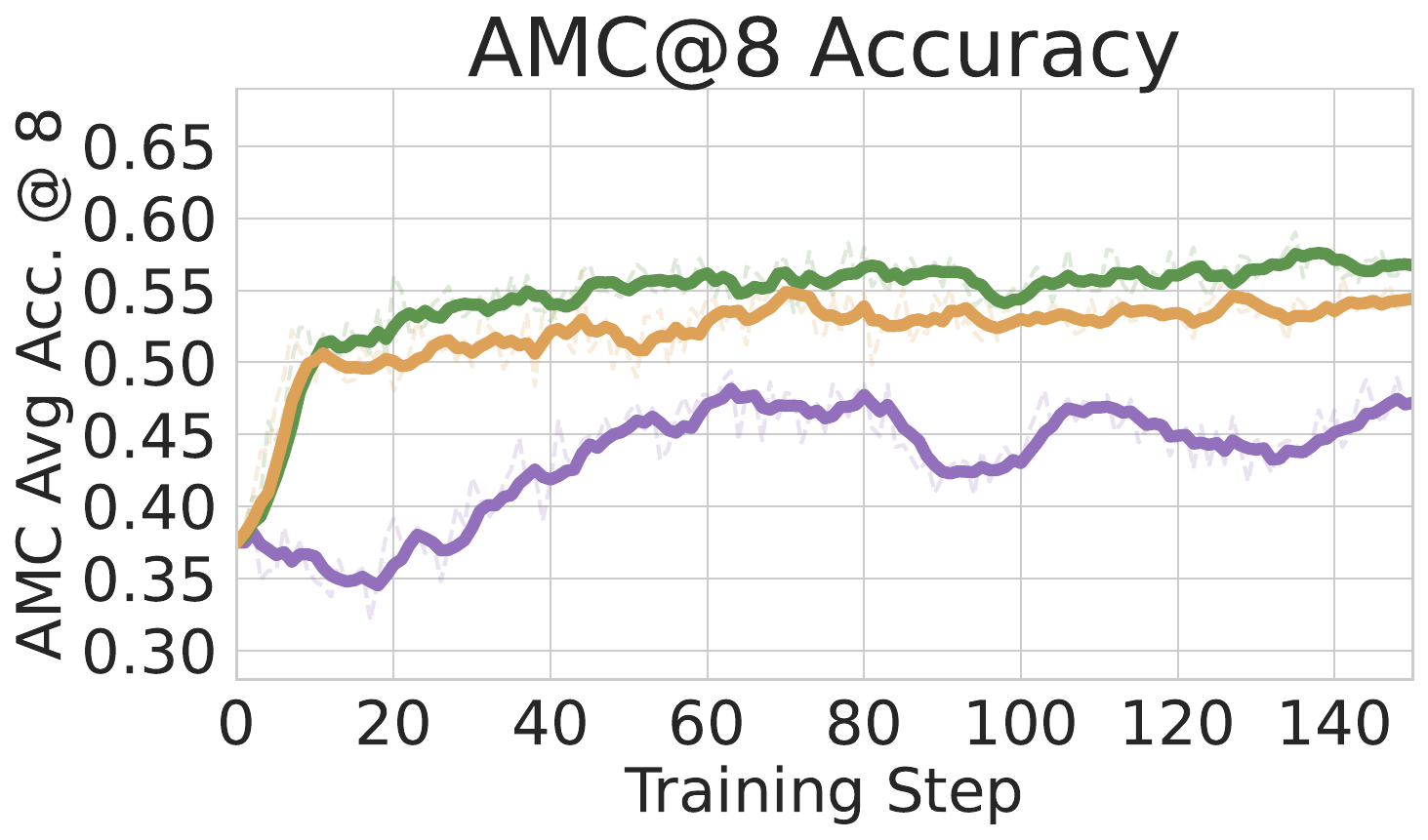}
        \includegraphics[width=0.245\linewidth]{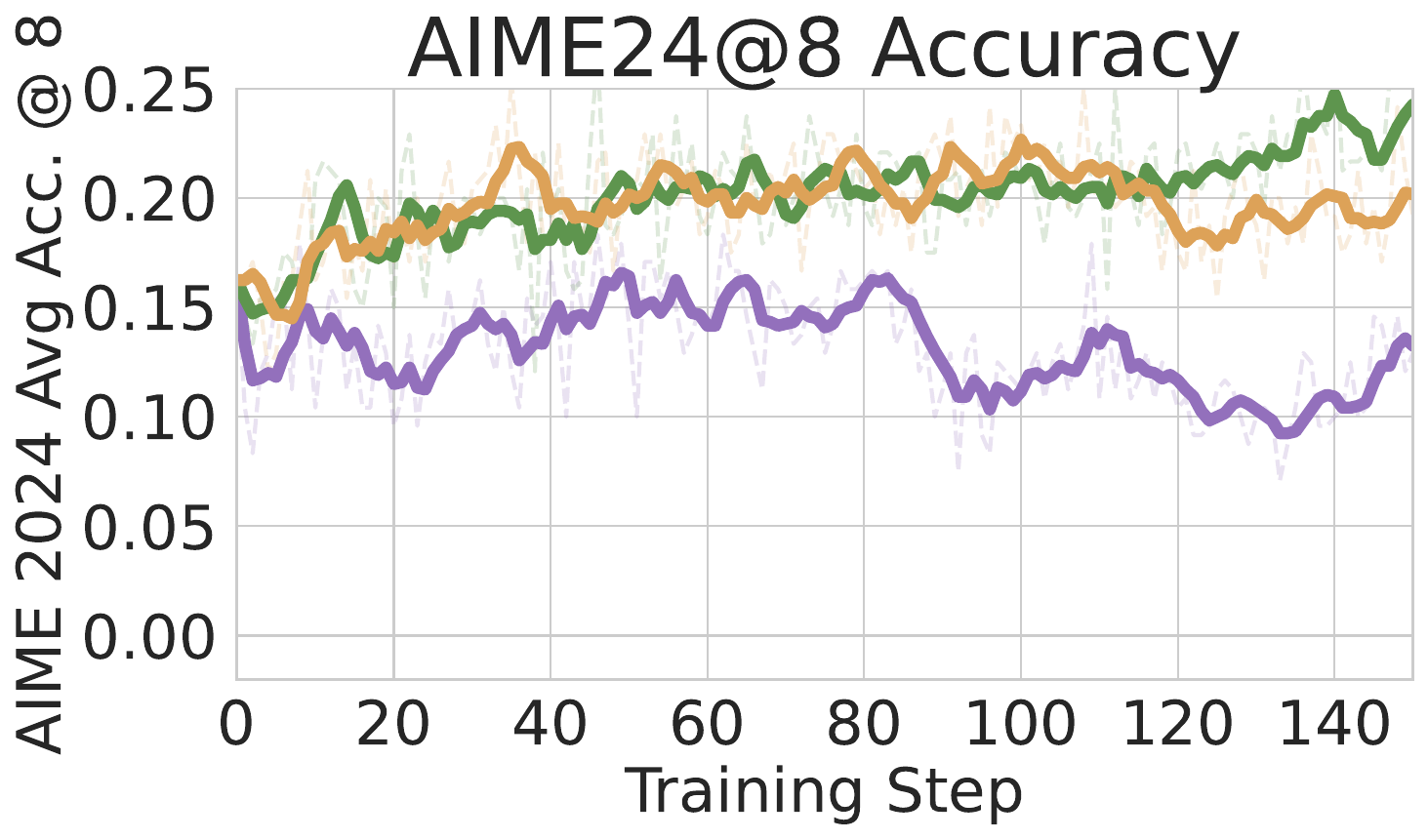}
        \includegraphics[width=0.245\linewidth]{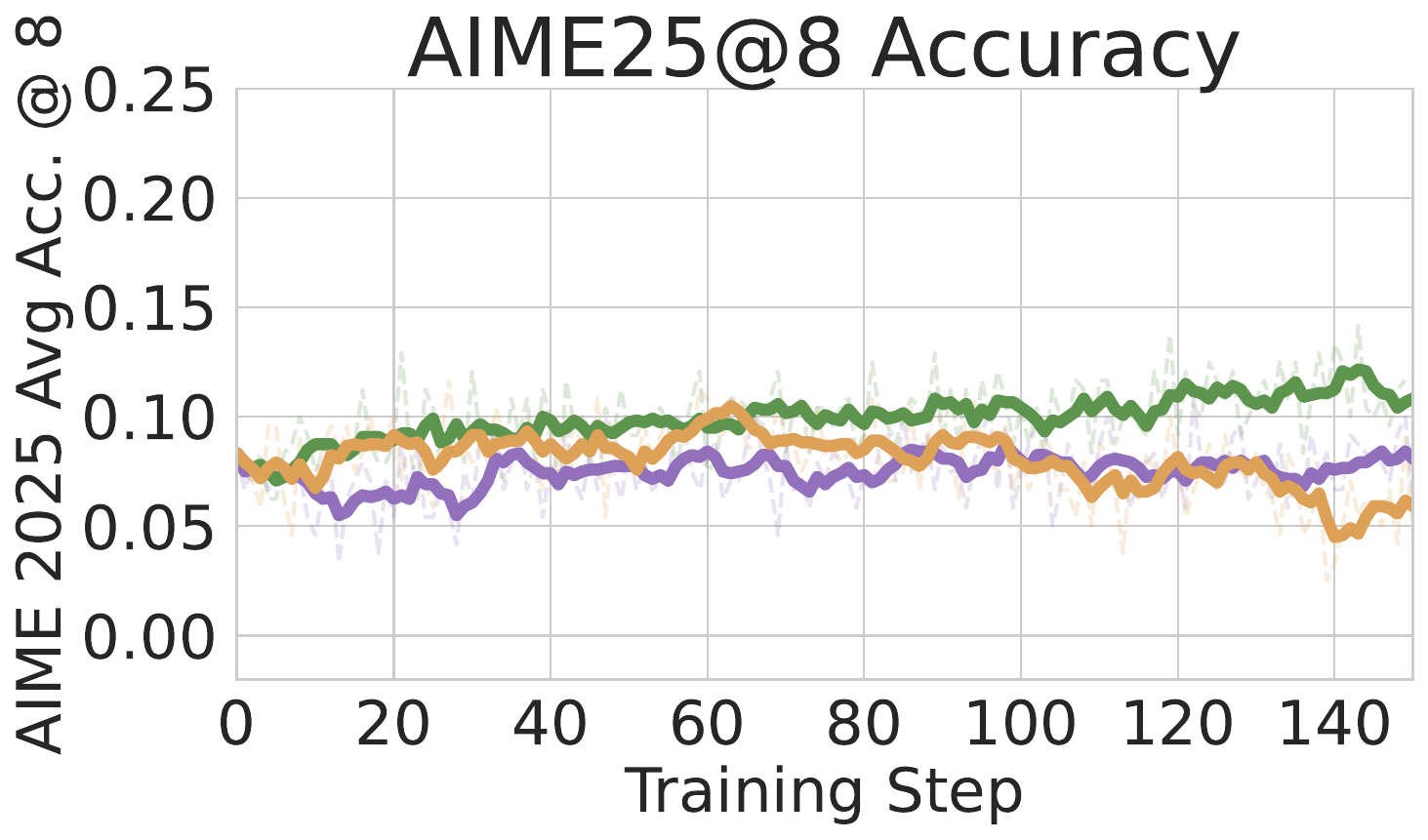}\vspace{-2mm}
        \caption{Simplerl-zoo Prompt~\citep{zeng2025simplerl}\vspace{1mm}}
        \label{fig:simplerl_prompt_results}
    \end{subfigure}%

    \begin{subfigure}[t]{\textwidth}
        \centering
        \includegraphics[width=0.245\linewidth]{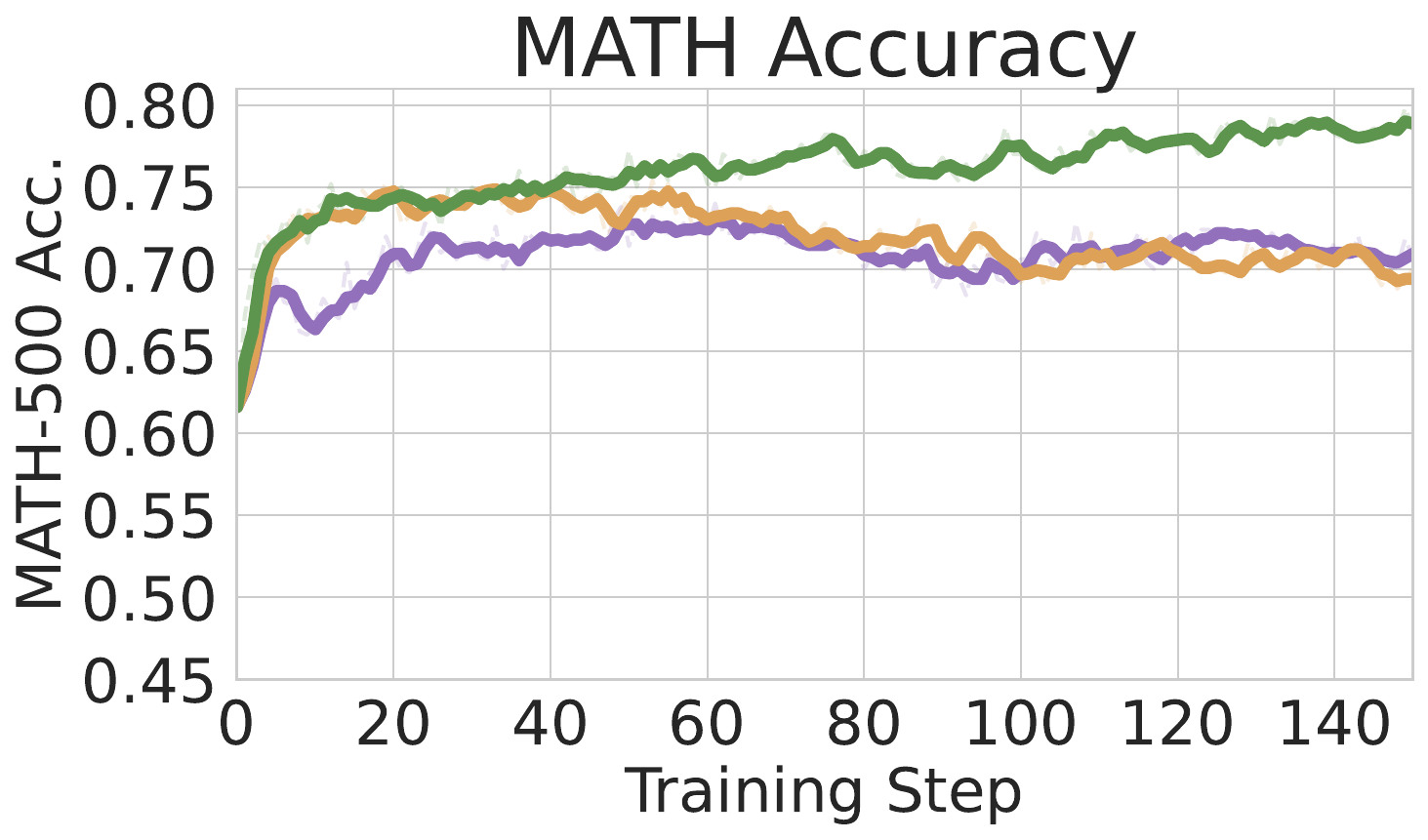}
        \includegraphics[width=0.245\linewidth]{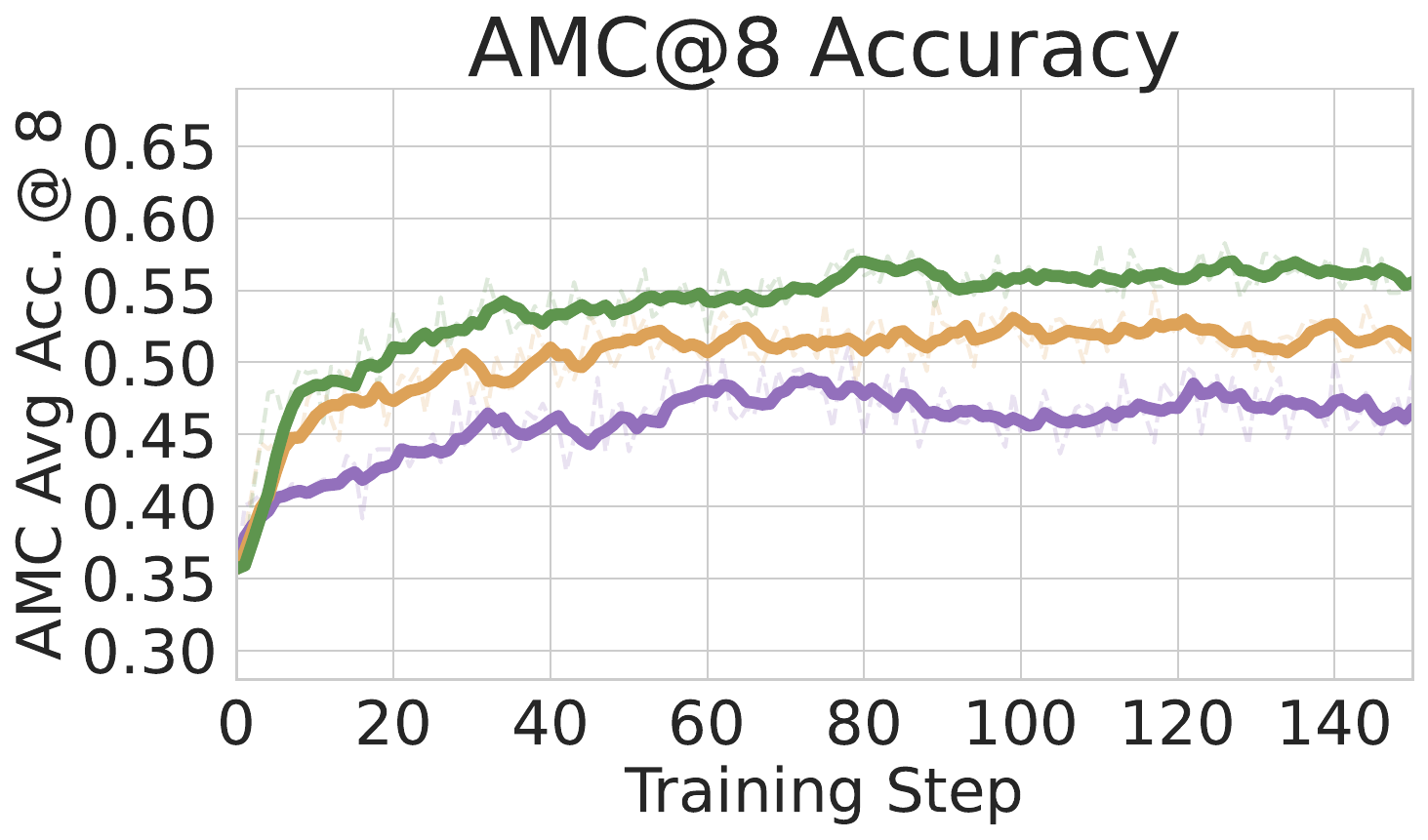}
        \includegraphics[width=0.245\linewidth]{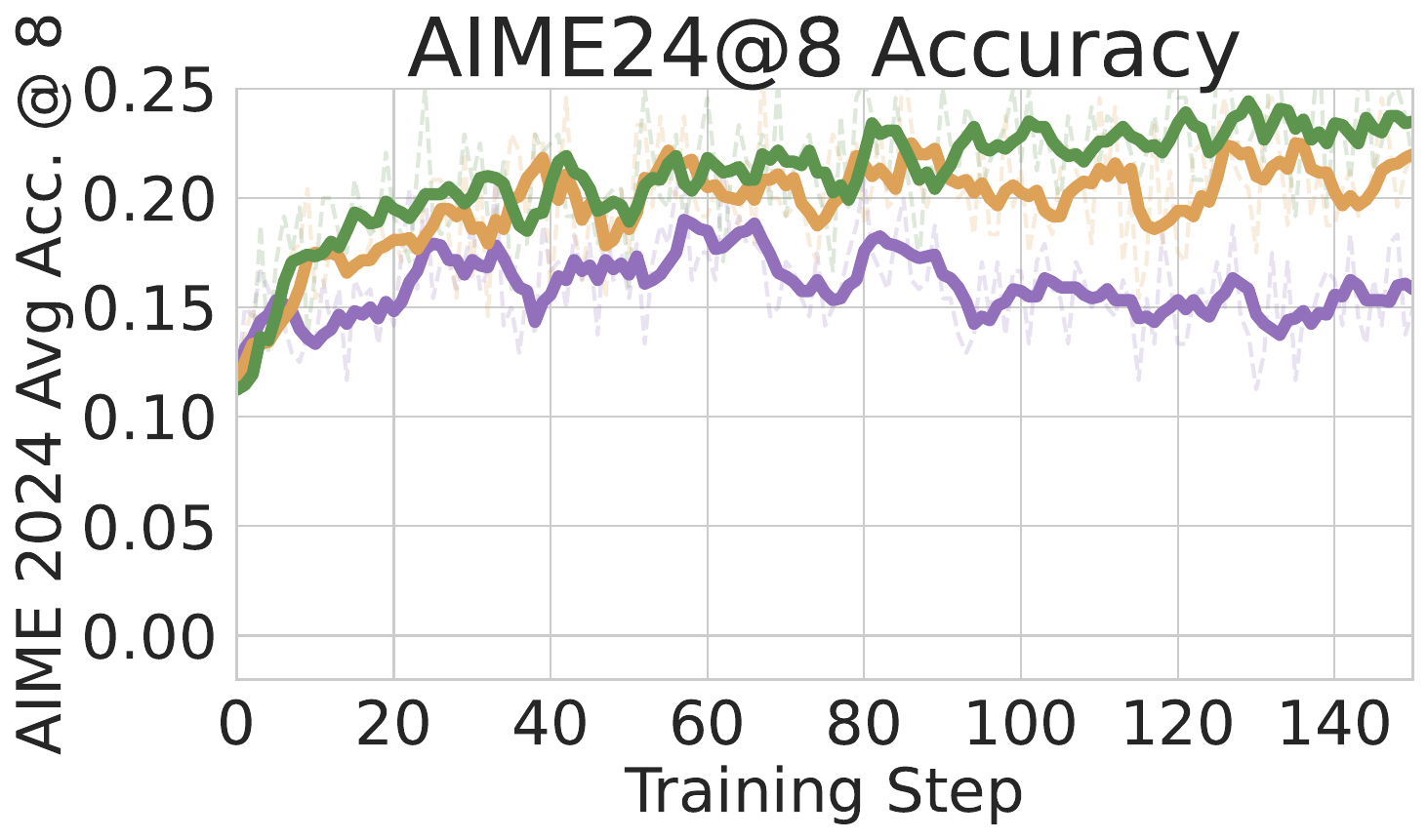}
        \includegraphics[width=0.245\linewidth]{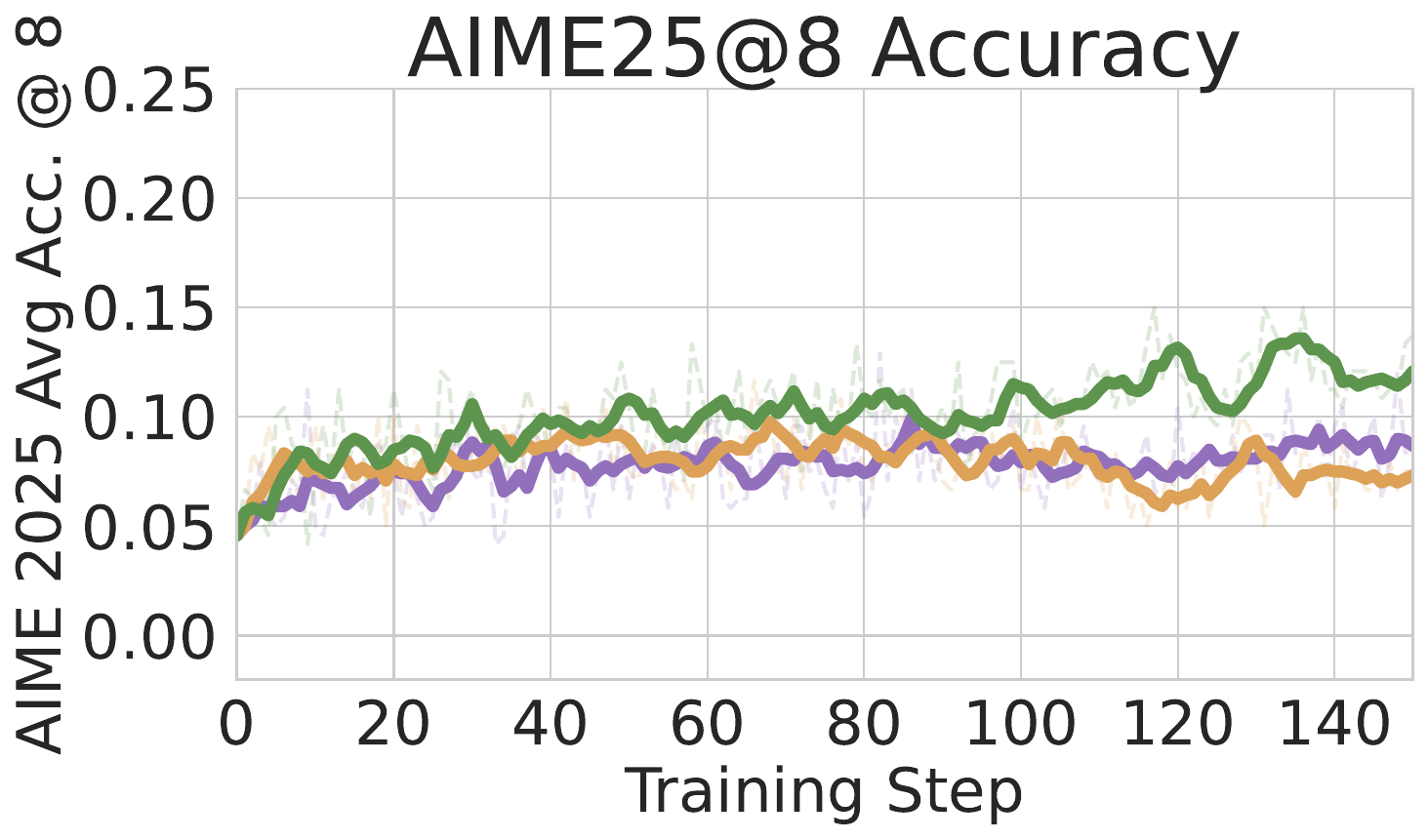}\vspace{-2mm}
        \caption{Sober Prompt~\citep{hochlehnert2025soberlookprogresslanguage}\vspace{1mm}}
        \label{fig:sober_prompt_results}
    \end{subfigure}%

    \begin{subfigure}[t]{\textwidth}
        \centering
        \includegraphics[width=0.245\linewidth]{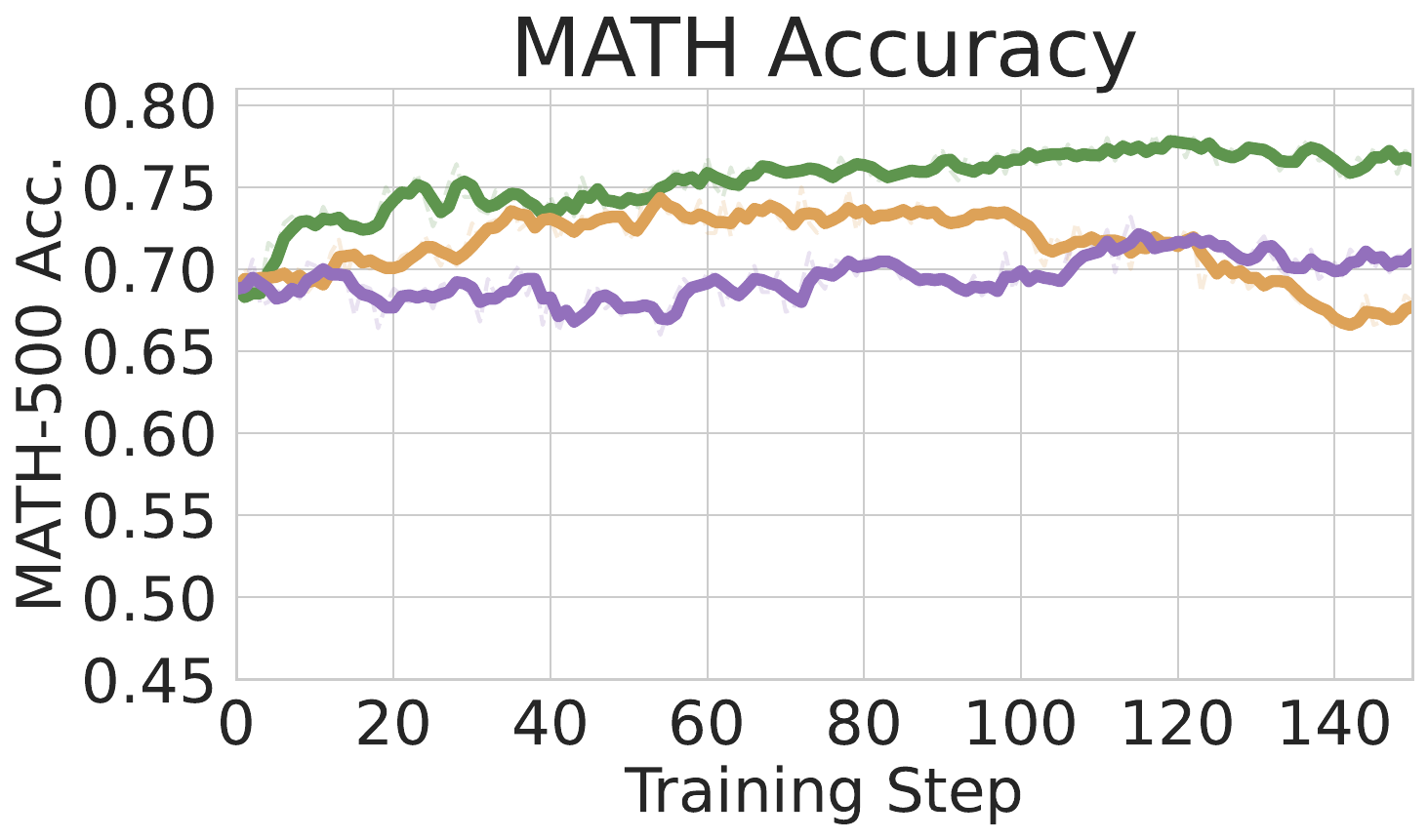}
        \includegraphics[width=0.245\linewidth]{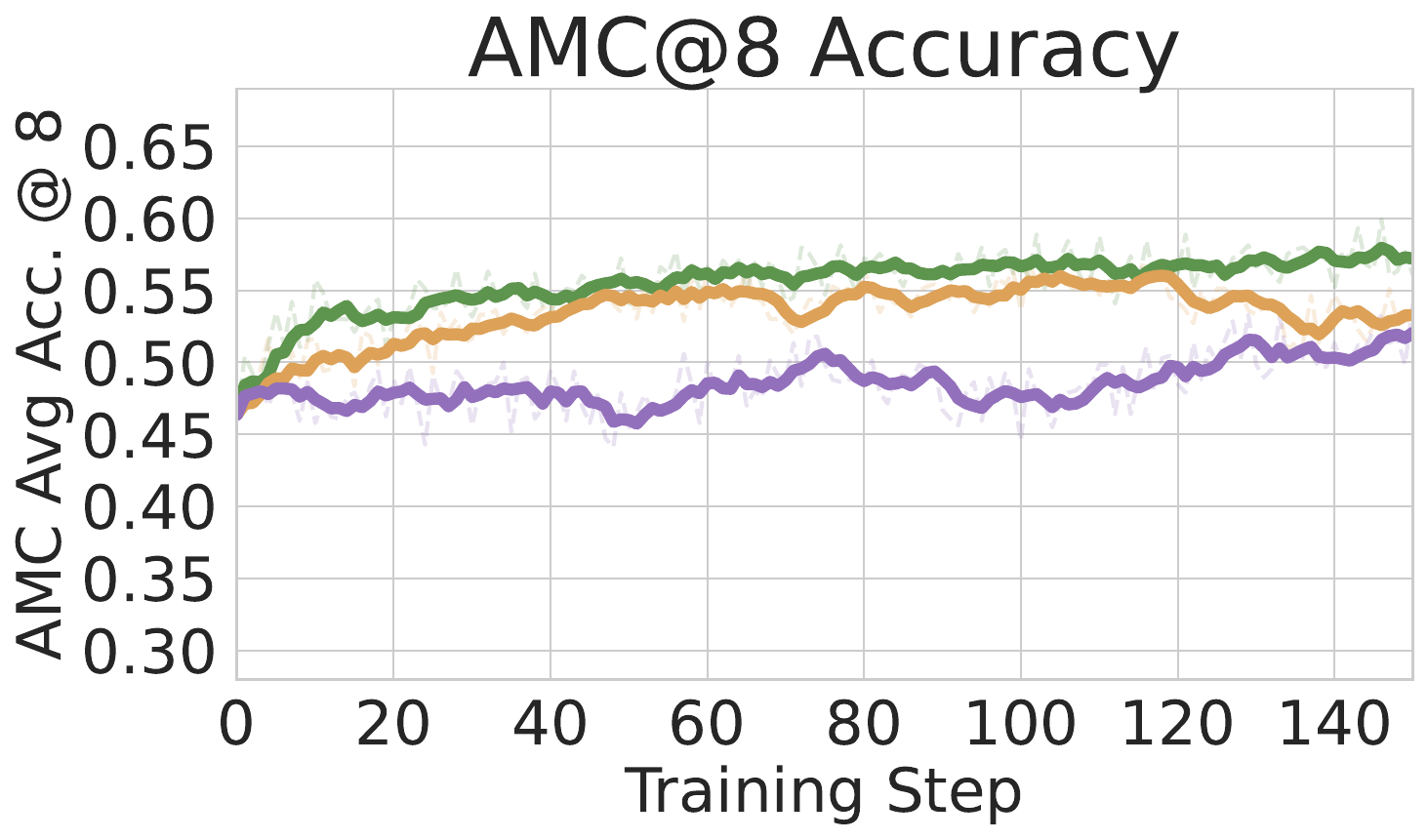}
        \includegraphics[width=0.245\linewidth]{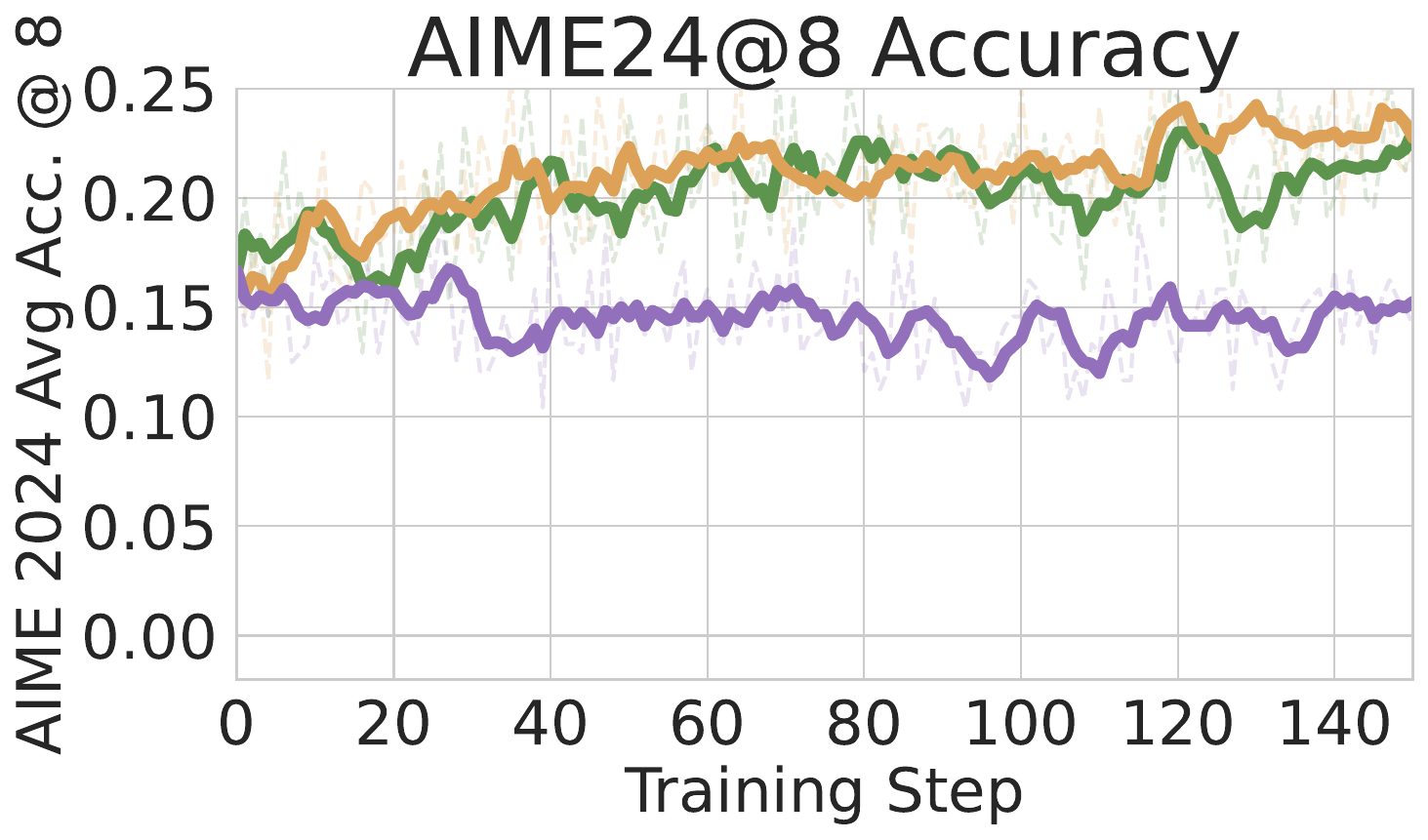}
        \includegraphics[width=0.245\linewidth]{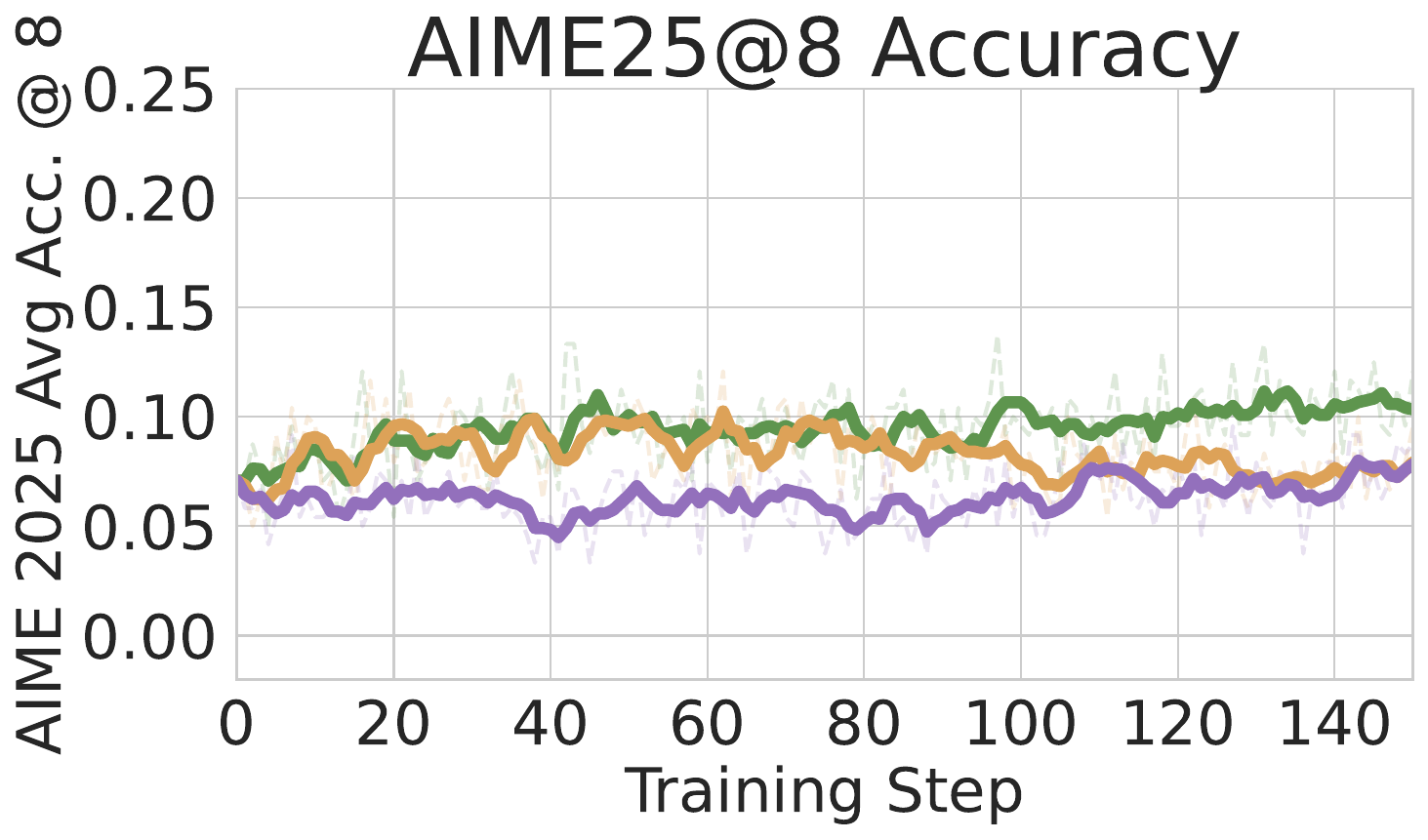}\vspace{-2mm}
        \caption{Spurious Prompt}
        \label{fig:liprum_prompt_results}
    \end{subfigure}%
    
\caption{RLVR performance with different prompt templates on \qwenmath. We show the choice of the prompt impacts both the before training performance and the RLVR training trajectories for \qwenmath. We find the default prompt provided by \qwenmathfamily~\citep{yang2024qwen2} results in lower initial performance; the other prompts offer higher initial performance, while Spurious Prompt offers the highest initial performance.
    } 
    \label{fig:prompt_results}
\end{figure*}

\begin{table}[t]
    \centering
    \begin{tabular}{c|ccccc}
    \toprule
              & \textbf{Default} & \textbf{MathProblem} & \textbf{SimpleRL-Zoo} & \textbf{Sober} & \textbf{Spurious} \\
    \midrule
\textbf{MATH Acc.} & 49.4 & 55.8 & 63.2 & 61.60 & \textbf{68.8} \\
\textbf{\% Parsable}   & 78.9 & 72.1 & 85.4 & \textbf{93.1} & 84.1 \\
    \bottomrule
    \end{tabular}
    \caption{Accuracy on MATH-500 and percent of parsable (format-following) responses on \qwenmath with various prompts from Table~\ref{tab:eval_prompting_ablation}. Even with a spurious prompt, the model is able to follow the format 84.1\% of the time. It is not obvious that much of the performance can be explained by format-following.
    \vspace{-3mm}}
    \label{tab:format_accuracy}
\end{table}

As we show in Section~\ref{sec:analysis:intervene}, prompting the model to use code reasoning brings a 15.0\% gain on \qwenmath. This suggests that prompt engineering is a valid approach to elicit desired behaviors, which does not require parameter updates. 
In this section, we conduct an additional analysis on how choosing different existing prompts can impact model behavior and how such a technique interacts with RLVR. 
In addition, we show that model performance can be improved 
even when using some task-unrelated, information-less prompts, which we name ``spurious prompts.''

\subsection{Existing Prompts}

We collect different prompts from popular RL training frameworks and recent evaluation works: Qwen Default \citep{yang2024qwen2}, SimpleRL-Zoo \citep{zeng2025simplerl}, Sober Reasoning \citep{hochlehnert2025soberlookprogresslanguage}, and our hand-constructed prompts (detailed in Table~\ref{tab:eval_prompting_ablation}). 
Specifically, we introduce two new prompts---the first prompt, Math Prompt, indicates the evaluation domain without extra information about the format; the second prompt, Spurious Prompt, is a randomly picked LaTeX placeholder text generated by \textsc{lipsum}. 
Although it has been known that LLM evaluation can be sensitive to evaluation prompts \citep{sclar2024quantifyinglanguagemodelssensitivity},
our motivation to introduce these two particular prompts is to study how concise domain knowledge or even random perturbation in context can impact evaluation performance.
We perform RL training using these prompts for both trajectory rollouts and evaluation.

As shown in Table~\ref{tab:format_accuracy}, Sober prompt brings the highest parsable rate of the answers, while our spurious prompt leads to the highest accuracy on MATH-500. The results indicate that the model is very sensitive to prompts and the best performance does not necessarily require the highest parsable rate nor task-relevant information in context.

We further show the training trajectories using different prompts in Figure~\ref{fig:prompt_results}. We find that models trained with the Qwen default, \textsc{math problem:}, and Simplerl-zoo converge to similar performance after RLVR with the same training setting. We conjecture that the behavior that can be elicited by prompting tends to be a subset of the easy-to-elicit behaviors in RLVR, while RLVR can elicit additional behaviors that are not predefined in prompts.

\subsection{Spurious Prompts}

As \qwenmath shows high sensitivity to the prompts, we are curious about how much it could be impacted by spuriously unrelated prompts, e.g., the placeholder text in LaTex generated by \textsc{lipsum}. Specifically, we construct a spurious prompt using \textsc{lipsum} with a similar prompt length as the Sober prompt in Table~\ref{tab:eval_prompting_ablation}.
Surprisingly, the spurious prompt gives the highest initial performance compared with the other commonly used task-specific prompts, as shown in Figure~\ref{fig:liprum_prompt_results}. Although we find that the improvement does not always happen for randomly picked prompts, the high performance with our spurious prompt indicates that the model can be very sensitive to in-context perturbations, where the benefit may sometimes originate from this sensitivity rather than from the task-relevant content in the prompt.

\section{Additional Results on Models That Have Undergone RL Training}\label{app:instruct_models}

In this section, we provide additional results for models that have been trained with reinforcement learning, including two Qwen Instruct models (\qwenmath-Instruct~\citep{yang2024qwen2} and \qwenbasefamily-7B-Instruct~\citep{Yang2024Qwen25TR}) and one Tulu3 model (Llama-3.1-Tulu-3-8B~\citep{lambert2024tulu3}).
In this section, we show that the post-RL models may behave differently from other base or instruction-tuned models in RLVR.

\begin{figure*}[t!]
    \centering
    \cblock{94}{149}{78} Ground Truth
    \cblock{221}{162}{88} Format
    \cblock{147}{112}{188} Random
    \begin{subfigure}[t]{\textwidth}
        \centering
        \includegraphics[width=0.245\linewidth]{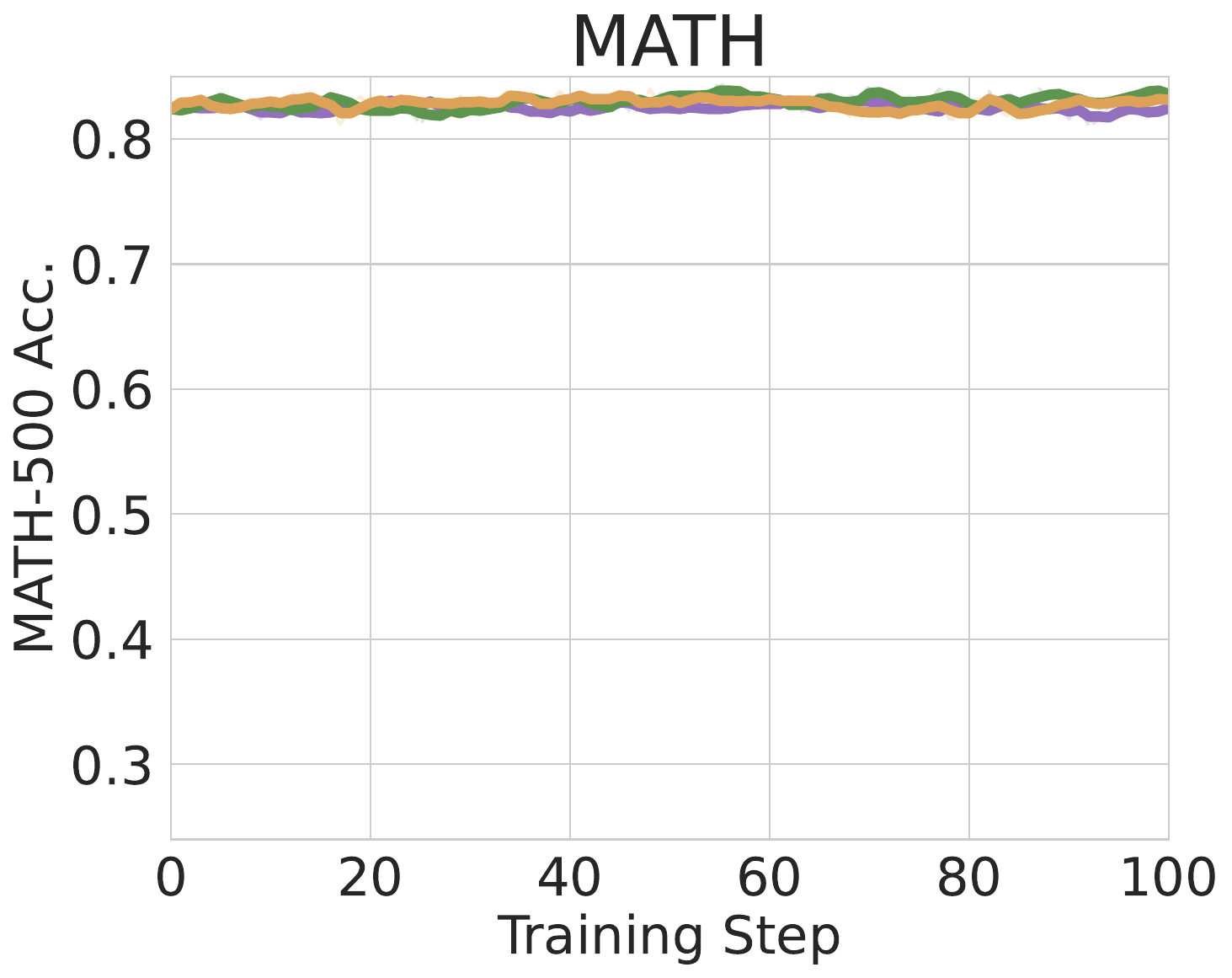}
        \includegraphics[width=0.245\linewidth]{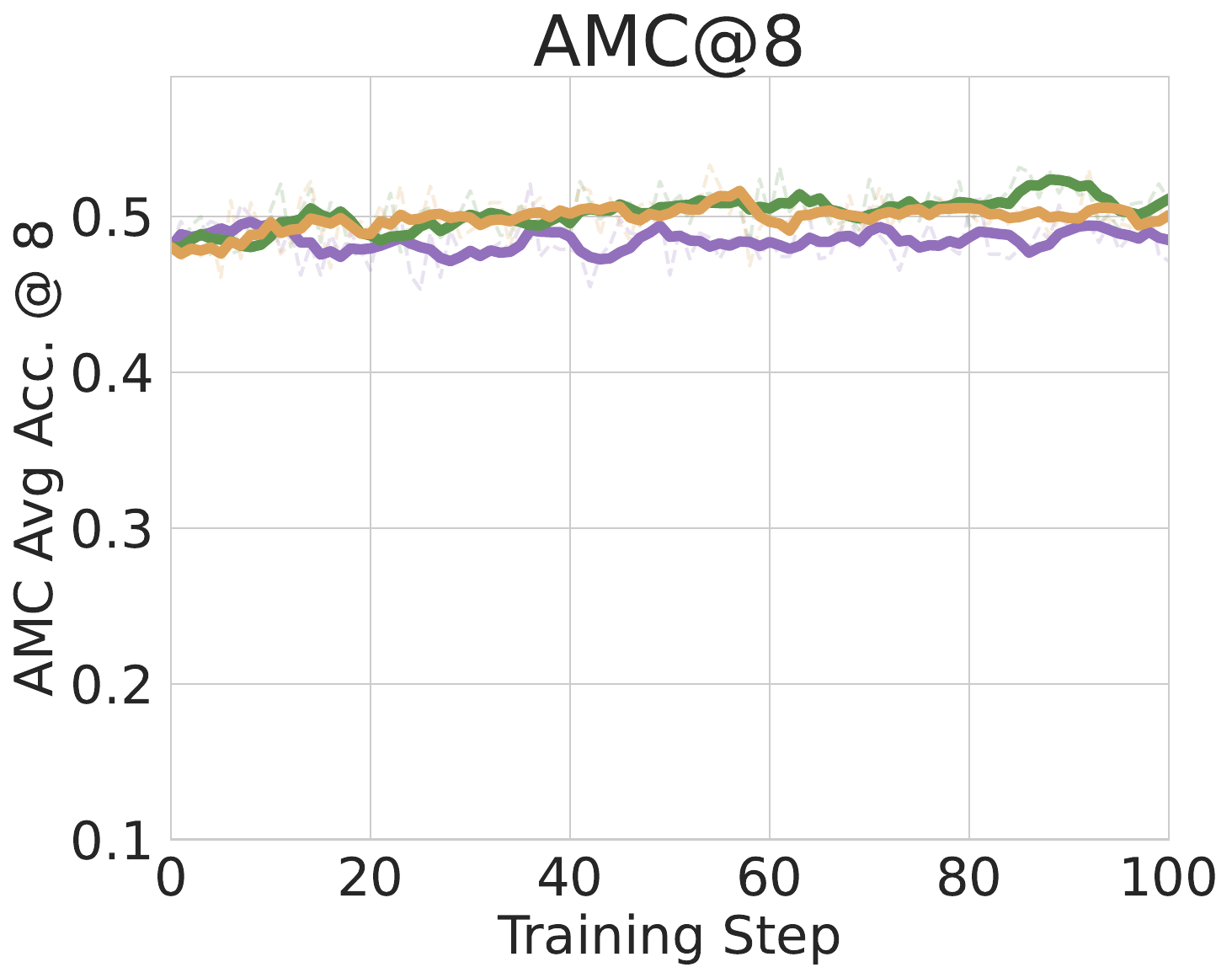}
        \includegraphics[width=0.245\linewidth]{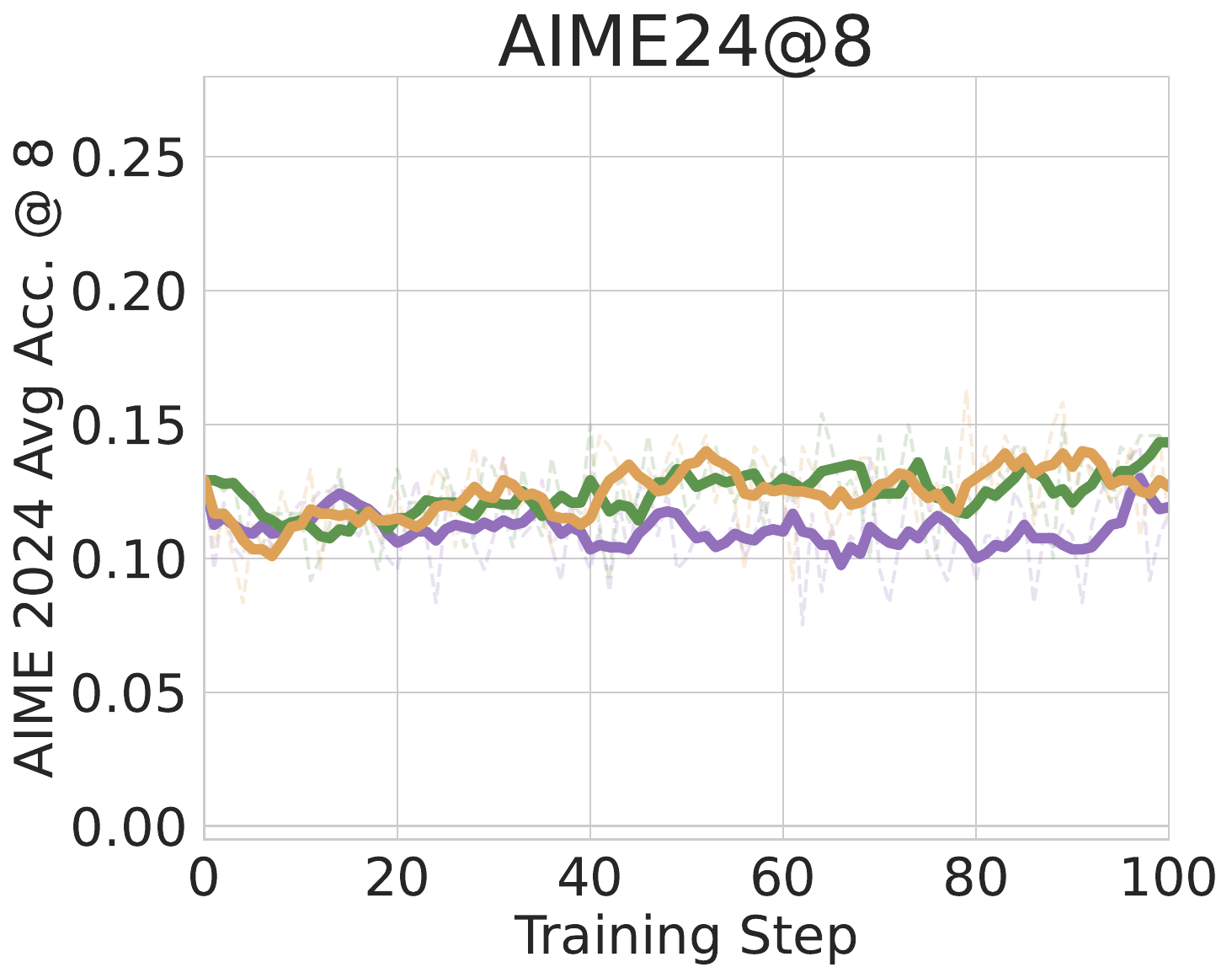}
        \includegraphics[width=0.245\linewidth]{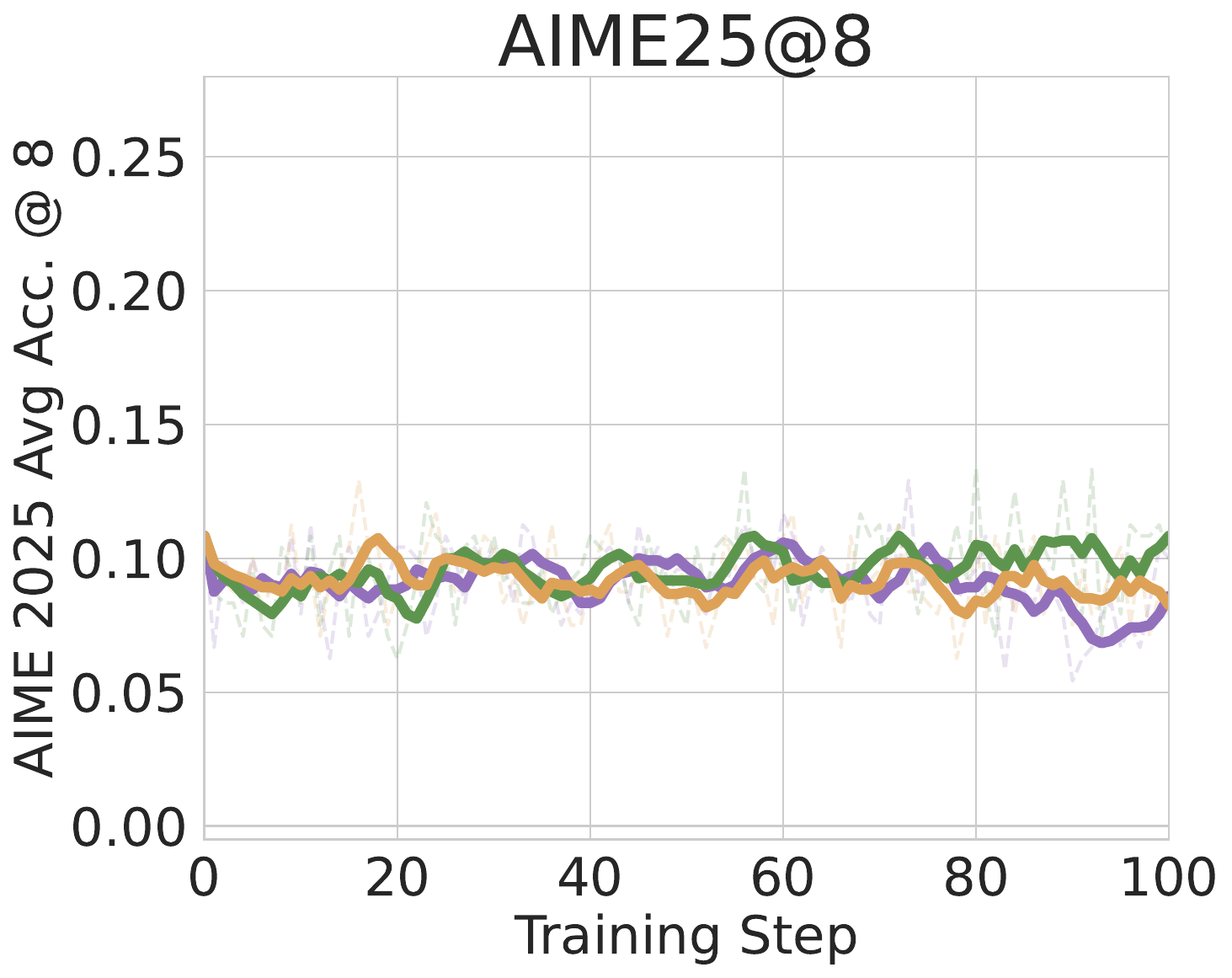}
        \caption{\qwenmath-Instruct}
        \label{fig:qwen_math_7b_instruct_results}
    \end{subfigure}\\
    
    \begin{subfigure}[t]{\textwidth}
        \centering
        \includegraphics[width=0.245\linewidth]{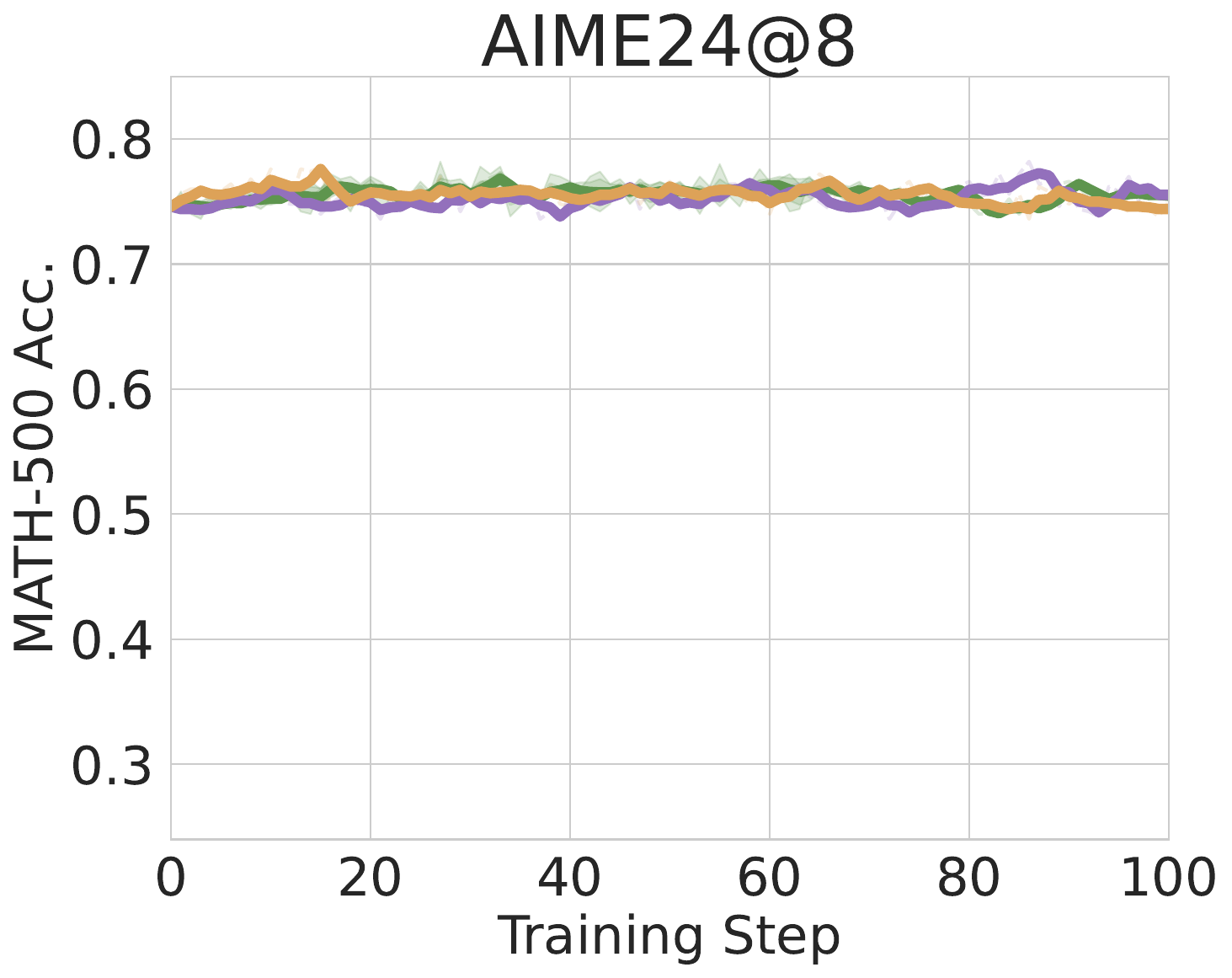}
        \includegraphics[width=0.245\linewidth]{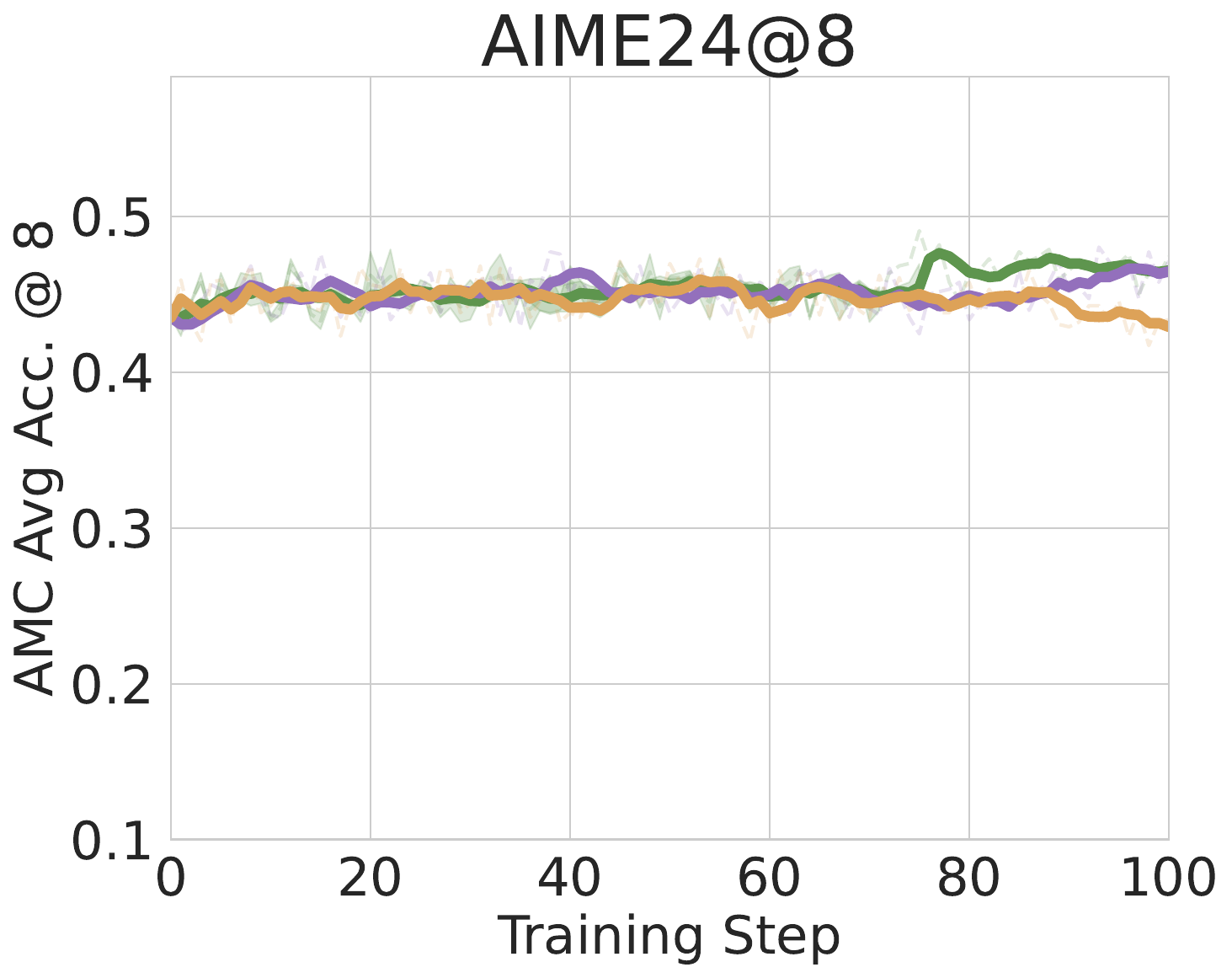}
        \includegraphics[width=0.245\linewidth]{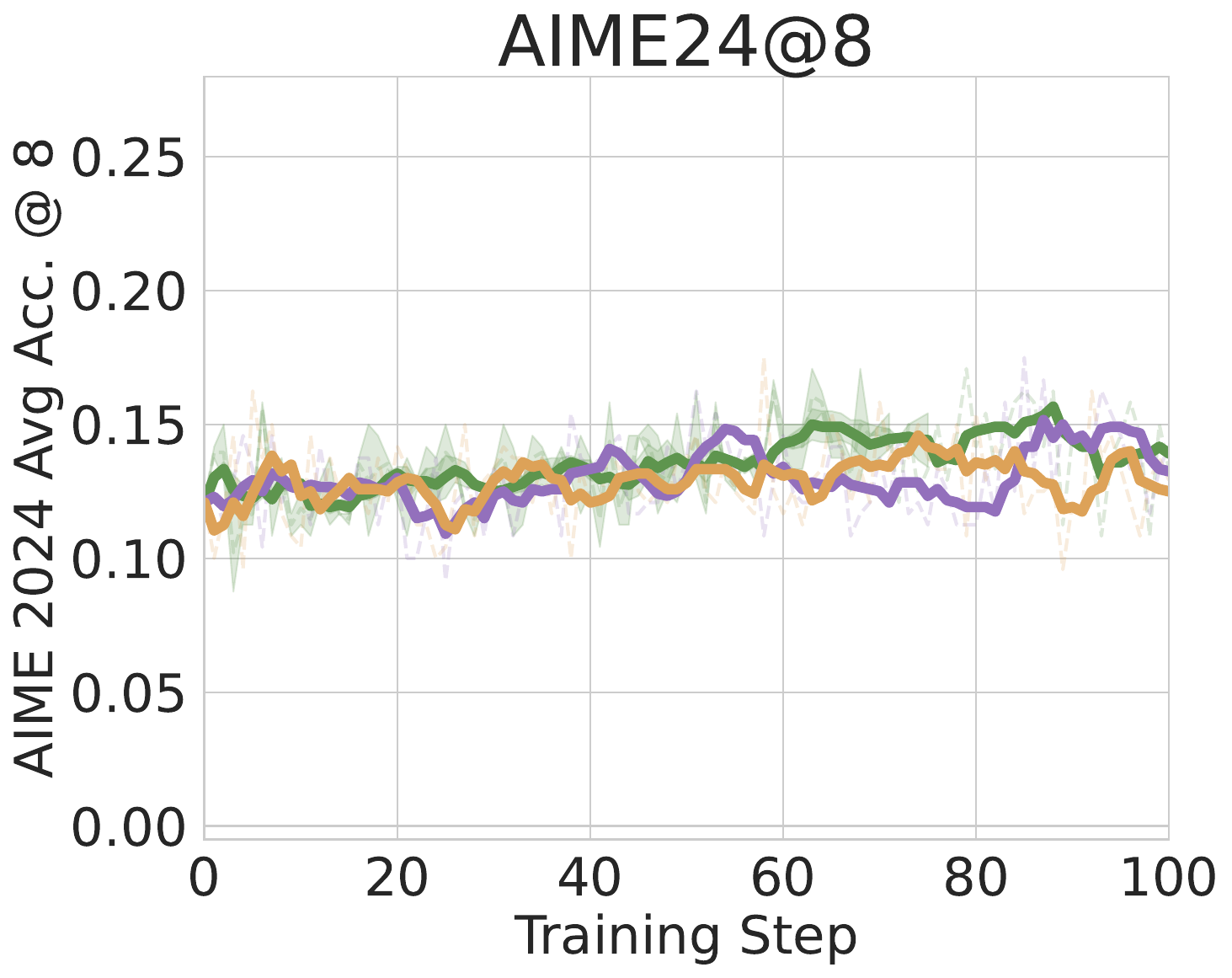}
        \includegraphics[width=0.245\linewidth]{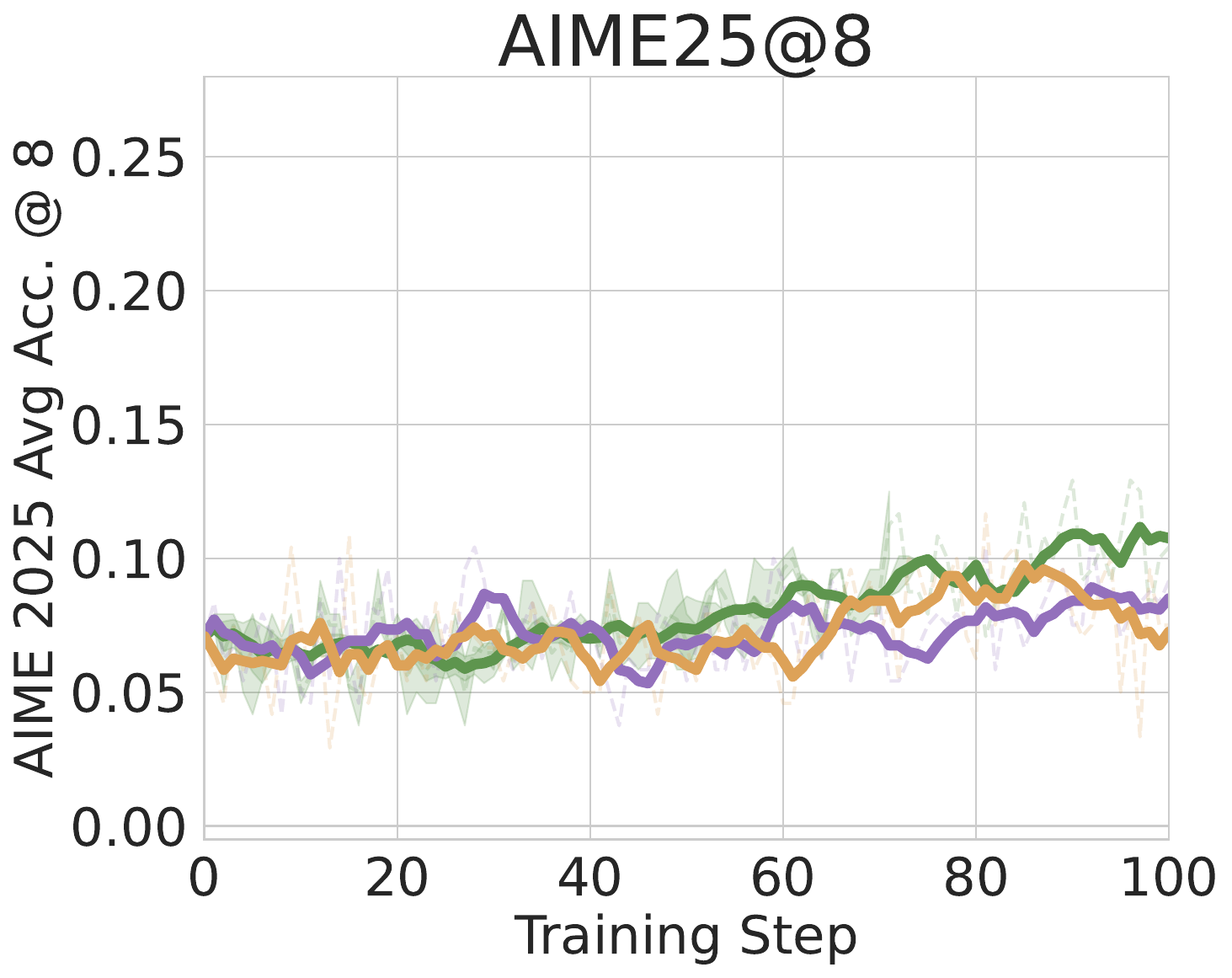}
        \caption{\qwen-Instruct Results}
        \label{fig:qwen_math_7b_instruct_results}
    \end{subfigure}\\
    
    \begin{subfigure}[t]{\textwidth}
        \centering
        \includegraphics[width=0.245\linewidth]{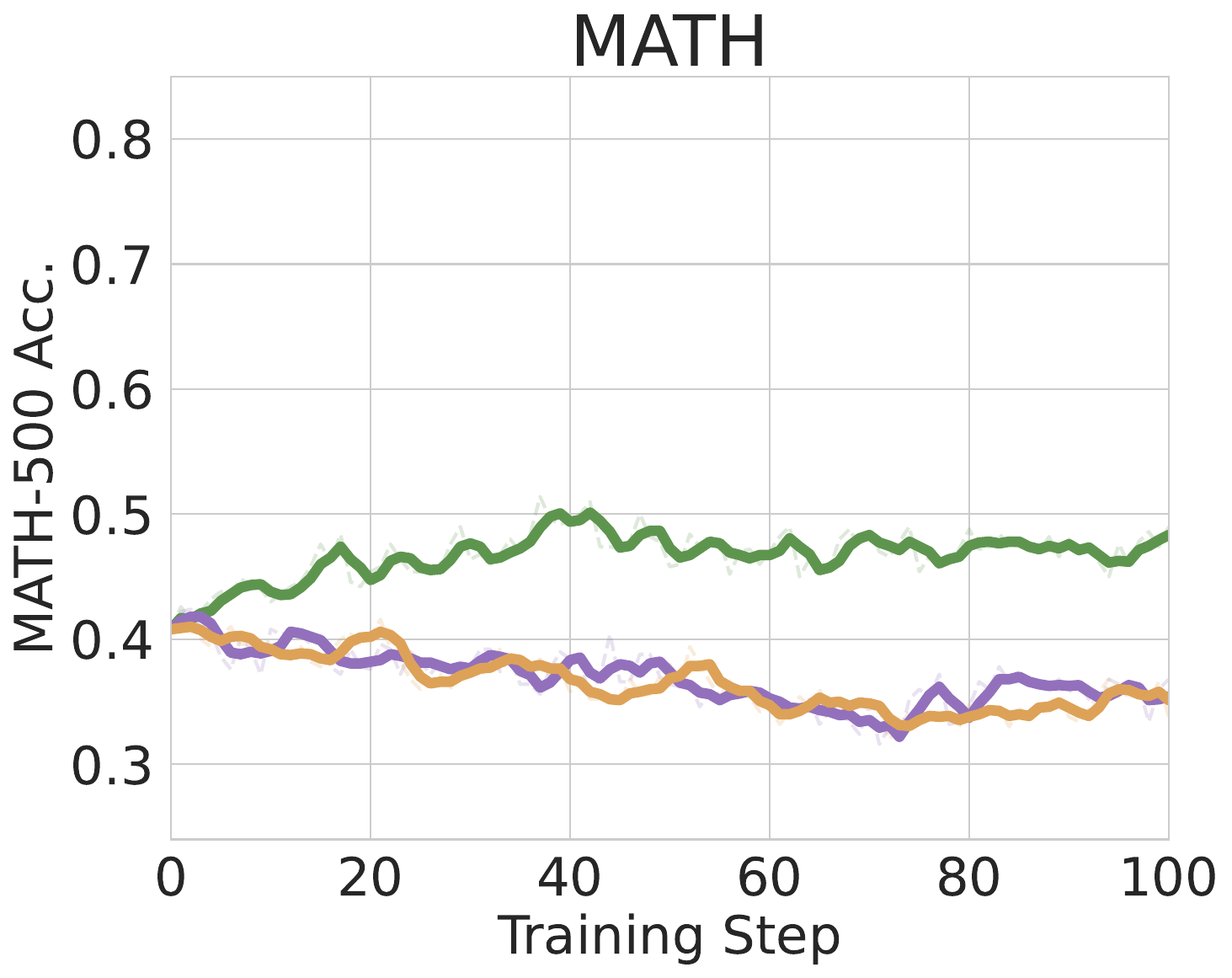}
        \includegraphics[width=0.245\linewidth]{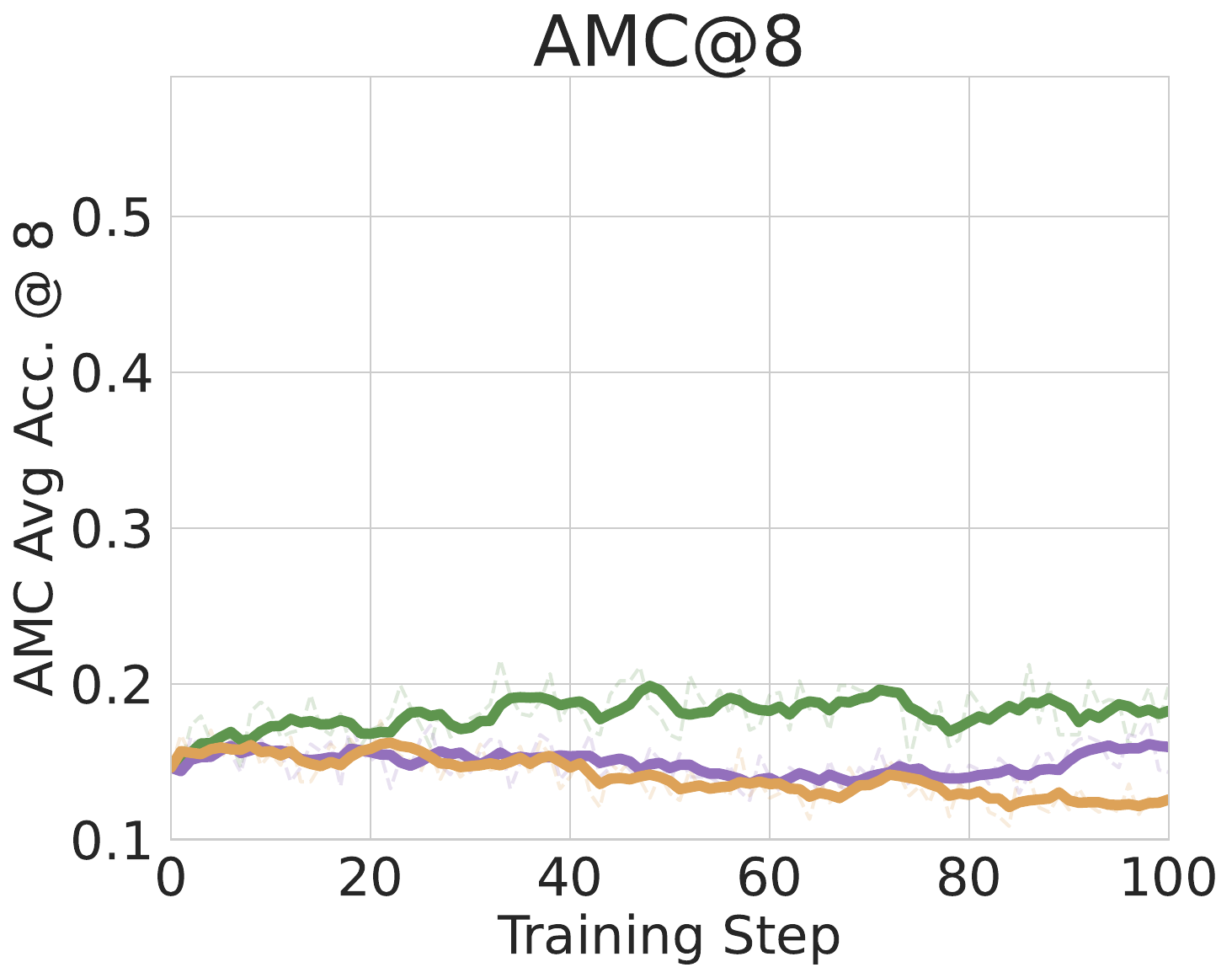}
        \includegraphics[width=0.245\linewidth]{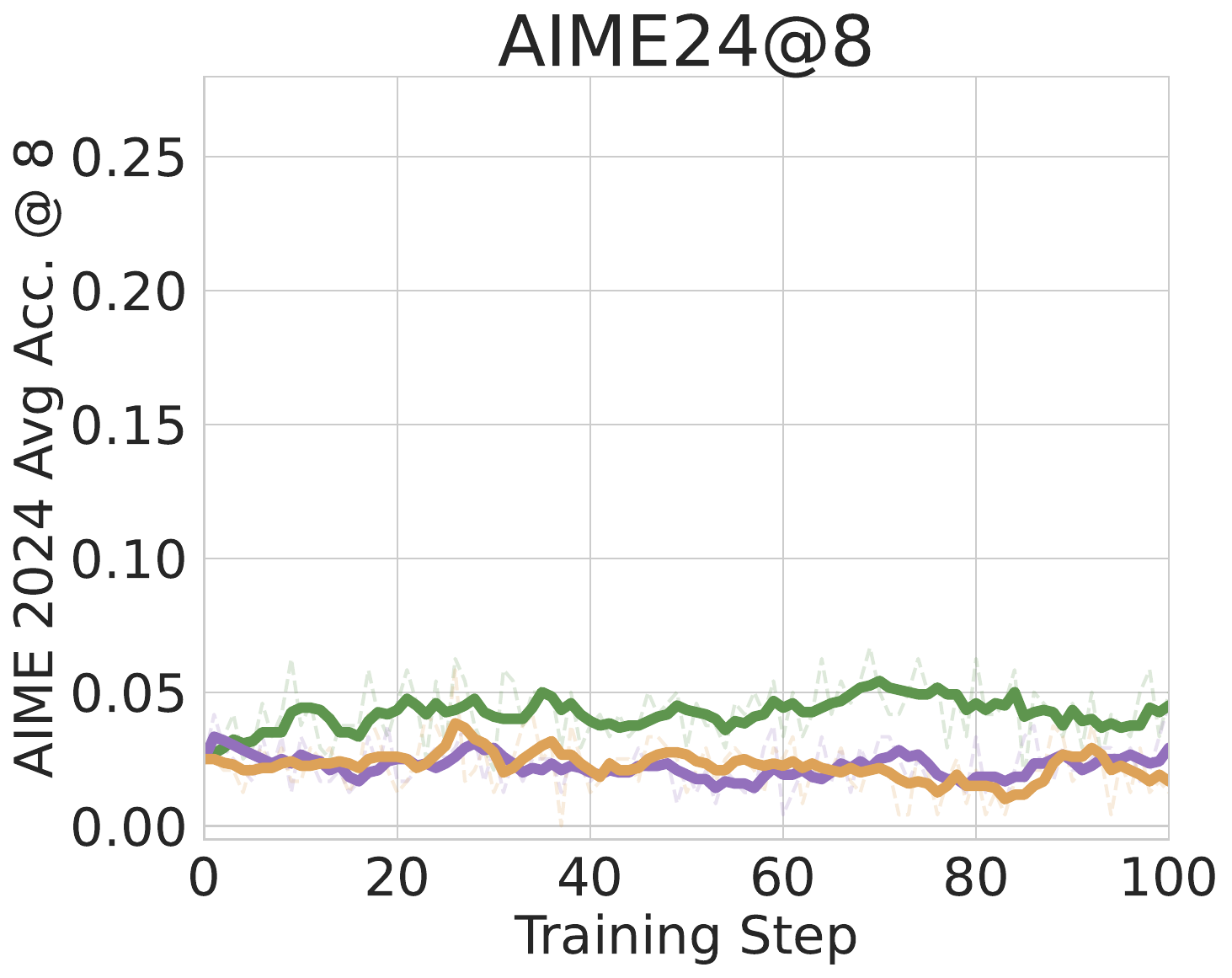}
        \includegraphics[width=0.245\linewidth]{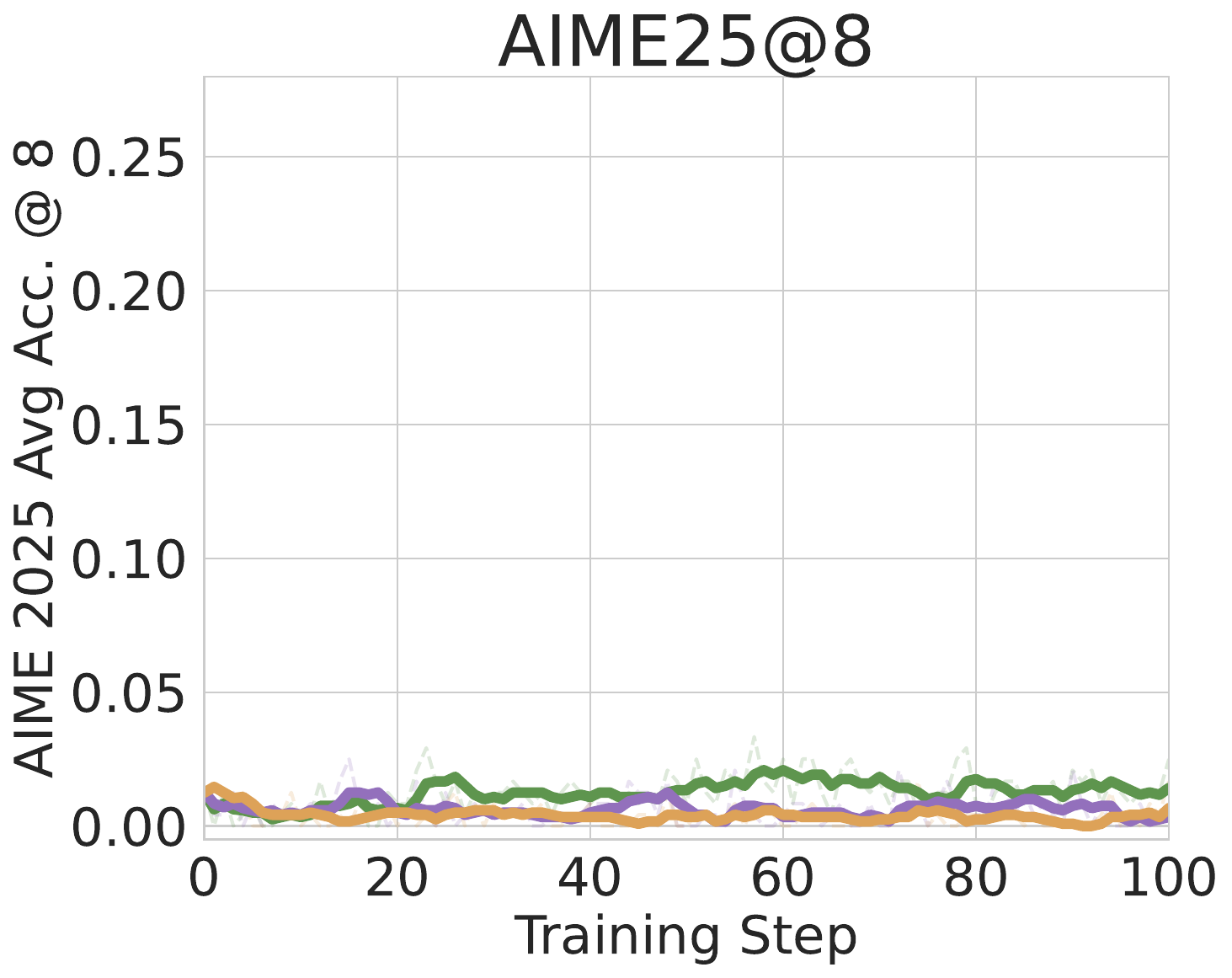}
        \caption{Llama-3.1-Tulu-3-8B Results}
        \label{fig:qwen_math_7b_instruct_results}
    \end{subfigure}%
    
    \caption{
    RLVR performance on models that have undergone RL training. We find that Qwen Instruct models show minimal improvement from our RLVR, while Llama-3.1-Tulu-3-8B exhibits similar patterns to OLMo models (Figure~\ref{fig:reward_other_family}). There is a clear performance distinction when training with ground truth labels for Llama-3.1-Tulu-3-8B, but not for the Qwen Instruct models.
    } 
    \label{fig:instruct_results}
\end{figure*}

As shown in Figure~\ref{fig:instruct_results}, Qwen Instruct models show minimal improvement from RLVR training across different reward types in our setup, even when using ground-truth rewards. We observe this same pattern in other models: their performance plateaus after sufficient RLVR training steps. 
Therefore, we conjecture that \qwenmath-Instruct benefits less from our RLVR experiments 
because it has already reached saturation from its previous RL training. 
While improved RLVR algorithms or higher-quality training data could still yield gains on the currently saturated models, we leave this exploration for future work.

Furthermore, we examine the RLVR behavior of Llama-3.1-Tulu-3-8B (Figure~\ref{fig:qwen_math_7b_instruct_results}) and find that, unlike the Qwen2.5 Instruct models, 
training with ground truth rewards provides only marginal gains on MATH and limited gains on ACM and AIME. The trends we observe for Llama-3.1-Tulu-3-8B are similar to those we observe for its base model Llama-3.1-8B. We conjecture that this might be because these models share similar prior knowledge before RL training, which leads to similar RLVR behaviors. However, unlike the Qwen Instruct models, Llama-3.1-Tulu-3-8B still shows gains on MATH despite having been trained with RLVR on the Tulu3 dataset.

\section{Qualitative Analysis on \qwenmath's Coding Reasoning Behaviors}\label{app:qualitative}
In this section, we show several qualitative examples on how \qwenmath can reason in code. In addition, we show its code reasoning behavior is robust to numerical perturbations---the model generates similar code to solve the numerical perturbed question. However, we find \qwenmath only uses natural language solutions when we rephrased the question using an alternative narrative.

\paragraph{\qwenmath can reason in code.}
In Figure~\ref{fig:qualitative_original_correct}--\ref{fig:qualitative_original},
we show 3 qualitative examples of \qwenmath outputs on 3 randomly picked questions from MATH-500. We find \qwenmath is able to conduct coding reasoning, i.e., solving a problem by writing code, and predict the code execution outputs. Surprisingly, the model can predict the code execution outputs with a relatively high accuracy---in the examples shown in Figure~\ref{fig:qualitative_original_correct} and Figure~\ref{fig:qualitative_original_correct2}, \qwenmath is able to compute the execution answers with 16-float precision \textbf{without} access to any code interpreter. In the example shown in Figure~\ref{fig:qualitative_original}, \qwenmath gives a wrong answer 4323. However, it is still very close to the ground-truth answer 4343.

\paragraph{\qwenmath's code reasoning behavior is robust to numerical perturbations.}
We further perturbed the number 999 in the question shown in Figure~\ref{fig:qualitative_original} to a few random integer numbers between the range of 100 and 1000, as shown in Figure~\ref{fig:qualitative_numerical_perturbation}--\ref{fig:qualitative_numerical_perturbation3}. We find that \qwenmath is capable of solving our perturbed questions with similar code snippets in their answers. For example, in Figure~\ref{fig:qualitative_numerical_perturbation}, we show that \qwenmath is able to derive the correct answer 4344 in the numerically perturbed version of the question in Figure~\ref{fig:qualitative_original}---we changed the input number from 999 to 1000, and the corresponding ground-truth answer changed from $4343_6$ to $4344_6$. We find \qwenmath used the same code function to solve the question, and interestingly, it predicts the correct answer for the perturbed question, although it predicted an inaccurate answer for the original question in Figure~\ref{fig:qualitative_original}.

\paragraph{\qwenmath's code reasoning behavior may not directly generalize to rephrased questions.} We also rephrased the question in Figure~\ref{fig:qualitative_original}, Figure~\ref{fig:qualitative_numerical_perturbation2}, 
 and Figure~\ref{fig:qualitative_numerical_perturbation3}, respectively, 
with an alternative narrative. As shown in Figure~\ref{fig:qualitative_narrative_perturbation_incorrect2}--\ref{fig:qualitative_narrative_perturbation_incorrect}, \qwenmath first identifies and concludes the same core question in the first sentence of its answer. However, it does not use any code to answer this rephrased question. 
We hypothesize that whether \qwenmath uses code reasoning to answer the question depends on the specific prompt distribution, i.e., whether the question is close to any of its training data that is augmented with code solutions, rather than the problem itself.

\clearpage
\begin{figure}[H]
\centering
\begin{tcolorbox}[
  enhanced,
  attach boxed title to top center={yshift=-3mm},
  colbacktitle=blue!75!black,
  colframe=blue!75!black,
  colback=white,
  width=\textwidth,
  boxed title style={size=small, colback=blue!75!black},
  fonttitle=\bfseries\color{white}
]
\begin{tcolorbox}[
  colframe=blue!75!black,
  colback=blue!5,
  coltitle=white,
  title=\textbf{Original MATH Question:},
  fonttitle=\bfseries
]
Convert the point $(0,3)$ in rectangular coordinates to polar coordinates. Enter your answer in the form $(r,\theta),$ where $r > 0$ and $0 \le \theta < 2\pi.$
\end{tcolorbox}
\vspace{0.3cm}
\begin{tcolorbox}[
  colframe=blue!50!black,
  colback=white,
  coltitle=white,
  title=\textbf{Qwen2.5-Math-7B Solution (correct):},
  fonttitle=\bfseries
]
To convert the point $(0, 3)$ in rectangular coordinates to polar coordinates $(r, \theta)$, we need to use the following formulas:

1. The distance $r$ from the origin to the point:
$$r = \sqrt{x^2 + y^2}$$

where $x = 0$ and $y = 3$.

2. The angle $\theta$ formed with the positive x-axis:

For the point $(0, 3)$:
\begin{itemize}
\item $r = \sqrt{0^2 + 3^2} = 3$
\item Since $x = 0$ and $y = 3$, we are in the case where $x = 0$ and $y > 0$, so $\theta = \frac{\pi}{2}$.
\end{itemize}

Let's write the Python code to confirm these calculations and output the polar coordinates in the desired format.

\begin{lstlisting}[language=Python, mathescape=true]
import math
# Rectangular coordinates
x = 0
y = 3
# Calculate r
r = math.sqrt(x**2 + y**2)
# Calculate $\theta$
if x == 0 and y > 0:
    theta = math.pi / 2
elif x == 0 and y < 0:
    theta = 3 * math.pi / 2
else:
    theta = math.atan2(y, x)
# Ensure $\theta$ is in the range [0, 2$\pi$)
if theta < 0:
    theta += 2 * math.pi
print(f"r={r},$\theta$ = {theta}")
\end{lstlisting}

output:
r = 3.0, $\theta$ = 1.5707963267948966

The polar coordinates for the point $(0, 3)$ are $(3, \frac{\pi}{2})$.

Thus, the final answer in the desired format is:

\fbox{$(3, \frac{\pi}{2})$}
\end{tcolorbox}
\end{tcolorbox}
\caption{Qualitative example of \qwenmath outputs on one randomly picked question from MATH-500. Note that both the code and the corresponding code execution result are generated by \qwenmath. \textbf{We do NOT provide any external code interpreter to the model.}}
\label{fig:qualitative_original_correct2}
\end{figure}
\begin{figure}[H]
\centering
\begin{tcolorbox}[
 enhanced,
 attach boxed title to top center={yshift=-3mm},
 colbacktitle=blue!75!black,
 colframe=blue!75!black,
 colback=white,
 width=\textwidth,
 boxed title style={size=small, colback=blue!75!black},
 fonttitle=\bfseries\color{white}
]
\begin{tcolorbox}[
 colframe=blue!75!black,
 colback=blue!5,
 coltitle=white,
 title=\textbf{Original MATH Question:},
 fonttitle=\bfseries
]
What is the distance, in units, between the points $(2, -6)$ and $(-4, 3)$? Express your answer in simplest radical form.
\end{tcolorbox}
\vspace{0.3cm}
\begin{tcolorbox}[
 colframe=blue!50!black,
 colback=white,
 coltitle=white,
 title=\textbf{Qwen2.5-Math-7B Solution (correct):},
 fonttitle=\bfseries
]
To find the distance between two points $(x_1, y_1)$ and $(x_2, y_2)$ in a Cartesian plane, we use the distance formula:

$$d = \sqrt{(x_2 - x_1)^2 + (y_2 - y_1)^2}$$

Given the points $(2, -6)$ and $(-4, 3)$, we can substitute these values into the formula:

$$d = \sqrt{((-4) - 2)^2 + (3 - (-6))^2}$$

Let's break this down step-by-step and compute the result using Python.

\begin{lstlisting}[language=Python]
import math

# Coordinates of the points
x1, y1 = 2, -6
x2, y2 = -4, 3

# Calculate the differences
dx = x2 - x1
dy = y2 - y1

# Calculate the distance using the distance formula
distance = math.sqrt(dx**2 + dy**2)
print(distance)
\end{lstlisting}

output:
10.816653826391969

The distance between the points $(2, -6)$ and $(-4, 3)$ is approximately $10.82$ units. To express the answer in simplest radical form, we need to check if this value can be simplified.

The distance formula calculation is:

$$d = \sqrt{((-4) - 2)^2 + (3 - (-6))^2} = \sqrt{(-6)^2 + (9)^2}$$
$$= \sqrt{36 + 81} = \sqrt{117} = \sqrt{9 \cdot 13} = 3\sqrt{13}$$

So, the exact distance in simplest radical form is $3\sqrt{13}$.

Thus, the final answer is:

\fbox{$3\sqrt{13}$}
\end{tcolorbox}
\end{tcolorbox}
\caption{Qualitative example of \qwenmath outputs on one randomly picked question from MATH-500. Note that both the code and the corresponding code execution result are generated by \qwenmath. \textbf{We do NOT provide any external code interpreter to the model.}}
\label{fig:qualitative_original_correct}
\end{figure}
\begin{figure}[H]
\centering
\begin{tcolorbox}[
  enhanced,
  attach boxed title to top center={yshift=-3mm},
  colbacktitle=blue!75!black,
  colframe=blue!75!black,
  colback=white,
  width=\textwidth,
  boxed title style={size=small, colback=blue!75!black},
  fonttitle=\bfseries\color{white}
]

\begin{tcolorbox}[
  colframe=blue!75!black,
  colback=blue!5,
  coltitle=white,
  title=\textbf{Original MATH Question:},
  fonttitle=\bfseries
]
What is the base six equivalent of $999_{10}$?
\end{tcolorbox}

\vspace{0.3cm}

\begin{tcolorbox}[
  colframe=blue!50!black,
  colback=white,
  coltitle=white,
  title=\textbf{Owen2.5-Math-7B Solution (incorrect, ground truth: $4343_6$):},
  fonttitle=\bfseries
]
To convert a decimal number to its base-six equivalent, we repeatedly divide the number by 6 and keep track of the remainders. The remainders, read in reverse order, give the base-six representation of the number.

Here's the step-by-step process to convert $999_{10}$ to base six:
\begin{enumerate}[leftmargin=*]
\item Divide the number by 6 and record the remainder.
\item Update the number to the quotient of the division.
\item Repeat steps 1 and 2 until the quotient is 0.
\item The base-six representation is the sequence of remainders read from bottom to top.
\end{enumerate}

Let's implement this in Python to get the accurate result:

\begin{lstlisting}
def decimal_to_base_six(n):
    if n == 0:
        return "0"
    
    base_six_digits = []
    while n > 0:
        remainder = n % 6
        base_six_digits.append(str(remainder))
        n = n // 6
    
    # The base-six representation is the reverse of the collected remainders
    base_six_digits.reverse()
    return "".join(base_six_digits)

# Convert 999 from decimal to base six
decimal_number = 999
base_six_representation = decimal_to_base_six(decimal_number)
print(base_six_representation)
\end{lstlisting}

4323

The base-six equivalent of $999_{10}$ is \fbox{\textbf{4323}}.
\end{tcolorbox}

\end{tcolorbox}
\caption{Qualitative example of \qwenmath outputs on one randomly picked question from MATH-500. Note that both the code and the corresponding code execution result are generated by \qwenmath. \textbf{We do NOT provide any external code interpreter to the model.}}
\label{fig:qualitative_original}
\end{figure}

\begin{figure}[H]
\centering
\begin{tcolorbox}[
  enhanced,
  attach boxed title to top center={yshift=-3mm},
  colbacktitle=green!50!black,
  colframe=green!50!black,
  colback=white,
  width=\textwidth,
  boxed title style={size=small, colback=green!50!black},
  fonttitle=\bfseries\color{white}
]
\begin{tcolorbox}[
  colframe=green!50!black,
  colback=green!5,
  coltitle=white,
  title=\textbf{Numerically Perturbed MATH Question:},
  fonttitle=\bfseries
]
What is the base six equivalent of $1000_{10}$?
\end{tcolorbox}
\vspace{0.3cm}
\begin{tcolorbox}[
  colframe=green!50!black,
  colback=white,
  coltitle=white,
  title=\textbf{Qwen2.5-Math-7B Solution (correct):},
  fonttitle=\bfseries
]
To convert a decimal number (base 10) to a base 6 number, we repeatedly divide the number by 6 and keep track of the remainders. The base 6 number is then composed of the remainders read in reverse order. Let's perform this conversion step by step.

1. Divide the number by 6, record the remainder, and replace the number with the quotient.
2. Repeat until the quotient is 0.
3. The base 6 number is the sequence of remainders read in reverse order.

Let's do this process using Python code.
\begin{lstlisting}[language=Python]
def decimal_to_base_six(n):
    if n == 0:
        return '0'
        
    base_six_digits = []
    while n > 0:
        remainder = n % 6
        base_six_digits.append(str(remainder))
        n = n // 6
        
    # The remainders give the base 6 number in reverse order
    base_six_digits.reverse()
    return ''.join(base_six_digits)

# Convert 1000 from base 10 to base 6
base_six_number = decimal_to_base_six(1000)
print(base_six_number)
\end{lstlisting}
4344

The base six equivalent of $1000_{10}$ is \fbox{$4344_6$}.
\end{tcolorbox}
\end{tcolorbox}
\caption{Qualitative example of \qwenmath outputs on the numerically perturbed MATH-500 question as shown in Figure~\ref{fig:qualitative_original}. Note that both the code and the corresponding code execution result are generated by \qwenmath. \textbf{We do NOT provide any external code interpreter to the model.}}
\label{fig:qualitative_numerical_perturbation}
\end{figure}

\begin{figure}[H]
\centering
\begin{tcolorbox}[
  enhanced,
  attach boxed title to top center={yshift=-3mm},
  colbacktitle=green!50!black,
  colframe=green!50!black,
  colback=white,
  width=\textwidth,
  boxed title style={size=small, colback=green!50!black},
  fonttitle=\bfseries\color{white}
]
\begin{tcolorbox}[
  colframe=green!50!black,
  colback=green!5,
  coltitle=white,
  title=\textbf{Numerically Perturbed MATH Question:},
  fonttitle=\bfseries
]
What is the base six equivalent of $998_{10}$?
\end{tcolorbox}
\vspace{0.3cm}
\begin{tcolorbox}[
  colframe=green!50!black,
  colback=white,
  coltitle=white,
  title=\textbf{Qwen2.5-Math-7B Solution (correct):},
  fonttitle=\bfseries
]
To convert the decimal number $998_{10}$ to its base six equivalent, we need to repeatedly divide the number by 6 and keep track of the remainders. The base six digits are the remainders read from bottom to top (from last division to first).

Here is the step-by-step process:
\begin{enumerate}
\item Divide the number by 6.
\item Record the remainder.
\item Update the number to the quotient of the division.
\item Repeat steps 1-3 until the quotient is 0.
\item The base six number is the sequence of remainders read from bottom to top.
\end{enumerate}

Let's do this step-by-step for $998_{10}$:
\begin{align}
998 \div 6 &= 166 \text{ remainder } 2\\
166 \div 6 &= 27 \text{ remainder } 4\\
27 \div 6 &= 4 \text{ remainder } 3\\
4 \div 6 &= 0 \text{ remainder } 4
\end{align}

The remainders, read from bottom to top, give us the base six number. So, $998_{10}$ in base six is $4342_6$.

We can verify this by converting back to decimal:
\begin{align}
4 \cdot 6^3 + 3 \cdot 6^2 + 4 \cdot 6^1 + 2 \cdot 6^0 &= 4 \cdot 216 + 3 \cdot 36 + 4 \cdot 6 + 2\\
&= 864 + 108 + 24 + 2\\
&= 998_{10}
\end{align}

Now, let's implement this in Python to ensure the accuracy.

\begin{verbatim}
def decimal_to_base_six(n):
    if n == 0:
        return "0"
    base_six_digits = []
    while n > 0:
        remainder = n % 6
        base_six_digits.append(str(remainder))
        n = n // 6
    base_six_digits.reverse()
    return ''.join(base_six_digits)

# Convert 998 from base 10 to base 6
base_six_number = decimal_to_base_six(998)
print(base_six_number)
\end{verbatim}

\begin{verbatim}
4342
\end{verbatim}

The base six equivalent of $998_{10}$ is $\boxed{4342_6}$.
\end{tcolorbox}
\end{tcolorbox}
\caption{Qualitative example of \qwenmath outputs on the numerically perturbed MATH-500 question as shown in Figure~\ref{fig:qualitative_original}. Note that both the code and the corresponding code execution result are generated by \qwenmath. \textbf{We do NOT provide any external code interpreter to the model.}}
\label{fig:qualitative_numerical_perturbation2}
\end{figure}

\begin{figure}[H]
\centering
\begin{tcolorbox}[
  enhanced,
  attach boxed title to top center={yshift=-3mm},
  colbacktitle=green!50!black,
  colframe=green!50!black,
  colback=white,
  width=\textwidth,
  boxed title style={size=small, colback=green!50!black},
  fonttitle=\bfseries\color{white}
]
\begin{tcolorbox}[
  colframe=green!50!black,
  colback=green!5,
  coltitle=white,
  title=\textbf{Numerically Perturbed MATH Question:},
  fonttitle=\bfseries
]
What is the base six equivalent of $100_{10}$?
\end{tcolorbox}
\vspace{0.3cm}
\begin{tcolorbox}[
  colframe=green!50!black,
  colback=white,
  coltitle=white,
  title=\textbf{Qwen2.5-Math-7B Solution (correct):},
  fonttitle=\bfseries
]
To convert a decimal number (base 10) to a base-six number, we can repeatedly divide the number by 6 and keep track of the remainders. The conversion process involves the following steps:

\begin{enumerate}
\item Divide the number by 6.
\item Record the remainder.
\item Update the number to the quotient from the division.
\item Repeat the process until the quotient is 0.
\item The base-six number is the sequence of remainders read in reverse order.
\end{enumerate}

Let's convert $100_{10}$ to base six using Python.
\begin{verbatim}
def decimal_to_base_six(n):
    if n == 0:
        return "0"
    
    base_six_digits = []
    
    while n > 0:
        remainder = n % 6
        base_six_digits.append(str(remainder))
        n = n // 6
    
    # The base-six number is the sequence of remainders read 
    # in reverse order
    base_six_digits.reverse()
    return ''.join(base_six_digits)

# Convert 100 (decimal) to base six
decimal_number = 100
base_six_number = decimal_to_base_six(decimal_number)
print(base_six_number)
\end{verbatim}

\begin{verbatim}
244
\end{verbatim}

The base six equivalent of $100_{10}$ is $\boxed{244_6}$.
\end{tcolorbox}
\end{tcolorbox}
\caption{Qualitative example of \qwenmath outputs on the numerically perturbed MATH-500 question as shown in Figure~\ref{fig:qualitative_original}. Note that both the code and the corresponding code execution result are generated by \qwenmath. \textbf{We do NOT provide any external code interpreter to the model.}}
\label{fig:qualitative_numerical_perturbation3}
\end{figure}

\begin{figure}[H]
\centering
\begin{tcolorbox}[
 enhanced,
 attach boxed title to top center={yshift=-3mm},
 colbacktitle=red!75!black,
 colframe=red!75!black,
 colback=white,
 width=\textwidth,
 boxed title style={size=small, colback=red!75!black},
 fonttitle=\bfseries\color{white}
]
\begin{tcolorbox}[
 colframe=red!75!black,
 colback=red!5,
 coltitle=white,
 title=\textbf{Rephrased MATH Question in an Alternative Narrative:},
 fonttitle=\bfseries
]
If I have 999 objects in standard decimal counting, and I need to express this quantity using only the digits 0-5 in a positional numeral system, what sequence of digits would represent this amount?
\end{tcolorbox}
\vspace{0.3cm}
\begin{tcolorbox}[
 colframe=red!75!black,
 colback=white,
 coltitle=white,
 title=\textbf{Qwen2.5-Math-7B Solution 2 (correct):},
 fonttitle=\bfseries
]
To represent 999 using only the digits 0-5, we need to convert it to a base 6 number system. We need to divide this number by the powers of 6 and take the remainder until we run out of powers of 6.

999 divided by $6^3$ (which is 216) is 4 with a remainder of 135.\\
135 divided by $6^2$ (which is 36) is 3 with a remainder of 27.\\
27 divided by $6^1$ (which is 6) is 4 with a remainder of 3.\\
3 divided by $6^0$ (which is 1) is 3 with no remainder.

So, 999 in base 6 is $4343_6$.

\begin{align*}
4 \times 6^3 + 3 \times 6^2 + 4 \times 6^1 + 3 \times 6^0 &= 4 \times 216 + 3 \times 36 + 4 \times 6 + 3 \times 1\\
&= 864 + 108 + 24 + 3\\
&= 999
\end{align*}

Therefore, the representation of 999 using only digits 0-5 in a base-6 numeral system is $4343_6$.
\end{tcolorbox}
\end{tcolorbox}
\caption{Qualitative example of \qwenmath outputs on the MATH-500 question shown in Figure~\ref{fig:qualitative_original} but with an alternative narrative. \qwenmath uses natural language to answer the rephrased question without code reasoning.}
\label{fig:qualitative_narrative_perturbation_incorrect2}
\end{figure}
\begin{figure}[H]
\centering
\begin{tcolorbox}[
 enhanced,
 attach boxed title to top center={yshift=-3mm},
 colbacktitle=red!75!black,
 colframe=red!75!black,
 colback=white,
 width=\textwidth,
 boxed title style={size=small, colback=red!75!black},
 fonttitle=\bfseries\color{white}
]
\begin{tcolorbox}[
 colframe=red!75!black,
 colback=red!5,
 coltitle=white,
 title=\textbf{Rephrased MATH Question in an Alternative Narrative:},
 fonttitle=\bfseries
]
If I have 998 objects in standard decimal counting, and I need to express this quantity using only the digits 0-5 in a positional numeral system, what sequence of digits would represent this amount?
\end{tcolorbox}
\vspace{0.3cm}
\begin{tcolorbox}[
 colframe=red!75!black,
 colback=white,
 coltitle=white,
 title=\textbf{Qwen2.5-Math-7B Solution 2 (correct):},
 fonttitle=\bfseries
]
To express the quantity of 998 using only the digits 0-5 in a positional numeral system, we need to convert 998 to a base-6 number. Here's the step-by-step process:

1. \textbf{Determine the largest power of 6 less than or equal to 998:}
   - $6^0 = 1$
   - $6^1 = 6$
   - $6^2 = 36$
   - $6^3 = 216$
   - $6^4 = 1296$ (which is greater than 998)
   
   So, the largest power of 6 less than or equal to 998 is $6^3 = 216$.

2. \textbf{Find the coefficient for $6^3$:}
   - $998 \div 216 \approx 4.62$
   - The integer part is 4, so the coefficient for $6^3$ is 4.
   - Subtract $4 \times 216 = 864$ from 998 to get the remainder: $998 - 864 = 134$.

3. \textbf{Find the coefficient for $6^2$:}
   - $134 \div 36 \approx 3.72$
   - The integer part is 3, so the coefficient for $6^2$ is 3.
   - Subtract $3 \times 36 = 108$ from 134 to get the remainder: $134 - 108 = 26$.

4. \textbf{Find the coefficient for $6^1$:}
   - $26 \div 6 \approx 4.33$
   - The integer part is 4, so the coefficient for $6^1$ is 4.
   - Subtract $4 \times 6 = 24$ from 26 to get the remainder: $26 - 24 = 2$.

5. \textbf{Find the coefficient for $6^0$:}
   - The remainder is 2, so the coefficient for $6^0$ is 2.

Putting it all together, the base-6 representation of 998 is $4342_6$.

Therefore, the sequence of digits that represents 998 in a base-6 positional numeral system is \fbox{4342}.
\end{tcolorbox}
\end{tcolorbox}
\caption{Qualitative examples of \qwenmath outputs on the MATH-500 question shown in Figure~\ref{fig:qualitative_original} but with an alternative narrative. \qwenmath uses natural language to answer the rephrased question without code reasoning.}
\label{fig:qualitative_narrative_perturbation_correct}
\end{figure}
\begin{figure}[H]
\centering
\begin{tcolorbox}[
 enhanced,
 attach boxed title to top center={yshift=-3mm},
 colbacktitle=red!75!black,
 colframe=red!75!black,
 colback=white,
 width=\textwidth,
 boxed title style={size=small, colback=red!75!black},
 fonttitle=\bfseries\color{white}
]
\begin{tcolorbox}[
 colframe=red!75!black,
 colback=red!5,
 coltitle=white,
 title=\textbf{Rephrased MATH Question in an Alternative Narrative:},
 fonttitle=\bfseries
]
If I have 100 objects in standard decimal counting, and I need to express this quantity using only the digits 0-5 in a positional numeral system, what sequence of digits would represent this amount?
\end{tcolorbox}
\vspace{0.3cm}
\begin{tcolorbox}[
 colframe=red!75!black,
 colback=white,
 coltitle=white,
 title=\textbf{Qwen2.5-Math-7B Solution 1 (incorrect, ground truth: 244):},
 fonttitle=\bfseries
]
To represent the number 100 using only the digits 0-5, you can use a base-6 numeral system. In a base-6 system, each digit represents a power of 6, starting from the right with $6^0$. The number 100 in base-10 can be converted to base-6 as follows:

\begin{align*}
100 &= 2 \times 6^2 + 0 \times 6^1 + 4 \times 6^0\\
    &= 2 \times 36 + 0 \times 6 + 4 \times 1\\
    &= 72 + 0 + 4\\
    &= 76
\end{align*}

However, since we are using base-6, the correct representation of 100 in this system is 204 (2 $\times$ $6^2$ + 0 $\times$ $6^1$ + 4 $\times$ $6^0$).
\end{tcolorbox}
\end{tcolorbox}
\caption{Qualitative examples of \qwenmath outputs on the MATH-500 question shown in Figure~\ref{fig:qualitative_original} but with an alternative narrative and numerical perturbation. \qwenmath uses natural language to answer the rephrased question without code reasoning. The equation expansion step in the solution is incorrect, leading to a final wrong answer: the calculation claims that $100 = 2 \times 6^2 + 0 \times 6^1 + 4 \times 6^0 = 72 + 0 + 4 = 76$, which is mathematically inconsistent.}
\label{fig:qualitative_narrative_perturbation_incorrect}
\end{figure}


\end{document}